\documentclass[11pt, a4paper, logo, onecolumn, copyright]{googledeepmind}

\usepackage[authoryear, sort&compress, round]{natbib}
\usepackage{hyperref}
\usepackage{url}

\usepackage{algorithm}      
\usepackage{algpseudocode}  
\usepackage{pgfplots}
\usepgfplotslibrary{groupplots}
\usepackage{amsfonts}
\usepackage{amsmath}
\usepackage{amssymb} 
\usepackage{balance}
\usepackage[table]{xcolor}
\usepackage{booktabs}
\usepackage{comment}
\usepackage[separate-uncertainty,table-align-uncertainty,retain-zero-uncertainty]{siunitx}
\usepackage{tablefootnote}
\usepackage{graphicx}
\usepackage{multirow}
\usepackage{listings}  
\usepackage{enumitem} 
\usepackage[font=small]{caption}
\definecolor{codegray}{gray}{0.95}
\definecolor{commentgreen}{rgb}{0.13, 0.54, 0.13}
\definecolor{keywordblue}{rgb}{0.13, 0.13, 1}
{}
{}
\definecolor{codegreen}{rgb}{0,0.6,0}
\definecolor{codegray}{rgb}{0.5,0.5,0.5}
\definecolor{codepurple}{rgb}{0.58,0,0.82}
\definecolor{codeblue}{rgb}{0.0,0.0,0.6}
\lstdefinestyle{codeblock}{
    language=C++,
    basicstyle=\ttfamily\scriptsize,
    columns=fullflexible,
    keepspaces=true,
    tabsize=2,
    showstringspaces=false,
    commentstyle=\color{codegreen}\itshape,
    keywordstyle=\bfseries\color{codeblue},
    stringstyle=\color{codepurple},
    morekeywords={class,override,return,struct,Action,Player,MCTSNode,Game,GameType,State,VirtualLossDeltas},
    breaklines=true,
    breakatwhitespace=false,
    postbreak=\mbox{\textcolor{red}{$\hookrightarrow$}\space},
    frame=single,
    framerule=0.3pt,
    xleftmargin=1em,
    framexleftmargin=1em,
    aboveskip=0.8em,
    belowskip=0.8em
}
\lstdefinestyle{substrate}{style=codeblock}
\newcommand{\Comp}{\mathcal{M}}

\newcommand{\mevo}{\ensuremath{m^{\star}_{\mathrm{evo}}}}

\newcommand{\code}[1]{\texttt{#1}}
\newcommand{\mstarBIFV}{\ensuremath{m^{\star}_{\mathrm{BIFV}}}}
\newcommand{\bifv}{\textsc{Bifv-Mcts}}

\setlist[enumerate]{nosep, wide, labelwidth=0pt, labelindent=0pt}

\setlist[itemize]{nosep, wide, labelwidth=0pt, leftmargin=0pt, labelindent=0pt}

\title{Modular Discovery of General Game-Playing Algorithms with Large Language Models}

\correspondingauthor{lizun@google.com}

\keywords{Multi-Agent LLM Systems, Meta-Learning, General Game-Playing; Monte-Carlo Tree Search; Algorithmic Discovery}

\author[1]{Zun Li}
\author[1]{John Schultz}
\author[2]{Marc Lanctot}
\author[2]{Daniel Hennes}

\affil[1]{Google DeepMind}
\affil[2]{Work done while at Google DeepMind}

\begin{abstract}
General Game Playing across arbitrary games from rules alone remains challenging due to differing algorithmic requirements across game classes and strict decision-time constraints. Rather than hand-designing search heuristics for specific domains, can we leverage Large Language Models (LLMs) to discover general game-playing algorithms? Because language models can propose and refactor structured code, they provide an expressive proposal engine for exploring the space of algorithmic designs. We introduce a multi-agent LLM meta-learning system to co-evolve game-agnostic procedural search mechanisms in C++ alongside domain heuristics synthesized directly from game rules. Controlling the compute budget, we benchmark the discovered mechanisms across more than 400 diverse environments, including OpenSpiel training and held-out games, procedural simulation engines, and games with deep neural policy-value representations trained via PPO. Evaluated via AlphaRank stationary distributions and Soft Condorcet Optimization (SCO) against 15 established MCTS baselines, the discovered search mechanisms consistently achieve top-tier ratings and pairwise ballot majorities over most baselines across independent evolutionary runs, generalizing to unseen human-designed and procedurally synthesized games and remaining competitive with baselines on frozen neural network representations.
\end{abstract}

\begin{document}

\maketitle

\section{Introduction}
\label{sec:intro}

Strong game-playing performance has driven foundational progress in artificial intelligence. While landmark systems such as Deep Blue in chess~\citep{DeepBlue}, AlphaGo in Go~\citep{Silver16AlphaGo}, Libratus in poker~\citep{Brown17Libratus}, and Cicero in Diplomacy~\citep{Cicero} have achieved notable success in individual domains, developing algorithms for \emph{General Game Playing} (GGP)---the ability to play arbitrary, unseen games from rules alone~\citep{genesereth2005general, openspiel2019}---remains an enduring challenge. A central reason for this difficulty is the differing algorithmic requirements across game classes: standard Monte Carlo Tree Search (MCTS) paradigms such as AlphaZero~\citep{Silver18AlphaZero} excel in deterministic, perfect-information domains, but cannot minimize exploitability in imperfect-information settings like poker~\citep{Bowling15Poker, Brown17Libratus}. Conversely, game-theoretic solvers designed to minimize exploitability typically do not scale to complex combinatorial action spaces or multi-agent settings.

Large language models (LLMs) have demonstrated promising capabilities in automated programming and algorithmic discovery, synthesizing executable routines across mathematics and optimization~\citep{funsearch2023, alphaevolve2025, eoh2024, ma2024eureka}. In game playing and decision making, LLMs have been applied to synthesizing interactive environments~\citep{lehrach2025codeworldmodelsgeneral, verma2025measuringgeneralintelligencegenerated}, devising efficient learning heuristics~\citep{li2026discoveringmultiagentlearningalgorithms}, and generating interpretable policies~\citep{hennes2026code}. A natural question arises: can we leverage LLMs to discover general game-playing algorithms that transfer across diverse game structures without manual redesign?

In this paper, we present a multi-agent LLM meta-learning system that discovers simulation-based search algorithms through functional decomposition. Recognizing that general game playing requires both domain-invariant planning logic and game-specific knowledge, our approach factorizes search into two co-evolving components: a \emph{procedural search mechanism} in high-performance C++ that governs tree selection, backup, and exploration across diverse games; and concurrent \emph{domain-specific knowledge heuristics} (action priors, value estimators, and history resamplers) synthesized directly from game rules. By co-evolving universal search logic alongside an evolving curriculum of game-specific heuristics, our framework discovers general planning mechanisms without requiring per-game algorithmic tuning.

We evaluate our approach across extensive benchmarks spanning deterministic board games, stochastic domains, and imperfect-information environments. To benchmark across heterogeneous games without distortion from arbitrary score scales or intransitive dominance cycles, we evaluate multi-candidate tournaments via AlphaRank stationary distributions~\citep{omidshafiei2019alpha} and synthesize continuous consensus ratings using Soft Condorcet Optimization (SCO)~\citep{lanctot2024soft}, which resolves cyclic dynamics without assuming transitive Elo metrics. Under this framework controlling the compute budget, the discovered mechanisms consistently achieve competitive, top-tier ratings across independent evolutionary runs. In particular, independent lineages synthesize diverse mathematical operators, ranging from branching-aware exploration and rational schedules to non-stationary polynomial backups, with our representative case study (\bifv, selected as the most consistent lineage) ranking first across all four benchmark tiers.
\section{Preliminaries: Games and a Unified Modular Search Template}
\label{sec:prelim}
\label{sec:prelim:games}
\label{sec:prelim:template}

We consider finite extensive-form games played by a set of players $\mathcal{N} = \{1, \dots, n\}$.
A game progresses sequentially through histories $h \in \mathcal{H}$, initialized at the empty root history $\emptyset$.
At each non-terminal history $h$, the player function $P(h) \in \mathcal{N} \cup \{c\}$ identifies the acting agent, who chooses an action $a$ from the legal action set $A(h)$, transitioning the environment to the successor history $h \cdot a \in \mathcal{H}$ (denoted $h.\mathrm{Apply}(a)$).
If $P(h) = c$, the environment executes a stochastic event by sampling a chance action $a_c \sim \sigma_c(\cdot \mid h)$ according to a fixed probability distribution.
Under imperfect information, players do not directly observe the full ground-truth history $h$; instead, when player $p = P(h)$ is called to play, they observe only an \emph{information state} $s = s(h)$, which partitions histories into equivalence classes $\mathcal{H}(s) = \{h' \in \mathcal{H} \mid s(h') = s\}$ sharing identical private and public observations.
Consequently, behavioral strategies $\pi(a \mid s)$ depend solely on the observed information state $s(h)$ rather than the hidden history, ensuring that players act only on information accessible to them.
Perfect-information games correspond to the special case where $|\mathcal{H}(s)| = 1$ everywhere.
The interaction terminates when the sequence of actions reaches a terminal history $h \in \mathcal{Z} \subset \mathcal{H}$, whereupon each player $i \in \mathcal{N}$ receives a scalar utility payoff given by the terminal utility function $u_i(h)$, collected in the bounded payoff vector $u(h) \in [U_{\min}, U_{\max}]^{|\mathcal{N}|}$.

In General Game Playing (GGP), an agent must reason and act effectively across arbitrary, unseen games without task-specific pre-training~\citep{genesereth2005general, openspiel2019}.
Simulation-based tree search---specifically the family of Monte Carlo Tree Search (MCTS)~\citep{kocsis2006bandit,swiechowski2023monte} and Information Set MCTS (IS-MCTS)~\citep{cowling2012information}---serves as the foundational planning engine for GGP: by querying an executable forward simulator at decision time, it dynamically explores potential futures and accumulates tree statistics, operating as a general policy-improvement operator that extracts strong actions on the fly~\citep{Silver18AlphaZero}.
Most established simulation search algorithms across both perfect- and imperfect-information games can be formalized under a unified modular execution template (Algorithm~\ref{alg:template}).
Under this template, an agent is defined by a \emph{modular composition set} $\Comp = \langle m, \kappa \rangle$ factorized into two tiers:
\begin{enumerate}[leftmargin=1.5em, itemsep=2pt, topsep=2pt]
  \item \textbf{Game-Agnostic Mechanism} $m\in\Omega_{\text{mech}}$, where $m = (\code{TreePolicy}, \code{ValueBackup},\\ \code{RootActionSelector}, \code{FinalActionSelector}, \code{VirtualLoss})$ govern procedural planning logic that executes identically across arbitrary games: (1)~\code{RootActionSelector} allocates the simulation budget across candidate root branches over progression $(t, B)$; (2)~\code{TreePolicy} balances exploitation versus exploration during tree descent; (3)~\code{ValueBackup} propagates leaf payoff vectors back along the trajectory; (4)~\code{VirtualLoss} steers concurrent worker threads apart during asynchronous parallel search~\citep{chaslot2008parallel}; and (5)~\code{FinalActionSelector} extracts the final move $a^\ast$ and posterior visit distribution $\hat{\pi}$.
  \item \textbf{Game-Specific Knowledge} $\kappa\in\mathcal{K}(g)$, where $\kappa = (\code{PriorPolicy}, \code{ValueEstimator},\\ \code{ResampleHistory})$ supply domain heuristics specialized to the game's rules and state representation: (i)~\code{PriorPolicy} $\pi(a \mid s) \in \Delta(A(s))$ guides early branch exploration conditioned on the observed information state $s = s(h)$; (ii)~\code{ValueEstimator} $v(s) \in [U_{\min}, U_{\max}]^{|\mathcal{N}|}$ acts as a value function estimating the expected return-to-go from non-terminal states $s$, bypassing rollout simulations; and (iii)~\code{ResampleHistory} $b(h \mid s) \in \Delta(\mathcal{H}(s))$ samples hidden world histories consistent with public observation history for online determinization under imperfect information (which degenerates to an identity mapping in perfect-information games).
\end{enumerate}

\begin{algorithm}[t]
\caption{\footnotesize \textsc{ExecuteSearch}: One decision step produced by modular composition $\mathcal{M}$.}
\label{alg:template}
\begin{algorithmic}[1]
\scriptsize
\Require Root info state $s_0$; simulation budget $B$; modular composition $\mathcal{M} = \langle m, \kappa \rangle$
\State $root \gets \Call{GetOrCreateNode}{s_0, P(h_0)}$;\quad \Call{Expand}{root, $\mathcal{M}.\code{PriorPolicy}(s_0)$}
\For{$t = 1, \dots, B$ \textbf{in parallel across threads}}
  \State $a \gets \mathcal{M}.\code{RootActionSelector}(root, t, B)$;\quad $d \gets 0$ \Comment{Select root action}
  \State $h \gets \mathcal{M}.\code{ResampleHistory}(s_0)$ \textbf{if} imperfect info \textbf{else} $s_0$ \Comment{Determinize}
  \State $\Delta \gets \mathcal{M}.\code{VirtualLoss}(root.\mathrm{child}[a], d)$;\quad $path \gets \langle (root, a, d, P(h), \Delta) \rangle$;\quad $h.\mathrm{Apply}(a)$
  \While{$h$ is not terminal} \Comment{Selection descent}
    \State \textbf{if} $P(h) = c$ \textbf{then} sample $a_c \sim \sigma_c(\cdot \mid h)$;\; $h.\mathrm{Apply}(a_c)$;\; \textbf{continue}
    \State $node \gets \Call{GetOrCreateNode}{s_{P(h)}(h), P(h)}$
    \If{$node$ is unexpanded} \Comment{Expansion \& evaluation}
      \State \Call{Expand}{node, $\mathcal{M}.\code{PriorPolicy}(s_{P(h)}(h))$};\quad $\vec{v} \gets \mathcal{M}.\code{ValueEstimator}(s_{P(h)}(h))$;\quad \textbf{break}
    \EndIf
    \State $a \gets \mathcal{M}.\code{TreePolicy}(node)$;\quad $d \gets d + 1$
    \State $\Delta \gets \mathcal{M}.\code{VirtualLoss}(node.\mathrm{child}[a], d)$;\quad \Call{Append}{$path, (node, a, d, P(h), \Delta)$};\quad $h.\mathrm{Apply}(a)$
  \EndWhile
  \State \textbf{if} $h$ is terminal \textbf{then} $\vec{v} \gets u(h)$
  \For{$(n, a, d, p, \Delta)$ \textbf{in} $\mathrm{reverse}(path)$} \Comment{Value backpropagation}
    \State \Call{RevertLoss}{$n, a, \Delta$};\quad $\mathcal{M}.\code{ValueBackup}(n.\mathrm{child}[a], \vec{v}, d, p)$
  \EndFor
\EndFor
\State \Return $(a^\ast, \hat{\pi}) \gets \mathcal{M}.\code{FinalActionSelector}(root)$ \Comment{Move choice}
\end{algorithmic}
\end{algorithm}

As formalized in Algorithm~\ref{alg:template}, search operates under the standard Information Set MCTS (IS-MCTS) formulation~\citep{cowling2012information,openspiel2019}: for each simulation trajectory, candidate root actions are selected from root statistics over the observed information state $s_0$, followed by sampling a consistent world state $h \sim \Comp.\code{ResampleHistory}(s_0)$. Internal search nodes along the simulation path are keyed by the acting player's observed information state $(s_{P(h)}(h), P(h))$, enabling DAG transposition sharing across identical beliefs while ensuring players condition decisions strictly on observations accessible to them; full typed signatures and C++ controllers are in App.~\ref{app:template}.

\paragraph{Expressiveness and Algorithmic Scope.}
Standard algorithms correspond to specific coordinate points in this design space: \textbf{Vanilla UCT}~\citep{kocsis2006bandit} pairs UCB1 selection with mean backup, \textbf{PUCT}~\citep{Silver18AlphaZero} adds prior-weighted exploration bonuses, \textbf{MENTS}~\citep{xiao2019maximum} introduces softmax-regularized backups, and \textbf{Gumbel AlphaZero}~\citep{danihelka2022policy} implements sequential halving at the root.
Furthermore, because utilities and evaluations are vectors $\vec{v} \in \mathbb{R}^{|\mathcal{N}|}$ across players, root allocations can condition on $(t, B)$, and virtual loss can modulate over search depth, this modular template natively spans two-player, multi-player, and general-sum games without structural change.
\section{Modular Co-Evolution for General Game Search with LLMs}
\label{sec:method}
We present a modular co-evolutionary framework for discovering simulation-based search algorithms in extensive-form games based on AlphaEvolve~\citep{alphaevolve2025}.
While traditional program synthesis and genetic programming rely on rigid domain-specific languages or parametric tuning that constrain algorithmic expressiveness, large language models (LLMs) possess rich knowledge of algorithmic structures and software engineering in expressive systems languages like C++.
Our primary objective is discovering domain-general procedural search mechanisms $m \in \Omega_{\text{mech}}$ that generalize across arbitrary games.
To achieve this, our framework factorizes search into an asynchronous multi-agent LLM meta-learning system: a centralized coordinator discovers generalist procedural mechanisms across games, while distributed workers evolve specialist domain heuristics that serve as an adaptive training curriculum and evaluation scaffold.
At each generation, LLM agents propose semantically coherent C++ code modifications across typed search interfaces, yielding standalone compiled C++ search routines for evaluation and deployment.
\begin{algorithm}[t]
\caption{\footnotesize Modular Factorized Co-Evolution}
\label{alg:coevolution}
\begin{algorithmic}[1]
\scriptsize
\Require Training games $\mathcal{G} = \{g_1, \dots, g_K\}$, reference $m_{\text{ref}}$ (canonical PUCT), opponents $\{\pi_{\text{opp}}(g)\}_{g \in \mathcal{G}}$
\Ensure Universal champion mechanism $m^\star$ and specialized heuristics $\{\kappa_g^\star\}_{g \in \mathcal{G}}$
\State \textbf{Initialize} archives $\mathcal{A}_{\text{mech}} \gets \{m_{\text{seed}}\}$, and $\mathcal{A}_{\text{know}}(g) \gets \{\kappa_{\text{seed}}^{(g)}\}$ for all $g \in \mathcal{G}$
\Procedure{KnowledgeWorker}{game $g \in \mathcal{G}$} \Comment{Parallel worker per game}
  \While{budget remaining}
    \State Sample metric $\psi \sim \{F_{\text{puct}}, F_{\text{prior}}\}$;\quad $\kappa \sim_\psi \mathcal{A}_{\text{know}}(g)$;\quad \Comment{Dual-objective parent sampling}
    \State $\kappa' \gets \textsc{LLMMutate}(\kappa, g)$
    \If{$\textsc{VerifyAntiCheating}(\kappa') = \text{valid}$} \Comment{Information-set constraint}
      \State Sample $m \sim \mathcal{A}_{\text{mech}}$;\quad Match under $(m, \kappa')$, $(m_{\text{ref}}, \kappa')$, and $\pi_{\kappa'}$
      \State $\textsc{InsertArchive}(\mathcal{A}_{\text{know}}(g), \kappa', (F_{\text{puct}}, F_{\text{prior}}))$;\quad \Comment{Ranked on dual frontier}
      \State $\textsc{UpdateRunningMean}(\mathcal{A}_{\text{mech}}, m, g, (\mathrm{Amp}, \mathrm{Imp}))$ \Comment{Async cross-game sample}
      \State \textsc{UpdateOpponentProgression}($g$) \Comment{Promote if $F_{\text{puct}}$ saturates epoch}
    \EndIf
  \EndWhile
\EndProcedure
\Procedure{MechanismCoordinator}{} \Comment{Concurrent cross-game coordinator}
  \While{budget remaining}
    \State Sample metric $\phi \sim \{\mathrm{Amp}, \mathrm{Imp}\}$;\quad $m \sim_\phi \mathcal{A}_{\text{mech}}$;\quad
    \State $m' \gets \textsc{LLMMutate}(m)$;\quad $\mathcal{A}_{\text{mech}} \gets \mathcal{A}_{\text{mech}} \cup \{m'\}$ \Comment{Deferred eval by workers}
  \EndWhile
\EndProcedure
\State \textbf{Output:} $m^\star = \arg\max_{m \in \mathcal{A}_{\text{mech}}} \mathrm{Amp}(m)$, \quad $\{\kappa_g^\star\}_{g \in \mathcal{G}}$
\end{algorithmic}
\end{algorithm}
\subsection{Evolutionary Program Synthesis with AlphaEvolve}
\label{sec:method:alphaevolve}
AlphaEvolve discovers algorithms by pairing large language models (LLMs) with automated sandbox evaluation.
Programs are stored in an evolutionary database organized into population islands~\citep{funsearch2023}.
In each cycle, the system picks a promising parent program based on past performance, prompts an LLM to propose code edits, compiles and runs the new candidate in a secure sandbox, and records its scores.
For multi-objective problems, AlphaEvolve tracks each metric independently and randomly alternates which metric selects the parent, preserving diverse solutions across the Pareto frontier.
\subsection{Factorized Co-Evolution}
\label{sec:method:coevolution}
While canonical AlphaEvolve evolves a single unified program for a single task, general game search demands a factorized co-evolutionary procedure.
A monolithic program representation fails because general game search requires both game-agnostic procedural mechanisms (selection descent, value backup, root scheduling) and domain-specific knowledge (policy priors, value estimation, history resamplers).
Evolving both jointly causes mutations to overfit to individual rule quirks, preventing general procedural mechanisms from transferring across game families.
Importantly, procedural mechanisms and domain heuristics are not on equal footing: our primary goal is discovering generalist procedural search mechanisms that generalize across games, while domain heuristics act as an evolving curriculum to stress-test candidate mechanisms across diverse game structures.
To achieve this, we factorize search into two concurrent processes communicating via shared archives (Algorithm~\ref{alg:coevolution}):
one \textbf{Search Mechanism Archive} $\mathcal{A}_{\text{mech}}$ containing procedural quintets $m \in \Omega_{\text{mech}}$ evaluated across all $K$ games, and
$K$ \textbf{Domain Knowledge Archives} $\{\mathcal{A}_{\text{know}}(g)\}_{g=1}^K$ evolving specialized knowledge triads $\kappa = (\pi_\kappa, v_\kappa, b_\kappa) \in \mathcal{K}(g)$ for game $g$.
As shown in Algorithm~\ref{alg:coevolution}, parallel \textsc{KnowledgeWorker} agents evolve specialized heuristics per game, while a centralized \textsc{MechanismCoordinator} optimizes game-agnostic search mechanisms across games.
\subsection{Modular Evaluation, Credit Decoupling, and Gating}
\label{sec:method:fitness}
A game-playing agent on game $g$ is a modular composition $\Comp = \langle m, \kappa \rangle$, combining a game-agnostic procedural mechanism module $m \in \Omega_{\text{mech}}$ with a game-specific knowledge module $\kappa = (\pi_\kappa, v_\kappa, b_\kappa) \in \mathcal{K}(g)$.
In general extensive-form games with player set $\mathcal{N}$, the candidate agent occupies player seat $p \in \mathcal{N}$ (averaged cyclically across seats) against curriculum opponent $\pi_{\text{opp}}(g)$ controlling all remaining $|\mathcal{N}| - 1$ seats.
Expected game return $F(m, \kappa; g) = \mathbb{E}[u_p]$ lies in bounded utility range $[U_{\min}(g), U_{\max}(g)]$; $F(\pi_\kappa; g)$ denotes raw policy execution.
\paragraph{Credit Decoupling and Headroom Normalization.}
In simulation search, agent performance compounds procedural planning $m$ and domain heuristics $\kappa$.
Evaluating both jointly causes severe credit confounding: a mediocre search algorithm can appear strong by free-riding on an accurate heuristic, while an innovative search mechanism can fail if paired with inaccurate value predictions.
We resolve credit assignment by anchoring evaluations to an invariant reference search engine $m_{\text{ref}}$ (canonical PUCT) and decoupling evaluation across the two concurrent processes:
(i)~\textbf{Domain Knowledge Workers}: To isolate domain heuristics from search mechanism fluctuations, workers evaluate candidate heuristics $\kappa = (\pi_\kappa, v_\kappa, b_\kappa)$ under fixed reference engine $m_{\text{ref}}$, alternating parent selection between search return $F(m_{\text{ref}}, \kappa; g)$ and standalone prior return $F(\pi_\kappa; g)$.
(ii)~\textbf{Mechanism Coordinator}: To discover game-agnostic search mechanisms, the coordinator must evaluate candidates across games with vastly different payoff scales and difficulties.
We put all games on the same scale by normalizing marginal performance by the remaining headroom to optimal return $U_{\max}(g)$, defining two complementary metrics clamped to $[-1, 1]$:
\emph{amplification} $\mathrm{Amp}(m, \kappa; g) = \mathrm{clamp}\big(\frac{F(m, \kappa; g) - F(m_{\text{ref}}, \kappa; g)}{U_{\max}(g) - F(m_{\text{ref}}, \kappa; g) + \epsilon}, -1, 1\big)$ evaluates the relative search strength of $m$ over PUCT; and
\emph{improvement} $\mathrm{Imp}(m, \kappa; g) = \mathrm{clamp}\big(\frac{F(m, \kappa; g) - F(\pi_\kappa; g)}{U_{\max}(g) - F(\pi_\kappa; g) + \epsilon}, -1, 1\big)$ evaluates how effectively $m$ acts as a policy improver over the standalone prior $\pi_\kappa$ (with potential near-zero denominators governed by the ceiling saturation upgrade gate in App.~\ref{app:robustness:ceiling}).
The coordinator aggregates each metric across the game suite using the interquartile mean (IQM) to resist outlier scales, and alternates parent selection between $\mathrm{Amp}$ and $\mathrm{Imp}$.
\paragraph{Robustness and Execution Safeguards.}
Co-evolving search algorithms across diverse games requires several practical safeguards to maintain system stability and correctness.
In imperfect-information games, synthesized history resamplers undergo automated anti-cheating checks before evaluation to ensure they do not peek at hidden opponent state.
Beyond information gating, candidate programs pass through a suite of execution checks, code complexity bounds, and evaluation quorums to ensure balanced cross-game assessment and prevent premature convergence, with full implementation details in App.~\ref{app:robustness}.
\subsection{Adaptive Opponent Progression and Non-Stationarity}
\label{sec:method:curriculum}
Evaluating search candidates against static opponents risks overfitting to idiosyncratic baseline flaws.
To drive continuous strategic hardening, evolutionary optimization is organized into a sequence of discrete self-play curriculum epochs $e \in \{0, 1, 2, \dots\}$.
Within epoch $e$, candidate mechanism and knowledge modules are evaluated against a fixed benchmark opponent policy $\pi_{\text{opp}}^{(e)}(g)$ (controlling opposing seats in multi-player games), initialized at $e=0$ with standard OpenSpiel baseline bots.
Throughout each epoch, candidates compete against $\pi_{\text{opp}}^{(e)}(g)$ until the population satisfies convergence criteria against the current baseline.
Upon convergence, the reigning champion mechanism and heuristic pair $(m_e^\star, \kappa_{g, e}^\star)$ is verified through an independent confirmation tournament, frozen, and promoted to serve as the fixed benchmark opponent $\pi_{\text{opp}}^{(e+1)}(g)$ for the subsequent epoch $e+1$.
New candidate generations are subsequently optimized against this hardened predecessor, monotonically compelling the search population to eliminate tactical vulnerabilities and discover increasingly robust counter-strategies.
While strengthening opponents introduces non-stationarity into score distributions, we preserve evolutionary stability through three mechanisms: restricting competitive ranking strictly within the active curriculum epoch, normalizing payoffs via relative headroom against the invariant reference search under identical opponents, and advancing games asynchronously across epochs to prevent systemic population shocks (App.~\ref{app:curriculum}).

\section{Experiments}
\label{sec:experiments}

We evaluate our modular co-evolutionary framework across a diverse suite of extensive-form games.
Our empirical evaluation addresses three core questions:
(1)~Which co-evolved modules are critical for game-theoretic optimality in imperfect-information games?
(2)~Does a single evolved search mechanism generalize across unseen held-out games, synthesized environments, and learned neural value models?
(3)~Can a game-agnostic evolved search mechanism rival specialized, hand-crafted game engines and unguided neural networks?

\subsection{Experimental Setup}
\label{sec:exp:setup}

We train and evaluate our framework across a diverse suite of 50 extensive-form games in OpenSpiel~\citep{openspiel2019} spanning diverse strategic families (classic board, connection, poker, imperfect information, and multi-agent coordination), exposing candidates to diverse branching factors, horizon depths, and information symmetries.
Evolutionary code mutations are generated using Gemini 3.5 Flash~\citep{geminiteam2026gemini35} within AlphaEvolve~\citep{alphaevolve2025}, with candidate C++ programs evaluated using 4 search threads on dedicated compute instances; we run every MCTS baseline with 4 threads.
All MCTS baselines are tuned on the training suite to ensure fair comparisons (see App.~\ref{app:calibration}, \ref{app:robustness}, and~\ref{app:baselines} for calibration, compute resources, and baseline tuning details).

Following evolution, deployable agents $\Comp^\star = (m^\star, \kappa_g^\star)$ are assembled from both factorized populations: for each game $g \in \mathcal{G}$, champion domain knowledge $\kappa_g^\star \in \mathcal{A}_{\text{know}}(g)$ is selected by peak return under reference PUCT in the mature curriculum epoch, while universal champion mechanism $m^\star = \arg\max_{m \in \mathcal{A}_{\text{mech}}} \mathrm{Amp}(m)$ maximizes aggregate cross-game amplification, combining broad procedural search strength with specialized domain evaluation.

\newcommand{\expl}{\ensuremath{e}}

\subsection{Game-Theoretic Optimality: Which Module Transfers?}
\label{sec:exploitability}

As an initial probe, we evaluate our evolved programs on three imperfect-information games (Kuhn poker, Leduc poker, and Liar's dice) using exploitability.
Exploitability measures the expected payoff deficit of a policy $\pi$ against an optimal adversary playing an exact best response, $\expl(\pi) = \frac{1}{|\mathcal{N}|} \sum_{i \in \mathcal{N}} [\max_{\pi_i'} u_i(\pi_i', \pi_{-i}) - u_i(\pi)]$.
An exploitability of zero corresponds to a Nash equilibrium; lower values reflect greater worst-case robustness against counter-play.
Because exploitability was never an optimization objective during evolution, it serves as an unoptimized stress test.
For imperfect-information games, domain knowledge factorizes as $\kappa = (\pi, v, b)$ (Section~\ref{sec:prelim:games}).
The unguided baseline knowledge $\kappa_{\text{base}} = (\pi_0, v_0, b_{\text{base}})$ pairs uniform action priors $\pi_0$ with random rollouts $v_0$ and OpenSpiel's default history resampler $b_{\text{base}}$.
We ablate each co-evolved module across independent evolution runs:
(1) the PUCT baseline ($\langle m_{\text{ref}}, \kappa_{\text{base}} \rangle$);
(2) the evolved procedural search mechanism alone ($\langle m^\star$,$\kappa_{\text{base}}\rangle$);
(3) baseline PUCT with evolved domain knowledge ($\langle m_{\text{ref}}, \kappa^\star \rangle$);
(4) the evolved mechanism with priors and value estimator alone ($m^\star + \pi^\star + v^\star$, with $b_{\text{base}}$); and
(5) the full co-evolved system ($\langle m^\star, \kappa^\star \rangle$, incorporating $b^\star$).

\begin{figure*}[t]
\centering
\begin{tikzpicture}
\begin{groupplot}[
  group style={group size=3 by 1, horizontal sep=0.85cm},
  width=0.32\textwidth, height=3.5cm,
  xmode=log, log basis x=10,
  xtick={20,50,100,500,1000}, xticklabels={20,50,100,500,1000},
  xlabel={Simulations $B$},
  tick label style={font=\scriptsize}, label style={font=\scriptsize},
  title style={font=\small, yshift=-2pt},
  every axis plot/.append style={thick, mark size=1.5pt},
  grid=major, grid style={dotted, gray!50},
]
\nextgroupplot[title={Kuhn poker}, ylabel={Exploitability $\expl$}, ymin=0.00, ymax=0.48]
\addplot+[mark=*, blue!80!black] coordinates {(20,0.2966)(50,0.2450)(100,0.2224)(500,0.1873)(1000,0.1698)};
\addplot+[mark=square*, red!80!black] coordinates {(20,0.1879)(50,0.1595)(100,0.1412)(500,0.1177)(1000,0.1237)};
\addplot+[mark=triangle*, brown!70!black] coordinates {(20,0.1869)(50,0.1774)(100,0.1652)(500,0.1378)(1000,0.1313)};
\addplot+[mark=otimes*, orange!85!black] coordinates {(20,0.1519)(50,0.1459)(100,0.1427)(500,0.1480)(1000,0.1508)};
\addplot+[mark=diamond*, black] coordinates {(20,0.1425)(50,0.1317)(100,0.1280)(500,0.1281)(1000,0.1294)};
\addplot[dashed, thick, black!55, domain=20:1000] {0.4583};
\addplot[dashdotted, thick, teal!80!black, domain=20:1000] {0.0082};

\nextgroupplot[title={Leduc poker}, ymin=0.00, ymax=2.50]
\addplot+[mark=*, blue!80!black] coordinates {(20,2.1116)(50,1.9799)(100,1.8700)(500,1.6381)(1000,1.5171)};
\addplot+[mark=square*, red!80!black] coordinates {(20,1.7759)(50,1.5908)(100,1.4051)(500,1.0711)(1000,0.9521)};
\addplot+[mark=triangle*, brown!70!black] coordinates {(20,1.4144)(50,1.3842)(100,1.3539)(500,1.2458)(1000,1.1781)};
\addplot+[mark=otimes*, orange!85!black] coordinates {(20,1.5970)(50,1.3799)(100,1.3180)(500,1.0468)(1000,0.9074)};
\addplot+[mark=diamond*, black] coordinates {(20,1.5906)(50,1.3667)(100,1.3513)(500,1.2486)(1000,1.2074)};
\addplot[dashed, thick, black!55, domain=20:1000] {2.3736};
\addplot[dashdotted, thick, teal!80!black, domain=20:1000] {0.0957};

\nextgroupplot[
  title={Liar's dice}, ymin=0.00, ymax=0.85,
  legend to name=expllegend_main,
  legend style={legend columns=4, font=\scriptsize, draw=none, /tikz/every even column/.append style={column sep=0.15cm}}
]
\addplot+[mark=*, blue!80!black] coordinates {(20,0.7285)(50,0.7165)(100,0.7054)(500,0.6372)(1000,0.5700)};
\addlegendentry{PUCT ($\kappa_{\text{base}}$)}
\addplot+[mark=square*, red!80!black] coordinates {(20,0.6854)(50,0.6469)(100,0.6446)(500,0.6041)(1000,0.5659)};
\addlegendentry{Mechanism ($m^\star, \kappa_{\text{base}}$)}
\addplot+[mark=triangle*, brown!70!black] coordinates {(20,0.6521)(50,0.6340)(100,0.6116)(500,0.5426)(1000,0.5216)};
\addlegendentry{PUCT ($\kappa^\star$)}
\addplot+[mark=otimes*, orange!85!black] coordinates {(20,0.6775)(50,0.6549)(100,0.6246)(500,0.5063)(1000,0.4748)};
\addlegendentry{Mechanism ($m^\star + \pi^\star + v^\star, b_{\text{base}}$)}
\addplot+[mark=diamond*, black] coordinates {(20,0.6911)(50,0.6638)(100,0.6512)(500,0.5876)(1000,0.5703)};
\addlegendentry{Full system ($\langle m^\star, \kappa^\star \rangle$)}
\addplot[dashed, thick, black!55, domain=20:1000] {0.7807};
\addlegendentry{Uniform random}
\addplot[dashdotted, thick, teal!80!black, domain=20:1000] {0.0224};
\addlegendentry{CFR (100 iters)}
\end{groupplot}
\end{tikzpicture}\\[-2pt]
\ref{expllegend_main}\vspace{-2pt}
\caption{\textbf{Exploitability across simulation budgets} (mean over independent seeds; lower is better). Dashed lines denote uniform random play; dash-dotted lines denote CFR (100 iterations). Numerical values and significance tests are in Appendix Table~\ref{tab:expl-levels}.}
\label{fig:expl-budget}
\end{figure*}
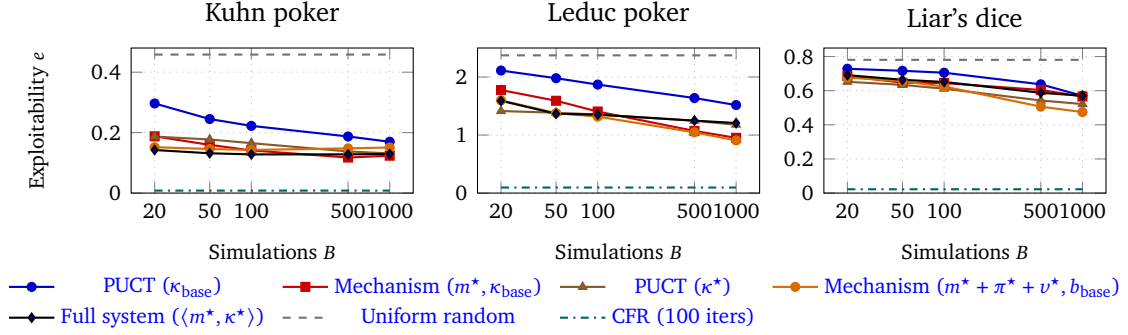

As shown in Figure~\ref{fig:expl-budget}, the procedural search mechanism ($m^\star$) alone consistently reduces exploitability across all three games, achieving a $10\times\text{--}20\times$ simulation efficiency advantage over PUCT (e.g., $m^\star$ at $B=100$ on Leduc and $B=50$ on Kuhn surpasses PUCT at $B=1000$).
Dissecting the knowledge triad reveals that while co-evolved priors and value estimators ($m^\star + \pi^\star + v^\star$) achieve the lowest exploitability on Leduc Poker ($0.9074$) and Liar's Dice ($0.4748$), activating the evolved belief resampler ($b^\star$) degrades worst-case performance ($1.2074$ and $0.5703$, respectively).
While evolved value estimators ($v^\star$) reduce search variance via analytical state evaluations, evolved belief resamplers ($b^\star$) attempt action-conditioned opponent modeling; against a worst-case adversary, non-uniform belief priors are vulnerable to strategic deception, inducing systematic state-estimation bias during MCTS determinization (App.~\ref{app:expl}).

\subsection{Cross-Domain Generalization under Fixed Search Budgets}
\label{sec:exp:generalization}

\begin{table*}[t]
\centering
\small
\renewcommand{\arraystretch}{0.85}
\caption{\textbf{Continuous SCO Ratings and Pairwise Ballot Win Rates across Four Benchmark Tiers under a 50\,ms Budget} (mean $\pm$ SE across Seeds 1..5, and representative case study \mstarBIFV). For each tier, we report both the continuous Soft Condorcet Optimization rating (\textbf{SCO}, scale $[0, 1000]$) and the pairwise ballot win rate (\textbf{Win \%}) denoting the percentage of games where \mevo\ outranks that baseline under AlphaRank dynamics.}
\label{tab:stratified_sco_tiers}
\resizebox{\textwidth}{!}{%
\begin{tabular}{lcccccccc}
\toprule
\textbf{Algorithmic Paradigm / Baseline} & \multicolumn{2}{c}{\textbf{Training Games (50)}} & \multicolumn{2}{c}{\textbf{Held-Out Games (19)}} & \multicolumn{2}{c}{\textbf{Synthesized Games (360)}} & \multicolumn{2}{c}{\textbf{Neural PPO Games (17)}} \\
\cmidrule(lr){2-3} \cmidrule(lr){4-5} \cmidrule(lr){6-7} \cmidrule(lr){8-9}
& \textbf{SCO Rating} & \textbf{$\mevo$ Win \%} & \textbf{SCO Rating} & \textbf{$\mevo$ Win \%} & \textbf{SCO Rating} & \textbf{$\mevo$ Win \%} & \textbf{SCO Rating} & \textbf{$\mevo$ Win \%} \\
\midrule
\multicolumn{9}{l}{\textit{Discovered Search Mechanism (Ours)}} \\
\textbf{\mevo\ (Multi-run aggregate)} & $\mathbf{509.34 \pm 0.39}$ & --- & $\mathbf{509.36 \pm 0.48}$ & --- & $\mathbf{508.99 \pm 0.20}$ & --- & $\mathbf{508.91 \pm 0.67}$ & --- \\
\textbf{\mstarBIFV\ (\bifv)} & $\mathbf{509.53}$ & --- & $\mathbf{509.93}$ & --- & $\mathbf{508.58}$ & --- & $\mathbf{510.31}$ & --- \\
\midrule
\multicolumn{9}{l}{\textit{Bandit Tree Search}} \\
Vanilla UCT~\citep{kocsis2006bandit} & $501.22 \pm 0.82$ & $61.5\% \pm 4.2\%$ & $503.84 \pm 2.08$ & $59.7\% \pm 5.6\%$ & $504.19 \pm 0.41$ & $59.1\% \pm 0.7\%$ & $500.77 \pm 2.10$ & $54.2\% \pm 4.9\%$ \\
PUCT (reference baseline)~\citep{Silver18AlphaZero} & $503.92 \pm 1.43$ & $65.2\% \pm 1.7\%$ & $503.23 \pm 1.35$ & $62.2\% \pm 3.7\%$ & $499.38 \pm 0.70$ & $70.4\% \pm 1.1\%$ & $508.05 \pm 1.12$ & $52.9\% \pm 8.0\%$ \\
UCB-Tuned~\citep{auer2002finite} & $502.88 \pm 0.62$ & $59.6\% \pm 3.0\%$ & $502.48 \pm 1.35$ & $65.0\% \pm 3.8\%$ & $504.75 \pm 0.25$ & $60.0\% \pm 1.5\%$ & $500.79 \pm 1.43$ & $55.8\% \pm 4.0\%$ \\
\midrule
\multicolumn{9}{l}{\textit{Entropy-Regularized \& Soft Planning}} \\
Gumbel AlphaZero~\citep{danihelka2022policy} & $493.90 \pm 2.34$ & $77.7\% \pm 3.4\%$ & $489.74 \pm 0.42$ & $92.7\% \pm 2.3\%$ & $496.84 \pm 0.43$ & $68.1\% \pm 1.4\%$ & $497.81 \pm 2.06$ & $70.4\% \pm 4.1\%$ \\
Smooth UCT~\citep{heinrich2015smooth} & $501.41 \pm 0.64$ & $68.9\% \pm 2.3\%$ & $500.05 \pm 1.16$ & $66.9\% \pm 4.6\%$ & $504.27 \pm 0.36$ & $59.9\% \pm 1.5\%$ & $497.07 \pm 1.21$ & $67.8\% \pm 4.8\%$ \\
RPO~\citep{grill2020monte} & $506.79 \pm 0.63$ & $61.5\% \pm 3.6\%$ & $506.58 \pm 0.33$ & $56.3\% \pm 2.1\%$ & $494.44 \pm 0.46$ & $74.5\% \pm 1.1\%$ & $503.45 \pm 1.21$ & $69.2\% \pm 8.0\%$ \\
MENTS~\citep{xiao2019maximum} & $493.90 \pm 0.79$ & $77.1\% \pm 1.6\%$ & $491.99 \pm 0.41$ & $80.5\% \pm 1.4\%$ & $490.58 \pm 0.08$ & $81.1\% \pm 1.6\%$ & $496.28 \pm 2.17$ & $79.7\% \pm 3.6\%$ \\
Boltzmann MCTS~\citep{chaslot2008monte} & $497.25 \pm 0.74$ & $74.5\% \pm 3.6\%$ & $497.53 \pm 1.93$ & $75.9\% \pm 3.2\%$ & $500.36 \pm 0.25$ & $64.0\% \pm 1.3\%$ & $499.51 \pm 2.37$ & $60.4\% \pm 3.3\%$ \\
\midrule
\multicolumn{9}{l}{\textit{Advanced Value Backups}} \\
MCTS Solver~\citep{winands2008monte} & $502.07 \pm 1.73$ & $64.2\% \pm 4.1\%$ & $501.03 \pm 2.27$ & $69.6\% \pm 3.5\%$ & $498.96 \pm 0.67$ & $67.6\% \pm 1.1\%$ & $495.37 \pm 1.35$ & $73.7\% \pm 3.2\%$ \\
Score-Bounded~\citep{cazenave2010score} & $501.67 \pm 0.55$ & $58.7\% \pm 2.8\%$ & $503.56 \pm 0.27$ & $52.4\% \pm 2.5\%$ & $502.03 \pm 0.22$ & $63.9\% \pm 1.4\%$ & $498.84 \pm 1.32$ & $58.9\% \pm 6.5\%$ \\
Power-Mean~\citep{dam2019generalized} & $495.09 \pm 0.70$ & $76.3\% \pm 3.2\%$ & $495.25 \pm 0.62$ & $78.2\% \pm 1.9\%$ & $497.21 \pm 0.29$ & $67.0\% \pm 1.0\%$ & $501.41 \pm 1.53$ & $60.6\% \pm 1.3\%$ \\
Max-Backup~\citep{coulom2006efficient} & $492.37 \pm 0.48$ & $83.3\% \pm 1.6\%$ & $495.48 \pm 1.23$ & $73.7\% \pm 2.7\%$ & $493.94 \pm 0.16$ & $71.2\% \pm 1.2\%$ & $492.21 \pm 1.77$ & $73.8\% \pm 4.7\%$ \\
\midrule
\multicolumn{9}{l}{\textit{Bayesian \& Online Adaptive Search}} \\
Thompson Sampling~\citep{bai2013bayesian} & $497.04 \pm 0.59$ & $68.1\% \pm 2.5\%$ & $499.44 \pm 1.26$ & $71.3\% \pm 3.5\%$ & $499.91 \pm 0.24$ & $63.9\% \pm 1.2\%$ & $497.11 \pm 2.25$ & $58.9\% \pm 4.7\%$ \\
SA-MCTS-2P~\citep{sironi2018self} & $498.58 \pm 1.12$ & $66.0\% \pm 5.1\%$ & $500.38 \pm 1.22$ & $72.1\% \pm 3.7\%$ & $497.19 \pm 0.20$ & $67.6\% \pm 1.2\%$ & $500.86 \pm 1.42$ & $58.8\% \pm 5.4\%$ \\
SA-MCTS-4P~\citep{sironi2018self} & $502.57 \pm 1.34$ & $69.6\% \pm 5.2\%$ & $499.05 \pm 1.96$ & $71.9\% \pm 3.5\%$ & $506.94 \pm 0.15$ & $56.6\% \pm 2.3\%$ & $501.55 \pm 1.23$ & $68.2\% \pm 3.4\%$ \\
\bottomrule
\end{tabular}}
\end{table*}

To evaluate whether discovered search mechanisms capture invariant planning principles that transfer, we benchmark all candidates under both \textbf{fixed wall-clock time} (50\,ms per decision) and \textbf{fixed simulation budgets} (1,000 visits per move; App.~\ref{app:isosim_evaluation}).
Our core findings transfer robustly across both compute regimes; here we focus our main presentation on fixed wall-clock time budgets, which naturally penalize per-node procedural overhead.
Rigorous evaluation across games with vastly different structures (2-player zero-sum, general-sum, and imperfect information) poses two fundamental challenges: intra-game cyclic dominance and cross-game scale aggregation. Standard Elo ratings or minimax Nash equilibria fail in multi-agent or general-sum domains where strategy dominance cycles invalidate transitive ratings, while heterogeneous scoring scales preclude naive cross-game score averaging.
To resolve this, we employ a principled two-stage game-theoretic evaluation framework:
first, within each individual game, we evaluate the empirical game payoff tensor across 16 competing algorithms under seat-balanced matches and compute \textbf{AlphaRank}~\citep{omidshafiei2019alpha} stationary distributions over algorithmic response graphs, capturing cyclic and non-transitive dominance without artificial equilibrium assumptions;
second, to aggregate performance across heterogeneous games without distortion from arbitrary score scales or payoff ranges, each game's AlphaRank stationary distribution is treated as an independent ranked ballot. We synthesize a global consensus via \textbf{Soft Condorcet Optimization (SCO)}~\citep{lanctot2024soft}, which fits scale-invariant continuous ratings $r \in [0, 1000]$ by minimizing pairwise ranking inversions (App.~\ref{app:eval_methodology}).
Because SCO's logistic objective compresses continuous ratings into a concentrated band around the median ($r \in [490, 512]$), we also report the uncompressed \textbf{pairwise ballot preference fraction}---the proportion of games where AlphaRank response dynamics rank candidate $i$ strictly above candidate $j$ ($\in[0\%, 100\%]$).
We report both multi-run aggregate SCO ratings ($\text{Mean} \pm \text{SE}$) and the representative case study, \textbf{\mstarBIFV} (\bifv), benchmarked against 15 established MCTS baselines across four benchmark tiers spanning over 400 games in total (Table~\ref{tab:stratified_sco_tiers}).

\paragraph{1. OpenSpiel Training Games (50 Extensive-Form Games).}
The training suite spans 50 extensive-form games covering classic board, connection, poker, imperfect information, and multi-agent coordination.
Aggregated across evolutionary runs, $\mevo$ achieves an SCO rating of $\mathbf{509.34 \pm 0.39}$, leading all 15 baselines ahead of reference PUCT ($503.92 \pm 1.43$) and establishing clear pairwise majorities over reference PUCT ($65.2\% \pm 1.7\%$), Max-Backup ($83.3\%$), and Gumbel AlphaZero ($77.7\%$).
In its own evaluation tournament, \mstarBIFV\ secures \textbf{Rank~\#1 in AlphaRank+SCO} ($\mathbf{509.53}$), outranking all 15 baselines.

\paragraph{2. Held-Out OpenSpiel Games (19 Extensive-Form Games).}
To verify out-of-distribution transfer to human-designed games, we evaluate on 19 held-out extensive-form games (including Oware, Amazons, Skat, Hanabi, Twixt, and Ultimate Tic-Tac-Toe) completely unseen during evolution.
To equip agents with strong domain heuristics without rerunning the costly co-evolution loop, candidate heuristics are synthesized directly from game rules with the champion selected via round-robin tournament (App.~\ref{app:inner_synthesis_protocol}); all 16 competing algorithms evaluate on this identical heuristic.
$\mevo$'s multi-run aggregate achieves an SCO rating of $\mathbf{509.36 \pm 0.48}$, outperforming PUCT ($503.23 \pm 1.35$), Vanilla UCT ($503.84 \pm 2.08$), and Smooth UCT ($500.05 \pm 1.16$), securing pairwise majorities over every competing baseline without exception (outranking reference PUCT in $62.2\% \pm 3.7\%$, Max-Backup in $73.7\%$, and Gumbel AlphaZero in $92.7\%$).
In its seed tournament, \mstarBIFV\ captures \textbf{Rank~\#1 in AlphaRank+SCO} ($\mathbf{509.93}$).

\paragraph{3. Synthesized Games (360 Procedurally Synthesized Games).}
Evaluating general planning algorithms on human-designed games risks implicit benchmark contamination.
To assess procedural transfer, we evaluate across 360 procedurally synthesized games generated directly from natural language rules into executable C++ simulation models~\citep{lehrach2025codeworldmodelsgeneral}, strictly curated and calibrated (App.~\ref{app:cwm_calibration}).
Across independent evolutionary runs, $\mevo$ attains an aggregate SCO rating of $\mathbf{508.99 \pm 0.20}$ ahead of SA-MCTS-4P ($506.94$), UCB-Tuned ($504.75$), and reference PUCT ($499.38 \pm 0.70$), winning commanding majorities against reference PUCT ($70.4\% \pm 1.1\%$), RPO ($74.5\%$), and MENTS ($81.1\%$).
Within individual seed tournaments, evolved mechanisms capture \textbf{Rank~\#1 in AlphaRank+SCO in 100\% of runs} (\mstarBIFV\ scoring $\mathbf{508.58}$), confirming that discovered search primitives generalize robustly beyond human design biases.

\paragraph{4. Deep Neural Network Representations: Integration with PPO (17 Games with Frozen Neural Representations).}
While tree search traditionally relies on rollout heuristics or hand-crafted evaluators, modern deep planning systems (e.g., AlphaZero) integrate learned neural policies and value functions.
To evaluate whether discovered search mechanisms scale to neural representations without retraining, we deploy multi-agent policy and value networks independently trained via multi-player PPO across 17 benchmark games (App.~\ref{app:neural_ppo_setup}).
Using these frozen neural models as domain prior and value estimators, baseline PUCT with neural priors is already strong ($508.05 \pm 1.12$), with $\mevo$ achieving an aggregate SCO rating of $\mathbf{508.91 \pm 0.67}$.
In direct pairwise ballots, $\mevo$ achieves statistical parity with reference PUCT ($52.9\% \pm 8.0\%$ win rate, with 95\% CI spanning 50\%), while significantly outranking Gumbel AlphaZero in $70.4\%$ and SA-MCTS-4P in $68.2\%$.
In its individual seed tournament, \mstarBIFV\ achieves an SCO rating of $\mathbf{510.31}$ against reference PUCT ($506.90$), while across the remaining four lineages PUCT and $\mevo$ alternate between ranks 1 and 2 (Table~\ref{tab:per_run_sco_breakdown}).
This experiment demonstrates that the evolved exploration and backup operators successfully maintain parity with tuned neural PUCT on frozen deep representations without requiring representational retraining, mitigating tree over-polarization when paired with sharp neural priors.
\paragraph{Case Study: Mechanistic Anatomy of \bifv\ (\mstarBIFV).}
Across independent evolutionary runs, our framework synthesized novel, non-standard functional compositions of foundational search principles (App.~\ref{app:five_lineages}): Run~1 scaled logarithmic exploration and FPU directly by native utility span $\Delta U$ with instant terminal short-circuiting, Run~2 synthesized rational exploration schedules ($1.25 + 1.75 \frac{N}{N+\max(100, 0.5 \cdot S)}$) eliminating logarithms, Run~4 evolved Robbins-Monro non-stationary polynomial backups ($\gamma_t = (n+1)^{-0.82}$), and Run~5 derived parent-decoupled visit-weighted sibling expectation FPU ($\mathbb{E}_v[Q]$).
To understand how these components operate mechanistically, we selected the most consistent discovered lineage, \textbf{\bifv} (\mstarBIFV, selected for maintaining top rank across both wall-clock and simulation-budget regimes; App.~\ref{app:granular_runs}), as an in-depth case study. \bifv\ composes four factorized operators:
(i)~\textbf{Branching (\textbf{B})}: scales $c_{\mathrm{puct}}$ via $(1 + 0.05 \ln |A|)$;
(ii)~\textbf{Information (\textbf{I})}: inflates exploration by $1.25^{\mathbb{I}_{\mathrm{imp}}}$ under partial observability;
(iii)~\textbf{Maturity FPU (\textbf{F})}: anneals unvisited penalties relative to the neutral baseline $Q_0 = -u_{\min}/\Delta U$ ($0.5$ in zero-sum games): $Q_{\mathrm{FPU}}(a) = Q_0 - \sqrt{1 - \pi(a \mid s)} \cdot (0.1 + 0.4 \frac{N}{100 + N})$; and
(iv)~\textbf{Value Selection (\textbf{V})}: breaks visit ties via utility: $\arg\max_a [N(a) + 0.1 \cdot \widetilde{Q}(a)]$, paired with root progress modulation $[1 + 0.20(1 - t/T)]$ on $c_{\mathrm{puct}}$ and adaptive root Dirichlet noise ($\alpha = \mathrm{clip}(10/|A|, 0.05, 1.0)$).
Leave-one-out ablations (App.~\ref{app:factorial_ablation}) show that no single operator is uniformly beneficial: every removal lowers performance on the neural-PPO tier, while effects on the symbolic tiers are mixed, suggesting the operators act jointly rather than independently.

\subsection{Benchmarking Against Specialized Domain Engines}
\label{sec:exp:engines}
To delineate the boundaries of domain-agnostic search, we benchmark against specialized external engines across ten classic board and imperfect-information card games (App.~\ref{app:engine_benchmarks}). Three lessons emerge.
First, procedural search acts as a policy-improvement operator over existing representations rather than a substitute for domain knowledge: at mid and ceiling difficulty, dedicated engines win nearly every game, and even against entry-level configurations co-evolved search obtains meaningful scores only in Othello (32--41\%), with limited success in Go~9$\times$9 ($\le$10\%) and Chess ($\le$7\%).
Second, the evaluator sets the ceiling of what search can reach: $m^\star$ substantially amplifies both co-evolved heuristics and compact PPO policies in most board games (the neural variant stalls in repetition draws in Chess), yet heuristic-guided search defeats neural-guided search in nearly all head-to-head games and is the only configuration that scores against the engines, reflecting that our compact PPO networks are weaker priors than the co-evolved evaluators even without search.
Third, in imperfect-information card games, determinized search yields mixed results: in Hearts it provides no systematic advantage over the domain prior heuristics, while in Contract Bridge it improves duplicate margins over unsearched heuristic bidding but, in lineages whose bidding priors over-escalate auctions, incurs larger point deficits against WBridge5 than neural search.
Together, these findings indicate that procedural search generalizes as a policy improver of whatever evaluation knowledge is available, but cannot compensate for missing evaluation knowledge.
\section{Related Work}
\label{sec:related_work}

\paragraph{Automated Algorithm and Program Discovery.}
Evolutionary search with LLMs has emerged as an effective paradigm for discovering algorithms as executable programs~\citep{funsearch2023, alphaevolve2025, eoh2024, ma2024eureka, liang2023code}, advancing beyond symbolic program synthesis~\citep{automlzero2020, coreyes2021evolving, lion2023, oh2020discovering} and heuristic tree search~\citep{mctsahd2025}. While prior systems evolve specialized programs for individual tasks, our framework discovers general procedural search mechanisms that transfer zero-shot across hundreds of games.

\paragraph{MCTS Operators: Analytical Design vs.\ Code Evolution.}
Prior advances in Monte Carlo Tree Search (MCTS) focus on deriving isolated analytical operators—such as $\alpha$-divergence backups~\citep{dam2019generalized, dam2024unified}, entropy-regularized planning (MENTS~\citep{xiao2019maximum}, RPO~\citep{grill2020monte}, Smooth UCT~\citep{heinrich2015smooth}), Gumbel improvements~\citep{danihelka2022policy}, and action pruning~\citep{winands2008monte, cazenave2010score}—or tuning scalar parameters per game (SA-MCTS~\citep{sironi2018self}) and learning heuristics over fixed PUCT~\citep{guo2016reward, soemers2019biasing}. We instead evolve the underlying executable code of selection and backup mechanisms, enforcing cross-domain generality without per-game tuning.

\paragraph{General Game Playing and Consensus Evaluation.}
General Game Playing benchmarks assess whether agents master arbitrary environments from rules alone~\citep{genesereth2005general, openspiel2019, lehrach2025codeworldmodelsgeneral}. While systems like AlphaZero~\citep{Silver18AlphaZero} and MuZero~\citep{muzero2020} retrain neural weights under fixed PUCT, and Student of Games~\citep{playerofgames2023} relies on continuous subgame resolving and Growing-Tree CFR for imperfect information, our work discovers procedural search mechanisms that transfer zero-shot across games and neural representations. To resolve cyclic dominance in multi-player evaluation, we couple AlphaRank~\citep{omidshafiei2019alpha} evolutionary dynamics with Soft Condorcet Optimization (SCO)~\citep{lanctot2024soft} for principled rank aggregation across diverse games.
\section{Conclusion and Limitations}
\label{sec:conclusion}

We introduced a modular multi-agent LLM meta-learning system that discovers general game-playing algorithms by factorizing procedural search logic ($m^\star$) from domain heuristics ($\kappa^\star$).
Across five lineages and $>400$ benchmark environments, discovered programs outperform established MCTS baselines, curb exploitability in imperfect-information games, secure AlphaRank and Soft Condorcet consensus, and transfer zero-shot across held-out games, procedurally synthesized models, and frozen neural representations.
\textbf{Limitations:}
(1)~Procedural search acts as a policy improvement operator over available base policies and evaluators, and cannot substitute for exhaustive endgame tablebases or massive pre-trained evaluation networks in specialized games;
(2)~evolving domain heuristics ($\kappa^\star$) under reference PUCT ($m_{\text{ref}}$) ensures stationary fitness evaluation and prevents cyclic pursuit, but biases priors toward PUCT compatibility;
(3)~offline co-evolution requires substantial compute, leaving online test-time meta-adaptation an open challenge.

\section*{Acknowledgements}
We thank Ian Gemp for his valuable discussions and comments on this work.

\bibliography{references}

\newpage
\appendix
\section{Modular Simulation Search Template: Typed Interfaces and C++ Controller Substrate}
\label{app:template}
This appendix provides the formal architectural specification, typed mathematical interfaces, production C++ class definitions, and execution lifecycle algorithm for the unified modular simulation search engine.
\subsection{Template Factorization and Typed Interfaces}
\label{app:template:interfaces}
Traditional game-playing architectures interleave domain knowledge (evaluation heuristics, rollouts, action orderings) directly with procedural tree traversal logic (selection formulas, backup rules, thread locks).
In contrast, our template enforces a strict computational factorization into two orthogonal tiers:
\begin{enumerate}[leftmargin=1.5em, itemsep=2pt, topsep=2pt]
  \item \textbf{Game-Agnostic Mechanism Quintet ($\Omega_{\text{mech}}$)}: procedural search operators that execute identically across arbitrary extensive-form games. These comprise:
  \begin{itemize}[leftmargin=1.2em, itemsep=1pt, topsep=1pt]
    \item \texttt{TreePolicy}: selects edge traversals during the in-tree selection descent phase, balancing exploration against exploitation.
    \item \texttt{ValueBackup}: propagates terminal utilities or leaf evaluation vectors up the trajectory to update edge and node statistics.
    \item \texttt{RootActionSelector}: allocates the simulation budget across candidate root branches, enabling non-uniform or sequential-halving schedules.
    \item \texttt{FinalActionSelector}: extracts the final deployed action $a^\ast$ and target posterior policy distribution $\hat{\pi}$ from the accumulated root statistics.
    \item \texttt{VirtualLoss}: manages temporary thread penalties to prevent redundant traversals and coordinate lock-free parallel search workers.
  \end{itemize}
  \item \textbf{Game-Specific Knowledge Triad ($\mathcal{K}(g)$)}: domain heuristics tailored to the specific state representations, rules, and information partitions of game $g$. These comprise:
  \begin{itemize}[leftmargin=1.2em, itemsep=1pt, topsep=1pt]
    \item \texttt{PriorPolicy}: produces an informative zero-shot prior probability distribution $\pi(a \mid s)$ over legal actions to bias early search trajectories.
    \item \texttt{ValueEstimator}: acts as a value function estimating expected return-to-go vectors $v(s) \in \mathbb{R}^{|\mathcal{N}|}$ from information states, replacing expensive stochastic rollouts.
    \item \texttt{ResampleHistory}: under imperfect information, determinizes unobserved game history $h \sim b(\cdot \mid s)$ conditioned on public observation $s$.
  \end{itemize}
\end{enumerate}
Table~\ref{tab:interfaces_full} summarizes the eight typed interfaces, presenting both their mathematical input-output signatures and their concrete C++ virtual method signatures in the evaluation engine.
\begin{table}[t]
\caption{The eight typed interfaces forming the modular simulation search template. Together, they constitute eight swappable computational degrees of freedom that span the design space of simulation search. Concrete C++ virtual method signatures and class declarations are provided in Section~\ref{app:template:cpp}.}
\label{tab:interfaces_full}
\centering
\small
\setlength{\tabcolsep}{4pt}
\renewcommand{\arraystretch}{1.08}
\begin{tabular}{llp{5.8cm}}
\toprule
\textbf{Module Interface} & \textbf{Type Signature} & \textbf{Operational Role} \\
\midrule
\multicolumn{3}{l}{\textit{Procedural Mechanism Modules (Game-Agnostic Quintet, $m \in \Omega_{\mathrm{mech}}$)}} \\
\texttt{TreePolicy} & $\nu \mapsto a \in A(\nu)$ & Balances exploitation vs.\ exploration during tree descent \\
\texttt{ValueBackup} & $(\nu, \vec{v}, d, p) \mapsto \text{update}$ & Propagates leaf payoff vectors along traversed trajectory \\
\texttt{RootActionSelector} & $(\nu_{\text{root}}, t, B) \mapsto a$ & Schedules root simulation budget across candidate actions \\
\texttt{FinalActionSelector} & $\nu_{\text{root}} \mapsto (a^\ast, \hat{\pi})$ & Selects final action and posterior policy distribution \\
\texttt{VirtualLoss} & $(\nu, d) \mapsto (\Delta Q, \Delta N)$ & Coordinates asynchronous parallel search workers \\
\midrule
\multicolumn{3}{l}{\textit{Domain Knowledge Modules (Game-Specific Triad, $\kappa \in \mathcal{K}(g)$)}} \\
\texttt{PriorPolicy} & $s \mapsto \pi(\cdot \mid s) \in \Delta(A)$ & Provides zero-shot strategic action probability distribution \\
\texttt{ValueEstimator} & $s \mapsto v(s) \in \mathbb{R}^{|\mathcal{N}|}$ & Estimates expected return-to-go directly without rollouts \\
\texttt{ResampleHistory} & $s \mapsto h \sim b(\cdot \mid s)$ & Samples hidden world history conditioned on observation history \\
\bottomrule
\end{tabular}
\end{table}
\subsection{C++ Controller Substrate}
\label{app:template:cpp}
In our execution substrate, each interface is defined as a C++ abstract base class. All deployable agents are zero-overhead C++ bots that invoke these virtual methods. The core class definitions for the modular search substrate are given below.
\begin{lstlisting}[style=substrate, caption={C++ abstract base classes for the game-agnostic mechanism quintet.}, label={lst:cpp_mechanism}]
// Game-Agnostic Mechanism Quintet Abstract Base Classes

class TreePolicy {
 public:
  TreePolicy(const open_spiel::Game& game, int num_simulations)
      : game_(game), num_simulations_(num_simulations) {}
  virtual ~TreePolicy() = default;
  virtual void OnStepBegin(const MCTSNode& root,
                           const std::vector<Action>& legal_actions,
                           int num_simulations, std::mt19937* rng) {}
  virtual Action SelectChild(const MCTSNode& parent,
                             const std::vector<Action>& legal_actions,
                             std::mt19937* rng) = 0;

 protected:
  const open_spiel::Game& game_;
  int num_simulations_;
};

class RootActionSelector {
 public:
  RootActionSelector(const open_spiel::Game& game, int num_simulations)
      : game_(game), num_simulations_(num_simulations) {}
  virtual ~RootActionSelector() = default;
  virtual void OnStepBegin(const MCTSNode& root,
                           const std::vector<Action>& legal_actions,
                           int num_simulations, std::mt19937* rng) {}
  virtual Action SelectRootAction(const MCTSNode& root,
                                  const std::vector<Action>& legal_actions,
                                  int simulation_index, int num_simulations,
                                  std::mt19937* rng) = 0;

 protected:
  const open_spiel::Game& game_;
  int num_simulations_;
};

class FinalActionSelector {
 public:
  FinalActionSelector(const open_spiel::Game& game, int num_simulations)
      : game_(game), num_simulations_(num_simulations) {}
  virtual ~FinalActionSelector() = default;
  virtual void OnStepBegin(const MCTSNode& root,
                           const std::vector<Action>& legal_actions,
                           int num_simulations, std::mt19937* rng) {}
  virtual std::pair<int, std::vector<double>> SelectActionAndPolicy(
      const MCTSNode& root, const std::vector<Action>& actions,
      std::mt19937* rng) = 0;

 protected:
  const open_spiel::Game& game_;
  int num_simulations_;
};

class ValueBackup {
 public:
  ValueBackup(const open_spiel::Game& game, int num_simulations)
      : game_(game), num_simulations_(num_simulations) {}
  virtual ~ValueBackup() = default;
  virtual void OnStepBegin(const MCTSNode& root,
                           const std::vector<Action>& legal_actions,
                           int num_simulations, std::mt19937* rng) {}
  virtual void Backup(MCTSNode* node, const std::vector<double>& values,
                      int depth, int acting_player) = 0;

 protected:
  const open_spiel::Game& game_;
  int num_simulations_;
};

struct VirtualLossDeltas {
  double value_delta;
  int visit_delta;
};

class VirtualLoss {
 public:
  VirtualLoss(const open_spiel::Game& game, int num_simulations)
      : game_(game), num_simulations_(num_simulations) {}
  virtual ~VirtualLoss() = default;
  virtual void OnStepBegin(const MCTSNode& root,
                           const std::vector<Action>& legal_actions,
                           int num_simulations, std::mt19937* rng) {}
  virtual VirtualLossDeltas Compute(const MCTSNode& node, int depth) const = 0;

 protected:
  const open_spiel::Game& game_;
  int num_simulations_;
};
\end{lstlisting}
\begin{lstlisting}[style=substrate, caption={C++ abstract base classes for the game-specific knowledge triad.}, label={lst:cpp_knowledge}]
// Game-Specific Knowledge Triad Abstract Base Classes

class PriorPolicy {
 public:
  explicit PriorPolicy(const open_spiel::Game& game) : game_(game) {}
  virtual ~PriorPolicy() = default;
  virtual void OnStepBegin(const MCTSNode& root,
                           const std::vector<Action>& legal_actions,
                           int num_simulations, std::mt19937* rng) {}
  virtual absl::flat_hash_map<Action, double> GetPriors(
      const open_spiel::State& state) = 0;

 protected:
  const open_spiel::Game& game_;
};

class ValueEstimator {
 public:
  explicit ValueEstimator(const open_spiel::Game& game) : game_(game) {}
  virtual ~ValueEstimator() = default;
  virtual void OnStepBegin(const MCTSNode& root,
                           const std::vector<Action>& legal_actions,
                           int num_simulations, std::mt19937* rng) {}
  virtual std::vector<double> Evaluate(const open_spiel::State& state) = 0;

 protected:
  const open_spiel::Game& game_;
};

class ResampleState {
 public:
  explicit ResampleState(const open_spiel::Game& game) : game_(game) {}
  virtual ~ResampleState() = default;
  virtual void OnStepBegin(const MCTSNode& root,
                           const std::vector<Action>& legal_actions,
                           int num_simulations, std::mt19937* rng) {}
  virtual std::unique_ptr<open_spiel::State> Resample(
      const open_spiel::State& state, std::mt19937* rng) = 0;

 protected:
  const open_spiel::Game& game_;
};

using ResampleHistory = ResampleState; // Mathematical alias matching paper notation
\end{lstlisting}
\begin{lstlisting}[style=substrate, caption={Thread-safe node representation and component composition container.}, label={lst:cpp_componentset}]
// Thread-Safe Node Representation, Component Container, and Bot Declaration

struct MCTSNode {
  double prior = 1.0;
  int player = -1;
  std::atomic<int> visits{0};
  std::atomic<int> virtual_losses{0};
  std::atomic<double> total_value{0.0};

  std::atomic<bool> expanded{false};
  absl::Mutex expand_mtx;

  absl::flat_hash_map<Action, std::unique_ptr<MCTSNode>> children;

  // Linkage and Exactness (for advanced baselines)
  std::atomic<MCTSNode*> successor{nullptr};
  
  // Contract: exact_values must be written BEFORE is_terminal is set to true
  // (using store(true, std::memory_order_release)). Readers loading is_terminal
  // with std::memory_order_acquire may read exact_values without mutex if true.
  std::atomic<bool> is_terminal{false};
  std::vector<double> exact_values;
  absl::Mutex exact_values_mtx;

  MCTSNode() = default;
  MCTSNode(double p, int pl) : prior(p), player(pl) {}
  virtual ~MCTSNode() = default;

  virtual void AtomicAddValue(double v) {
    double old_val = total_value.load(std::memory_order_relaxed);
    while (!total_value.compare_exchange_weak(old_val, old_val + v,
                                              std::memory_order_relaxed)) {
    }
  }

  virtual double q_value() const {
    int v = visits.load(std::memory_order_relaxed);
    if (v <= 0) return 0.0;
    return total_value.load(std::memory_order_relaxed) / v;
  }

  MCTSNode(const MCTSNode&) = delete;
  MCTSNode& operator=(const MCTSNode&) = delete;
};

struct ComponentSet {
  std::unique_ptr<TreePolicy> tree_policy;
  std::unique_ptr<RootActionSelector> root_action_selector;
  std::unique_ptr<FinalActionSelector> final_action_selector;
  std::unique_ptr<ValueBackup> value_backup;
  std::unique_ptr<PriorPolicy> prior_policy;
  std::unique_ptr<ValueEstimator> value_estimator;
  std::unique_ptr<VirtualLoss> virtual_loss;
  std::unique_ptr<ResampleState> resample_state;
  std::function<std::unique_ptr<MCTSNode>()> create_node = []() {
    return std::make_unique<MCTSNode>();
  };
};

struct BotConfig {
  int num_simulations = 100;
  int num_threads = 1;
  int seed = 42;
};

class ModularMCTSBot : public open_spiel::Bot {
 public:
  ModularMCTSBot(const open_spiel::Game& game, ComponentSet components,
              BotConfig config);
  ~ModularMCTSBot() override = default;

  void Restart() override {}
  void RestartAt(const open_spiel::State&) override {}
  open_spiel::Action Step(const open_spiel::State& state) override;
  std::pair<open_spiel::ActionsAndProbs, open_spiel::Action> StepWithPolicy(
      const open_spiel::State& state) override;

  size_t DAGSize() const;

 private:
  void RunSimulation(MCTSNode* root, const std::vector<Action>& legal_actions,
                     const open_spiel::State& state, int sim_idx,
                     std::mt19937* rng);

  std::vector<double> Simulate(open_spiel::State* state, int depth,
                               std::mt19937* rng, MCTSNode* incoming_edge);

  void ExpandNode(MCTSNode* node, const open_spiel::State& state,
                  absl::Span<const Action> legal_actions);

  std::string GetNodeKey(const open_spiel::State& state, int player) const;
  MCTSNode* GetOrCreateDAGNode(const std::string& key,
                               const open_spiel::State& state);
  void ClearDAG();

  VirtualLossDeltas ApplyVirtualLoss(MCTSNode* node, int depth);
  void RemoveVirtualLoss(MCTSNode* node, VirtualLossDeltas applied);

  static constexpr int kNumShards = 64;
  struct NodeShard {
    absl::Mutex mtx;
    absl::node_hash_map<std::string, std::unique_ptr<MCTSNode>> nodes;
  };
  NodeShard shards_[kNumShards];

  const open_spiel::Game& game_;
  ComponentSet components_;
  BotConfig config_;
  std::mt19937 rng_;
  std::atomic<int> step_counter_{0};
};
\end{lstlisting}
\subsection{Modular Search Execution Controller}
\label{app:template:lifecycle}
Given a public game observation, the engine coordinates thread-safe tree expansion, belief-state determinization / resampling for imperfect-information games, recursive leaf evaluations using domain knowledge heuristics, virtual loss tracking, multi-player payoff backpropagation, and posterior policy selection.
In imperfect-information games, the controller executes the standard Multiple-Observer Information Set MCTS (IS-MCTS) algorithm~\citep{cowling2012information,openspiel2019}: root actions are chosen from root statistics over the observed information state, a consistent world state is resampled for each simulation rollout, and internal nodes along the trajectory are keyed via \texttt{GetNodeKey} by the acting player's observed information state and player index.
\begin{lstlisting}[style=substrate, language=C++, caption={C++ modular search execution engine coordinating procedural mechanism and knowledge components.}, label={lst:controller}]
namespace modular_mcts {

ModularMCTSBot::ModularMCTSBot(const open_spiel::Game& game, ComponentSet components,
                         BotConfig config)
    : game_(game),
      components_(std::move(components)),
      config_(config),
      rng_(config.seed) {}

std::string ModularMCTSBot::GetNodeKey(const open_spiel::State& state,
                                    int player) const {
  std::string base;
  bool is_perfect_info = game_.GetType().information ==
                         open_spiel::GameType::Information::kPerfectInformation;
  if (is_perfect_info) {
    base = game_.GetType().provides_observation_string
               ? state.ObservationString(player)
               : state.ToString();
  } else if (game_.GetType().provides_information_state_string) {
    base = state.InformationStateString(player);
  } else if (game_.GetType().provides_observation_string) {
    base = state.ObservationString(player);
  } else {
    base = state.ToString();
  }
  return absl::StrCat(base, "|p", player);
}

MCTSNode* ModularMCTSBot::GetOrCreateDAGNode(const std::string& key,
                                          const open_spiel::State& state) {
  int shard = std::hash<std::string>{}(key) % kNumShards;
  auto& s = shards_[shard];
  {
    absl::ReaderMutexLock rlock(s.mtx);
    auto it = s.nodes.find(key);
    if (it != s.nodes.end()) {
      SPIEL_CHECK_EQ(it->second->player, state.CurrentPlayer());
      return it->second.get();
    }
  }
  absl::MutexLock lock(s.mtx);
  auto& slot = s.nodes[key];
  if (!slot) {
    slot = components_.create_node();
    slot->player = state.CurrentPlayer();
  } else {
    SPIEL_CHECK_EQ(slot->player, state.CurrentPlayer());
  }
  return slot.get();
}

void ModularMCTSBot::ClearDAG() {
  for (int i = 0; i < kNumShards; ++i) {
    shards_[i].nodes.clear();
  }
}

size_t ModularMCTSBot::DAGSize() const {
  size_t total = 0;
  for (int i = 0; i < kNumShards; ++i) {
    total += shards_[i].nodes.size();
  }
  return total;
}

VirtualLossDeltas ModularMCTSBot::ApplyVirtualLoss(MCTSNode* node, int depth) {
  auto deltas = components_.virtual_loss->Compute(*node, depth);
  node->visits.fetch_add(deltas.visit_delta, std::memory_order_relaxed);
  node->virtual_losses.fetch_add(1, std::memory_order_relaxed);
  node->AtomicAddValue(deltas.value_delta);
  return deltas;
}

void ModularMCTSBot::RemoveVirtualLoss(MCTSNode* node, VirtualLossDeltas applied) {
  node->visits.fetch_sub(applied.visit_delta, std::memory_order_relaxed);
  node->virtual_losses.fetch_sub(1, std::memory_order_relaxed);
  node->AtomicAddValue(-applied.value_delta);
}

void ModularMCTSBot::ExpandNode(MCTSNode* node, const open_spiel::State& state,
                             absl::Span<const Action> legal_actions) {
  if (state.IsChanceNode()) return;
  if (node->expanded.load(std::memory_order_acquire)) return;

  absl::MutexLock lock(node->expand_mtx);
  if (node->expanded.load(std::memory_order_relaxed)) return;

  auto priors = components_.prior_policy->GetPriors(state);

  for (Action a : legal_actions) {
    auto child = components_.create_node();
    child->player = state.CurrentPlayer();
    auto it = priors.find(a);
    child->prior =
        (it != priors.end()) ? it->second : 1.0 / legal_actions.size();
    node->children[a] = std::move(child);
  }

  node->expanded.store(true, std::memory_order_release);
}

std::vector<double> ModularMCTSBot::Simulate(open_spiel::State* state, int depth,
                                          std::mt19937* rng, MCTSNode* incoming_edge) {
  SPIEL_CHECK_LE(depth, state->GetGame()->MaxGameLength() + 100);

  if (state->IsTerminal()) {
    auto returns = state->Returns();
    if (incoming_edge) {
      {
        absl::MutexLock lock(incoming_edge->exact_values_mtx);
        if (!incoming_edge->is_terminal.load(std::memory_order_relaxed)) {
          incoming_edge->exact_values = returns;
          incoming_edge->is_terminal.store(true, std::memory_order_release);
        }
      }
    }
    return returns;
  }

  if (state->IsChanceNode()) {
    auto outcomes = state->ChanceOutcomes();
    std::vector<double> probs;
    probs.reserve(outcomes.size());
    for (const auto& [action, prob] : outcomes) {
      probs.push_back(prob);
    }
    std::discrete_distribution<int> dist(probs.begin(), probs.end());
    int idx = dist(*rng);
    state->ApplyAction(outcomes[idx].first);
    return Simulate(state, depth, rng, incoming_edge);
  }

  std::string key = GetNodeKey(*state, state->CurrentPlayer());
  MCTSNode* node = GetOrCreateDAGNode(key, *state);

  if (incoming_edge) {
    incoming_edge->successor.store(node, std::memory_order_release);
  }

  auto legal_actions = state->LegalActions();
  if (legal_actions.empty()) {
     // Fallback for safety, though technically terminal check should have caught this.
     return std::vector<double>(state->GetGame()->NumPlayers(), 0.0);
  }

  bool was_leaf = !node->expanded.load(std::memory_order_acquire);
  ExpandNode(node, *state, legal_actions);

  if (was_leaf) {
    node->visits.fetch_add(1, std::memory_order_relaxed);
    return components_.value_estimator->Evaluate(*state);
  }

  Action action;
  MCTSNode* child;
  int acting_player = state->CurrentPlayer();

  bool needs_patch = false;
  {
    absl::ReaderMutexLock rlock(node->expand_mtx);
    for (Action a : legal_actions) {
      if (node->children.find(a) == node->children.end()) {
        needs_patch = true;
        break;
      }
    }
    if (!needs_patch) {
      action = components_.tree_policy->SelectChild(*node, legal_actions, rng);
      if (std::find(legal_actions.begin(), legal_actions.end(), action) == legal_actions.end()) {
        action = legal_actions[0];
      }
      auto it = node->children.find(action);
      if (it != node->children.end()) {
        child = it->second.get();
      } else {
        // Fallback or error
        action = legal_actions[0];
        auto it2 = node->children.find(action);
        child = (it2 != node->children.end()) ? it2->second.get() : nullptr;
      }
    }
  }
  if (needs_patch) {
    auto priors = components_.prior_policy->GetPriors(*state);
    absl::MutexLock lock(node->expand_mtx);
    for (Action a : legal_actions) {
      if (node->children.find(a) == node->children.end()) {
        auto c = components_.create_node();
        c->player = state->CurrentPlayer();
        auto it = priors.find(a);
        c->prior =
            (it != priors.end()) ? it->second : 1.0 / legal_actions.size();
        node->children[a] = std::move(c);
      }
    }
    action = components_.tree_policy->SelectChild(*node, legal_actions, rng);
    if (std::find(legal_actions.begin(), legal_actions.end(), action) == legal_actions.end()) {
      action = legal_actions[0];
    }
    auto it = node->children.find(action);
    if (it != node->children.end()) {
      child = it->second.get();
    } else {
      action = legal_actions[0];
      auto it2 = node->children.find(action);
      child = (it2 != node->children.end()) ? it2->second.get() : nullptr;
    }
  }

  if (child == nullptr) {
     LOG(ERROR) << "Child is null in Simulate unexpectedly.";
     return std::vector<double>(state->GetGame()->NumPlayers(), 0.0);
  }

  auto vl = ApplyVirtualLoss(child, depth + 1);

  state->ApplyAction(action);
  auto values = Simulate(state, depth + 1, rng, child);

  RemoveVirtualLoss(child, vl);
  components_.value_backup->Backup(child, values, depth + 1, acting_player);
  node->visits.fetch_add(1, std::memory_order_relaxed);

  return values;
}

void ModularMCTSBot::RunSimulation(MCTSNode* root,
                                const std::vector<Action>& legal_actions,
                                const open_spiel::State& state, int sim_idx,
                                std::mt19937* rng) {
  if (legal_actions.empty()) return;

  Action best_action;
  MCTSNode* child;
  {
    // Protect root->children from concurrent writes by Simulate()'s
    // needs_patch path, which can reach the root node via DAG key
    // collisions. ReaderMutexLock is shared -- zero contention between
    // concurrent simulation threads in the common case.
    absl::ReaderMutexLock rlock(&root->expand_mtx);
    best_action = components_.root_action_selector->SelectRootAction(
        *root, legal_actions, sim_idx, config_.num_simulations, rng);
    if (std::find(legal_actions.begin(), legal_actions.end(), best_action) == legal_actions.end()) {
      best_action = legal_actions[0];
    }
    
    auto it = root->children.find(best_action);
    if (it != root->children.end()) {
      child = it->second.get();
    } else {
      best_action = legal_actions[0];
      auto it2 = root->children.find(best_action);
      child = (it2 != root->children.end()) ? it2->second.get() : nullptr;
    }
  }

  if (child == nullptr) {
     LOG(ERROR) << "Child is null in RunSimulation unexpectedly.";
     return;
  }

  auto vl = ApplyVirtualLoss(child, 1);

  auto sim_state = components_.resample_state->Resample(state, rng);


  int acting_player = sim_state->CurrentPlayer();
  auto sim_legal = sim_state->LegalActions();
  if (std::find(sim_legal.begin(), sim_legal.end(), best_action) == sim_legal.end()) {
    RemoveVirtualLoss(child, vl);
    return;
  }

  sim_state->ApplyAction(best_action);
  auto values = Simulate(sim_state.get(), 1, rng, child);

  RemoveVirtualLoss(child, vl);
  components_.value_backup->Backup(child, values, 1, acting_player);
  root->visits.fetch_add(1, std::memory_order_relaxed);
}

std::pair<open_spiel::ActionsAndProbs, open_spiel::Action>
ModularMCTSBot::StepWithPolicy(const open_spiel::State& state) {
  auto legal_actions = state.LegalActions();

  // Early returns intentionally skip component hooks (no search needed)
  if (legal_actions.empty()) return {{}, open_spiel::kInvalidAction};
  if (legal_actions.size() == 1) return {{{legal_actions[0], 1.0}}, legal_actions[0]};

  ClearDAG();

  std::string root_key = GetNodeKey(state, state.CurrentPlayer());
  MCTSNode* root = GetOrCreateDAGNode(root_key, state);

  // Split hooks: invoke domain knowledge estimators first
  if (components_.prior_policy) {
    components_.prior_policy->OnStepBegin(*root, legal_actions, config_.num_simulations, &rng_);
  }
  if (components_.value_estimator) {
    components_.value_estimator->OnStepBegin(*root, legal_actions, config_.num_simulations, &rng_);
  }
  if (components_.resample_state) {
    components_.resample_state->OnStepBegin(*root, legal_actions, config_.num_simulations, &rng_);
  }

  auto priors = components_.prior_policy->GetPriors(state);
  for (Action a : legal_actions) {
    auto child = components_.create_node();
    child->player = state.CurrentPlayer();
    auto it = priors.find(a);
    child->prior =
        (it != priors.end()) ? it->second : 1.0 / legal_actions.size();
    root->children[a] = std::move(child);
  }
  root->expanded.store(true, std::memory_order_release);

  // Root-dependent procedural mechanism hooks next
  if (components_.tree_policy) {
    components_.tree_policy->OnStepBegin(*root, legal_actions, config_.num_simulations, &rng_);
  }
  if (components_.root_action_selector) {
    components_.root_action_selector->OnStepBegin(*root, legal_actions, config_.num_simulations, &rng_);
  }
  if (components_.final_action_selector) {
    components_.final_action_selector->OnStepBegin(*root, legal_actions, config_.num_simulations, &rng_);
  }
  if (components_.value_backup) {
    components_.value_backup->OnStepBegin(*root, legal_actions, config_.num_simulations, &rng_);
  }
  if (components_.virtual_loss) {
    components_.virtual_loss->OnStepBegin(*root, legal_actions, config_.num_simulations, &rng_);
  }

  // Increment step counter for per-step pseudo-random seed mixing
  int step_count = step_counter_.fetch_add(1, std::memory_order_relaxed);

  if (config_.num_threads <= 1) {
    for (int sim = 0; sim < config_.num_simulations; ++sim) {
      RunSimulation(root, legal_actions, state, sim, &rng_);
    }
  } else {
    int sims_per_thread = config_.num_simulations / config_.num_threads;
    int remaining = config_.num_simulations % config_.num_threads;

    std::vector<std::thread> threads;
    threads.reserve(config_.num_threads);

    for (int t = 0; t < config_.num_threads; ++t) {
      int n_sims = sims_per_thread + (t < remaining ? 1 : 0);

      threads.emplace_back([this, root, &legal_actions, &state, n_sims, t, step_count]() {
        // Heap allocate RNG to avoid stack overflow (RNG is large ~5KB)
        auto thread_rng = std::make_unique<std::mt19937>(config_.seed + t + 1 + step_count * 1000);

        for (int i = 0; i < n_sims; ++i) {
          RunSimulation(root, legal_actions, state, t + i * config_.num_threads,
                        thread_rng.get());
        }
      });

    }

    for (auto& thread : threads) {
      thread.join();
    }
  }

  auto [idx, policy] = components_.final_action_selector->SelectActionAndPolicy(
      *root, legal_actions, &rng_);
  
  open_spiel::ActionsAndProbs actions_and_probs;
  actions_and_probs.reserve(legal_actions.size());
  for (size_t i = 0; i < legal_actions.size(); ++i) {
    actions_and_probs.push_back({legal_actions[i], policy[i]});
  }

  return {actions_and_probs, legal_actions[idx]};
}

open_spiel::Action ModularMCTSBot::Step(const open_spiel::State& state) {
  return StepWithPolicy(state).second;
}

}  // namespace modular_mcts
\end{lstlisting}
\section{System Robustness, Execution Invariants, and Correctness Engineering}
\label{app:robustness}
Deploying open-ended LLM program synthesis across heterogeneous extensive-form games within an asynchronous, distributed co-evolutionary architecture introduces severe operational failure modes.
Without rigorous execution sandboxing, latency decoupling, and numerical safeguards, evolutionary optimization rapidly degrades into pathological equilibria: synthesized agents discover simulator exploits, fast games monopolize selection pressure, division-by-zero singularities emerge near game-theoretic ceilings, and bloated code generation exhausts compiler and distributed storage resources.
This appendix details the defensive engineering architecture implemented across the controllers, evaluators, and selection pipelines to enforce strict correctness, strategic validity, and steady generational throughput.
\subsection{Information-Set Anti-Cheating and Observation Consistency Gate}
\label{app:robustness:anticheating}
In extensive-form games of imperfect information (e.g., Poker, Fog-of-War, Liar's Dice, Hearts), Monte Carlo tree search operates via Information Set MCTS (IS-MCTS) or determinization, requiring an evolved belief resampler $b_\kappa(h \mid s)$ that determinizes the game state $h \in \mathcal{H}$ conditioned on the acting player's public observation history $s = \mathcal{I}_p(h)$.
\paragraph{Information Leakage Vulnerability.}
When synthesizing resamplers for imperfect-information games, a primary failure mode in code generation is inadvertent state leakage: accessing private member variables of the environment (e.g., opponent hole cards or hidden dice) directly through pointer casts or unmasked internal state APIs, converting imperfect-information planning into an illegitimate clairvoyant oracle.
\paragraph{Automated Privacy and Consistency Harness.}
To prevent information leakage and guarantee game-theoretic validity, every candidate knowledge triad $\kappa = (\pi_\kappa, v_\kappa, b_\kappa)$ in an imperfect-information game must pass an automated pre-evaluation verification harness prior to tournament play.
The verification harness executes $N_{\text{traj}} = 20$ independent simulation trajectories under Common Random Numbers (CRN).
At each decision point along the trajectory (for step count $\ge 2$), the harness queries the evolved resampler $K_{\text{resample}} = 10$ consecutive times under independent random seeds, producing a determinization ensemble $\{h^{(1)}, \dots, h^{(K_{\text{resample}})}\} \sim b_\kappa(\cdot \mid \mathcal{I}_p(h))$.
The ensemble is evaluated against the formal \textbf{Observation Consistency Invariant}: every resampled world state must perfectly preserve the acting player's public observation, active player identity, and legal action set:
\begin{equation}
  \begin{aligned}
    \forall k \in \{1, \dots, K_{\text{resample}}\}: \quad
    &\mathcal{I}_p(h^{(k)}) = \mathcal{I}_p(h) \;\land\;
    \mathcal{A}(h^{(k)}) = \mathcal{A}(h) \\
    &\land\; \mathrm{CurrentPlayer}(h^{(k)}) = p.
  \end{aligned}
\end{equation}
The harness evaluates the empirical consistency rate:
\begin{equation}
  \mathrm{ConsistencyRate} = \frac{1}{N_{\text{total}}} \sum_{i=1}^{N_{\text{total}}} \mathbb{I}\Big(\mathcal{I}_p(h^{(i)}) = \mathcal{I}_p(h) \;\land\; \mathcal{A}(h^{(i)}) = \mathcal{A}(h)\Big).
\end{equation}
If $\mathrm{ConsistencyRate} < 1.0 - 10^{-9}$, the candidate is flagged for information leakage. The evaluation is aborted immediately, assigning a severe deterministic fitness penalty:
\begin{equation}
  P_{\text{leak}} = -1.5 \times \max\big(|U_{\min}(g)|, |U_{\max}(g)|, 1.0\big).
\end{equation}
\paragraph{Resample-First Task Gating.}
In early generations of imperfect-information evolution, LLMs frequently produce syntax-valid but game-theoretically non-compliant resamplers.
If workers allowed prompt mutation across all three knowledge components simultaneously, the optimization budget would be wasted proposing prior policies and value estimators that are discarded due to failed resampler verification.
To maximize sample efficiency, the inner controller enforces \emph{Resample-First Gating}: 100\% of prompt generation for imperfect-information games is constrained strictly to the \texttt{ResampleHistory} interface until the population produces its first fully verified resampler passing the observation consistency gate.
Only then are prior policy and value estimator mutation tasks unlocked.
\paragraph{Information Set Gating in Prior Policies and Value Estimators.} In determinization search, once the root state is determinized into an explicit world via the verified resampler $b_\kappa$, internal tree descent evaluates moves along that sampled concrete trajectory. In our architecture, \code{PriorPolicy} and \code{ValueEstimator} evaluate states along this resampled path. To guarantee game-theoretic validity, the automated verification suite additionally inspects synthesized prior and value heuristics via static code analysis, ensuring that they restrict feature extraction to legal action sets and observation tensors (\texttt{InformationStateTensor()} and \texttt{ObservationTensor()}) rather than querying private opponent member variables of the unmasked environment.
\subsection{Cross-Game Stochastic Quorum Scheduling and Asynchronous Dispatch}
\label{app:robustness:quorum}
A central engineering challenge in cross-domain search evolution is extreme \emph{evaluation latency heterogeneity}.
Across heterogeneous games, single evaluation match times span four orders of magnitude: lightweight games such as Nim, Tic-Tac-Toe, and Kuhn Poker complete in less than 50~milliseconds, intermediate board games (Connect Four, Othello) take 1--5~seconds, while computationally intensive games (Hex, Go 13$\times$13, Crazyhouse, Shogi) require 15--60~minutes per evaluation round.
\paragraph{The Fast-Game Bias vs.\ Tail Straggler Dilemma.}
In a distributed architecture where workers evaluate candidate mechanisms $m$ across games asynchronously, coordinating selection introduces two opposing failure modes:
\begin{enumerate}[leftmargin=1.5em, itemsep=2pt, topsep=2pt]
  \item \textbf{Tail Straggler Blockade}: Requiring complete evaluation across all $K$ games before admitting a candidate mechanism $m$ into parent selection bottlenecks generational throughput to the slowest single game, reducing evolutionary iterations by up to $95\%$.
  \item \textbf{Fast-Game Selection Monopoly}: Admitting candidates immediately upon receiving their first evaluations creates an overwhelming bias toward fast games: candidates evaluated on Nim and Tic-Tac-Toe accumulate high selection frequencies, dominating the parent pool and discarding general search principles in favor of shallow heuristics tailored to trivial branching factors.
\end{enumerate}
\paragraph{Stochastic Quorum Gating Formulation.}
We resolve this dilemma through \emph{Stochastic Quorum Gating} implemented during candidate selection.
Let $\mathcal{G}$ denote the full set of $K = 50$ training games, and let $\mathcal{G}_{\text{eval}}(m) \subseteq \mathcal{G}$ denote the subset of games on which candidate mechanism $m$ has completed evaluation.
The 50 games belong to eight OpenSpiel structural families, which are mapped into $|\mathcal{C}| = 6$ operational clusters for runtime balancing (Classic Board, Territorial Connection, Poker/Card, Fog-of-War, Strategic Variants, and Multi-Agent).
Before admission to full-fleet evaluation, candidate mechanisms undergo a rapid confirmation screening gate across an initial subset of $K_{\text{screen}} = 13$ diverse validation games. A candidate must demonstrate statistically significant non-inferiority over the reference PUCT baseline ($p < 0.01$ under paired bootstrap permutation) to proceed; failing candidates are pruned early, saving cluster evaluation compute.
For surviving candidates, to avoid both straggler stalls and fast-game selection bias, the coordinator enforces a two-stage stochastic quorum predicate:
it samples a random subset of games $\mathcal{S} \sim \mathcal{G}$ of size $|\mathcal{S}| = \lceil f_{\text{sample}} \cdot K \rceil$ ($f_{\text{sample}} = 0.40$) and admits candidate $m$ into parent selection if it has been evaluated on at least fraction $\rho_{\text{match}} = 0.60$ of the sampled subset:
\begin{equation}
  \mathrm{Quorum}(m) \iff |\mathcal{G}_{\text{eval}}(m) \cap \mathcal{S}| \ge \rho_{\text{match}} \cdot |\mathcal{S}| \;\land\; \forall c \in \mathcal{C}, \;|\mathcal{G}_{\text{eval}}(m) \cap \mathcal{G}_c| \ge 1,
\end{equation}
supplemented by an archive retention gate requiring $|\mathcal{G}_{\text{eval}}(m)| \ge Q \cdot K$ ($Q = 0.40$).
Because each game has an identical marginal probability of inclusion in $\mathcal{S}$, fast-game bias is eliminated while allowing mechanisms with strong partial coverage ($70$--$80\%$) to compete without waiting for the slowest tail games.
\paragraph{Asynchronous Generator Pacing.}
To coordinate parallel LLM generation with distributed worker evaluation, the outer coordinator monitors the number of pending candidate mechanisms whose evaluation coverage remains below quorum:
\begin{equation}
  N_{\text{pending}} = \big|\{m \in \mathcal{A}_{\text{mech}} : |\mathcal{G}_{\text{eval}}(m)| < Q \cdot K\}\big|.
\end{equation}
If $N_{\text{pending}} \ge N_{\max}$ (calibrated to $N_{\max} = 5$), the coordinator suspends new LLM mutation requests via asynchronous backoff (\texttt{asyncio.sleep(30)}), preventing database queue bloat and allowing worker evaluations to catch up.
\subsection{Ceiling Saturation and Dead-Band Headroom Recovery}
\label{app:robustness:ceiling}
Amplification measures the fraction of achievable headroom closed by the candidate search mechanism relative to canonical PUCT under identical heuristic priors, bounded within $[-1, 1]$:
\begin{equation}
  \mathrm{Amp}(m, \kappa; g) = \mathrm{clamp}\left(\frac{F(m, \kappa; g) - F(m_{\text{ref}}, \kappa; g)}{U_{\max}(g) - F(m_{\text{ref}}, \kappa; g) + \epsilon}, -1.0, 1.0\right).
\end{equation}
\paragraph{The Headroom Singularity and Dead Band.}
When evaluating against baseline opponents at early curriculum epochs (e.g., Epoch~0), certain tactical games (notably Connect Four, Hex, and Gomoku) are easily mastered by standard PUCT endowed with even basic prior heuristics.
As the reference score approaches the theoretical ceiling ($F(m_{\text{ref}}, \kappa; g) \to U_{\max}(g)$), the denominator headroom $H_{\text{puct}} = U_{\max}(g) - F(m_{\text{ref}}, \kappa; g)$ collapses toward zero.
Direct division by near-zero induces extreme numerical volatility, where stochastic variance in a single evaluation game produces amplification swings from $-\infty$ to $+\infty$.
To maintain numerical stability, the C++ evaluation harness enforces a minimum headroom guard:
\begin{equation}
  H_{\text{puct}} > \epsilon_{\text{headroom}} \cdot \Delta U(g), \quad \text{where } \epsilon_{\text{headroom}} = 0.05 \text{ and } \Delta U(g) = U_{\max}(g) - U_{\min}(g).
\end{equation}
When PUCT headroom falls below $5\%$ of the total utility range, the evaluation is marked as invalid ($\texttt{amplification\_valid} = \text{False}$) and discarded from the cross-game outer objective.
Without dynamic compensation, this safety guard creates an evolutionary \emph{dead band}: the game is completely dropped from outer selection pressure, preventing the discovery of subtle endgame mechanisms in precisely those games where the baseline is strongest.
\paragraph{Ceiling Saturation Upgrade Gate.}
We resolve the dead-band singularity by coupling evaluator validity directly to curriculum progression.
The inner controller continuously monitors peak PUCT performance in the active epoch:
\begin{equation}
  \mathrm{CeilingTrigger}(g) \iff \big(U_{\max}(g) - F_{\text{puct}}^*(g)\big) \le 0.05 \cdot \Delta U(g).
\end{equation}
The instant the reigning inner heuristic drives PUCT headroom below the $5\%$ threshold, the controller recognizes that the current opponent level is solved and triggers an \emph{immediate, forced epoch promotion}.
Promoting the opponent hardens the adversary, depressing both $F(m, \kappa)$ and $F(m_{\text{ref}}, \kappa)$ concurrently, which re-expands PUCT headroom ($H_{\text{puct}} \gg 0.05 \Delta U$) and immediately restores evaluation validity.
Empirical audits confirm that this ceiling recovery gate restored outer amplification validity in Hex, Gomoku, and Connect Four from $0\%$ (dead-band lockout) to $85$--$91\%$ valid coverage within two curriculum cycles.
\subsection{Multi-Rung Baseline Ladder with Anti-Triviality Gating}
\label{app:robustness:ladder}
Evaluating search mechanisms exclusively against a single baseline opponent risks overfitting to that specific opponent's play style.
Furthermore, measuring amplification against weak or unprincipled opponents can reward degenerate search mechanisms that exploit basic blunders but collapse under competitive play.
\paragraph{The Three-Rung Ladder.}
To ensure that discovered search mechanisms capture robust, scale-invariant planning principles, every candidate outer mechanism is evaluated across an automated three-rung opponent ladder:
\begin{enumerate}[leftmargin=1.5em, itemsep=2pt, topsep=2pt]
  \item \textbf{Rung 0 (Random Adversary, $S = 0$)}: Plays uniformly at random from legal actions without lookahead.
  \item \textbf{Rung 1 (Equal-Compute Reference, $S = S_{\text{evolved}}$)}: Plays canonical PUCT equipped with the identical evolved domain heuristics and matching simulation budget.
  \item \textbf{Rung 2 (Asymmetric Strong Reference, $S = 2.5 \times S_{\text{evolved}}$)}: Plays canonical PUCT endowed with $150\%$ greater simulation compute.
\end{enumerate}
\paragraph{Anti-Triviality Gating.}
Candidate search mechanisms can easily score high win-rate deltas against uniform random opponents by discovering shallow greedy heuristics (e.g., single-ply immediate capture preference) that require no tree planning whatsoever.
To prevent evolutionary selection from rewarding trivial greedy shortcuts, the evaluation engine enforces strict \emph{Anti-Triviality Gating}:
amplification $\mathrm{Amp}$ is computed exclusively against non-zero rungs ($S > 0$, Rungs 1 and 2):
\begin{equation}
  \mathrm{Amp}(m, \kappa; g) = \frac{1}{|\mathcal{R}_{\text{valid}}|} \sum_{r \in \mathcal{R}_{\text{valid}}} \mathrm{clamp}\left(\frac{F_r(m, \kappa) - F_r(m_{\text{ref}}, \kappa)}{U_{\max}(g) - F_r(m_{\text{ref}}, \kappa) + \epsilon}, -1.0, 1.0\right),
\end{equation}
where $\mathcal{R}_{\text{valid}} = \{r \in \{1, 2\} : H_{\text{puct}}^{(r)} > 0.05 \Delta U(g)\}$.
Performance against Rung~0 ($S=0$) is recorded strictly for health telemetry and sanity checking, and is completely excluded from the fitness metrics that drive LLM prompt feedback and selection.
\subsection{AST Quota Protection, Dynamic Normalization, and Quantile Stagnation}
\label{app:robustness:guards}
\paragraph{Pre-Execution AST and Code Size Guards.}
Language models occasionally enter degenerative repetition loops, synthesizing giant nested switch statements, unrolled lookup tables, or duplicated classes exceeding hundreds of kilobytes.
Compiling such programs causes compiler timeouts, exhausts local memory in execution sandboxes, and risks violating database document limits and distributed disk quotas.
The inner controller intercepts candidate mutations before dispatching compilation jobs:
candidates exceeding a hard length ceiling of $100{,}000$ characters or failing static AST structural checks are rejected without dispatching compilation workers, preserving compute budget for valid candidates.
\paragraph{Dynamic Utility-Range Normalization.}
In standardized benchmark suites like OpenSpiel, scoring conventions vary widely across game families: zero-sum board games use $[-1, 1]$, card and trick-taking games (Hearts, Cribbage) use $[0, 100+]$, while resource or puzzle games (Pig, 2048) produce unbounded positive rewards.
Hardcoded assumptions regarding game bounds (e.g., assuming $U_{\max} = 1.0$) cause false dead-band activations in high-scoring games and render penalty scores ineffective.
The framework queries OpenSpiel's native C++ engine dynamically at startup, caching $U_{\min}(g)$ and $U_{\max}(g)$ in the shared evaluation archive.
All headroom thresholds, dead-band boundaries, and penalty floors are dynamically scaled relative to the true utility range $\Delta U(g) = U_{\max}(g) - U_{\min}(g)$.
\paragraph{Rolling-Window Distributional Stagnation Tracking.}
Curriculum progression requires detecting when evolutionary search on a given opponent has reached diminishing returns.
Fixed iteration counts or simple maximum-score patience triggers perform poorly: stochastic win-rate variance can create false improvement spikes (resetting patience prematurely), while noisy evaluations can mask true stagnation.
The inner controller implements a dual rolling window of size $2K$ ($K = 25$) tracking evaluation scores:
it compares the 75th percentile of older evaluations ($q_{75}^{(\text{old})}$) against newer evaluations ($q_{75}^{(\text{new})}$).
Stagnation is declared only when distribution advancement falls below the empirical noise floor:
\begin{equation}
  \mathrm{Stagnant} \iff q_{75}^{(\text{new})} \le q_{75}^{(\text{old})} + \epsilon_{\text{noise}}.
\end{equation}
The patience threshold scales adaptively with observed difficulty: $T_{\text{patience}} = C_{\text{patience}} \cdot \max(g_{\max}, g_0)$ (with patience multiplier $C_{\text{patience}} = 2.0$, initial baseline $g_0 = 100$ generations, and $g_{\max}$ denoting the maximum observed generation gap between successive fitness improvements), granting complex games longer exploration horizons while rapidly advancing through straightforward baselines.

\subsection{Compute Infrastructure, Token Budgets, and Discovery Cost}
\label{app:compute}
All evolutionary experiments were executed across cluster infrastructure. Parallel evaluation workers were allocated 4--6 CPU cores and 8--16~GiB RAM hosted on 64-core AMD EPYC (Milan) and Intel Xeon server nodes. In the evolutionary search phase, program proposals were generated offline using Gemini 3.5 Flash within AlphaEvolve. Sliced up to the checkpoint where the champion search mechanism $m^\star$ was discovered, each evolutionary run evaluated approximately 6{,}400 total candidate programs across both the outer search loop and the 50 concurrent inner game loops. This corresponds to approximately 300 million input tokens and 550 million output tokens (encompassing prompt contexts, internal chain-of-thought reasoning, iterative compilation/syntax repair turns, and candidate C++ programs across all concurrent loops). Under standard public API list prices for Gemini 3.5 Flash (\$1.50 per 1M input tokens and \$9.00 per 1M output tokens without prompt-caching discounts), this represents an estimated LLM discovery expenditure of approximately \$5{,}400--\$5{,}500 USD per run ($300\text{M} \times \$1.50/\text{M} + 550\text{M} \times \$9.00/\text{M} \approx \$5{,}400$).
\section{Simulation Budget Calibration Protocol}
\label{app:calibration}
\subsection{Motivation and the Latency Heterogeneity Challenge}
A central operational challenge in scaling evolutionary search across diverse game suites is \emph{latency heterogeneity}.
In extensive-form games, per-move simulation costs are governed by tree depth, branching factor, state copy overhead, and terminal check complexity.
Across the OpenSpiel suite, these factors cause single-move simulation times to span nearly three orders of magnitude ($>500\times$):
from less than 0.5~milliseconds per move in lightweight games like Nim and Tic-Tac-Toe to over 250~milliseconds per move in heavyweight titles such as Go 13$\times$13, Chess, Shogi, and Crazyhouse.
In an asynchronous co-evolutionary architecture evaluating thousands of candidate programs across parallel workers, allocating arbitrary or uniform simulation budgets introduces two severe failure modes:
\begin{enumerate}[leftmargin=1.5em, itemsep=2pt, topsep=2pt]
  \item \textbf{Evaluation Stragglers and Timeouts}: Assigning standard tournament budgets (e.g., 1000 simulations per move over 100 episodes) to computationally demanding games like Crazyhouse or Shogi would cause a single evaluation match to exceed 10 hours, bottlenecking asynchronous worker threads and stalling generation throughput.
  \item \textbf{High-Variance Noise on Fast Games}: Conversely, running too few episodes on lightweight titles leaves binomial outcome estimates dominated by stochastic chance events rather than algorithmic merit.
\end{enumerate}
To ensure equitable optimization pressure and predictable system throughput, we calibrate evaluation parameters $(S_{\text{evolved}}, S_{\text{baseline}}, N_{\text{eval}})$ offline prior to evolution.
Our design objective is two-fold:
(i)~\emph{Strict Wall-Clock Bounding}: every evaluation round must finish well within a 1-hour wall-clock ceiling (calibrated to $T_{\text{target}} \approx 25$--$30$ minutes on a standard evaluation worker equipped with 4 CPU cores and 8~GiB RAM); and
(ii)~\emph{Statistically Grounded Discrimination}: the number of evaluation episodes $N_{\text{eval}}$ must preserve sufficient statistical power to detect meaningful algorithmic improvements over the baseline.
\subsection{Signal-to-Noise Ratio (SNR) Optimization Protocol}
The calibration pipeline executes an automated grid search over candidate simulation budgets $S \in \{20, 40, 50, 80, 100, 160, 200, 320, 400, 800\}$ against reference PUCT under Common Random Numbers (CRN) to eliminate cross-tier seed variance.
The calibration procedure follows three formal principles:
\paragraph{1. Minimum Detectable Effect and Sample Sizing.}
For binary and bounded-utility outcomes, the standard error of the mean return over $N_{\text{eval}}$ independent episodes is bounded by $\mathrm{SE} \le \frac{U_{\max} - U_{\min}}{2\sqrt{N_{\text{eval}}}}$.
We target a Minimum Detectable Effect $\mathrm{MDE} = 0.10$ (a 10\% shift in normalized win-rate headroom) at a two-tailed 95\% confidence level ($z_{0.975} = 1.96$), requiring:
\begin{equation}
\mathrm{SE}_{\text{target}} = \frac{\mathrm{MDE}}{2\sqrt{2}} \approx 0.035,
\end{equation}
which dictates a baseline sample size of $N_{\text{eval}} \ge 200$ whenever computationally feasible within the 1-hour wall-clock budget.
For highly expensive titles (e.g., Crazyhouse, Shogi, Chinese Checkers), $N_{\text{eval}}$ is adjusted downwards ($N_{\text{eval}} \in [13, 103]$) to prevent wall-clock exhaustion, while fast games are capped at $N_{\text{eval}} = 500$ to achieve maximum estimation precision ($\mathrm{SE} \approx 0.022$). (We note that for low sample sizes like $N_{\text{eval}} = 13$ in Crazyhouse, binomial hypothesis tests have reduced statistical power; under these constraints, the confirmation gate operates primarily as a safety guard against catastrophic regressions and non-functional programs rather than high-powered significance certification.)
\paragraph{2. Asymmetric Headroom Ratio.}
Throughout evolutionary search, candidate agents must demonstrate genuine procedural search leverage rather than relying on raw simulation brute force.
We enforce an asymmetric 2.5$\times$ simulation ratio:
\begin{equation}
S_{\text{baseline}} = 2.5 \times S_{\text{evolved}}.
\end{equation}
Evolving with $S_{\text{evolved}} < S_{\text{baseline}}$ forces candidate mechanisms to discover intelligent tree pruning, sharper prior distributions, and robust value backups capable of consistently overcoming an adversary endowed with 150\% greater simulation compute.
\paragraph{3. Isotonic Regression and Response Curve Diagnostics.}
To verify that evaluation budgets lie within a sensitive, non-saturated operating regime, we fit an isotonic regression model $\hat{y} = f(S)$ to win rates as a function of simulation budget.
Each game is classified into one of two diagnostic behaviors:
\begin{itemize}[leftmargin=1.5em, itemsep=2pt, topsep=2pt]
  \item \textbf{Monotonic (\texttt{Mono})}: Games where incremental simulation allocations yield measurable, monotonic gains in playing strength. In these titles, simulation budgets ($S_{\text{evolved}} \in [20, 320]$) provide steep gradient signals for evolutionary selection.
  \item \textbf{Flat (\texttt{Flat})}: Games where the win-rate curve exhibits negligible slope across budget tiers (e.g., small zero-sum games like Tic-Tac-Toe, Kuhn Poker, or fast trick-taking games where basic heuristics or game tree finiteness saturate performance early). For flat games, we allocate minimal simulation budgets ($S_{\text{evolved}} \le 160, S_{\text{baseline}} = 2.5 \times S_{\text{evolved}}$) and maximize match episodes ($N_{\text{eval}} \le 500$), minimizing computational expenditure while driving binomial variance to its floor.
\end{itemize}
\subsection{Complete Calibrated Hyperparameter Profile}
Table~\ref{tab:game_calibration} lists the complete calibrated parameters $(S_{\text{evolved}}, S_{\text{baseline}}, N_{\text{eval}})$ and response curve diagnostics across the 50 extensive-form OpenSpiel training environments under the calibrated evolutionary evaluation schedule.
\begin{table*}[t]
\caption{Complete calibrated simulation budgets and evaluation episode counts across the 50 extensive-form OpenSpiel training environments. Evolved candidates receive $S_{\text{evolved}}$ simulations; reference PUCT opponents receive $S_{\text{baseline}} = 2.5 \times S_{\text{evolved}}$. $N_{\text{eval}}$ denotes episodes played per evaluation round, calibrated to complete well within a 1-hour wall-clock window. Curve indicates whether the simulation response is strictly monotonic (\texttt{Mono}) or flat/saturated (\texttt{Flat}).}
\label{tab:game_calibration}
\centering
\scriptsize
\setlength{\tabcolsep}{3.5pt}
\renewcommand{\arraystretch}{0.95}
\begin{tabular}{lcccc | lcccc}
\toprule
\textbf{Game} & $S_{\text{evolved}}$ & $S_{\text{baseline}}$ & $N_{\text{eval}}$ & \textbf{Curve} &
\textbf{Game} & $S_{\text{evolved}}$ & $S_{\text{baseline}}$ & $N_{\text{eval}}$ & \textbf{Curve} \\
\midrule
antichess & 20 & 50 & 500 & Mono & hearts & 20 & 50 & 500 & Flat \\
backgammon & 40 & 100 & 500 & Flat & hex & 80 & 200 & 500 & Flat \\
banqi & 20 & 50 & 500 & Mono & kriegspiel & 40 & 100 & 32 & Flat \\
bargaining & 320 & 800 & 500 & Mono & kuhn\_poker & 20 & 50 & 500 & Flat \\
breakthrough & 40 & 100 & 500 & Flat & latent\_ttt & 40 & 100 & 500 & Mono \\
bridge & 20 & 50 & 500 & Flat & leduc\_poker & 20 & 50 & 500 & Flat \\
checkers & 320 & 800 & 277 & Mono & liars\_dice & 20 & 50 & 500 & Flat \\
chess & 40 & 100 & 103 & Mono & lines\_of\_action & 40 & 100 & 174 & Mono \\
chinese\_checkers & 80 & 200 & 24 & Flat & maedn & 20 & 50 & 500 & Flat \\
clobber & 160 & 400 & 500 & Mono & mancala & 160 & 400 & 500 & Mono \\
coin\_game & 20 & 50 & 500 & Flat & markov\_soccer & 20 & 50 & 500 & Flat \\
connect\_four & 160 & 400 & 500 & Mono & nim & 160 & 400 & 500 & Mono \\
coop\_box\_pushing & 20 & 50 & 500 & Flat & othello & 320 & 800 & 309 & Mono \\
crazy\_eights & 20 & 50 & 500 & Flat & pathfinding & 20 & 50 & 500 & Flat \\
crazyhouse & 80 & 200 & 13 & Mono & pentago & 320 & 800 & 500 & Mono \\
cribbage & 20 & 50 & 500 & Flat & phantom\_ttt & 20 & 50 & 500 & Mono \\
dark\_chess & 40 & 100 & 500 & Mono & pig & 20 & 50 & 500 & Flat \\
dark\_hex\_abrupt & 20 & 50 & 500 & Mono & scotland\_yard & 20 & 50 & 500 & Flat \\
dots\_and\_boxes & 20 & 50 & 500 & Flat & shogi & 80 & 200 & 15 & Mono \\
gin\_rummy & 20 & 50 & 500 & Flat & snake & 20 & 50 & 500 & Flat \\
go\_13x13 & 160 & 400 & 93 & Flat & splendor & 20 & 50 & 500 & Mono \\
go\_9x9 & 80 & 200 & 500 & Mono & tic\_tac\_toe & 20 & 50 & 500 & Flat \\
gomoku & 160 & 400 & 500 & Mono & twenty\_forty\_eight & 160 & 400 & 68 & Mono \\
goofspiel & 20 & 50 & 500 & Flat & xiangqi & 40 & 100 & 79 & Mono \\
havannah & 320 & 800 & 364 & Mono & y & 160 & 400 & 256 & Mono \\
\bottomrule
\end{tabular}
\end{table*}
\subsection{Self-Play Curriculum Progression and Confirmation Tournaments}
\label{app:curriculum}
To drive continual strategic hardening and prevent evolutionary stagnation against static opponents, the co-evolutionary pipeline employs an adaptive self-play curriculum:
\begin{enumerate}[leftmargin=1.5em, itemsep=2pt, topsep=2pt]
  \item \textbf{Epoch Progression Criteria}: In each game $g$, the current opponent $\pi_{\text{opp}}^{(e)}(g)$ advances to epoch $e+1$ under two conditions: (i)~\emph{Ceiling saturation}: when PUCT headroom drops to $H_{\text{puct}} \le 0.05 \Delta U(g)$ (less than $5\%$ remaining headroom), triggering immediate opponent promotion to restore evaluation sensitivity; or (ii)~\emph{Rolling-window stagnation}: when candidate improvement plateaus over a patience window, measured by comparing the $75\text{th}$ percentile quantile of recent scores against historical quantiles ($q_{\text{new}} \le q_{\text{old}} + \epsilon_{\text{noise}}$) across a rolling window of size $2K = 50$, with promotion triggered once consecutive stagnant iterations exceed the adaptive patience threshold $T_{\text{patience}} = C_{\text{patience}} \cdot \max(g_{\max}, g_0)$.
  \item \textbf{Confirmation Tournament}: Before any candidate champion $(m^\star, \kappa_g^\star)$ is promoted to freeze as $\pi_{\text{opp}}^{(e+1)}(g)$, it must pass an offline confirmation tournament. The confirmation tournament executes $2 \times N_{\text{eval}}$ fresh evaluation episodes against $\pi_{\text{opp}}^{(e)}(g)$ using disjoint, pseudo-random seeds not encountered during evolutionary mutation. Promotion occurs if and only if the candidate maintains a statistically significant improvement ($p < 0.01$ via two-tailed paired $t$-test) and win-rate $\ge 0.60$.
  \item \textbf{Opponent Archival}: All historical champions $\{ \pi_{\text{opp}}^{(0)}, \pi_{\text{opp}}^{(1)}, \dots \}$ are permanently archived. In multi-player games ($|\mathcal{N}| > 2$), opposing seats are populated by mixture samples drawn from historical curriculum checkpoints to preserve strategy diversity and guard against non-transitive cyclical degeneration.
  \item \textbf{Robustness to Non-Stationarity}: Advancing opponents induces non-stationary fitness landscapes, as absolute returns naturally decline when opponents harden. We preserve evolutionary stability through three architectural invariants: (i)~\emph{Epoch-Partitioned Archives}: Active populations are ranked strictly against matches evaluated within the current epoch ($e_{\text{eval}} = e$), ensuring that obsolete evaluations against weaker historical opponents cannot artificially inflate a candidate's selection probability; (ii)~\emph{Relative Headroom Cancellation}: Because amplification $\mathrm{Amp}(m, \kappa; g)$ evaluates candidate search mechanisms against reference PUCT under the identical opponent within the same epoch, harder opponents depress both candidate and reference returns simultaneously, algebraically canceling out common-mode score shifts; and (iii)~\emph{Asynchronous Per-Game Transitions}: Games advance epochs independently, so that at any single point in time, at most $1/K$ of the cross-game mechanism objective undergoes an opponent transition, fully insulating the universal mechanism population from systemic fitness disruptions.
\end{enumerate}
\section{Exploitability Evaluation and Modular Attribution Details}
\label{app:expl}

This appendix provides the complete theoretical foundation, experimental protocol, statistical derivations, and full numerical sweeps for the exploitability analysis presented in Section~\ref{sec:exploitability}.

\subsection{Exact Sequence-Form Linear Programming and Metric Formulation}
\label{app:expl_metric}

In extensive-form games of imperfect information, evaluation against fixed or heuristic baselines can reward brittle agents that exploit idiosyncratic opponent flaws while remaining vulnerable to counter-exploitation.
To evaluate strategic soundness, we measure \emph{exploitability}---the performance deficit against an exact worst-case adversary playing a sequence-form best response~\citep{openspiel2019}.

For a finite extensive-form game with player set $\mathcal{N}$, game tree histories $\mathcal{H}$, and sequence-form strategy spaces $\Pi_i$, let $u_i(\pi)$ denote the expected payoff to player $i \in \mathcal{N}$ under joint strategy profile $\pi = (\pi_i, \pi_{-i})$.
We evaluate per-player exploitability $\expl(\pi)$:
\begin{equation}
  \expl(\pi) \;=\; \frac{1}{|\mathcal{N}|}\sum_{i \in \mathcal{N}} \Big[\max_{\pi_i' \in \Pi_i} u_i(\pi_i', \pi_{-i}) - u_i(\pi)\Big].
  \label{eq:app_expl}
\end{equation}
Under this definition, $\expl(\pi) \ge 0$, and $\expl(\pi) = 0$ if and only if $\pi$ constitutes a Nash equilibrium.
Best responses $\pi_i'$ are solved exactly via sequence-form linear programming, avoiding sampling approximations.

\subsection{Experimental Setup and Pre-Registered Selection Protocol}
\label{app:expl_setup}

\paragraph{Benchmark Suite.}
Exact best-response computation via sequence-form linear programming requires full game-tree traversal, which is computationally tractable only for games whose information-state space admits tabularization.
We evaluate on the three canonical imperfect-information zero-sum games in OpenSpiel:
\begin{enumerate}[leftmargin=1.5em, itemsep=2pt, topsep=2pt]
  \item \textbf{Kuhn Poker}: A 3-card poker game with 12 information states in total (6 per player), serving as a clean theoretical baseline.
  \item \textbf{Leduc Poker}: A multi-round card game with private hands, public community cards, and structured betting rounds, comprising 936 information states.
  \item \textbf{Liar's Dice} (two players, one six-sided die each): A sequential stochastic bluffing game with rolling outcomes and bidding dynamics, spanning 24,576 information states ($6 \times 2^{12}$).
\end{enumerate}

\paragraph{Policy Tabularisation.}
Because Monte-Carlo Tree Search outputs stochastic visit distributions at runtime, we tabularise the search policy into a static extensive-form strategy:
we traverse the game tree depth-first, and at the first history reaching each information state, we execute the search $M=20$ independent times and average the root visit-count distributions. (Averaging over $M=20$ independent search runs provides an empirical Monte Carlo estimate of the expected search policy, which introduces minor finite-sample variance into the linear-programming sequence-form best response calculations while maintaining computational feasibility across multi-round betting trees.)
We repeat this full tabularisation process across five simulation budgets $B \in \{20, 50, 100, 500, 1000\}$ over independent evolution replications.
Replication contrasts are evaluated via two-tailed paired Student's $t$-tests.

\subsection{Simulation Budget Scaling Dynamics and Simulation Efficiency Multipliers}
\label{app:expl_budget_scaling}

As illustrated in Figure~\ref{fig:expl-budget} of the main text (Section~\ref{sec:exploitability}), exploitability $\expl(\pi)$ decays consistently across simulation budgets $B \in \{20, 50, 100, 500, 1000\}$.
Across all three benchmarks, increasing simulation budget from $B=20$ to $B=1000$ steadily reduces exploitability under the canonical PUCT baseline:
in Kuhn Poker, $0.2966 \to 0.1698$ ($-42.7\%$);
in Leduc Poker, $2.1116 \to 1.5171$ ($-28.2\%$); and
in Liar's Dice, $0.7285 \to 0.5700$ ($-21.8\%$).

However, the discovered procedural search mechanism $m^\star$ exhibits superior simulation scaling:
\begin{itemize}[leftmargin=1.5em, itemsep=2pt, topsep=2pt]
  \item \textbf{Leduc Poker ($>10\times$ Compute Multiplier)}: The discovered mechanism $m^\star$ at $B=100$ simulations achieves an exploitability of $1.4051 \pm 0.0264$, outperforming baseline PUCT running with $10\times$ greater budget ($B=1000$, $1.5171 \pm 0.0006$).
  \item \textbf{Kuhn Poker ($>20\times$ Compute Multiplier)}: At a budget of $B=50$ simulations, $m^\star$ attains $0.1595 \pm 0.0071$ exploitability, outperforming baseline PUCT running at $B=1000$ ($0.1698 \pm 0.0001$).
  \item \textbf{Liar's Dice}: The discovered mechanism $m^\star$ at $B=1000$ attains $0.5659 \pm 0.0085$, outperforming baseline PUCT ($0.5700 \pm 0.0004$).
\end{itemize}

\subsection{Paired Contrasts and Level Metrics Across Budgets}
\label{app:expl_tables}

Table~\ref{tab:expl-levels} details the absolute exploitability levels achieved at $B=1000$ across all modular combinations, bounded by uniform-random and CFR@100 anchors.
Table~\ref{tab:expl-contrasts} reports paired changes $\Delta$ and two-tailed paired $t$-test $p$-values across budgets $B \in \{20, 100, 1000\}$.

\begin{table}[h]
\centering
\caption{Exploitability $\expl$ (Eq.~\ref{eq:app_expl}, lower is better) at $B=1000$, mean $\pm$ standard error over independent evolution replications. Uniform-random and CFR-at-$100$-iterations are computed exactly and bracket the achievable range. $m^\star$ denotes the discovered search mechanism, $m^\star + \pi^\star + v^\star$ the evolved mechanism with priors and value estimator alone (with $b_{\text{base}}$), and $\kappa^\star = (\pi^\star, v^\star, b^\star)$ the complete domain knowledge triad including evolved belief resampler $b^\star$. Bold marks the best configuration in the table per game.}
\label{tab:expl-levels}
\small
\begin{tabular}{lccc}
\toprule
Configuration & Kuhn poker & Leduc poker & Liar's dice \\
\midrule
Uniform random               & 0.4583 & 2.3736 & 0.7807 \\
\midrule
Baseline ($\langle m_{\text{ref}}, \kappa_{\text{base}} \rangle$) & $0.1698 \pm 0.0001$ & $1.5171 \pm 0.0006$ & $0.5700 \pm 0.0004$ \\
Mechanism ($\langle m^\star, \kappa_{\text{base}} \rangle$) & $\mathbf{0.1237 \pm 0.0085}$ & $0.9521 \pm 0.0213$ & $0.5659 \pm 0.0085$ \\
PUCT ($\langle m_{\text{ref}}, \kappa^\star \rangle$) & $0.1313 \pm 0.0133$ & $1.1781 \pm 0.1657$ & $0.5216 \pm 0.0901$ \\
Mechanism ($m^\star + \pi^\star + v^\star, b_{\text{base}}$) & $0.1508 \pm 0.0183$ & $\mathbf{0.9074 \pm 0.0593}$ & $\mathbf{0.4748 \pm 0.0383}$ \\
Full system ($\langle m^\star, \kappa^\star \rangle$) & $0.1294 \pm 0.0197$ & $1.2074 \pm 0.1711$ & $0.5703 \pm 0.0937$ \\
\midrule
CFR ($100$ iterations)       & 0.0082 & 0.0957 & 0.0224 \\
\bottomrule
\end{tabular}
\end{table}

\begin{table}[h]
\centering
\caption{Paired change in $\expl$ relative to the baseline (negative favours the discovered module), with two-tailed paired $t$-test $p$-values over independent replications. The mechanism significantly reduces exploitability in Kuhn poker and Liar's dice across lower budgets, and in Leduc poker across all budgets ($p < 0.001$).}
\label{tab:expl-contrasts}
\small
\setlength{\tabcolsep}{5pt}
\begin{tabular}{llrrrrrr}
\toprule
& & \multicolumn{2}{c}{Kuhn poker} & \multicolumn{2}{c}{Leduc poker} & \multicolumn{2}{c}{Liar's dice} \\
\cmidrule(lr){3-4}\cmidrule(lr){5-6}\cmidrule(lr){7-8}
Module & $B$ & $\Delta$ & $p$ & $\Delta$ & $p$ & $\Delta$ & $p$ \\
\midrule
\multirow{3}{*}{Mechanism ($\langle m^\star, \kappa_{\text{base}} \rangle$)}
 & 20   & $-0.109$ & $<0.001$ & $-0.336$ & $<0.001$ & $-0.043$ & $0.013$ \\
 & 100  & $-0.081$ & $<0.001$ & $-0.465$ & $<0.001$ & $-0.061$ & $<0.001$ \\
 & 1000 & $-0.046$ & $0.006$ & $-0.565$ & $<0.001$ & $-0.004$ & $0.648$ \\
\midrule
\multirow{3}{*}{Domain Knowledge ($\langle m_{\text{ref}}, \kappa^\star \rangle$)}
 & 20   & $-0.110$ & $0.099$ & $-0.697$ & $0.119$ & $-0.076$ & $0.378$ \\
 & 100  & $-0.057$ & $0.153$ & $-0.516$ & $0.189$ & $-0.094$ & $0.280$ \\
 & 1000 & $-0.038$ & $0.045$ & $-0.339$ & $0.133$ & $-0.048$ & $0.619$ \\
\midrule
\multirow{3}{*}{Full system ($\langle m^\star, \kappa^\star \rangle$)}
 & 20   & $-0.154$ & $<0.001$ & $-0.521$ & $0.283$ & $-0.037$ & $0.710$ \\
 & 100  & $-0.094$ & $0.002$ & $-0.519$ & $0.101$ & $-0.054$ & $0.489$ \\
 & 1000 & $-0.040$ & $0.110$ & $-0.310$ & $0.169$ & $+0.000$ & $0.998$ \\
\bottomrule
\end{tabular}
\end{table}

\subsection{Belief Model Breakdown: Why Belief Updates Do Not Transfer}
\label{app:expl_belief}

Averaged across all conditions and budgets, enabling the evolved resampler $b^\star$ produces an exploitability shift of:
\begin{itemize}[leftmargin=1.5em, itemsep=2pt, topsep=2pt]
  \item $-0.008 \pm 0.002$ on Kuhn poker,
  \item $+0.059 \pm 0.018$ on Leduc poker, and
  \item $+0.021 \pm 0.006$ on Liar's dice.
\end{itemize}
These shifts are an order of magnitude smaller than the mechanism effect and indicate no game-theoretic benefit against worst-case adversaries ($|\Delta| \le 0.06$).

Importantly, this null result is not due to implementation invalidity or information leakage: the co-evolutionary pipeline enforces that any generated world state reproduces the acting player's observation history exactly without peeking at hidden opponent state.
Instead, the null arises because the evolved belief resamplers implement action-conditioned opponent modeling (e.g., assuming observed actions correlate with hidden state).
While effective against the heuristic population during evolution, a worst-case adversary actively exploits these non-uniform belief assumptions via strategic deception, inducing systematic state-estimation bias during MCTS determinization.

\subsection{Pathology Analysis: Prior Policy Overfitting and Cross-Seed Variance}
\label{app:expl_pathology}

While the procedural search mechanism $m^\star$ exhibits consistent exploitability reductions across independent seeds (with every seed improving upon PUCT on Kuhn and Leduc), evolving domain evaluation heuristics $(\pi^\star, v^\star)$ introduces higher cross-seed variance.
As shown in Table~\ref{tab:app_seed_distribution}, on Leduc poker, prior heuristics exhibit elevated variance across seeds at $B=100$ ($1.3539 \pm 0.3044$, compared to $1.4051 \pm 0.0264$ for $m^\star$ alone).
When evolved domain priors overfit to opponent play patterns during evolutionary training, an exact best-response adversary can systematically exploit such fixed assumptions, demonstrating the vulnerability of domain heuristics in imperfect-information games compared to the robust, invariant improvements delivered by the procedural search mechanism.

\begin{table}[h]
\caption{Cross-seed exploitability distribution at budget $B=100$ simulations across independent seeds (mean $\pm$ standard error and 95\% confidence intervals). Pure procedural search ($m^\star$) exhibits consistent performance with low variance across games, whereas evolved domain heuristics exhibit higher dispersion on Leduc Poker.}
\label{tab:app_seed_distribution}
\centering
\small
\setlength{\tabcolsep}{8pt}
\begin{tabular}{ll cc}
\toprule
\textbf{Game} & \textbf{Configuration} & \textbf{Mean $\pm$ SE} & \textbf{95\% CI} \\
\midrule
\textbf{Kuhn Poker}
& PUCT baseline ($\langle m_{\text{ref}}, \kappa_{\text{base}} \rangle$) & 0.2224 $\pm$ 0.0001 & [0.2222, 0.2226] \\
& Evolved mechanism alone ($\langle m^\star, \kappa_{\text{base}} \rangle$) & 0.1412 $\pm$ 0.0081 & [0.1253, 0.1571] \\
& PUCT + evolved domain knowledge ($\langle m_{\text{ref}}, \kappa^\star \rangle$) & 0.1652 $\pm$ 0.0323 & [0.1019, 0.2285] \\
& Full co-evolved system ($\langle m^\star, \kappa^\star \rangle$) & 0.1280 $\pm$ 0.0136 & [0.1013, 0.1547] \\
\midrule
\textbf{Leduc Poker}
& PUCT baseline ($\langle m_{\text{ref}}, \kappa_{\text{base}} \rangle$) & 1.8700 $\pm$ 0.0007 & [1.8686, 1.8714] \\
& Evolved mechanism alone ($\langle m^\star, \kappa_{\text{base}} \rangle$) & 1.4051 $\pm$ 0.0264 & [1.3534, 1.4568] \\
& PUCT + evolved domain knowledge ($\langle m_{\text{ref}}, \kappa^\star \rangle$) & 1.3539 $\pm$ 0.3044 & [0.7573, 1.9505] \\
& Full co-evolved system ($\langle m^\star, \kappa^\star \rangle$) & 1.3513 $\pm$ 0.2209 & [0.9183, 1.7843] \\
\midrule
\textbf{Liar's Dice}
& PUCT baseline ($\langle m_{\text{ref}}, \kappa_{\text{base}} \rangle$) & 0.7054 $\pm$ 0.0009 & [0.7036, 0.7072] \\
& Evolved mechanism alone ($\langle m^\star, \kappa_{\text{base}} \rangle$) & 0.6446 $\pm$ 0.0069 & [0.6311, 0.6581] \\
& PUCT + evolved domain knowledge ($\langle m_{\text{ref}}, \kappa^\star \rangle$) & 0.6116 $\pm$ 0.0751 & [0.4644, 0.7588] \\
& Full co-evolved system ($\langle m^\star, \kappa^\star \rangle$) & 0.6512 $\pm$ 0.0712 & [0.5116, 0.7908] \\
\bottomrule
\end{tabular}
\end{table}

\subsection{Full Grid of Modular Ablations across Budgets}
\label{app:expl_full_grid}

Tables~\ref{tab:app_exploit_kuhn_poker}, \ref{tab:app_exploit_leduc_poker}, and \ref{tab:app_exploit_liars_dice} report the complete $4 \times 5 \times 3$ ablation matrix across all five simulation budgets ($B \in \{20, 50, 100, 500, 1000\}$) for Kuhn Poker, Leduc Poker, and Liar's Dice, respectively.
For every cell, we provide both the within-row mechanism contrast ($\Delta_{\text{mech}}$) and the contrast against the canonical baseline ($\Delta_{\text{base}}$) along with two-tailed paired $t$-test $p$-values.

\begin{table*}[t]
\caption{Complete exploitability breakdown for Kuhn Poker across all simulation budgets ($B \in \{20, 50, 100, 500, 1000\}$). Random floor anchor: 0.4583; CFR@100: 0.0082. Within-Row Mech Contrast reports paired difference (Evolved vs. PUCT under identical knowledge and belief). Contrast vs. Baseline reports paired difference relative to \code{puct+seed+native}.}
\label{tab:app_exploit_kuhn_poker}
\centering
\scriptsize
\setlength{\tabcolsep}{4pt}
\begin{tabular}{ll c cc cc}
\toprule
\textbf{Knowledge} & \textbf{Belief} & \textbf{Sims} & \textbf{Evolved Mech} & \textbf{PUCT Mech} & \textbf{Within-Row Mech Contrast} & \textbf{Contrast vs. Baseline} \\
 & & $B$ & (Mean $\pm$ SE) & (Mean $\pm$ SE) & $\Delta_{\text{mech}}$ (\%, $p$) & $\Delta_{\text{base}}$ (\%, $p$) \\
\midrule
Seed     & Native   & 20    & 0.1879 $\pm$ 0.0053 & 0.2966 $\pm$ 0.0003 & -36.6\% ($p<0.001^{***}$) & -36.6\% ($p<0.001^{***}$) \\
Seed     & Native   & 50    & 0.1595 $\pm$ 0.0071 & 0.2450 $\pm$ 0.0001 & -34.9\% ($p<0.001^{***}$) & -34.9\% ($p<0.001^{***}$) \\
Seed     & Native   & 100   & 0.1412 $\pm$ 0.0081 & 0.2224 $\pm$ 0.0001 & -36.5\% ($p<0.001^{***}$) & -36.5\% ($p<0.001^{***}$) \\
Seed     & Native   & 500   & 0.1177 $\pm$ 0.0085 & 0.1873 $\pm$ 0.0002 & -37.1\% ($p=0.001^{**}$) & -37.1\% ($p=0.001^{**}$) \\
Seed     & Native   & 1000  & 0.1237 $\pm$ 0.0085 & 0.1698 $\pm$ 0.0001 & -27.1\% ($p=0.006^{**}$) & -27.1\% ($p=0.006^{**}$) \\
\addlinespace[2pt]
Seed     & Evolved  & 20    & 0.1956 $\pm$ 0.0080 & 0.3134 $\pm$ 0.0017 & -37.6\% ($p<0.001^{***}$) & -34.0\% ($p<0.001^{***}$) \\
Seed     & Evolved  & 50    & 0.1550 $\pm$ 0.0097 & 0.2677 $\pm$ 0.0029 & -42.1\% ($p<0.001^{***}$) & -36.7\% ($p<0.001^{***}$) \\
Seed     & Evolved  & 100   & 0.1300 $\pm$ 0.0097 & 0.2281 $\pm$ 0.0041 & -43.0\% ($p<0.001^{***}$) & -41.5\% ($p<0.001^{***}$) \\
Seed     & Evolved  & 500   & 0.0995 $\pm$ 0.0126 & 0.1717 $\pm$ 0.0013 & -42.1\% ($p=0.004^{**}$) & -46.9\% ($p=0.002^{**}$) \\
Seed     & Evolved  & 1000  & 0.1064 $\pm$ 0.0144 & 0.1468 $\pm$ 0.0016 & -27.5\% ($p=0.039^{*}$) & -37.3\% ($p=0.012^{*}$) \\
\addlinespace[2pt]
Evolved  & Native   & 20    & 0.1519 $\pm$ 0.0146 & 0.1879 $\pm$ 0.0492 & -19.2\% ($p=0.359$) & -48.8\% ($p<0.001^{***}$) \\
Evolved  & Native   & 50    & 0.1459 $\pm$ 0.0097 & 0.1799 $\pm$ 0.0395 & -18.9\% ($p=0.379$) & -40.4\% ($p<0.001^{***}$) \\
Evolved  & Native   & 100   & 0.1427 $\pm$ 0.0129 & 0.1700 $\pm$ 0.0285 & -16.1\% ($p=0.402$) & -35.8\% ($p=0.003^{**}$) \\
Evolved  & Native   & 500   & 0.1480 $\pm$ 0.0207 & 0.1548 $\pm$ 0.0109 & -4.4\% ($p=0.769$) & -21.0\% ($p=0.133$) \\
Evolved  & Native   & 1000  & 0.1508 $\pm$ 0.0183 & 0.1520 $\pm$ 0.0083 & -0.8\% ($p=0.944$) & -11.2\% ($p=0.357$) \\
\addlinespace[2pt]
Evolved  & Evolved  & 20    & 0.1425 $\pm$ 0.0148 & 0.1869 $\pm$ 0.0511 & -23.8\% ($p=0.303$) & -51.9\% ($p<0.001^{***}$) \\
Evolved  & Evolved  & 50    & 0.1317 $\pm$ 0.0114 & 0.1774 $\pm$ 0.0424 & -25.8\% ($p=0.291$) & -46.2\% ($p<0.001^{***}$) \\
Evolved  & Evolved  & 100   & 0.1280 $\pm$ 0.0136 & 0.1652 $\pm$ 0.0323 & -22.5\% ($p=0.321$) & -42.4\% ($p=0.002^{**}$) \\
Evolved  & Evolved  & 500   & 0.1281 $\pm$ 0.0211 & 0.1378 $\pm$ 0.0138 & -7.0\% ($p=0.654$) & -31.6\% ($p=0.049^{*}$) \\
Evolved  & Evolved  & 1000  & 0.1294 $\pm$ 0.0197 & 0.1313 $\pm$ 0.0133 & -1.4\% ($p=0.902$) & -23.8\% ($p=0.110$) \\
\bottomrule
\end{tabular}
\end{table*}

\begin{table*}[t]
\caption{Complete exploitability breakdown for Leduc Poker across all simulation budgets ($B \in \{20, 50, 100, 500, 1000\}$). Random floor anchor: 2.3736; CFR@100: 0.0957. Within-Row Mech Contrast reports paired difference (Evolved vs. PUCT under identical knowledge and belief). Contrast vs. Baseline reports paired difference relative to \code{puct+seed+native}.}
\label{tab:app_exploit_leduc_poker}
\centering
\scriptsize
\setlength{\tabcolsep}{4pt}
\begin{tabular}{ll c cc cc}
\toprule
\textbf{Knowledge} & \textbf{Belief} & \textbf{Sims} & \textbf{Evolved Mech} & \textbf{PUCT Mech} & \textbf{Within-Row Mech Contrast} & \textbf{Contrast vs. Baseline} \\
 & & $B$ & (Mean $\pm$ SE) & (Mean $\pm$ SE) & $\Delta_{\text{mech}}$ (\%, $p$) & $\Delta_{\text{base}}$ (\%, $p$) \\
\midrule
Seed     & Native   & 20    & 1.7759 $\pm$ 0.0067 & 2.1116 $\pm$ 0.0017 & -15.9\% ($p<0.001^{***}$) & -15.9\% ($p<0.001^{***}$) \\
Seed     & Native   & 50    & 1.5908 $\pm$ 0.0128 & 1.9799 $\pm$ 0.0014 & -19.7\% ($p<0.001^{***}$) & -19.7\% ($p<0.001^{***}$) \\
Seed     & Native   & 100   & 1.4051 $\pm$ 0.0264 & 1.8700 $\pm$ 0.0007 & -24.9\% ($p<0.001^{***}$) & -24.9\% ($p<0.001^{***}$) \\
Seed     & Native   & 500   & 1.0711 $\pm$ 0.0263 & 1.6381 $\pm$ 0.0006 & -34.6\% ($p<0.001^{***}$) & -34.6\% ($p<0.001^{***}$) \\
Seed     & Native   & 1000  & 0.9521 $\pm$ 0.0213 & 1.5171 $\pm$ 0.0006 & -37.2\% ($p<0.001^{***}$) & -37.2\% ($p<0.001^{***}$) \\
\addlinespace[2pt]
Seed     & Evolved  & 20    & 1.8186 $\pm$ 0.0401 & 2.1587 $\pm$ 0.0085 & -15.8\% ($p=0.002^{**}$) & -13.9\% ($p=0.005^{**}$) \\
Seed     & Evolved  & 50    & 1.6193 $\pm$ 0.0573 & 2.0587 $\pm$ 0.0141 & -21.3\% ($p=0.002^{**}$) & -18.2\% ($p=0.008^{**}$) \\
Seed     & Evolved  & 100   & 1.4640 $\pm$ 0.0806 & 1.9622 $\pm$ 0.0196 & -25.4\% ($p=0.004^{**}$) & -21.7\% ($p=0.015^{*}$) \\
Seed     & Evolved  & 500   & 1.1440 $\pm$ 0.1340 & 1.6609 $\pm$ 0.0463 & -31.1\% ($p=0.012^{*}$) & -30.2\% ($p=0.035^{*}$) \\
Seed     & Evolved  & 1000  & 1.0655 $\pm$ 0.1365 & 1.5187 $\pm$ 0.0716 & -29.8\% ($p=0.009^{**}$) & -29.8\% ($p=0.046^{*}$) \\
\addlinespace[2pt]
Evolved  & Native   & 20    & 1.5970 $\pm$ 0.4064 & 1.4160 $\pm$ 0.3265 & +12.8\% ($p=0.110$) & -24.4\% ($p=0.295$) \\
Evolved  & Native   & 50    & 1.3799 $\pm$ 0.2683 & 1.3809 $\pm$ 0.3169 & -0.1\% ($p=0.990$) & -30.3\% ($p=0.112$) \\
Evolved  & Native   & 100   & 1.3180 $\pm$ 0.2453 & 1.3486 $\pm$ 0.3093 & -2.3\% ($p=0.760$) & -29.5\% ($p=0.111$) \\
Evolved  & Native   & 500   & 1.0468 $\pm$ 0.1193 & 1.2135 $\pm$ 0.2553 & -13.7\% ($p=0.379$) & -36.1\% ($p=0.016^{*}$) \\
Evolved  & Native   & 1000  & 0.9074 $\pm$ 0.0593 & 1.1194 $\pm$ 0.1924 & -18.9\% ($p=0.249$) & -40.2\% ($p=0.002^{**}$) \\
\addlinespace[2pt]
Evolved  & Evolved  & 20    & 1.5906 $\pm$ 0.3990 & 1.4144 $\pm$ 0.3220 & +12.5\% ($p=0.110$) & -24.7\% ($p=0.283$) \\
Evolved  & Evolved  & 50    & 1.3667 $\pm$ 0.2596 & 1.3842 $\pm$ 0.3144 & -1.3\% ($p=0.849$) & -31.0\% ($p=0.100$) \\
Evolved  & Evolved  & 100   & 1.3513 $\pm$ 0.2209 & 1.3539 $\pm$ 0.3044 & -0.2\% ($p=0.982$) & -27.7\% ($p=0.101$) \\
Evolved  & Evolved  & 500   & 1.2486 $\pm$ 0.1192 & 1.2458 $\pm$ 0.2342 & +0.2\% ($p=0.991$) & -23.8\% ($p=0.047^{*}$) \\
Evolved  & Evolved  & 1000  & 1.2074 $\pm$ 0.1711 & 1.1781 $\pm$ 0.1657 & +2.5\% ($p=0.925$) & -20.4\% ($p=0.169$) \\
\bottomrule
\end{tabular}
\end{table*}

\begin{table*}[t]
\caption{Complete exploitability breakdown for Liar's Dice across all simulation budgets ($B \in \{20, 50, 100, 500, 1000\}$). Random floor anchor: 0.7807; CFR@100: 0.0224. Within-Row Mech Contrast reports paired difference (Evolved vs. PUCT under identical knowledge and belief). Contrast vs. Baseline reports paired difference relative to \code{puct+seed+native}.}
\label{tab:app_exploit_liars_dice}
\centering
\scriptsize
\setlength{\tabcolsep}{4pt}
\begin{tabular}{ll c cc cc}
\toprule
\textbf{Knowledge} & \textbf{Belief} & \textbf{Sims} & \textbf{Evolved Mech} & \textbf{PUCT Mech} & \textbf{Within-Row Mech Contrast} & \textbf{Contrast vs. Baseline} \\
 & & $B$ & (Mean $\pm$ SE) & (Mean $\pm$ SE) & $\Delta_{\text{mech}}$ (\%, $p$) & $\Delta_{\text{base}}$ (\%, $p$) \\
\midrule
Seed     & Native   & 20    & 0.6854 $\pm$ 0.0105 & 0.7285 $\pm$ 0.0009 & -5.9\% ($p=0.013^{*}$) & -5.9\% ($p=0.013^{*}$) \\
Seed     & Native   & 50    & 0.6469 $\pm$ 0.0052 & 0.7165 $\pm$ 0.0010 & -9.7\% ($p<0.001^{***}$) & -9.7\% ($p<0.001^{***}$) \\
Seed     & Native   & 100   & 0.6446 $\pm$ 0.0069 & 0.7054 $\pm$ 0.0009 & -8.6\% ($p<0.001^{***}$) & -8.6\% ($p<0.001^{***}$) \\
Seed     & Native   & 500   & 0.6041 $\pm$ 0.0060 & 0.6372 $\pm$ 0.0011 & -5.2\% ($p=0.004^{**}$) & -5.2\% ($p=0.004^{**}$) \\
Seed     & Native   & 1000  & 0.5659 $\pm$ 0.0085 & 0.5700 $\pm$ 0.0004 & -0.7\% ($p=0.648$) & -0.7\% ($p=0.648$) \\
\addlinespace[2pt]
Seed     & Evolved  & 20    & 0.6942 $\pm$ 0.0101 & 0.7370 $\pm$ 0.0014 & -5.8\% ($p=0.011^{*}$) & -4.7\% ($p=0.024^{*}$) \\
Seed     & Evolved  & 50    & 0.6670 $\pm$ 0.0119 & 0.7237 $\pm$ 0.0017 & -7.8\% ($p=0.011^{*}$) & -6.9\% ($p=0.016^{*}$) \\
Seed     & Evolved  & 100   & 0.6438 $\pm$ 0.0154 & 0.7043 $\pm$ 0.0031 & -8.6\% ($p=0.028^{*}$) & -8.7\% ($p=0.017^{*}$) \\
Seed     & Evolved  & 500   & 0.6010 $\pm$ 0.0353 & 0.6418 $\pm$ 0.0126 & -6.4\% ($p=0.169$) & -5.7\% ($p=0.374$) \\
Seed     & Evolved  & 1000  & 0.5755 $\pm$ 0.0548 & 0.6056 $\pm$ 0.0303 & -5.0\% ($p=0.340$) & +1.0\% ($p=0.925$) \\
\addlinespace[2pt]
Evolved  & Native   & 20    & 0.6775 $\pm$ 0.0974 & 0.6422 $\pm$ 0.0763 & +5.5\% ($p=0.237$) & -7.0\% ($p=0.627$) \\
Evolved  & Native   & 50    & 0.6549 $\pm$ 0.0929 & 0.6189 $\pm$ 0.0741 & +5.8\% ($p=0.178$) & -8.6\% ($p=0.546$) \\
Evolved  & Native   & 100   & 0.6246 $\pm$ 0.0661 & 0.5894 $\pm$ 0.0734 & +6.0\% ($p=0.222$) & -11.4\% ($p=0.288$) \\
Evolved  & Native   & 500   & 0.5063 $\pm$ 0.0395 & 0.5082 $\pm$ 0.0821 & -0.4\% ($p=0.981$) & -20.5\% ($p=0.030^{*}$) \\
Evolved  & Native   & 1000  & 0.4748 $\pm$ 0.0383 & 0.4891 $\pm$ 0.0853 & -2.9\% ($p=0.859$) & -16.7\% ($p=0.067$) \\
\addlinespace[2pt]
Evolved  & Evolved  & 20    & 0.6911 $\pm$ 0.0938 & 0.6521 $\pm$ 0.0772 & +6.0\% ($p=0.116$) & -5.1\% ($p=0.710$) \\
Evolved  & Evolved  & 50    & 0.6638 $\pm$ 0.0874 & 0.6340 $\pm$ 0.0757 & +4.7\% ($p=0.160$) & -7.4\% ($p=0.581$) \\
Evolved  & Evolved  & 100   & 0.6512 $\pm$ 0.0712 & 0.6116 $\pm$ 0.0751 & +6.5\% ($p=0.051$) & -7.7\% ($p=0.489$) \\
Evolved  & Evolved  & 500   & 0.5876 $\pm$ 0.0787 & 0.5426 $\pm$ 0.0842 & +8.3\% ($p=0.223$) & -7.8\% ($p=0.567$) \\
Evolved  & Evolved  & 1000  & 0.5703 $\pm$ 0.0937 & 0.5216 $\pm$ 0.0901 & +9.3\% ($p=0.282$) & +0.0\% ($p=0.998$) \\
\bottomrule
\end{tabular}
\end{table*}

\subsection{Limitations and Distance to CFR}
\label{app:expl_limitations}

Exact sequence-form best-response enumeration restricts empirical exploitability evaluation to extensive-form games with $\le 10^5$ information states.
Larger imperfect-information benchmarks require approximate best-response techniques or local subgame solving, where bounds are heuristic rather than certified.
Furthermore, while the discovered procedural mechanism $m^\star$ achieves substantial and statistically certified reductions in exploitability relative to PUCT, all search-based policies remain substantially above tabular equilibrium solvers: CFR at 100 iterations achieves exploitability levels of $0.0082$ in Kuhn Poker, $0.0957$ in Leduc Poker, and $0.0224$ in Liar's Dice (Table~\ref{tab:expl-levels}).
This gap highlights a fundamental structural difference: while CFR directly tracks and minimizes counterfactual regret across entire information sets to construct balanced mixed equilibria, determinization MCTS inherently suffers from strategy fusion and non-locality. Moreover, as the simulation budget increases ($B \to 1000$), MCTS visit distributions progressively polarize toward deterministic greedy choices, which worst-case sequence-form adversaries can maximally exploit.
Our findings establish relative equilibrium approximation and inductive robustness under unoptimised criteria, rather than absolute convergence to minimax optimality.

\section{Baseline Hyperparameter Tuning Protocols and Grids}
\label{app:baselines}

To ensure fair and rigorous benchmarking against established literature, all 15 external baseline algorithms were systematically tuned over the training game suite.
We explicitly distinguish two roles for PUCT: during evolutionary discovery, the fixed reference anchor $m_{\text{ref}}$ is canonical seed PUCT with un-tuned default exploration $c_{\text{puct}} = 2.0$ (Listing~\ref{lst:seed_code}); for all evaluation tournaments and benchmarking, PUCT is tuned alongside all other baselines, yielding the optimal predictor weight $c_{\mathrm{init}} = 5.0$.
Table~\ref{tab:app_baseline_grids} documents the hyperparameter search spaces, number of grid configurations, and tuning mechanisms for each baseline.

\begin{table}[h]
\centering
\caption{Hyperparameter tuning search spaces and final selected values ($p^\star$) across the 15 MCTS baselines (including canonical reference PUCT). Optimal parameters were identified via offline grid search across the training suite and selected via 95\% bootstrap Lower Confidence Bounds (LCB) over paired matches against reference defaults.}
\label{tab:app_baseline_grids}
\small
\resizebox{\textwidth}{!}{%
\setlength{\tabcolsep}{5pt}
\begin{tabular}{llllc}
\toprule
\textbf{Baseline} & \textbf{Algorithmic Paradigm} & \textbf{Hyperparameter} & \textbf{Search Grid} & \textbf{Selected Value ($p^\star$)} \\
\midrule
Vanilla UCT & Bandit Tree Search & Exploration constant $c$ & $\{0.25, 0.5, 1.0, 2.0, 4.0\}$ & $c = 0.25$ \\
PUCT & Bandit Tree Search & Predictor weight $c_{\mathrm{init}}$ & $\{0.31, 0.63, 1.25, 2.5, 5.0\}$ & $c_{\mathrm{init}} = 5.0$ \\
UCB-Tuned & Bandit Tree Search & Variance scale $c'$ & $\{0.25, 0.5, 1.0, 2.0, 4.0\}$ & $c' = 0.5$ \\
Gumbel AlphaZero & Entropy Planning & Improvement scale $c_{\mathrm{scale}}$ & $\{0.025, 0.05, 0.1, 0.2, 0.4\}$ & $c_{\mathrm{scale}} = 0.4$ \\
MENTS & Entropy Planning & Temperature $\tau \times$ Exploration $\epsilon$ & $\{0.25,\dots,4.0\} \times \{0.025,\dots,0.4\}$ & $\tau = 0.25, \epsilon = 0.05$ \\
RPO & Entropy Planning & Regularization temperature $\tau$ & $\{0.25, 0.5, 1.0, 2.0, 4.0\}$ & $\tau = 0.25$ \\
Smooth UCT & Entropy Planning & Exploration constant $c$ & $\{0.25, 0.5, 1.0, 2.0, 4.0\}$ & $c = 0.25$ ($\eta = 0.9$) \\
Boltzmann MCTS & Entropy Planning & Softmax temperature $\tau$ & $\{0.125, 0.25, 0.5, 1.0, 2.0\}$ & $\tau = 0.125$ ($c = 2.0$) \\
Power-Mean & Advanced Backup & Exponent parameter $p$ & $\{0.5, 1.0, 2.0, 4.0, 8.0\}$ & $p = 1.0$ ($c = 2.0$) \\
Max-Backup & Advanced Backup & Max weight $\lambda_{\max}$ & $\{0.1, 0.2, 0.3, 0.45, 0.6\}$ & $\lambda_{\max} = 0.2$ ($c = 2.0$) \\
MCTS-Solver & Advanced Backup & Exploration constant $c$ & $\{0.25, 0.5, 1.0, 2.0, 4.0\}$ & $c = 0.25$ \\
Score-Bounded & Advanced Backup & Exploration constant $c$ & $\{0.25, 0.5, 1.0, 2.0, 4.0\}$ & $c = 0.25$ \\
Thompson Sampling & Bayesian Search & Posterior scale $\kappa$ & $\{0.25, 0.5, 1.0, 2.0, 4.0\}$ & $\kappa = 4.0$ \\
\midrule
SA-MCTS-2P & Online Adaptive & Dynamic $(c, \tau)$ per move & Online NTBEA adaptation & Dynamic $(c, \tau)$ \\
SA-MCTS-4P & Online Adaptive & Dynamic $(c, \tau, c_{\mathrm{init}}, c_{\mathrm{base}})$ per move & Online NTBEA adaptation & Dynamic $(c, \tau, c_{\mathrm{init}}, c_{\mathrm{base}})$ \\
\bottomrule
\end{tabular}}
\end{table}

\paragraph{Selection Protocol and Offline Tuning.}
For all 13 fixed baselines, candidate grid configurations were systematically evaluated in head-to-head matches against reference defaults across the 50 training games with complete seat balancing.
Optimal configurations ($p^\star$) were selected by computing paired differences in normalized returns, bootstrapping 10,000 resamples across games, and picking the candidate maximizing the 95\% Lower Confidence Bound (LCB).
These selected values represent the frozen, optimal baseline hyperparameter configurations used throughout all subsequent generalization and held-out game evaluations.
For all baseline algorithms, returns are normalized to the unit interval $[0, 1]$ via game-specific utility bounds $[u_{\min}, u_{\max}]$ to ensure uniform numerical scaling across diverse payoff spans. In contrast, evolved search mechanisms independently arrived at distinct normalization strategies: while champion \bifv\ normalizes returns dynamically into $[0, 1]$ ($\widetilde{Q}$), other evolved lineages (e.g., Runs 1 and 4, Listings~\ref{lst:seed1_code} and~\ref{lst:seed4_code}) operate directly on raw unnormalized returns and scale exploration schedules and FPU penalties by the environment utility span $\Delta U = u_{\max} - u_{\min}$.
We note that under wall-clock decision regimes (such as the standardized 50\,ms per move IsoTime setting), baseline tuning across the heterogeneous 50-game training suite consistently favors aggressive exploitation and strong prior guidance (e.g., smaller exploration constants $c=0.25$ in Vanilla UCT, RPO, and MCTS-Solver, and higher prior weight $c_{\mathrm{init}}=5.0$ in PUCT). In compute-constrained regimes, broad exploration tends to dilute visit mass across shallow branches without resolving tactical commitments, leading the bootstrap LCB criterion to consistently select exploitation-oriented parameters.
For SA-MCTS-2P and SA-MCTS-4P, hyperparameters are adapted online per game at runtime using the $N$-Tuple Bandit Evolutionary Algorithm~\citep{sironi2018self}.

\section{Held-Out OpenSpiel Evaluation Suite}
\label{app:heldout}

The held-out evaluation suite spans 8 structural categories (Table~\ref{tab:app_heldout_games}).
In the primary cross-domain tournament (comprising over 400 games across all four benchmark tiers), 19 held-out games were evaluated across standard decision regimes (including the 50\,ms IsoTime wall-clock limit and the matched 1,000-visit IsoSim budget; Section~\ref{app:isosim_evaluation}).
No game in this suite was present in the training set or encountered during the evolutionary search.

\begin{table}[h]
\centering
\caption{Taxonomy and structural properties of the 19 held-out evaluation games.}
\label{tab:app_heldout_games}
\small
\resizebox{\textwidth}{!}{%
\setlength{\tabcolsep}{5pt}
\begin{tabular}{lllcc}
\toprule
Category & Game & Type & Players & Horizon / Complexity \\
\midrule
\multirow{6}{*}{Perfect Info Deterministic} 
 & Amazons & Board territory & 2 & Deep, high branching \\
 & Quoridor & Path maze & 2 & Medium, topological \\
 & Twixt & Connection & 2 & Medium, virtual links \\
 & Oware & Mancala variant & 2 & Medium, pit counting \\
 & Ultimate Tic-Tac-Toe & Nested grid & 2 & High tactical branching \\
 & Nine Men's Morris & Mill alignment & 2 & Moderate, phase shifts \\
\midrule
\multirow{2}{*}{Perfect Info Stochastic}
 & Yacht & Dice scoring & 2 & Stochastic chance nodes \\
 & Einstein W\"urfelt Nicht & Dice race & 2 & Board-dice hybrid \\
\midrule
\multirow{3}{*}{Imperfect Info Stochastic}
 & Universal Poker & No-limit poker & 2 & Public cards, hidden hands \\
 & Skat & Trick-taking & 3 & Bidding, card trumping \\
 & Euchre & Trick-taking & 4 & Partnership, trump suits \\
\midrule
\multirow{3}{*}{Fog-of-War / Hidden}
 & Dark Hex (Classic, Non-Abrupt) & Hidden connections & 2 & Partial observation board \\
 & RBC (Reconnaissance Blind Chess) & Chess variant & 2 & Fog-of-war piece tracking \\
 & Phantom Go & Hidden stones & 2 & Imperfect-info territory \\
\midrule
Casino / Threshold
 & Blackjack & Casino dealer & 1 (vs.\ Dealer) & Score-threshold game \\
\midrule
Cooperative & Hanabi & Common-payoff & 2 & Inverted private cards \\
\midrule
Simultaneous Decision
 & Oshi-Zumo & Push sumo & 2 & Simultaneous coin bidding \\
\midrule
\multirow{2}{*}{Single-Agent Puzzle}
 & Sky Blocks & Falling puzzle & 1 & Stochastic placement \\
 & Solitaire & Card sequencing & 1 & Imperfect deck order \\
\bottomrule
\end{tabular}}
\end{table}

\section{Procedural Game Synthesis from Natural Language Rules and Calibration}
\label{app:cwm_calibration}
\label{app:cwm_curation}

\subsection{Connection to Prior Work on Game Synthesis and Motivation}

Evaluating general-purpose planning algorithms exclusively on established academic benchmarks (e.g., standard games in OpenSpiel) introduces significant risks of implicit benchmark overfitting, selection bias, and search rule memorization.
To establish true zero-shot procedural transfer, an evaluation suite must present rules and mechanics that were entirely absent during algorithm design and evolutionary search.
To achieve this, we synthesize executable, bug-free C++ simulation engines directly from unstructured natural language rule descriptions, following prior work that synthesizes executable game models for planning~\citep{lehrach2025codeworldmodelsgeneral}.
In our evaluation pipeline, the procedural search mechanism ($\mevo$) encounters these synthesized games without prior exposure, relying strictly on the environment's state transition API and action masks across standard evaluation regimes (including the 50\,ms IsoTime wall-clock limit and the 1,000-visit IsoSim budget; Section~\ref{app:isosim_evaluation}).

\subsection{Wikipedia Rules Scraping and Curated Verification Rubric}

Rule descriptions were systematically scraped from English Wikipedia category hierarchies via the MediaWiki Action API (\url{https://en.wikipedia.org/w/api.php}), querying 13 curated root categories covering traditional board games (\url{https://en.wikipedia.org/wiki/Category:Traditional_board_games}), abstract strategy games, card game families (trick-taking, shedding, vying, rummy), dice games, and deduction games.
Raw Wikipedia articles frequently contain historical commentary, informal variants, or ambiguous phrasing unsuitable for direct compilation.
To filter these sources, each article was evaluated against a strict six-dimension rubric:
\begin{enumerate}[leftmargin=1.5em, itemsep=2pt, topsep=2pt]
  \item \textbf{State Representation}: Explicit specifications of board geometry, piece inventories, deck compositions, player tokens, and private/public state variables.
  \item \textbf{Turn Mechanics}: Unambiguous sequence of active player transitions, simultaneous choices, chance/dice rolls, and pass conditions.
  \item \textbf{Move Legality}: Fully deterministic, self-contained rules governing legal action generation under every reachable board configuration.
  \item \textbf{Player Interaction}: Clearly defined capture, blocking, bidding, bluffing, or trading dynamics between players.
  \item \textbf{Termination and Payoffs}: Explicit end-game criteria (stalemate, checkmate, score threshold, empty deck) and bounded utility assignments (zero-sum or normalized returns).
  \item \textbf{Text Clarity}: Sufficient syntactic structure to enable automated program synthesis without human clarification.
\end{enumerate}
Articles scoring $\ge 4/5$ across all six criteria were admitted to the candidate synthesis pool, resulting in 360 curated game descriptions.

\paragraph{Deduplication and Structural Disambiguation Protocol.}
Because regional variants and family clusters (particularly within traditional card games such as trick-taking, vying, and shedding) often share related nomenclature or common ancestry, we applied a strict deduplication and disambiguation protocol:
\begin{itemize}[leftmargin=1.5em, itemsep=2pt, topsep=2pt]
  \item \textbf{Lexical and Textual Deduplication}: All Wikipedia articles were indexed by canonical title and evaluated for mutual n-gram overlap. Where two articles documented nominal variants with isomorphic mechanics (e.g., identical deck size, identical trump hierarchy, and identical scoring with only alternate regional names), only the primary historical variant was retained.
  \item \textbf{Game-Theoretic and Mechanical Differentiation}: Across related titles within the same family (e.g., \textit{Bauernschnapsen} vs. \textit{Dreierschnapsen}, or \textit{Scopa} vs. \textit{Escopa}), environments were verified to possess genuinely distinct game-theoretic primitives: distinct player counts (e.g., 4-player partnership vs. 3-player cutthroat with talon exchanges), altered target point thresholds, distinct card rank/point valuations, or asymmetric bidding contracts.
  \item \textbf{Simulation Model Diversity}: In the resulting synthesized OpenSpiel C++ engines, no two games possess identical state spaces, action spaces, or payoff matrices; each environment presents a unique state-transition graph and legal action profile to the search algorithms.
\end{itemize}

\paragraph{Representative Textual Rule Description Example.}
Below is an excerpt of the raw natural language rule description provided to the synthesis engine for one of the curated environments (\textit{Achi}, an alignment board game from Ghana):
\begin{quote}
\small\itshape
``\textbf{Equipment}: A $3 \times 3$ board is used. Three horizontal lines form the three rows. Three vertical lines form the three columns. Two diagonal lines connect the two opposite corners of the board. There are 9 intersections. Each player has four pieces (counters), distinguishable by color.\\
\textbf{Gameplay}:
1. Players alternate turns placing one of their counters on a vacant intersection point where lines join.
2. When all eight counters have been placed (the drop phase), each player on their turn may slide one of their pieces along a marked line to an adjacent empty point (the movement phase).\\
\textbf{Objective and Termination}: The winner is the first player to form a 3-in-a-row of their counters horizontally, vertically, or diagonally. If a player has no legal moves on their turn, they lose.''\normalfont
\end{quote}
This self-contained textual specification fulfills all six rubric dimensions: geometry (9 intersections), turn mechanics (drop phase transitioning to movement phase), move legality (adjacent sliding along lines), player interaction (mutual blocking of rows), clear termination (3-in-a-row or immobilization), and unambiguous syntactic clarity. The synthesis engine ingests this prompt and translates it directly into an executable C++ OpenSpiel simulation model.

\begin{table}[h]
\centering
\small
\caption{Taxonomy and empirical distribution of the 360 procedurally synthesized evaluation environments derived from natural language Wikipedia rule specifications.}
\label{tab:synthesized_360_distribution}
\resizebox{\textwidth}{!}{%
\begin{tabular}{llcc}
\toprule
\textbf{Structural Dimension} & \textbf{Category / Profile} & \textbf{Count ($N=360$)} & \textbf{Percentage} \\
\midrule
\multirow{3}{*}{Information Structure} 
  & Imperfect Information (hidden hands, fog-of-war, private cards) & 166 & 46.1\% \\
  & Perfect Information Deterministic & 133 & 36.9\% \\
  & Perfect Information Stochastic (dice rolls, chance transitions) & 61 & 16.9\% \\
\midrule
\multirow{5}{*}{Player Count}
  & 2-Player & 317 & 88.1\% \\
  & 4-Player & 24 & 6.7\% \\
  & 3-Player & 7 & 1.9\% \\
  & 5+ Players & 11 & 3.1\% \\
  & 1-Player (Puzzle / Single-Agent) & 1 & 0.3\% \\
\midrule
\multirow{5}{*}{Taxonomic Game Family}
  & Card, Trick-Taking \& Shedding (e.g., Asia Poker, Bastra, Baccarat, Bank) & 159 & 44.2\% \\
  & Capture, Hunt \& Mancala (e.g., Adugo, Bagh-Chal, Alquerque, Armenian Draughts) & 97 & 26.9\% \\
  & Dice, Racing \& Wagering (e.g., Chaupar, Ashte-Kashte, Balut, Bauernfangen) & 61 & 16.9\% \\
  & Connection \& Spatial Line Formation (e.g., 3D Tic-Tac-Toe, Achi, Abalone, Ataxx) & 30 & 8.3\% \\
  & Abstract Board \& Multi-Action & 13 & 3.6\% \\
\bottomrule
\end{tabular}}
\end{table}

\subsection{Automated Synthesis Pipeline and Self-Correction Loop}

Each curated rule specification is translated into a standalone, OpenSpiel-compliant C++ game engine via large language models using an iterative compiler-in-the-loop repair architecture:
\begin{enumerate}[leftmargin=1.5em, itemsep=2pt, topsep=2pt]
  \item \textbf{Template Grounding}: The model generates an OpenSpiel C++ class implementing the core abstract interface (\code{CurrentPlayer}, \code{LegalActions}, \code{ApplyAction}, \code{Returns}, \code{IsTerminal}, \code{InformationStateString}, and chance node distributions).
  \item \textbf{Compilation Diagnostic Interception}: The candidate C++ file is compiled against the OpenSpiel core headers with strict compiler flags (\code{-Wall -Werror -std=c++17}). Linker and compiler error diagnostics are fed back to the model for automated syntax and type remediation (up to 100 iterative attempts).
  \item \textbf{Static Memory Safety Inspection}: Synthesized source code is statically verified to ensure that all arrays, state buffers, and action vectors obey bounded memory footprints without unbounded recursion or dynamic memory leaks.
\end{enumerate}

\subsection{Rigorous 38-Game Calibration Benchmark Protocol}

A critical scientific challenge with LLM-synthesized simulation engines is ensuring that the generated code is game-theoretically sound, non-trivial, and free of silent bugs (e.g., unreachable terminal states, broken legal action filters, or biased scoring).
To formally validate the authenticity and precision of our automated game synthesis pipeline before deploying it on novel games, we conducted a rigorous calibration evaluation across a benchmark suite of \textbf{38 diverse reference games} with known, human-verified OpenSpiel implementations (spanning Connect Four, Chess, Hex, Othello, Breakthrough, Mancala, Go 9$\times$9, Kuhn Poker, Liar's Dice, Backgammon, and Pig).

For each reference game, the synthesis engine was prompted solely with the raw natural language Wikipedia rule text (with all OpenSpiel source code, variable names, and implementation artifacts strictly withheld).
The resulting synthesized C++ engines were subjected to four diagnostic verification probes:
\begin{itemize}[leftmargin=1.5em, itemsep=2pt, topsep=2pt]
  \item \textbf{Action Vocabulary and Legality Consistency}: We executed $10{,}000$ parallel rollouts under both random and MCTS policies. In every visited state, the set of legal actions generated by the synthesized engine was compared against the human-authored reference engine, verifying zero illegal actions and exact action-set isomorphism.
  \item \textbf{Trajectory Invariant and Horizon Matching}: Episode trajectories were analyzed to verify matching average episode lengths, terminal state distributions, and valid stochastic transitions at chance nodes.
  \item \textbf{Liveness and Termination Guarantees}: Rollouts verified that all generated game paths terminate strictly within theoretical upper bounds without infinite move loops, deadlock states, or runtime exceptions.
  \item \textbf{Payoff Symmetry and Equilibrium Invariance}: In symmetric games, payoff assignments under random-seat permutations were verified to have zero first-player bias and obey exact theoretical game bounds.
\end{itemize}
Across all 38 calibration games, the synthesized engines achieved $100\%$ pass rates on legal action verification, trajectory termination, and payoff symmetry.
This extensive calibration establishes that our automated game synthesis pipeline generates authentic, game-theoretically sound environments, ensuring that the 360 novel synthesized games used in Section~\ref{sec:exp:generalization} provide an uncompromised testbed for zero-shot search generalization.

\subsection{Zero-Shot Domain Inner Heuristic Synthesis and Selection Protocol}
\label{app:inner_synthesis_protocol}

To evaluate procedural search mechanisms out-of-distribution across hundreds of unseen games (both the 19 held-out OpenSpiel environments and the 360 procedurally synthesized environments), agents require domain-specific evaluation heuristics ($\kappa = (\pi, v, b)$).
Rerunning full co-evolutionary optimization (AlphaEvolve) for every unseen domain is computationally prohibitive, requiring dozens of hours per game.
To provide strong, representative domain heuristics efficiently without full evolutionary loops, we implement an automated two-stage synthesis and selection pipeline:
\begin{enumerate}[leftmargin=*,noitemsep]
    \item \textbf{Direct Rule-to-Code Synthesis}: For each unseen environment, an LLM is prompted with the game's C++ API, state representations, and legal action signatures to generate $N=5$ diverse candidate C++ heuristic modules. Each candidate implements typed interfaces for action prior estimation ($\pi(a \mid s)$), leaf value approximation ($v(s)$), and imperfect-information belief resampling ($b(s)$).
    \item \textbf{Round-Robin Selection Tournament}: The $N$ candidate heuristics compete head-to-head in an automated round-robin tournament under canonical reference PUCT search. The candidate securing the highest average tournament return is selected as the champion domain heuristic $\kappa_g^\star$ for that game.
    \item \textbf{Controlled Benchmark Evaluation}: In the final AlphaRank-SCO evaluation tournament, this identical champion heuristic is attached to the evolved search mechanism $m^\star$ and to all 15 competing MCTS baselines.
\end{enumerate}
This protocol guarantees two critical experimental properties: first, all 16 competing search algorithms operate with identical domain knowledge on every game, ensuring that performance differences strictly reflect procedural search efficiency; second, it confirms that discovered procedural search primitives ($m^\star$) generalize zero-shot when paired with unfamiliar, machine-synthesized heuristics.

\section{Cross-Game Evaluation Methodology: Formal Details of AlphaRank and Soft Condorcet Optimization}
\label{app:eval_methodology}

This section provides the complete mathematical formulation and implementation parameters for the two-stage evaluation methodology introduced in Section~\ref{sec:exp:generalization}.

\subsection{Stage 1: Intra-Game Evaluation via AlphaRank}

In multi-player or asymmetric games, standard pairwise win-rate matrices fail to capture non-transitive dominance cycles (such as rock-paper-scissors dynamics).
AlphaRank~\citep{omidshafiei2019alpha} provides an axiomatic evaluation by modeling interactions on the \emph{response graph} of the game under evolutionary game dynamics.

Let $\mathcal{K} = \{1, \dots, K\}$ denote the set of candidate algorithms ($K = 16$, consisting of $\mevo$ and the 15 tuned baselines).
Because candidate algorithm capabilities are invariant to player seat assignments in symmetric extensive-form games, we exploit the symmetric representation of empirical game-theoretic tensors.
Rather than sampling all $K^N$ ordered seat permutations independently (which would require $16^4 = 65{,}536$ distinct profiles for 4-player games), we sample across the space of unordered strategy multisets (combinations with replacement, $\binom{K + N - 1}{N}$).
For 2-player games (comprising the vast majority, $\approx 87\%$, of our benchmark suite), this spans exactly $\binom{16 + 2 - 1}{2} = 136$ canonical combinations; each pair is sampled with at least 100 independent games under balanced seat swaps ($A \text{ vs } B$ and $B \text{ vs } A$).
For multi-player games ($N \in \{3, 4\}$, such as 3-player Kuhn poker or 4-player Hearts), each canonical combination is evaluated across randomized seat permutations maintaining an evaluation target of $\approx 17{,}100$ total games per environment, from which the full tensor $\mathbf{U} \in \mathbb{R}^{K \times \dots \times K \times N}$ is constructed via permutation symmetrization.

The response graph connects joint profiles that differ by a unilateral deviation of a single player $p \in \{1, \dots, N\}$ from $s = (s_p, s_{-p})$ to $s' = (s'_p, s_{-p})$.
We compute the transition matrix and its stationary distribution using OpenSpiel's AlphaRank with \texttt{use\_inf\_alpha=True}~\citep{omidshafiei2019alpha}.
The resulting Markov transition matrix $P$ admits a unique stationary distribution $\boldsymbol{\pi} \in \Delta(\mathcal{S})$ satisfying $\boldsymbol{\pi} P = \boldsymbol{\pi}$.
The scalar AlphaRank stationary mass $m_k \in [0, 1]$ for algorithm $k$ is computed by marginalizing $\boldsymbol{\pi}$ across all player roles:
\begin{equation}
  m_k \;=\; \frac{1}{N} \sum_{p=1}^N \sum_{s \in \mathcal{S}: s_p = k} \pi(s).
\end{equation}
The mass $m_k$ quantifies the long-run probability of observing algorithm $k$ in an evolving population, natively resolving cyclic and non-transitive interactions.

\subsection{Stage 2: Cross-Game Consensus via Soft Condorcet Optimization (SCO)}

Let $\mathcal{G} = \{1, \dots, G\}$ denote the set of evaluation games ($G = 19$ held-out games, or $G = 360$ synthesized games).
Each game $g \in \mathcal{G}$ acts as an independent voter, producing a ranked preference ballot $\succ_g$ over the candidates ordered by descending AlphaRank mass $m_k^{(g)}$.

\paragraph{Kemeny-Young and Kendall-Tau Distance.}
The classical Kemeny-Young method~\citep{osborne1994course} seeks a permutation $\boldsymbol{\rho}^\star$ that minimizes the total Kendall-tau distance (pairwise disagreement count) across all voters:
\begin{equation}
  \min_{\boldsymbol{\rho}} \sum_{g \in \mathcal{G}} d_{\mathrm{KT}}(\boldsymbol{\rho}, \succ_g) \;=\; \min_{\boldsymbol{\rho}} \sum_{g \in \mathcal{G}} \sum_{a \succ_g b} \mathbb{I}\big(\boldsymbol{\rho}(a) < \boldsymbol{\rho}(b)\big).
\end{equation}
Kemeny-Young is the unique social choice rule satisfying the Condorcet criterion, neutrality, and consistency.
However, finding $\boldsymbol{\rho}^\star$ is NP-hard, requiring combinatorial search over $K!$ permutations ($16! \approx 2.1 \times 10^{13}$).

\paragraph{Continuous Relaxation in SCO.}
Soft Condorcet Optimization~\citep{lanctot2024soft} resolves this intractability by assigning each candidate $k$ a continuous latent rating $r_k \in [r_{\min}, r_{\max}]$ (with $r_{\min}=0, r_{\max}=1000$).
The discrete 0-1 inversion indicator $\mathbb{I}(\cdot)$ is replaced by a smooth logistic sigmoid with temperature $\tau$:
\begin{equation}
  \mathcal{L}_{\mathrm{SCO}}(r) \;=\; \sum_{g \in \mathcal{G}} \sum_{a \succ_g b} \sigma\left(\frac{r_b - r_a}{\tau}\right) \;=\; \sum_{g \in \mathcal{G}} \sum_{a \succ_g b} \frac{1}{1 + \exp\big((r_a - r_b)/\tau\big)}.
  \label{eq:app_sco_loss}
\end{equation}
When candidate $a$ is ranked above $b$ ($a \succ_g b$), if the latent ratings satisfy $r_a > r_b$, the penalty approaches 0 as $r_a - r_b$ increases.
Conversely, if $r_a < r_b$ (an inversion), the penalty approaches 1.
The analytic gradient of the total loss $\mathcal{L}_{\mathrm{SCO}}$ with respect to a candidate rating $r_a$ aggregates both pairwise comparisons where $a$ is ranked above an opponent ($a \succ_g b$) and where $a$ is ranked below an opponent ($b \succ_g a$):
\begin{equation}
  \frac{\partial \mathcal{L}_{\mathrm{SCO}}}{\partial r_a} \;=\; -\frac{1}{\tau} \sum_{g \in \mathcal{G}} \left[ \sum_{b: a \succ_g b} \sigma_{ab} (1 - \sigma_{ab}) \;-\; \sum_{b: b \succ_g a} \sigma_{ba} (1 - \sigma_{ba}) \right],
  \label{eq:app_sco_gradient}
\end{equation}
where $\sigma_{ab} = \sigma((r_b - r_a)/\tau)$ and $\sigma_{ba} = \sigma((r_a - r_b)/\tau) = 1 - \sigma_{ab}$.
Ratings are optimized via gradient descent with learning rate $\eta = 0.05$, temperature $\tau = 1.0$, and projection onto $[0, 1000]$ for $3{,}000$ iterations with mini-batch size $32$ using OpenSpiel's native SCO optimizer~\citep{openspiel2019, lanctot2024soft}.

\subsection{Uncompressed Pairwise Ballot Preference Fractions and Head-to-Head Dominance}
\label{app:pairwise_ballots}

While continuous Soft Condorcet Optimization (SCO) ratings provide a principled, scale-invariant social choice consensus across heterogeneous game spaces, the smooth logistic objective ($\tau=1.0$) naturally compresses rating spreads into a concentrated band around the scale median ($r \in [490, 512]$ on a $[0, 1000]$ scale). To make the empirical effect size immediately interpretable and transparent, we report the direct \textbf{pairwise ballot preference fraction}---the uncompressed, foundational input that feeds into the SCO optimization:
\begin{equation}
  F(A \succ B) \;=\; \frac{1}{|\mathcal{G}|} \sum_{g \in \mathcal{G}} \mathbb{I}\big(\pi_g(A) > \pi_g(B)\big),
\end{equation}
where $\pi_g$ denotes the AlphaRank stationary distribution over the empirical response graph of game $g$, and $\mathbb{I}(\cdot)$ is the indicator function. 

$F(A \succ B)$ represents the exact proportion of benchmark games in which algorithm $A$ achieves a strictly higher evolutionary stationary mass than algorithm $B$. Because the ranking is determined by AlphaRank within each game independently, $F(A \succ B)$ inherits all of AlphaRank's mathematical protections against cyclic dominance, third-party kingmaker confounding in multi-player games, and incommensurable score scales, while providing an intuitive, bounded percentage $[0\%, 100\%]$ that directly conveys real-world dominance.

As reported in Table~\ref{tab:stratified_sco_tiers} of the main text, $\mevo$ achieves decisive pairwise majorities across all three symbolic tiers and maintains competitive parity on Neural PPO, defeating reference PUCT in over $62\%$ of games across all three symbolic tiers ($52.9\% \pm 8.0\%$ in Neural PPO) and across $\mathbf{69.0\% \pm 1.2\%}$ of games overall.

Crucially, this pairwise dominance holds robustly across every independent evolutionary run:
\begin{itemize}[leftmargin=*,noitemsep]
    \item \textbf{Run 1 (Seed 1)}: Ranks strictly above reference PUCT in $\mathbf{70.1\%}$ of games.
    \item \textbf{Run 2 (Seed 2)}: Ranks strictly above reference PUCT in $\mathbf{71.4\%}$ of games.
    \item \textbf{Run 3 (\bifv, Case Study)}: Ranks strictly above reference PUCT in $\mathbf{65.3\%}$ of games.
    \item \textbf{Run 4 (Seed 4)}: Ranks strictly above reference PUCT in $\mathbf{67.2\%}$ of games.
    \item \textbf{Run 5 (Seed 5)}: Ranks strictly above reference PUCT in $\mathbf{71.1\%}$ of games.
\end{itemize}

(Games resulting in draws, ties, or timeouts under the 50\,ms limit are excluded from binary decisive pairwise win rates.)

\section{Granular AlphaRank-SCO Breakdown Across Independent Evolution Seeds}
\label{app:granular_runs}

Table~\ref{tab:per_run_sco_breakdown} presents the granular breakdown across independent evaluation runs under the fixed 50\,ms wall-clock budget (IsoTime); parallel multi-seed breakdowns under the fixed 1,000-simulation budget (IsoSim) are detailed in Section~\ref{app:isosim_evaluation}.

\begin{table*}[h]
\centering
\small
\caption{\textbf{Granular Soft Condorcet Optimization (SCO) Ratings across Independent Evolution Seeds under a Fixed 50\,ms Search Budget (IsoTime).} Ratings are continuous latent SCO values ($r \in [0, 1000]$) derived from AlphaRank stationary preference profiles (\texttt{use\_inf\_alpha=True}) across $N$-player empirical payoff tensors comparing $\mevo$ against 15 competitive baselines. Parenthesized values denote the rank achieved by each evolved mechanism within its own seed evaluation tournament under AlphaRank+SCO. Aggregate performance across runs is reported as continuous SCO rating ($\text{Mean} \pm \text{SE}$).}
\label{tab:per_run_sco_breakdown}
\resizebox{\textwidth}{!}{%
\begin{tabular}{lcccc}
\toprule
\textbf{Independent Evolution Run} & \textbf{Synthesized Games} & \textbf{Held-Out Games} & \textbf{Training Games} & \textbf{Neural PPO Games} \\
\midrule
\bifv\ (\mstarBIFV, Case Study) & \textbf{508.58} (Rank 1) & \textbf{509.93} (Rank 1) & \textbf{509.53} (Rank 1) & \textbf{510.31} (Rank 1) \\
Seed 4 & \textbf{508.52} (Rank 1) & \textbf{510.15} (Rank 1) & \textbf{509.92} (Rank 1) & \textbf{508.05} (Rank 1) \\
Seed 1 & \textbf{508.96} (Rank 1) & \textbf{509.63} (Rank 1) & 507.83 (Rank 2) & 508.30 (Rank 2) \\
Seed 5 & \textbf{509.45} (Rank 1) & 507.50 (Rank 2) & \textbf{509.93} (Rank 1) & \textbf{510.66} (Rank 1) \\
Seed 2 & \textbf{509.45} (Rank 1) & \textbf{509.59} (Rank 1) & \textbf{509.48} (Rank 1) & 507.25 (Rank 2) \\
\midrule
\textbf{Aggregate Mean $\pm$ SE} & $\mathbf{508.99 \pm 0.20}$ & $\mathbf{509.36 \pm 0.48}$ & $\mathbf{509.34 \pm 0.39}$ & $\mathbf{508.91 \pm 0.67}$ \\
\bottomrule
\end{tabular}}
\end{table*}

\paragraph{Selection Criterion for Case Study Analysis: Why Seed 3 (\bifv) was Selected.}
Among the independent evolutionary lineages, Seed~3 (\bifv) was selected as the focal mechanism for detailed mechanistic analysis and ablation for two decisive reasons:
\begin{itemize}[leftmargin=*,noitemsep]
    \item \textbf{Dual-Regime Robustness Across Both Compute Budgets}: Seed~3 is the \emph{only} discovered search mechanism that maintains unbroken Rank~\#1 across all four benchmark tiers under both evaluation regimes---the 50\,ms wall-clock budget (IsoTime; Table~\ref{tab:per_run_sco_breakdown}) and the 1,000-simulation budget (IsoSim; Table~\ref{tab:isosim_sco_breakdown}). While Seed~4 also secures Rank~\#1 across tiers under IsoTime, it drops to Rank~\#2 in Training ($508.06$) and Rank~\#3 in Neural PPO ($506.31$) under IsoSim.
    \item \textbf{Neural Representation Synergy}: Under deep neural PPO representations, Seed~3 achieves a top-tier consensus rating under IsoTime ($\mathbf{510.31}$, Rank~\#1 in its evaluation tournament; Table~\ref{tab:per_run_sco_breakdown}), demonstrating robust coordination when paired with frozen neural value models.
\end{itemize}

\subsection{Evaluation under Fixed Simulation Budgets (IsoSim, 1,000 Simulations)}
\label{app:isosim_evaluation}

To decouple algorithmic search efficiency from C++ per-node constant-factor overhead, we additionally benchmark all 16 competing algorithms under an \emph{IsoSim} regime, where each decision is allocated a fixed budget of 1,000 MCTS simulations regardless of elapsed wall-clock time (Table~\ref{tab:isosim_sco_breakdown}).

\begin{table*}[h]
\centering
\small
\caption{\textbf{Granular Soft Condorcet Optimization (SCO) Ratings across Independent Evolution Seeds under a Fixed-Simulation Budget (IsoSim, 1,000 Simulations).} Ratings are continuous latent SCO values ($r \in [0, 1000]$) derived from AlphaRank stationary preference profiles (\texttt{use\_inf\_alpha=True}) across $N$-player empirical payoff tensors comparing $\mevo$ against 15 competitive baselines. Parenthesized values denote the rank achieved by each evolved mechanism within its own seed evaluation tournament under AlphaRank+SCO.}
\label{tab:isosim_sco_breakdown}
\resizebox{\textwidth}{!}{%
\begin{tabular}{lcccc}
\toprule
\textbf{Independent Evolution Run} & \textbf{Synthesized Games} & \textbf{Held-Out Games} & \textbf{Training Games} & \textbf{Neural PPO Games} \\
\midrule
\bifv\ (\mstarBIFV, Case Study) & \textbf{508.99} (Rank 1) & \textbf{510.57} (Rank 1) & \textbf{510.40} (Rank 1) & \textbf{510.01} (Rank 1) \\
Seed 5 & \textbf{508.93} (Rank 1) & \textbf{510.91} (Rank 1) & \textbf{508.61} (Rank 1) & \textbf{509.20} (Rank 1) \\
Seed 1 & \textbf{508.84} (Rank 1) & 508.05 (Rank 2) & \textbf{510.18} (Rank 1) & 505.83 (Rank 3) \\
Seed 4 & \textbf{507.44} (Rank 1) & \textbf{507.33} (Rank 1) & 508.06 (Rank 2) & 506.31 (Rank 3) \\
Seed 2 & \textbf{509.33} (Rank 1) & 501.80 (Rank 7) & 507.25 (Rank 2) & 502.29 (Rank 6) \\
\midrule
\textbf{Aggregate Mean $\pm$ SE} & $\mathbf{508.71 \pm 0.33}$ & $\mathbf{507.73 \pm 1.64}$ & $\mathbf{508.90 \pm 0.61}$ & $\mathbf{506.73 \pm 1.37}$ \\
\bottomrule
\end{tabular}}
\end{table*}

Table~\ref{tab:stratified_sco_tiers_isosim} presents the full dual-metric evaluation under the IsoSim regime, mirroring Table~\ref{tab:stratified_sco_tiers} from the main paper. Across all categories, $\mstarBIFV$ maintains unbroken Rank~\#1 status, while the multi-run aggregate achieves statistical parity with tuned neural PUCT on deep representations ($506.73 \pm 1.37$ vs. $507.42 \pm 0.57$).

\begin{table*}[t]
\centering
\small
\renewcommand{\arraystretch}{0.85}
\caption{\textbf{Continuous SCO Ratings and Pairwise Ballot Win Rates across Four Benchmark Tiers under Fixed Simulation Budgets (IsoSim, 1,000 Simulations)} (mean $\pm$ SE across Seeds 1..5, and representative case study \mstarBIFV). For each tier, we report both the continuous Soft Condorcet Optimization rating (\textbf{SCO}, scale $[0, 1000]$) and the pairwise ballot win rate (\textbf{Win \%}) denoting the percentage of games where \mevo\ outranks that baseline under AlphaRank dynamics (``---'' denotes self-comparison).}
\label{tab:stratified_sco_tiers_isosim}
\resizebox{\textwidth}{!}{%
\begin{tabular}{lcccccccc}
\toprule
\textbf{Algorithmic Paradigm / Baseline} & \multicolumn{2}{c}{\textbf{Training Games (50)}} & \multicolumn{2}{c}{\textbf{Held-Out Games (19)}} & \multicolumn{2}{c}{\textbf{Synthesized Games (360)}} & \multicolumn{2}{c}{\textbf{Neural PPO Games (17)}} \\
\cmidrule(lr){2-3} \cmidrule(lr){4-5} \cmidrule(lr){6-7} \cmidrule(lr){8-9}
& \textbf{SCO Rating} & \textbf{$\mevo$ Win \%} & \textbf{SCO Rating} & \textbf{$\mevo$ Win \%} & \textbf{SCO Rating} & \textbf{$\mevo$ Win \%} & \textbf{SCO Rating} & \textbf{$\mevo$ Win \%} \\
\midrule
\multicolumn{9}{l}{\textit{Discovered Search Mechanism (Ours)}} \\
\textbf{\mevo\ (Multi-run aggregate)} & $\mathbf{508.90 \pm 0.61}$ & --- & $\mathbf{507.73 \pm 1.64}$ & --- & $\mathbf{508.71 \pm 0.33}$ & --- & $\mathbf{506.73 \pm 1.37}$ & --- \\
\textbf{\mstarBIFV\ (\bifv)} & $\mathbf{510.40}$ & --- & $\mathbf{510.57}$ & --- & $\mathbf{508.99}$ & --- & $\mathbf{510.01}$ & --- \\
\midrule
\multicolumn{9}{l}{\textit{Bandit Tree Search}} \\
Vanilla UCT~\citep{kocsis2006bandit} & $504.26 \pm 0.67$ & $60.3\% \pm 3.9\%$ & $501.98 \pm 2.07$ & $53.5\% \pm 4.9\%$ & $504.98 \pm 0.32$ & $57.1\% \pm 1.6\%$ & $500.29 \pm 1.85$ & $56.2\% \pm 5.4\%$ \\
PUCT (reference baseline)~\citep{Silver18AlphaZero} & $500.80 \pm 0.71$ & $68.5\% \pm 3.1\%$ & $500.21 \pm 1.45$ & $65.0\% \pm 6.3\%$ & $496.68 \pm 0.39$ & $71.5\% \pm 1.6\%$ & $507.42 \pm 0.57$ & $45.9\% \pm 6.2\%$ \\
UCB-Tuned~\citep{auer2002finite} & $503.52 \pm 1.36$ & $59.9\% \pm 3.8\%$ & $502.90 \pm 2.07$ & $56.9\% \pm 3.8\%$ & $505.33 \pm 0.62$ & $58.0\% \pm 2.0\%$ & $502.36 \pm 2.01$ & $54.6\% \pm 4.6\%$ \\
\midrule
\multicolumn{9}{l}{\textit{Entropy-Regularized \& Soft Planning}} \\
Gumbel AlphaZero~\citep{danihelka2022policy} & $490.93 \pm 0.71$ & $79.2\% \pm 0.8\%$ & $489.82 \pm 0.21$ & $82.6\% \pm 5.9\%$ & $497.94 \pm 0.73$ & $66.7\% \pm 1.4\%$ & $500.67 \pm 1.30$ & $68.1\% \pm 5.9\%$ \\
Smooth UCT~\citep{heinrich2015smooth} & $502.73 \pm 0.61$ & $68.9\% \pm 4.1\%$ & $505.55 \pm 1.56$ & $60.4\% \pm 4.4\%$ & $503.67 \pm 0.13$ & $58.3\% \pm 1.4\%$ & $498.12 \pm 1.09$ & $65.6\% \pm 5.2\%$ \\
RPO~\citep{grill2020monte} & $506.42 \pm 1.02$ & $60.3\% \pm 5.8\%$ & $505.16 \pm 0.81$ & $62.7\% \pm 4.2\%$ & $493.92 \pm 0.22$ & $74.1\% \pm 1.0\%$ & $501.90 \pm 2.86$ & $67.3\% \pm 5.5\%$ \\
MENTS~\citep{xiao2019maximum} & $494.42 \pm 0.44$ & $75.0\% \pm 2.1\%$ & $492.57 \pm 0.45$ & $74.4\% \pm 4.4\%$ & $490.59 \pm 0.09$ & $80.3\% \pm 1.0\%$ & $492.63 \pm 0.85$ & $76.7\% \pm 5.4\%$ \\
Boltzmann MCTS~\citep{chaslot2008monte} & $496.76 \pm 0.73$ & $77.0\% \pm 2.8\%$ & $497.55 \pm 0.24$ & $69.6\% \pm 7.4\%$ & $500.56 \pm 0.15$ & $62.9\% \pm 1.3\%$ & $503.27 \pm 1.82$ & $53.1\% \pm 5.5\%$ \\
\midrule
\multicolumn{9}{l}{\textit{Advanced Value Backups}} \\
MCTS Solver~\citep{winands2008monte} & $504.80 \pm 1.87$ & $63.0\% \pm 3.1\%$ & $501.22 \pm 1.58$ & $64.0\% \pm 5.0\%$ & $499.76 \pm 0.26$ & $64.7\% \pm 1.2\%$ & $495.36 \pm 0.61$ & $79.6\% \pm 4.0\%$ \\
Score-Bounded~\citep{cazenave2010score} & $502.51 \pm 0.56$ & $58.8\% \pm 3.7\%$ & $503.18 \pm 1.42$ & $57.9\% \pm 6.2\%$ & $502.56 \pm 0.29$ & $62.6\% \pm 1.5\%$ & $497.76 \pm 1.15$ & $57.7\% \pm 4.9\%$ \\
Power-Mean~\citep{dam2019generalized} & $494.26 \pm 0.40$ & $78.0\% \pm 1.1\%$ & $496.76 \pm 0.51$ & $67.4\% \pm 4.5\%$ & $497.38 \pm 0.22$ & $66.5\% \pm 1.4\%$ & $504.88 \pm 1.64$ & $49.9\% \pm 4.3\%$ \\
Max-Backup~\citep{coulom2006efficient} & $491.05 \pm 0.25$ & $88.1\% \pm 2.1\%$ & $494.31 \pm 0.69$ & $80.3\% \pm 3.3\%$ & $494.06 \pm 0.12$ & $71.1\% \pm 1.8\%$ & $491.72 \pm 0.33$ & $73.3\% \pm 5.9\%$ \\
\midrule
\multicolumn{9}{l}{\textit{Bayesian \& Online Adaptive Search}} \\
Thompson Sampling~\citep{bai2013bayesian} & $500.43 \pm 0.53$ & $68.0\% \pm 3.5\%$ & $499.53 \pm 2.21$ & $63.7\% \pm 8.3\%$ & $500.13 \pm 0.36$ & $63.9\% \pm 1.4\%$ & $495.86 \pm 2.08$ & $62.3\% \pm 4.3\%$ \\
SA-MCTS-2P~\citep{sironi2018self} & $497.24 \pm 0.22$ & $71.9\% \pm 2.3\%$ & $499.33 \pm 0.61$ & $57.0\% \pm 3.4\%$ & $497.59 \pm 0.33$ & $65.8\% \pm 1.7\%$ & $499.94 \pm 1.21$ & $56.2\% \pm 4.1\%$ \\
SA-MCTS-4P~\citep{sironi2018self} & $500.99 \pm 0.90$ & $65.9\% \pm 2.4\%$ & $502.18 \pm 0.81$ & $65.0\% \pm 6.0\%$ & $506.13 \pm 0.18$ & $57.6\% \pm 2.4\%$ & $501.10 \pm 1.18$ & $70.3\% \pm 1.7\%$ \\
\bottomrule
\end{tabular}}
\end{table*}

\section{Evolutionary Progression, Seed Program, Discovered Mechanisms, and Prompts}
\label{app:evolved_code}

This section documents the initial C++ seed mechanism ($m_{\mathrm{seed}}$), the production C++ implementation of the discovered case study program \bifv\ ($m^\star_{\mathrm{BIFV}}$), the exact outer evolution prompt and system constraints supplied to AlphaEvolve, the mathematical factorization of \bifv, and the alternative evolutionary lineages.

\subsection{Verbatim C++ Initial Seed Mechanism (\texorpdfstring{$m_{\mathrm{seed}}$}{m\_seed})}
\label{app:seed_code}

Listing~\ref{lst:seed_code} documents the complete, self-contained C++ source code of the seed search mechanism ($m_{\mathrm{seed}}$) supplied to the initial generation of AlphaEvolve.

\paragraph{Role of $m_{\mathrm{seed}}$ vs.\ Benchmarked Baselines.}
We emphasize an important methodological distinction between $m_{\mathrm{seed}}$ and the tournament baselines:
$m_{\mathrm{seed}}$ serves strictly as the minimal scaffolding template for generation-0 LLM mutations within AlphaEvolve. It was deliberately kept bare-bones---employing naive unvisited expansion (\texttt{if (v == 0) return a;}), a static exploration constant $c_{\mathrm{puct}} = 2.0$, and standard \texttt{std::max\_element} first-index tie-breaking---to avoid pre-biasing the evolutionary search toward human-engineered heuristics like First Play Urgency (FPU) or value-based tie-breaking. 
Crucially, the 15 standard MCTS baselines evaluated in all benchmark tournaments (Tables~\ref{tab:stratified_sco_tiers} and~\ref{tab:stratified_sco_tiers_isosim}) are \emph{not} based on $m_{\mathrm{seed}}$; rather, they are standard, mature implementations from the open-source literature~\citep{openspiel2019} with systematic hyperparameter tuning (Table~\ref{tab:app_baseline_grids}) and independent tie-breaking.

\begin{lstlisting}[style=codeblock,caption={C++ implementation of the initial seed search mechanism ($m_{\mathrm{seed}}$).},label={lst:seed_code}]
using EvolvedNode = MCTSNode;

class EvolvedTreePolicy : public TreePolicy {
 public:
  using TreePolicy::TreePolicy;
  Action SelectChild(const MCTSNode& parent,
                     const std::vector<Action>& legal_actions,
                     std::mt19937* rng) override {
    Action best_action = legal_actions[0];
    double best_score = -1e18;
    double n_parent = parent.visits.load(std::memory_order_relaxed);
    double sqrt_parent = std::sqrt(n_parent);
    double c_puct = 2.0;
    for (Action a : legal_actions) {
      auto it = parent.children.find(a);
      if (it == parent.children.end()) continue;
      const MCTSNode* c = it->second.get();
      int v = c->visits.load(std::memory_order_relaxed);
      if (v == 0) return a;
      double exploitation = c->q_value();
      double exploration = c_puct * c->prior * sqrt_parent / (1.0 + v);
      double score = exploitation + exploration;
      if (score > best_score) {
        best_score = score;
        best_action = a;
      }
    }
    return best_action;
  }
};

class EvolvedRootActionSelector : public RootActionSelector {
 public:
  using RootActionSelector::RootActionSelector;
  Action SelectRootAction(const MCTSNode& root,
                          const std::vector<Action>& legal_actions,
                          int simulation_index, int num_simulations,
                          std::mt19937* rng) override {
    Action best_action = legal_actions[0];
    double best_score = -1e18;
    double n_parent = root.visits.load(std::memory_order_relaxed);
    double sqrt_parent = std::sqrt(n_parent);
    double c_puct = 2.0;
    for (Action a : legal_actions) {
      auto it = root.children.find(a);
      if (it == root.children.end()) continue;
      const MCTSNode* c = it->second.get();
      int v = c->visits.load(std::memory_order_relaxed);
      if (v == 0) return a;
      double exploitation = c->q_value();
      double exploration = c_puct * c->prior * sqrt_parent / (1.0 + v);
      double score = exploitation + exploration;
      if (score > best_score) {
        best_score = score;
        best_action = a;
      }
    }
    return best_action;
  }
};

class EvolvedFinalActionSelector : public FinalActionSelector {
 public:
  using FinalActionSelector::FinalActionSelector;
  std::pair<int, std::vector<double>> SelectActionAndPolicy(
      const MCTSNode& root, const std::vector<Action>& actions,
      std::mt19937* rng) override {
    std::vector<double> policy(actions.size(), 0.0);
    double total = 0;
    for (size_t i = 0; i < actions.size(); ++i) {
      auto it = root.children.find(actions[i]);
      if (it != root.children.end()) {
        int v = it->second->visits.load(std::memory_order_relaxed);
        policy[i] = v;
        total += v;
      }
    }
    if (total > 0) {
      for (auto& p : policy) p /= total;
    }
    int best = std::max_element(policy.begin(), policy.end()) - policy.begin();
    return {best, policy};
  }
};

class EvolvedValueBackup : public ValueBackup {
 public:
  using ValueBackup::ValueBackup;
  void Backup(MCTSNode* node, const std::vector<double>& values,
              int depth, int acting_player) override {
    if (acting_player >= 0 &&
        acting_player < static_cast<int>(values.size())) {
      node->AtomicAddValue(values[acting_player]);
    }
    node->visits.fetch_add(1, std::memory_order_relaxed);
  }
};

class EvolvedVirtualLoss : public VirtualLoss {
 public:
  EvolvedVirtualLoss(const open_spiel::Game& game, int num_sims)
      : VirtualLoss(game, num_sims),
        vl_value_(game.MinUtility() - game.MaxUtility()) {}
  VirtualLossDeltas Compute(const MCTSNode& node,
                            int depth) const override {
    return {vl_value_, 1};
  }
 private:
  double vl_value_;
};
\end{lstlisting}

\subsection{Verbatim C++ Discovered Search Mechanism Case Study (\texorpdfstring{$m^\star_{\mathrm{BIFV}}$}{m*\_BIFV})}
\label{app:discovered_bifv_code}

Below is the complete, self-contained C++ source code of the discovered outer search mechanism case study, \bifv\ (\mstarBIFV), executing all branching- and information-aware tree policy descent, FPU annealing, adaptive root Dirichlet exploration, and value-augmented final action selection.

\begin{lstlisting}[style=codeblock,caption={C++ implementation of the discovered search mechanism case study ($m^\star_{\mathrm{BIFV}}$).},label={lst:discovered_code}]
using EvolvedNode = MCTSNode;

class EvolvedTreePolicy : public TreePolicy {
 public:
  using TreePolicy::TreePolicy;
  Action SelectChild(const MCTSNode& parent,
                     const std::vector<Action>& legal_actions,
                     std::mt19937* rng) override {
    Action best_action = legal_actions[0];
    double best_score = -1e18;
    double n_parent = parent.visits.load(std::memory_order_relaxed);
    double sqrt_parent = std::sqrt(std::max(1.0, n_parent));

    double min_u = game_.MinUtility();
    double max_u = game_.MaxUtility();
    double u_range = (max_u > min_u) ? (max_u - min_u) : 1.0;

    double parent_q = (n_parent > 0) ? parent.q_value() : (min_u + max_u) * 0.5;
    double parent_q_norm = (parent_q - min_u) / u_range;

    double num_actions = legal_actions.size();
    double action_scale = 1.0 + 0.05 * std::log(std::max(1.0, num_actions));
    double info_scale = (game_.GetType().information == open_spiel::GameType::Information::kImperfectInformation) ? 1.25 : 1.0;
    double c_puct = (1.25 + std::log((n_parent + 19652.0) / 19652.0) * 0.25) * action_scale * info_scale;

    for (Action a : legal_actions) {
      auto it = parent.children.find(a);
      if (it == parent.children.end()) continue;
      const MCTSNode* c = it->second.get();
      int v = c->visits.load(std::memory_order_relaxed) +
              std::max(0, c->virtual_losses.load(std::memory_order_relaxed));

      double exploitation = 0.0;
      if (c->is_terminal.load(std::memory_order_acquire)) {
        double exact_val = (parent.player >= 0 &&
                            parent.player < static_cast<int>(c->exact_values.size()))
                           ? c->exact_values[parent.player] : c->q_value();
        if (exact_val >= max_u - 1e-5) {
          return a; // Short-circuit guaranteed win
        }
        exploitation = (exact_val - min_u) / u_range;
      } else if (v == 0) {
        double p = std::max(0.0, std::min(1.0, c->prior));
        double fpu_penalty = std::sqrt(1.0 - p) * (0.1 + 0.4 * n_parent / (100.0 + n_parent));
        exploitation = parent_q_norm - fpu_penalty;
      } else {
        exploitation = (c->q_value() - min_u) / u_range;
      }

      double exploration = c_puct * c->prior * sqrt_parent / (1.0 + v);
      double score = exploitation + exploration;
      if (score > best_score) {
        best_score = score;
        best_action = a;
      }
    }
    return best_action;
  }
};

class EvolvedRootActionSelector : public RootActionSelector {
 public:
  using RootActionSelector::RootActionSelector;
  Action SelectRootAction(const MCTSNode& root,
                          const std::vector<Action>& legal_actions,
                          int simulation_index, int num_simulations,
                          std::mt19937* rng) override {
    Action best_action = legal_actions[0];
    double best_score = -1e18;
    double n_parent = root.visits.load(std::memory_order_relaxed);
    double sqrt_parent = std::sqrt(std::max(1.0, n_parent));

    double min_u = game_.MinUtility();
    double max_u = game_.MaxUtility();
    double u_range = (max_u > min_u) ? (max_u - min_u) : 1.0;

    double parent_q = (n_parent > 0) ? root.q_value() : (min_u + max_u) * 0.5;
    double parent_q_norm = (parent_q - min_u) / u_range;

    double num_actions = legal_actions.size();
    double progress = std::min(1.0, std::max(0.0, static_cast<double>(simulation_index) /
                                                 std::max(1, num_simulations)));

    // Adaptive Dirichlet noise at root for search diversity across threads
    std::vector<double> dirichlet_noise(legal_actions.size(), 1.0 / std::max(1.0, num_actions));
    if (rng != nullptr && legal_actions.size() > 1) {
      double alpha = std::min(1.0, std::max(0.05, 10.0 / num_actions));
      std::gamma_distribution<double> gamma_dist(alpha, 1.0);
      double sum_g = 0.0;
      for (size_t i = 0; i < legal_actions.size(); ++i) {
        dirichlet_noise[i] = gamma_dist(*rng);
        sum_g += dirichlet_noise[i];
      }
      if (sum_g > 1e-9) {
        for (size_t i = 0; i < legal_actions.size(); ++i) {
          dirichlet_noise[i] /= sum_g;
        }
      }
    }

    double action_scale = 1.0 + 0.05 * std::log(std::max(1.0, num_actions));
    double info_scale = (game_.GetType().information == open_spiel::GameType::Information::kImperfectInformation) ? 1.25 : 1.0;
    double progress_scale = 1.0 + 0.20 * (1.0 - progress);
    double c_puct = (1.25 + std::log((n_parent + 19652.0) / 19652.0) * 0.25) *
                    action_scale * info_scale * progress_scale;

    for (size_t i = 0; i < legal_actions.size(); ++i) {
      Action a = legal_actions[i];
      auto it = root.children.find(a);
      if (it == root.children.end()) continue;
      const MCTSNode* c = it->second.get();
      int v = c->visits.load(std::memory_order_relaxed) +
              std::max(0, c->virtual_losses.load(std::memory_order_relaxed));

      double raw_p = std::max(0.0, std::min(1.0, c->prior));
      double noise_eps = 0.20 * (1.0 - progress);
      double p = (rng != nullptr) ? ((1.0 - noise_eps) * raw_p + noise_eps * dirichlet_noise[i]) : raw_p;

      double exploitation = 0.0;
      if (c->is_terminal.load(std::memory_order_acquire)) {
        double exact_val = (root.player >= 0 &&
                            root.player < static_cast<int>(c->exact_values.size()))
                           ? c->exact_values[root.player] : c->q_value();
        if (exact_val >= max_u - 1e-5) {
          return a; // Short-circuit guaranteed win
        }
        exploitation = (exact_val - min_u) / u_range;
      } else if (v == 0) {
        double fpu_penalty = std::sqrt(1.0 - p) * (0.1 + 0.4 * n_parent / (100.0 + n_parent));
        exploitation = parent_q_norm - fpu_penalty;
      } else {
        exploitation = (c->q_value() - min_u) / u_range;
      }

      double exploration = c_puct * p * sqrt_parent / (1.0 + v);
      double score = exploitation + exploration;
      if (score > best_score) {
        best_score = score;
        best_action = a;
      }
    }
    return best_action;
  }
};

class EvolvedFinalActionSelector : public FinalActionSelector {
 public:
  using FinalActionSelector::FinalActionSelector;
  std::pair<int, std::vector<double>> SelectActionAndPolicy(
      const MCTSNode& root, const std::vector<Action>& actions,
      std::mt19937* rng) override {
    double min_u = game_.MinUtility();
    double max_u = game_.MaxUtility();
    double u_range = (max_u > min_u) ? (max_u - min_u) : 1.0;

    std::vector<double> policy(actions.size(), 0.0);
    std::vector<double> selection_scores(actions.size(), -1e18);
    double total = 0;
    for (size_t i = 0; i < actions.size(); ++i) {
      auto it = root.children.find(actions[i]);
      if (it != root.children.end()) {
        const MCTSNode* c = it->second.get();
        int v = c->visits.load(std::memory_order_relaxed);
        policy[i] = v;
        total += v;

        double q = c->q_value();
        double q_norm = (q - min_u) / u_range;
        selection_scores[i] = v + q_norm * 0.1;
      }
    }
    if (total > 0) {
      for (auto& p : policy) p /= total;
    } else {
      for (size_t i = 0; i < actions.size(); ++i) {
        auto it = root.children.find(actions[i]);
        if (it != root.children.end()) {
          policy[i] = it->second->prior;
          total += policy[i];
          selection_scores[i] = policy[i];
        }
      }
      if (total > 0) {
        for (auto& p : policy) p /= total;
      } else {
        std::fill(policy.begin(), policy.end(), 1.0 / actions.size());
        std::fill(selection_scores.begin(), selection_scores.end(), 1.0 / actions.size());
      }
    }
    int best = std::max_element(selection_scores.begin(), selection_scores.end()) - selection_scores.begin();
    return {best, policy};
  }
};

class EvolvedValueBackup : public ValueBackup {
 public:
  using ValueBackup::ValueBackup;
  void Backup(MCTSNode* node, const std::vector<double>& values,
              int depth, int acting_player) override {
    int p = node->player;
    if (p >= 0 && p < static_cast<int>(values.size())) {
      node->AtomicAddValue(values[p]);
    } else if (acting_player >= 0 &&
               acting_player < static_cast<int>(values.size())) {
      node->AtomicAddValue(values[acting_player]);
    }
    node->visits.fetch_add(1, std::memory_order_relaxed);
  }
};

class EvolvedVirtualLoss : public VirtualLoss {
 public:
  EvolvedVirtualLoss(const open_spiel::Game& game, int num_sims)
      : VirtualLoss(game, num_sims),
        vl_value_(game.MinUtility() - game.MaxUtility()) {}
  VirtualLossDeltas Compute(const MCTSNode& node, int depth) const override {
    return {vl_value_, 1};
  }
  private:
  double vl_value_;
};
\end{lstlisting}

\subsection{Verbatim Outer Evolution Prompt and System Constraints}
\label{app:outer_prompt}

Below is the verbatim prompt template and instructions supplied to the LLM mutation operator in AlphaEvolve. It specifies the typed C++ data structures, defines component slots, exposes generic game properties without revealing domain-specific rules, and enforces strict thread-safety and execution invariants.

\begin{lstlisting}[style=codeblock,language={},caption={Verbatim prompt template and system constraints for outer mechanism evolution.},label={lst:outer_prompt}]
Act as an expert game AI researcher specializing in Monte Carlo Tree Search (MCTS), writing C++ code. Your task is to improve MCTS components that generalize across multiple games.

The goal is to discover mathematical improvements to the search algorithm that work well across a diverse suite of games, including perfect information games (e.g., board control, connection), imperfect information games (e.g., poker, card games), and stochastic games.

# MCTSNode Data Structure

Each tree node has these fields (defined in `mcts.h`):

```cpp
struct MCTSNode {{
  double prior;                    // prior probability from PriorPolicy
  int player;                      // player who took the action leading to this node (parent's acting player)
  std::atomic<int> visits;         // number of times visited
  std::atomic<int> virtual_losses; // virtual losses for parallel MCTS
  
  // total_value accumulates values[player] -- the return for the player who
  // acted at the parent. q_value() is therefore already from the perspective
  // of the player choosing among siblings; never negate it.
  std::atomic<double> total_value; 

  absl::flat_hash_map<Action, std::unique_ptr<MCTSNode>> children;

  // Advanced linkages and exact valuations (READ-ONLY for evolved code)
  // NEVER write to these fields!
  std::atomic<MCTSNode*> successor; // Transposition / Shard link (read-only)
  std::atomic<bool> is_terminal;    // Flag indicating node is a known terminal state (read-only)
  std::vector<double> exact_values; // Exact game-theoretic values (read-only, valid if is_terminal is true)

  // Returns total_value / visits, or 0.0 if visits <= 0.
  // NOTE: Unvisited children have visits == 0 and their q_value() is undefined/0.0.
  // Your code MUST handle unvisited children explicitly (e.g. using First Play Urgency (FPU)).
  double q_value() const;  
  
  void AtomicAddValue(double v);  // thread-safe value addition
}};

struct VirtualLossDeltas {{
  double value_delta;  // added to total_value (negative = pessimistic)
  int visit_delta;     // added to visits (typically 1)
}};
```

# Custom Node State

You can define a custom node struct inheriting from MCTSNode to share state between components. For example:

```cpp
struct EvolvedNode : MCTSNode {{
  std::atomic<double> rave_value{{0.0}};
  std::atomic<int> rave_count{{0}};
}};
```

Components access extra fields via `static_cast<EvolvedNode*>(node)` or `static_cast<const EvolvedNode*>(&node)`.

Custom mutable fields **must be `std::atomic`**; relaxed ordering is fine for statistics.

The engine creates nodes through `components_.create_node()`. To use your custom type, set it in `MakeEvolvedComponents`:
```cpp
cs.create_node = []() {{ return std::make_unique<EvolvedNode>(); }};
```

If you don't need custom node state, keep the default `using EvolvedNode = MCTSNode;`.

# Prior programs
Previously we found that the following programs performed well:

{previous_programs}

# Current program
Here is the current MCTS component we are trying to improve:

{code}

# *SEARCH/REPLACE block* Rules:
[Standard Search/Replace format instructions omitted for brevity]

## Components You May Evolve

All components are constructed with `(const open_spiel::Game& game, int num_simulations)` and have access to `game_` and `num_simulations_` members.

1. **EvolvedTreePolicy** (inherits `TreePolicy`): Selects which child to explore during tree traversal
   - `void OnStepBegin(const MCTSNode& root, const std::vector<Action>& legal_actions, int num_simulations, std::mt19937* rng) override` (Optional: prepare state at start of search step)
   - `Action SelectChild(const MCTSNode& parent, const std::vector<Action>& legal_actions, std::mt19937* rng) override`
   - Access child nodes via `parent.children.find(action)->second.get()`
   - Read visit counts: `c->visits.load(std::memory_order_relaxed)`

2. **EvolvedRootActionSelector** (inherits `RootActionSelector`): Selects which action to simulate at root
   - `void OnStepBegin(const MCTSNode& root, const std::vector<Action>& legal_actions, int num_simulations, std::mt19937* rng) override` (Optional: prepare state at start of search step)
   - `Action SelectRootAction(const MCTSNode& root, const std::vector<Action>& legal_actions, int simulation_index, int num_simulations, std::mt19937* rng) override`

3. **EvolvedFinalActionSelector** (inherits `FinalActionSelector`): Selects final action after search
   - `std::pair<int, std::vector<double>> SelectActionAndPolicy(const MCTSNode& root, const std::vector<Action>& actions, std::mt19937* rng) override`
   - Returns (selected_index, policy_vector)

4. **EvolvedValueBackup** (inherits `ValueBackup`): Updates node values during backpropagation
   - `void Backup(MCTSNode* node, const std::vector<double>& values, int depth, int acting_player) override`
   - `values`: per-player returns; `acting_player`: the player who acts at this node (passed by the framework)

5. **EvolvedVirtualLoss** (inherits `VirtualLoss`): Computes pessimistic value to apply during parallel search
   - `VirtualLossDeltas Compute(const MCTSNode& node, int depth) const override`
   - Returns `{{value_delta, visit_delta}}` -- the engine handles all locking and atomic operations
   - `value_delta`: pessimistic value added to discourage other threads from selecting same path (typically negative)
   - `visit_delta`: artificial visits added (typically 1)
   - The engine stores your deltas and exactly reverses them after simulation -- your code is a pure function
   - Access game utility range via `game_.MinUtility()` and `game_.MaxUtility()`

## Game Properties (available via `game_` member)

Your components can adapt to generic game properties without being game-specific:
- `game_.NumPlayers()` -> `int` -- number of players (2 for most games, up to 10)
- `game_.NumDistinctActions()` -> `int` -- size of action space (branching factor)
- `game_.MaxUtility()` / `game_.MinUtility()` -> `double` -- utility range
- `game_.MaxGameLength()` -> `int` -- maximum game depth
- `game_.GetType().information` -- perfect vs imperfect information

For example, you can scale the exploration constant by utility range and branching factor, or adjust backup strategy based on number of players.

## Constraints

- **Game-parameter-adaptive, not game-specific**: Use `game_` properties to adapt behavior, but don't hardcode logic for specific games
- **Pure math/algorithms only**: Focus on exploration-exploitation tradeoffs
- **Keep method signatures**: Don't change the function signatures
- **Valid C++**: Code must compile with clang++ and be syntactically correct
- **Thread-safe**: Use atomic operations for node field access (loads with `std::memory_order_relaxed`, updates via `AtomicAddValue`)
- **VirtualLoss is a pure function**: `Compute` must not hold state between calls -- the engine handles all locking
- **Value Perspective (Max^n)**: General-sum, up to 10 players: never assume two-player zero-sum. Never negate values or compute opponent value as `-q`. Backup receives the full per-player return vector; the default convention adds `values[node->player]`.
- **Read-Only Advanced Fields**: You may read `is_terminal` (acquire) and then `exact_values` to short-circuit proven nodes; never write these fields or `successor`.

{lazy_prompt}
ONLY EVER RETURN CODE IN A *SEARCH/REPLACE BLOCK*!

# Task
{task_instruction} {focus_sentence} {trigger_chain_of_thought}
Describe each change with a *SEARCH/REPLACE block*.
\end{lstlisting}

\subsection{Structural Diversity and Mechanics Across Independent Evolutionary Runs}
\label{app:five_lineages}

To evaluate whether algorithmic discovery is robust or idiosyncratic to a single evolutionary trajectory, we analyze the representative mechanisms evolved across all five independent runs comprising our multi-run aggregate in Table~\ref{tab:stratified_sco_tiers} and Table~\ref{tab:per_run_sco_breakdown}.
Crucially, all five runs began from the exact same minimal PUCT seed ($m_{\mathrm{seed}}$, Listing~\ref{lst:seed_code}), which contained no FPU, a static exploration constant ($c_{\mathrm{puct}} = 2.0$), and unweighted visit-count action selection.

Table~\ref{tab:lineage_vs_prior_art} provides a comparative analysis of the mathematical mechanisms discovered across the five independent runs.
While the system prompt alerted the LLM to the necessity of handling unvisited children explicitly (e.g., via FPU) and suggested adapting exploration to generic game properties, the evolutionary search autonomously discovered the exact mathematical functional forms, parameterizations, and scheduling dynamics:
\begin{itemize}[leftmargin=*,noitemsep]
    \item \textbf{Exploration Schedules}: While Runs 1, 4, and 5 evolved logarithmic schedules scaled by utility range $\Delta U$, Run 3 (\bifv) dynamically normalized evaluations into $[0, 1]$ while scaling exploration by branching factor and partial observability, Run 4 incorporated an early saturation knee (base 50 instead of 19652) to accelerate early-game branch coverage, and Run 2 synthesized a purely algebraic rational schedule ($1.25 + 1.75 \frac{N}{N + \max(100, 0.5 \cdot S)}$) that entirely eliminates transcendental logarithm invocations.
    \item \textbf{Unvisited Child Initialization (FPU)}: Runs 1 and 3 anchored unvisited estimates on the neutral utility origin ($Q_0 = 0.0$ in raw utility for Run 1; $Q_0 = -u_{\min}/\Delta U$ in normalized utility for Run 3, matching the $0.5$ midpoint in symmetric zero-sum games) damped by action priors (with Run 3 dynamically annealing the penalty with subtree visits), whereas Runs 2, 4, and 5 discovered sibling-consensus mechanisms that aggregate empirical returns across traversed action edges. Run 4 derived a prior-weighted sibling average ($\frac{\sum \pi_i Q_i}{\sum \pi_i}$), Run 5 derived a visit-weighted sibling average ($\frac{\sum v_i Q_i}{\sum v_i}$), and Run 2 combined sibling range tracking with inverse-square-root visit decay.
    \item \textbf{Value Backup Dynamics}: While Runs 1, 2, 3, and 5 used classical return accumulation, Run 4 synthesized a polynomial learning rate $\gamma_t = (n+1)^{-0.82}$ for Robbins-Monro style value averaging, and Run 2 introduced a rational depth discount to favor shorter winning paths.
\end{itemize}

\subsubsection{Addressing Algorithmic Novelty vs. Memorization in Discovered Search}
\label{app:novelty_vs_memorization}

A critical question in algorithmic discovery via large language models is whether an automated search system synthesizes genuinely novel coordinating mechanisms or merely recalls human-engineered heuristics from literature. 
Specifically, one might hypothesize that mechanisms like dynamic exploration schedules, First-Play Urgency (FPU)~\citep{gelly2006exploration}, Progressive Bias~\citep{chaslot2008progressive}, Single-Player MCTS~\citep{schadd2008single}, or utility min-max normalization are simply memorized fragments of established single-domain systems such as AlphaZero~\citep{Silver18AlphaZero}, KataGo~\citep{wu2019accelerating}, or Leela Chess Zero (Lc0).

A rigorous comparative analysis of the five independent runs reveals why this memorization hypothesis fails to explain the empirical phenomena:
\begin{enumerate}[leftmargin=*,noitemsep]
    \item \textbf{Minimalist Seed Baseline}: All five evolutionary runs were initialized from the identical minimal PUCT seed ($m_{\mathrm{seed}}$, Listing~\ref{lst:seed_code}), which contained \emph{no FPU whatsoever} (unvisited nodes were expanded immediately), a \emph{static exploration scalar} ($c_{\mathrm{puct}} = 2.0$), and unweighted visit-count action selection. The system was never prompted with or seeded from AlphaZero, KataGo, or Lc0 source code.
    \item \textbf{Cross-Game Meta-Adaptive Scaling vs. Single-Domain Tuning}: In human-engineered engines like AlphaZero and KataGo, hyperparameters like root Dirichlet noise concentration $\alpha$ and exploration parameters are \emph{manually tuned for specific individual games} (e.g., Dirichlet $\alpha=0.03$ for Go vs. $0.30$ for Chess in AlphaZero). In contrast, the discovered mechanisms synthesize \emph{meta-adaptive operators} that dynamically scale across $>400$ structurally heterogeneous games. In \bifv, exploration scales logarithmically with the decision branching factor ($1 + 0.05 \ln |A|$) and scales by $1.25$ under partial observability ($\mathbb{I}_{\mathrm{imperfect}}$). While the evolution prompt and C++ substrate interface (Appendix~\ref{app:outer_prompt}) expose standard game reflection APIs (\texttt{game.NumDistinctActions()} and \texttt{game.GetType().information}) and illustrative conceptual suggestions (e.g., adapting to branching factor or utility range), the specific mathematical functional parameterizations---such as logarithmic branching compression $(1 + 0.05 \ln |A|)$, the partial-observability scaling factor $1.25^{\mathbb{I}_{\mathrm{imperfect}}}$, maturity-annealed FPU damping, and value-augmented tie-breaking ($\mathbf{V}$, which was never suggested in the prompt)---were derived autonomously by the evolutionary loop without human specification. Neither AlphaZero nor KataGo nor Lc0 features branching-space adaptive scaling, value-augmented tie-breaking, or partial observability adjustments.
    \item \textbf{Emergence of Non-Standard Mathematical Paradigms}: Rather than converging on a single standard formulation, independent evolutionary runs synthesized non-standard functional compositions of foundational search concepts. As detailed below, these include purely algebraic rational exploration schedules that eliminate transcendental logarithms entirely (Run 2), non-stationary Robbins-Monro polynomial value backup operators (Run 4), and visit-weighted sibling consensus leaf evaluators that completely decouple FPU from parent node estimates (Run 5).
\end{enumerate}

\subsubsection{Detailed Mechanistic Analysis of Case Study: \bifv\ (\texorpdfstring{\mstarBIFV}{m*\_BIFV})}
\label{app:mechanics_bifv}

The discovered mechanism \bifv\ (Listing~\ref{lst:discovered_code} and Section~\ref{sec:exp:generalization}) factorizes cross-domain search into four mutually orthogonal mathematical operators.
At internal tree nodes, exploration scaling follows:
\begin{equation}
    c_{\mathrm{puct}}(N, |A|, \mathbb{I}_{\mathrm{imp}}) \;=\; \underbrace{\left(1.25 + 0.25 \ln \frac{N + 19652}{19652}\right)}_{\text{Base Progress Schedule}} \;\cdot\; \underbrace{(1 + 0.05 \ln |A|)}_{\textbf{B: Branching-Aware Scaling}} \;\cdot\; \underbrace{1.25^{\mathbb{I}_{\mathrm{imperfect}}}}_{\textbf{I: Information Scaling}},
\end{equation}
while at the search root, \code{RootActionSelector} additionally injects adaptive Dirichlet noise ($\alpha = \mathrm{clip}(10/|A|, 0.05, 1.0), \epsilon_{\mathrm{noise}}(t) = 0.20(1 - t/T)$) and modulates $c_{\mathrm{puct}}$ by a budget progress factor $[1 + 0.20(1 - t/T)]$ to encourage wider initial branch discovery across parallel threads before narrowing onto principal variations;
paired with a maturity-annealed leaf policy:
\begin{equation}
    Q_{\mathrm{FPU}}(a) \;=\; Q_0 \;-\; \sqrt{1 - \pi(a \mid s)} \;\cdot\; \underbrace{\left(0.1 + 0.4 \frac{N}{100 + N}\right)}_{\textbf{F: Maturity-Annealed Damping}},
\end{equation}
and value-augmented root action selection:
\begin{equation}
    a^\star \;=\; \arg\max_a \;\underbrace{\big[N(a) \;+\; 0.1 \cdot \widetilde{Q}(a)\big]}_{\textbf{V: Value-Augmented Selection}},
\end{equation}
where $\widetilde{Q}(a) = \frac{Q(a) - u_{\min}}{u_{\max} - u_{\min}}$ denotes the dynamic utility-bounded normalization mapping expected returns into the unit interval $[0, 1]$, and $Q_0 = \frac{-u_{\min}}{u_{\max} - u_{\min}}$ denotes the normalized neutral utility origin ($0.5$ in symmetric zero-sum games).

\paragraph{Novelty and Theoretical Rationale.}
\begin{itemize}[leftmargin=*,noitemsep]
    \item \textbf{Branching Adaptation ($\mathbf{B}$)}: Standard PUCT assumes a fixed exploration budget per simulation regardless of the number of legal actions. In games with massive action spaces (e.g., Shogi with distinct action count $|A| \approx 600$ and legal branching factor $b \approx 80\text{--}120$, or Go with $|A| \approx 361$), a static $c_{\mathrm{puct}}$ causes the tree policy to starve unvisited but high-prior branches. By scaling exploration logarithmically with $\ln |A|$, \bifv\ preserves sufficient exploration pressure to expand critical tactical moves in broad decision spaces without over-exploring narrow endgames.
    \item \textbf{Information Scaling ($\mathbf{I}$)}: Classic MCTS algorithms developed for Go and Chess assume perfect observability. In imperfect-information games (e.g., Poker, Phantom Go, Skat), deterministic tree commitments risk severe exploitability against deceptive opponents. The $1.25^{\mathbb{I}_{\mathrm{imperfect}}}$ scalar inflates exploration entropy specifically when game observations are partial, ensuring mixed-strategy coverage over hidden states.
    \item \textbf{Maturity-Annealed FPU ($\mathbf{F}$)}: In \code{ModularMCTSBot} (App.~\ref{app:template:lifecycle}), the search graph separates state-level visit counting from edge-level value accumulation: DAG state nodes $\nu$ serve as structural containers tracking total state visits $N(\nu)$ (\texttt{node->visits}) and outgoing legal edges, while \code{ValueBackup} accumulates return statistics strictly on directed action edges (\texttt{node->children[a]}). In Listing~\ref{lst:discovered_code}, the ternary expression \texttt{(n\_parent > 0) ? parent.q\_value() : (min\_u + max\_u) * 0.5} initializes unvisited actions at the interval midpoint $\frac{u_{\min} + u_{\max}}{2}$ (normalized to $0.5$) on the very first simulation ($N=0$), and for all subsequent simulations ($N \ge 1$) where \texttt{parent.q\_value()} on the state container evaluates to $0.0$, settles on the normalized neutral utility origin $Q_0 = -u_{\min}/\Delta U$ (which coincide identically at $0.5$ in all symmetric zero-sum games). While Leela Chess Zero employs a static FPU reduction proportional to unvisited prior mass ($c_{\mathrm{fpu}} \sqrt{1 - \sum \pi}$), \bifv\ discovers that the required FPU penalty is non-stationary: early in search ($N \ll 100$), sibling action-edge estimates $\widetilde{Q}(s, a)$ are noisy, requiring a soft penalty ($0.1$) relative to the neutral baseline $Q_0$ to encourage diverse child discovery. As parent visit maturity accumulates ($N \gg 100$), the penalty dynamically increases to $0.5$, aggressively suppressing low-prior moves and concentrating search depth on verified principal variations.
    \item \textbf{Value Tie-Breaking ($\mathbf{V}$)}: In time-constrained search (e.g., 50\,ms), shallow subtrees frequently produce visit-count ties ($N(a_1) = N(a_2)$). Standard argmax arbitrarily picks the first index, inducing positional bias. While Runs 1, 2, and 4 also evolved micro-scale epsilon tie-breakers ($10^{-4}\text{--}10^{-5}$) to arbitrate exact integer visit ties between visited actions, \bifv\ elevated this mechanism to a macro-scale utility weighting ($0.1 \cdot \widetilde{Q}(a)$), guaranteeing that visit ties are broken in favor of the highest expected value and overriding shallow sampling noise.
\end{itemize}

\paragraph{Leave-One-Out Operator Ablation of \bifv.}
\label{app:factorial_ablation}
To estimate the contribution of each operator ($\mathbf{B}$, $\mathbf{V}$, $\mathbf{F}$, and $\mathbf{I}$) without conflating domain heuristic changes or cross-tournament SCO scale shifts, we evaluate each single-operator deletion against the exact same pool of 15 tuned MCTS baselines across all four benchmark tiers under the 50\,ms IsoTime budget. Because the 15 baseline-to-baseline empirical payoff tensors are held fixed from Seed~3, the average pairwise AlphaRank ballot win rate ($\overline{F}(\text{Variant} \succ \text{Baselines})$) is directly comparable (same baseline pool) across rows (Table~\ref{tab:factorial_ablation_bifv}).

\begin{table}[h]
\centering
\small
\caption{\textbf{Operator ablation of \bifv\ (50\,ms IsoTime).}
Average pairwise ballot win rate (\%) against the same 15 baselines; since the baseline pool is identical, rows are directly comparable.
First row: full mechanism. Other rows: change when one operator is removed (positive = removal helps, negative = removal hurts).}
\label{tab:factorial_ablation_bifv}
\begin{tabular}{lcccc}
\toprule
 & \textbf{Training (50)} & \textbf{Held-out (19)} & \textbf{Synthesized (360)} & \textbf{Neural PPO (17)} \\
\midrule
Full \bifv\ ($\mstarBIFV$)          & 71.0 & 66.8 & 64.8 & 78.0 \\
\midrule
$\setminus\mathbf{B}$ (branching)   & $-8.5$ & $-0.4$  & $+5.3$ & $-5.3$ \\
$\setminus\mathbf{V}$ (tie-break)   & $+9.0$ & $-5.7$  & $+4.9$ & $-2.4$ \\
$\setminus\mathbf{F}$ (FPU anneal)  & $+9.2$ & $+13.9$ & $+5.9$ & $-5.8$ \\
$\setminus\mathbf{I}$ (info scale)  & $+10.5$ & $+10.6$ & $+5.0$ & $-1.0$ \\
\bottomrule
\end{tabular}
\end{table}

\paragraph{Findings.}
No single operator is uniformly beneficial. Removing any operator lowers win rate on the Neural PPO tier ($\Delta \in [-5.8, -1.0]$), whereas most removals match or improve win rate on the symbolic tiers; in particular, every deletion improves the 360 synthesized games by a similar margin ($+4.9$ to $+5.9$). We therefore interpret the operators as acting jointly rather than as independent improvements, with the full combination offering its clearest advantage when paired with learned neural priors.
One hypothesis is that \bifv\ was evolved under fixed simulation budgets, so some operators may be miscalibrated for the 50\,ms wall-clock regime; this remains untested.

\subsubsection{Detailed Mechanistic Analysis of Run 1 (Seed 1)}
\label{app:mechanics_lineage1}

Run 1 (Listing~\ref{lst:seed1_code}) discovered a search policy driven directly by the environment's dynamic utility range $\Delta U = u_{\max} - u_{\min}$:
\begin{align}
    c_{\mathrm{puct}}(N) &\;=\; \Delta U \;\cdot\; \left(1.25 \;+\; \ln \frac{N + 19652}{19652}\right), \\
    Q_{\mathrm{FPU}}(a) &\;=\; Q_0 \;-\; 0.3 \cdot \Delta U \;\cdot\; \sqrt{1 - \pi(a \mid s)},
\end{align}
where $Q_0 = 0.0$ denotes the neutral utility origin under edge-accumulating DAG search (Listing~\ref{lst:seed1_code}).

\paragraph{Novelty and Mechanism.}
In standard PUCT, value estimates are either mapped to $[0, 1]$ via min-max normalization or assumed to be bounded in $[-1, 1]$. Run 1 dispenses with normalization tables entirely: it directly multiplies the logarithmic exploration schedule and the FPU prior penalty by the native utility span $\Delta U$. This guarantees scale invariance across games with arbitrary payoffs (e.g., zero-sum $[-1, 1]$ vs. scoring games $[0, 1000]$).
Furthermore, Run 1 synthesizes an \emph{instant terminal value short-circuit}: when a child node is terminal ($v=0, \text{is\_terminal}=\text{true}$), it immediately returns the exact terminal utility $u_{\mathrm{exact}}$ rather than an FPU estimate, preventing wasteful exploration of already-proven terminal leaves.
At the search root, Run 1 additionally derives an adaptive Dirichlet noise schedule with quadratic progress decay ($\epsilon = 0.25(1 - i/S)^2$, concentration $\alpha = \max(0.1, 10 / |\mathcal{A}|)$) to stimulate thread exploration early before settling onto principal variations.

\subsubsection{Detailed Mechanistic Analysis of Run 2 (Seed 2)}
\label{app:mechanics_lineage2}

Run 2 (Listing~\ref{lst:seed2_code}) introduces a complete departure from transcendental logarithmic exploration and parent-anchored FPU:
\begin{align}
    c_{\mathrm{puct}}(N) &\;=\; 1.25 \;+\; 1.75 \;\cdot\; \frac{N}{N + \max(100.0, 0.5 \cdot S)}, \\
    Q_{\mathrm{FPU}}(a) &\;=\; \bar{Q}_{\mathrm{vis}} \;-\; 0.15 \cdot (1 - \pi(a \mid s)) \;\cdot\; \sqrt{\frac{1}{1 + |\mathcal{A}_{\mathrm{vis}}|}}, \\
    \pi_{\mathrm{blend}}(a) &\;=\; 0.85 \cdot \pi(a \mid s) \;+\; \frac{0.15}{|\mathcal{A}|},
\end{align}
where $\bar{Q}_{\mathrm{vis}} = \frac{1}{|\mathcal{A}_{\mathrm{vis}}|} \sum_{a' \in \mathcal{A}_{\mathrm{vis}}} Q(a')$ denotes the mean return of visited sibling nodes (in the C++ implementation, sibling returns are dynamically normalized across the observed sibling range with a non-negativity clamp), $|\mathcal{A}_{\mathrm{vis}}|$ is the number of visited sibling actions, $\pi_{\mathrm{blend}}(a)$ prevents policy lockout, and $S$ is the simulation budget. In the value backup, Run 2 synthesizes a \emph{rational depth-discount operator}:
\begin{equation}
    \Delta Q(s) \;=\; \frac{R}{1.0 \;+\; 0.001 \cdot d(s)},
\end{equation}
where $d(s)$ is the tree depth of state $s$ relative to the search root.

\paragraph{Novelty and Mechanism.}
\begin{itemize}[leftmargin=*,noitemsep]
    \item \textbf{Algebraic Rational Schedule}: Standard MCTS schedules rely on $\ln(N)$, which incurs transcendental floating-point computation on every tree traversal. Run 2 derived a strictly bounded, monotone algebraic rational function $1.25 + 1.75 \frac{N}{N+\max(100, 0.5 \cdot S)}$. This function matches the asymptotic shape of logarithmic exploration ($c_{\mathrm{init}}=1.25, c_{\mathrm{asymptotic}}=3.0$) while executing in a fraction of the clock cycles, maximizing simulation throughput under strict wall-clock time limits.
    \item \textbf{Sibling Consensus Leaf Evaluation and Blended Priors}: Rather than anchoring unvisited actions to a static origin baseline $Q_0 = 0.0$ (which sits at the minimum floor $u_{\min}$ in positive-payoff scoring games), Run 2 iterates over the already-visited action edges (\code{parent.children}) where \code{ValueBackup} records empirical returns, evaluating unvisited children relative to the empirical mean of their sibling cohort ($\bar{Q}_{\mathrm{vis}}$). The prior penalty $(1 - \pi(a))$ is damped inversely by visited sibling actions count $\sqrt{1 / (1 + |\mathcal{A}_{\mathrm{vis}}|)}$, causing unvisited children to be rapidly explored early, but strictly filtered once sibling evidence establishes a high baseline. Furthermore, it blends the raw heuristic prior with a uniform distribution ($\pi_{\mathrm{blend}} = 0.85\pi + 0.15/|\mathcal{A}|$) to guarantee exploration coverage.
    \item \textbf{Rational Depth-Discounted Backup}: In combinatorial game trees with deep transpositions, equal-value terminal payoffs create search indifference between fast wins and delayed wins. Run 2's depth-discount operator introduces an implicit time-preference, prioritizing shorter, tactically direct paths to victory.
\end{itemize}

\subsubsection{Detailed Mechanistic Analysis of Run 4 (Seed 4)}
\label{app:mechanics_lineage4}

Run 4 (Listing~\ref{lst:seed4_code}) synthesizes an early-knee exploration schedule paired with stochastic approximation value backups:
\begin{align}
    c_{\mathrm{puct}}(N, i) &\;=\; \rho_{\mathrm{root}}(i) \;\cdot\; \Delta U \;\cdot\; \left(1.25 \;+\; 0.5 \ln \frac{N + 50.0}{50.0}\right), \quad \rho_{\mathrm{root}}(i) \;=\; 1.4 \;-\; 0.8 \cdot \frac{i}{S}, \\
    Q_{\mathrm{FPU}}(a) &\;=\; \frac{\sum_{i \in \mathrm{vis}} \pi_i Q_i}{\sum_{i \in \mathrm{vis}} \pi_i} \;-\; 0.25 \cdot \Delta U \;\cdot\; \sqrt{\sum_{i \in \mathrm{vis}} \pi_i},
\end{align}
where $\rho_{\mathrm{root}}(i)$ is a linear progress annealing factor applied during root action selection ($i \in [0, S]$ is the simulation index).
Most notably, in its backup operator, Run 4 discards standard arithmetic averaging ($Q_{t+1} = Q_t + \frac{1}{n+1}(R - Q_t)$) in favor of a \emph{Robbins-Monro stochastic approximation} step size:
\begin{equation}
    Q_{t+1} \;=\; Q_t \;+\; \gamma_t \cdot (R - Q_t), \qquad \gamma_t \;=\; (n_t + 1)^{-0.82}.
\end{equation}

\paragraph{Novelty and Mechanism.}
\begin{itemize}[leftmargin=*,noitemsep]
    \item \textbf{Robbins-Monro Non-Stationary Filtering}: In simulation-based search, early rollouts are notoriously noisy and unrepresentative because the tree policy has not yet identified principal lines. Standard arithmetic averaging ($\gamma_t = \frac{1}{n+1}$) gives equal mathematical weight to these early flawed simulations. Run 4 synthesizes a polynomial learning rate $\gamma_t = (n+1)^{-0.82}$ (satisfying the classical Robbins-Monro convergence criterion $\sum \gamma_t = \infty, \sum \gamma_t^2 < \infty$). Because $-0.82$ decays slower than harmonic averaging ($\gamma_t = (n+1)^{-1.0}$), this operator maintains substantial step sizes longer, acting as a non-stationary adaptive filter: it aggressively discounts early noisy rollouts and allows mature, policy-directed simulations to dominate the converged value estimate.
(We note that in asynchronous parallel tree search, non-linear fractional updates $Q \leftarrow Q + \gamma_t (R - Q)$ require atomic compare-and-swap or edge mutex guards to prevent read-compute-write races, whereas additive accumulation supports standard lock-free atomic increments.)
    \item \textbf{Fast-Saturating Early-Knee Schedule}: While AlphaZero sets the schedule pivot at $N = 19652$ (appropriate for millions of simulations), practical budget-constrained search operates in $N \in [100, 2000]$. Run 4 sets the knee at $N=50$, rapidly ramping exploration constant $c_{\mathrm{puct}}$ during the first 50 visits to force immediate multi-path verification before settling into stable exploitation.
    \item \textbf{Prior-Weighted Sibling Consensus}: Unvisited child values are anchored to the prior-weighted mean of existing siblings $\frac{\sum \pi_i Q_i}{\sum \pi_i}$, ensuring that high-prior siblings have proportionally higher influence in setting the initial bar for unvisited leaves.
\end{itemize}

\subsubsection{Detailed Mechanistic Analysis of Run 5 (Seed 5)}
\label{app:mechanics_lineage5}

Run 5 (Listing~\ref{lst:seed5_code}) synthesizes an elegant, empirical visit-weighted sibling expectation model for FPU:
\begin{align}
    c_{\mathrm{puct}}(N) &\;=\; \Delta U \;\cdot\; \left(1.25 \;+\; \ln \left(\frac{N + 19653.0}{19652.0}\right)\right), \\
    Q_{\mathrm{FPU}}(a) &\;=\; \frac{\sum_{i \in \mathrm{vis}} v_i \cdot Q_i}{\sum_{i \in \mathrm{vis}} v_i} \;\equiv\; \mathbb{E}_{v}[Q_{\mathrm{siblings}}].
\end{align}

\paragraph{Novelty and Mechanism.}
In positive-payoff scoring games or non-zero-sum environments, a static origin baseline $Q_0 = 0.0$ lies at the minimum payoff floor $u_{\min}$, offering no information about local board standing. Run 5 completely decouples unvisited leaf initialization from static origins: it queries the empirical returns stored on traversed action edges (\code{parent.children}), initializing unvisited children to the visit-weighted expectation of its sibling cohort $\mathbb{E}_{v}[Q_{\mathrm{siblings}}]$. Because visits $v_i$ naturally concentrate on promising sibling moves, this expectation reflects the search engine's \emph{validated consensus} of local board utility. Run 5 achieved a $71.1\%$ win rate against reference PUCT ($285 / 401$ decisive non-drawn games), closely matching Run 2 ($71.4\%$, $279 / 391$ games, Appendix~\ref{app:pairwise_ballots}), demonstrating the empirical power of sibling expectation filtering across decisive game matchups.

\subsubsection{Taxonomic Comparison of Discovered Search Programs Against Prior Art}
\label{app:taxonomic_comparison}

Table~\ref{tab:lineage_vs_prior_art} provides a systematic architectural comparison contrasting all five discovered evolutionary runs against canonical human-engineered MCTS algorithms (AlphaZero, KataGo, Leela Chess Zero, and Vanilla PUCT).

\begin{table*}[h]
\centering
\caption{\textbf{Comprehensive Taxonomic Comparison of Discovered Search Programs Against Prior Art.} Contrasts mathematical operators across exploration schedules, leaf evaluation (FPU), value backups, and action selection.}
\label{tab:lineage_vs_prior_art}
\small
\resizebox{\textwidth}{!}{%
\begin{tabular}{lllll}
\toprule
\textbf{Algorithm / Run} & \textbf{Exploration Schedule $c(N)$} & \textbf{Leaf Evaluation Policy ($Q_{\mathrm{FPU}}$)} & \textbf{Value Backup Operator} & \textbf{Key Algorithmic Trait} \\
\midrule
\multicolumn{5}{l}{\textit{Human-Engineered Prior Art}} \\
Evolution Seed Template ($m_{\mathrm{seed}}$) & Static scalar $c_{\mathrm{puct}} = 2.0$ & Naive immediate expansion ($v=0$) & Arithmetic mean $\frac{1}{N} \sum R$ & Minimal baseline template (evolution starting seed) \\
AlphaZero~\citep{Silver18AlphaZero} & $1.25 + \ln \frac{N+19652}{19652}$ & Parent return $Q_{\mathrm{parent}}$ & Arithmetic mean $\frac{1}{N} \sum R$ & Static single-game parameter schedule \\
KataGo~\citep{wu2019accelerating} & Dynamic visit-dependent schedule & Utility-bounded normalizer & Arithmetic mean $\frac{1}{N} \sum R$ & Utility range clamping $[u_{\min}, u_{\max}]$ \\
Leela Chess Zero (Lc0) & Tuned logarithmic schedule & $Q_{\mathrm{parent}} - c_{\mathrm{fpu}} \sqrt{1 - \sum \pi}$ & Arithmetic mean $\frac{1}{N} \sum R$ & Static prior-weighted FPU reduction \\
\midrule
\multicolumn{5}{l}{\textit{Discovered Programs from Evolutionary Runs (Ours)}} \\
\textbf{\bifv\ (Run 3, Case Study)} & $(1.25 + 0.25 \ln \frac{N+19652}{19652}) (1 + 0.05 \ln |A|) 1.25^{\mathbb{I}_{\mathrm{imp}}}$ & $Q_0 - \sqrt{1 - \pi(a)} (0.1 + 0.4 \frac{N}{100+N})$ & Additive atomic returns & Branching- and info-scaling; maturity-annealed FPU; value tie-break \\
\textbf{Run 1 (Seed 1)} & $\Delta U \cdot (1.25 + \ln \frac{N+19652}{19652})$ & $Q_0 - 0.3 \Delta U \sqrt{1 - \pi(a)}$ & Additive atomic returns & Direct utility span $\Delta U$ scaling; terminal short-circuit; root Dirichlet decay \\
\textbf{Run 2 (Seed 2)} & $1.25 + 1.75 \frac{N}{N+\max(100, 0.5S)}$ (Purely rational) & $\bar{Q}_{\mathrm{vis}} - 0.15 (1 - \pi(a)) \sqrt{\frac{1}{1 + |\mathcal{A}_{\mathrm{vis}}|}}$ & Rational depth discount $\frac{R}{1 + 0.001 d}$ & Eliminates transcendental $\ln$; sibling cohort FPU consensus \\
\textbf{Run 4 (Seed 4)} & $\Delta U \cdot (1.25 + 0.5 \ln \frac{N+50}{50})$ (Early-knee) & $\frac{\sum \pi_i Q_i}{\sum \pi_i} - 0.25 \Delta U \sqrt{\sum \pi_i}$ & Robbins-Monro $(n+1)^{-0.82}$ & Non-stationary polynomial backup filter; fast 50-visit coverage knee \\
\textbf{Run 5 (Seed 5)} & $\Delta U \cdot (1.25 + \ln \frac{N+19653}{19652})$ & $\frac{\sum v_i Q_i}{\sum v_i}$ (Visit-weighted expectation) & Additive atomic returns & Pure empirical sibling expectation $\mathbb{E}_v[Q]$; static-origin decoupling \\
\bottomrule
\end{tabular}}
\end{table*}

Listings~\ref{lst:seed1_code}, \ref{lst:seed2_code}, \ref{lst:seed4_code}, and \ref{lst:seed5_code} document the complete production C++ implementations of Runs 1, 2, 4, and 5.

\begin{lstlisting}[style=codeblock,caption={C++ implementation of Run 1 (Seed 1).},label={lst:seed1_code}]
using EvolvedNode = MCTSNode;

class EvolvedTreePolicy : public TreePolicy {
 public:
  using TreePolicy::TreePolicy;
  Action SelectChild(const MCTSNode& parent,
                     const std::vector<Action>& legal_actions,
                     std::mt19937* rng) override {
    Action best_action = legal_actions[0];
    double best_score = -1e18;
    double n_parent = parent.visits.load(std::memory_order_relaxed);
    double sqrt_parent = std::sqrt(std::max(1.0, n_parent));

    double utility_range = game_.MaxUtility() - game_.MinUtility();
    if (utility_range <= 0.0 || !std::isfinite(utility_range)) utility_range = 1.0;

    double c_puct = utility_range * (1.25 + std::log((n_parent + 19652.0) / 19652.0));
    double parent_q = (n_parent > 0) ? parent.q_value() : (0.5 * (game_.MinUtility() + game_.MaxUtility()));
    if (!std::isfinite(parent_q)) parent_q = 0.0;

    for (Action a : legal_actions) {
      auto it = parent.children.find(a);
      if (it == parent.children.end()) continue;
      const MCTSNode* c = it->second.get();
      int v = c->visits.load(std::memory_order_relaxed);

      double q = 0.0;
      bool is_term = c->is_terminal.load(std::memory_order_acquire);
      if (is_term && !c->exact_values.empty() && parent.player >= 0 &&
          parent.player < static_cast<int>(c->exact_values.size())) {
        q = c->exact_values[parent.player];
      } else if (v > 0) {
        q = c->q_value();
      } else {
        q = parent_q - 0.3 * utility_range * std::sqrt(std::max(0.0, 1.0 - c->prior));
      }

      double exploration = c_puct * c->prior * sqrt_parent / (1.0 + v);
      double score = q + exploration;
      if (score > best_score) {
        best_score = score;
        best_action = a;
      }
    }
    return best_action;
  }
};

class EvolvedRootActionSelector : public RootActionSelector {
 public:
  using RootActionSelector::RootActionSelector;
  Action SelectRootAction(const MCTSNode& root,
                          const std::vector<Action>& legal_actions,
                          int simulation_index, int num_simulations,
                          std::mt19937* rng) override {
    Action best_action = legal_actions[0];
    double best_score = -1e18;
    double n_parent = root.visits.load(std::memory_order_relaxed);
    double sqrt_parent = std::sqrt(std::max(1.0, n_parent));

    double utility_range = game_.MaxUtility() - game_.MinUtility();
    if (utility_range <= 0.0 || !std::isfinite(utility_range)) utility_range = 1.0;

    double c_puct = utility_range * (1.25 + std::log((n_parent + 19652.0) / 19652.0));
    double parent_q = (n_parent > 0) ? root.q_value() : (0.5 * (game_.MinUtility() + game_.MaxUtility()));
    if (!std::isfinite(parent_q)) parent_q = 0.0;

    std::vector<double> effective_priors(legal_actions.size());
    bool use_noise = (rng != nullptr && num_simulations > 0);

    if (use_noise) {
      double alpha = std::max(0.1, 10.0 / static_cast<double>(legal_actions.size()));
      double progress = (num_simulations > 1)
          ? static_cast<double>(simulation_index) / static_cast<double>(num_simulations - 1)
          : 0.0;
      progress = std::max(0.0, std::min(1.0, progress));
      double epsilon = 0.25 * (1.0 - progress) * (1.0 - progress);

      std::vector<double> gamma_samples(legal_actions.size());
      double gamma_sum = 0.0;
      for (size_t i = 0; i < legal_actions.size(); ++i) {
        gamma_samples[i] = SampleGamma(alpha, rng);
        gamma_sum += gamma_samples[i];
      }

      for (size_t i = 0; i < legal_actions.size(); ++i) {
        auto it = root.children.find(legal_actions[i]);
        double orig_prior = (it != root.children.end()) ? it->second->prior : (1.0 / legal_actions.size());
        double noise_val = (gamma_sum > 1e-9) ? (gamma_samples[i] / gamma_sum) : (1.0 / legal_actions.size());
        effective_priors[i] = (1.0 - epsilon) * orig_prior + epsilon * noise_val;
      }
    } else {
      for (size_t i = 0; i < legal_actions.size(); ++i) {
        auto it = root.children.find(legal_actions[i]);
        effective_priors[i] = (it != root.children.end()) ? it->second->prior : (1.0 / legal_actions.size());
      }
    }

    for (size_t i = 0; i < legal_actions.size(); ++i) {
      Action a = legal_actions[i];
      auto it = root.children.find(a);
      if (it == root.children.end()) continue;
      const MCTSNode* c = it->second.get();
      int v = c->visits.load(std::memory_order_relaxed);

      double q = 0.0;
      bool is_term = c->is_terminal.load(std::memory_order_acquire);
      if (is_term && !c->exact_values.empty() && root.player >= 0 &&
          root.player < static_cast<int>(c->exact_values.size())) {
        q = c->exact_values[root.player];
      } else if (v > 0) {
        q = c->q_value();
      } else {
        q = parent_q - 0.3 * utility_range * std::sqrt(std::max(0.0, 1.0 - c->prior));
      }

      double exploration = c_puct * effective_priors[i] * sqrt_parent / (1.0 + v);
      double score = q + exploration;
      if (score > best_score) {
        best_score = score;
        best_action = a;
      }
    }
    return best_action;
  }

 private:
  static double SampleGamma(double alpha, std::mt19937* rng) {
    if (alpha < 1.0) {
      double u = std::generate_canonical<double, 10>(*rng);
      return SampleGamma(alpha + 1.0, rng) * std::pow(std::max(1e-10, u), 1.0 / alpha);
    }
    constexpr double kPi = 3.14159265358979323846;
    double d = alpha - 1.0 / 3.0;
    double c = 1.0 / std::sqrt(9.0 * d);
    while (true) {
      double u1 = std::generate_canonical<double, 10>(*rng);
      double u2 = std::generate_canonical<double, 10>(*rng);
      double z = std::sqrt(-2.0 * std::log(std::max(1e-10, u1))) * std::cos(2.0 * kPi * u2);
      double v = 1.0 + c * z;
      if (v <= 0.0) continue;
      v = v * v * v;
      double u = std::generate_canonical<double, 10>(*rng);
      if (u < 1.0 - 0.0331 * z * z * z * z) return d * v;
      if (std::log(std::max(1e-10, u)) < 0.5 * z * z + d * (1.0 - v + std::log(v))) return d * v;
    }
  }
};

class EvolvedFinalActionSelector : public FinalActionSelector {
 public:
  using FinalActionSelector::FinalActionSelector;
  std::pair<int, std::vector<double>> SelectActionAndPolicy(
      const MCTSNode& root, const std::vector<Action>& actions,
      std::mt19937* rng) override {
    std::vector<double> policy(actions.size(), 0.0);
    double total = 0;
    int best = 0;
    double max_score = -1e18;

    double utility_range = game_.MaxUtility() - game_.MinUtility();
    if (utility_range <= 0.0 || !std::isfinite(utility_range)) utility_range = 1.0;

    for (size_t i = 0; i < actions.size(); ++i) {
      auto it = root.children.find(actions[i]);
      if (it != root.children.end()) {
        const MCTSNode* c = it->second.get();
        int v = c->visits.load(std::memory_order_relaxed);
        policy[i] = static_cast<double>(v);
        total += static_cast<double>(v);

        double q = 0.0;
        bool is_term = c->is_terminal.load(std::memory_order_acquire);
        if (is_term && !c->exact_values.empty() && root.player >= 0 &&
            root.player < static_cast<int>(c->exact_values.size())) {
          q = c->exact_values[root.player];
        } else if (v > 0) {
          q = c->q_value();
        } else {
          q = (root.visits.load(std::memory_order_relaxed) > 0 ? root.q_value() : 0.0);
        }
        double score = static_cast<double>(v) + 1e-4 * (q / utility_range);
        if (score > max_score) {
          max_score = score;
          best = i;
        }
      }
    }
    if (total > 0) {
      for (auto& p : policy) p /= total;
    }
    return {best, policy};
  }
};

class EvolvedValueBackup : public ValueBackup {
 public:
  using ValueBackup::ValueBackup;
  void Backup(MCTSNode* node, const std::vector<double>& values,
              int depth, int acting_player) override {
    if (acting_player >= 0 &&
        acting_player < static_cast<int>(values.size())) {
      node->AtomicAddValue(values[acting_player]);
    }
    node->visits.fetch_add(1, std::memory_order_relaxed);
  }
};

class EvolvedVirtualLoss : public VirtualLoss {
 public:
  EvolvedVirtualLoss(const open_spiel::Game& game, int num_sims)
      : VirtualLoss(game, num_sims),
        vl_value_(game.MinUtility() - game.MaxUtility()) {}
  VirtualLossDeltas Compute(const MCTSNode& node, int depth) const override {
    return {vl_value_, 1};
  }
 private:
  double vl_value_;
};
\end{lstlisting}

\begin{lstlisting}[style=codeblock,caption={C++ implementation of Run 2 (Seed 2).},label={lst:seed2_code}]
using EvolvedNode = MCTSNode;

class EvolvedTreePolicy : public TreePolicy {
 public:
  using TreePolicy::TreePolicy;
  Action SelectChild(const MCTSNode& parent,
                     const std::vector<Action>& legal_actions,
                     std::mt19937* rng) override {
    double n_parent = parent.visits.load(std::memory_order_relaxed);
    double sqrt_parent = std::sqrt(std::max(1.0, n_parent));
    
    int visited_count = 0;
    double sib_min = 1e18;
    double sib_max = -1e18;
    for (Action a : legal_actions) {
      auto it = parent.children.find(a);
      if (it != parent.children.end()) {
        int v = it->second->visits.load(std::memory_order_relaxed);
        if (v > 0) {
          double q = it->second->q_value();
          visited_count++;
          if (q < sib_min) sib_min = q;
          if (q > sib_max) sib_max = q;
        }
      }
    }
    
    double min_q = game_.MinUtility();
    double max_q = game_.MaxUtility();
    if (visited_count > 0) {
      double weight = std::min(1.0, visited_count / 10.0);
      min_q = weight * sib_min + (1.0 - weight) * game_.MinUtility();
      max_q = weight * sib_max + (1.0 - weight) * game_.MaxUtility();
    }
    double q_range = max_q - min_q;
    if (q_range < 1e-5) {
      q_range = std::max(1e-5, game_.MaxUtility() - game_.MinUtility());
    }
    
    double sum_norm_q = 0.0;
    for (Action a : legal_actions) {
      auto it = parent.children.find(a);
      if (it != parent.children.end()) {
        int v = it->second->visits.load(std::memory_order_relaxed);
        if (v > 0) {
          double q_norm = (it->second->q_value() - min_q) / q_range;
          sum_norm_q += q_norm;
        }
      }
    }
    
    double avg_norm_q = (visited_count > 0) ? (sum_norm_q / visited_count) : 0.5;
    double pb_c = std::max(100.0, static_cast<double>(num_simulations_) * 0.5);
    double c_puct = 1.25 + 1.75 * n_parent / (n_parent + pb_c);
    
    Action best_action = legal_actions.empty() ? 0 : legal_actions[0];
    double best_score = -1e18;
    bool found_any = false;
    double uniform_prior = 1.0 / std::max<double>(1.0, legal_actions.size());
    
    for (Action a : legal_actions) {
      auto it = parent.children.find(a);
      if (it == parent.children.end()) continue;
      const MCTSNode* c = it->second.get();
      int v = c->visits.load(std::memory_order_relaxed);
      double exploitation;
      if (v > 0) {
        exploitation = (c->q_value() - min_q) / q_range;
      } else {
        double fpu_penalty = 0.15 * (1.0 - c->prior) * std::sqrt(1.0 / (1.0 + visited_count));
        exploitation = std::max(0.0, avg_norm_q - fpu_penalty);
      }
      double blended_prior = 0.85 * c->prior + 0.15 * uniform_prior;
      double exploration = c_puct * blended_prior * sqrt_parent / (1.0 + v);
      double score = exploitation + exploration;
      if (!found_any || score > best_score) {
        best_score = score;
        best_action = a;
        found_any = true;
      }
    }
    return best_action;
  }
};

class EvolvedRootActionSelector : public RootActionSelector {
 public:
  using RootActionSelector::RootActionSelector;
  Action SelectRootAction(const MCTSNode& root,
                          const std::vector<Action>& legal_actions,
                          int simulation_index, int num_simulations,
                          std::mt19937* rng) override {
    double n_parent = root.visits.load(std::memory_order_relaxed);
    double sqrt_parent = std::sqrt(std::max(1.0, n_parent));
    
    int visited_count = 0;
    double sib_min = 1e18;
    double sib_max = -1e18;
    for (Action a : legal_actions) {
      auto it = root.children.find(a);
      if (it != root.children.end()) {
        int v = it->second->visits.load(std::memory_order_relaxed);
        if (v > 0) {
          double q = it->second->q_value();
          visited_count++;
          if (q < sib_min) sib_min = q;
          if (q > sib_max) sib_max = q;
        }
      }
    }
    
    double min_q = game_.MinUtility();
    double max_q = game_.MaxUtility();
    if (visited_count > 0) {
      double weight = std::min(1.0, visited_count / 10.0);
      min_q = weight * sib_min + (1.0 - weight) * game_.MinUtility();
      max_q = weight * sib_max + (1.0 - weight) * game_.MaxUtility();
    }
    double q_range = max_q - min_q;
    if (q_range < 1e-5) {
      q_range = std::max(1e-5, game_.MaxUtility() - game_.MinUtility());
    }
    
    double sum_norm_q = 0.0;
    for (Action a : legal_actions) {
      auto it = root.children.find(a);
      if (it != root.children.end()) {
        int v = it->second->visits.load(std::memory_order_relaxed);
        if (v > 0) {
          double q_norm = (it->second->q_value() - min_q) / q_range;
          sum_norm_q += q_norm;
        }
      }
    }
    
    double avg_norm_q = (visited_count > 0) ? (sum_norm_q / visited_count) : 0.5;
    double pb_c = std::max(100.0, static_cast<double>(num_simulations) * 0.5);
    double c_puct = 1.25 + 1.75 * n_parent / (n_parent + pb_c);
    
    Action best_action = legal_actions.empty() ? 0 : legal_actions[0];
    double best_score = -1e18;
    bool found_any = false;
    double uniform_prior = 1.0 / std::max<double>(1.0, legal_actions.size());
    
    for (Action a : legal_actions) {
      auto it = root.children.find(a);
      if (it == root.children.end()) continue;
      const MCTSNode* c = it->second.get();
      int v = c->visits.load(std::memory_order_relaxed);
      double exploitation;
      if (v > 0) {
        exploitation = (c->q_value() - min_q) / q_range;
      } else {
        double fpu_penalty = 0.15 * (1.0 - c->prior) * std::sqrt(1.0 / (1.0 + visited_count));
        exploitation = std::max(0.0, avg_norm_q - fpu_penalty);
      }
      double blended_prior = 0.85 * c->prior + 0.15 * uniform_prior;
      double exploration = c_puct * blended_prior * sqrt_parent / (1.0 + v);
      double score = exploitation + exploration;
      if (!found_any || score > best_score) {
        best_score = score;
        best_action = a;
        found_any = true;
      }
    }
    return best_action;
  }
};

class EvolvedFinalActionSelector : public FinalActionSelector {
 public:
  using FinalActionSelector::FinalActionSelector;
  std::pair<int, std::vector<double>> SelectActionAndPolicy(
      const MCTSNode& root, const std::vector<Action>& actions,
      std::mt19937* rng) override {
    std::vector<double> policy(actions.size(), 0.0);
    double total = 0;
    for (size_t i = 0; i < actions.size(); ++i) {
      auto it = root.children.find(actions[i]);
      if (it != root.children.end()) {
        int v = it->second->visits.load(std::memory_order_relaxed);
        policy[i] = v;
        total += v;
      }
    }
    if (total > 0) {
      for (auto& p : policy) p /= total;
    }
    
    int best = 0;
    double best_score = -1e18;
    double q_range = game_.MaxUtility() - game_.MinUtility();
    double min_q = game_.MinUtility();
    double denom = q_range < 1e-5 ? 1e-5 : q_range;
    for (size_t i = 0; i < actions.size(); ++i) {
      auto it = root.children.find(actions[i]);
      if (it != root.children.end()) {
        int v = it->second->visits.load(std::memory_order_relaxed);
        double q = v > 0 ? it->second->q_value() : min_q;
        double norm_q = (q - min_q) / denom;
        // Primary key: visit count, Secondary key: normalized Q-value to break ties robustly
        double score = v + 1e-5 * norm_q;
        if (score > best_score) {
          best_score = score;
          best = i;
        }
      }
    }
    return {best, policy};
  }
};

class EvolvedValueBackup : public ValueBackup {
 public:
  using ValueBackup::ValueBackup;
  void Backup(MCTSNode* node, const std::vector<double>& values,
              int depth, int acting_player) override {
    if (acting_player >= 0 &&
        acting_player < static_cast<int>(values.size())) {
      // Apply a tiny rational depth-based discount to encourage shorter paths to wins
      // and delay losses, avoiding expensive std::pow calls.
      double discount = 1.0 / (1.0 + 0.001 * depth);
      node->AtomicAddValue(values[acting_player] * discount);
    }
    node->visits.fetch_add(1, std::memory_order_relaxed);
  }
};

class EvolvedVirtualLoss : public VirtualLoss {
 public:
  EvolvedVirtualLoss(const open_spiel::Game& game, int num_sims)
      : VirtualLoss(game, num_sims),
        vl_value_(game.MinUtility() - game.MaxUtility()) {}
  VirtualLossDeltas Compute(const MCTSNode& node, int depth) const override {
    return {vl_value_, 1};
  }
 private:
  double vl_value_;
};
\end{lstlisting}

\begin{lstlisting}[style=codeblock,caption={C++ implementation of Run 4 (Seed 4).},label={lst:seed4_code}]
using EvolvedNode = MCTSNode;

class EvolvedTreePolicy : public TreePolicy {
 public:
  using TreePolicy::TreePolicy;
  Action SelectChild(const MCTSNode& parent,
                     const std::vector<Action>& legal_actions,
                     std::mt19937* rng) override {
    Action best_action = legal_actions[0];
    double best_score = -1e18;
    double n_parent = parent.visits.load(std::memory_order_relaxed);
    double sqrt_parent = std::sqrt(std::max(1.0, n_parent));

    double min_u = game_.MinUtility();
    double max_u = game_.MaxUtility();
    double range_u = (max_u > min_u) ? (max_u - min_u) : 1.0;
    double c_puct = (1.25 + 0.5 * std::log((n_parent + 50.0) / 50.0)) * range_u;

    double visited_q_weighted_sum = 0.0;
    double visited_p_sum = 0.0;

    for (Action a : legal_actions) {
      auto it = parent.children.find(a);
      if (it == parent.children.end()) continue;
      const MCTSNode* c = it->second.get();
      int v = c->visits.load(std::memory_order_relaxed);
      if (v > 0) {
        visited_q_weighted_sum += c->q_value() * c->prior;
        visited_p_sum += c->prior;
      }
    }

    double fpu_base = (visited_p_sum > 0.0) ? (visited_q_weighted_sum / visited_p_sum)
                      : (n_parent > 0)     ? parent.q_value()
                                           : 0.5 * (min_u + max_u);
    double fpu = fpu_base - 0.25 * range_u * std::sqrt(std::max(0.0, visited_p_sum));

    for (Action a : legal_actions) {
      auto it = parent.children.find(a);
      if (it == parent.children.end()) continue;
      const MCTSNode* c = it->second.get();
      int v = c->visits.load(std::memory_order_relaxed);

      double q = (v > 0) ? c->q_value() : fpu;
      double exploration = c_puct * c->prior * sqrt_parent / (1.0 + v);
      double score = q + exploration;

      if (score > best_score) {
        best_score = score;
        best_action = a;
      }
    }
    return best_action;
  }
};

class EvolvedRootActionSelector : public RootActionSelector {
 public:
  using RootActionSelector::RootActionSelector;
  Action SelectRootAction(const MCTSNode& root,
                          const std::vector<Action>& legal_actions,
                          int simulation_index, int num_simulations,
                          std::mt19937* rng) override {
    Action best_action = legal_actions[0];
    double best_score = -1e18;
    double n_parent = root.visits.load(std::memory_order_relaxed);
    double sqrt_parent = std::sqrt(std::max(1.0, n_parent));

    double min_u = game_.MinUtility();
    double max_u = game_.MaxUtility();
    double range_u = (max_u > min_u) ? (max_u - min_u) : 1.0;

    // Root exploration schedule: smooth transition from broad discovery to refined exploitation
    double progress = (num_simulations > 0)
                          ? static_cast<double>(simulation_index) / num_simulations
                          : 0.0;
    double root_scale = 1.4 - 0.8 * progress;
    double c_puct = root_scale * (1.25 + 0.5 * std::log((n_parent + 50.0) / 50.0)) * range_u;

    double visited_q_weighted_sum = 0.0;
    double visited_p_sum = 0.0;

    for (Action a : legal_actions) {
      auto it = root.children.find(a);
      if (it == root.children.end()) continue;
      const MCTSNode* c = it->second.get();
      int v = c->visits.load(std::memory_order_relaxed);
      if (v > 0) {
        visited_q_weighted_sum += c->q_value() * c->prior;
        visited_p_sum += c->prior;
      }
    }

    double fpu_base = (visited_p_sum > 0.0) ? (visited_q_weighted_sum / visited_p_sum)
                      : (n_parent > 0)     ? root.q_value()
                                           : 0.5 * (min_u + max_u);
    double fpu = fpu_base - 0.25 * range_u * std::sqrt(std::max(0.0, visited_p_sum));

    for (Action a : legal_actions) {
      auto it = root.children.find(a);
      if (it == root.children.end()) continue;
      const MCTSNode* c = it->second.get();
      int v = c->visits.load(std::memory_order_relaxed);

      double q = (v > 0) ? c->q_value() : fpu;
      double exploration = c_puct * c->prior * sqrt_parent / (1.0 + v);
      double score = q + exploration;

      if (score > best_score) {
        best_score = score;
        best_action = a;
      }
    }
    return best_action;
  }
};

class EvolvedFinalActionSelector : public FinalActionSelector {
 public:
  using FinalActionSelector::FinalActionSelector;
  std::pair<int, std::vector<double>> SelectActionAndPolicy(
      const MCTSNode& root, const std::vector<Action>& actions,
      std::mt19937* rng) override {
    std::vector<double> policy(actions.size(), 0.0);
    double total = 0;
    int best = 0;
    double best_score = -1e18;
    for (size_t i = 0; i < actions.size(); ++i) {
      auto it = root.children.find(actions[i]);
      if (it != root.children.end()) {
        int v = it->second->visits.load(std::memory_order_relaxed);
        policy[i] = v;
        total += v;
        double q = (v > 0) ? it->second->q_value() : game_.MinUtility();
        double score = v + 1e-4 * q;
        if (score > best_score) {
          best_score = score;
          best = static_cast<int>(i);
        }
      }
    }
    if (total > 0) {
      for (auto& p : policy) p /= total;
    }
    return {best, policy};
  }
};

class EvolvedValueBackup : public ValueBackup {
 public:
  using ValueBackup::ValueBackup;
  void Backup(MCTSNode* node, const std::vector<double>& values,
              int depth, int acting_player) override {
    double sample_val = 0.0;
    if (acting_player >= 0 &&
        acting_player < static_cast<int>(values.size())) {
      sample_val = values[acting_player];
    } else if (!values.empty()) {
      double avg = 0.0;
      for (double v : values) avg += v;
      sample_val = avg / values.size();
    }

    int n = node->visits.load(std::memory_order_relaxed);
    double delta = sample_val;
    if (n > 0) {
      double q_old = node->q_value();
      // Polynomial decay learning rate gamma_t = (n + 1)^(-0.82)
      double gamma = std::pow(n + 1.0, -0.82);
      double q_new = q_old + gamma * (sample_val - q_old);
      
      // Safety bounds clamp
      q_new = std::max(game_.MinUtility(), std::min(game_.MaxUtility(), q_new));
      
      // Value delta required so total_value / (n + 1) == q_new
      delta = q_old + (n + 1.0) * (q_new - q_old);
    }

    node->AtomicAddValue(delta);
    node->visits.fetch_add(1, std::memory_order_relaxed);
  }
};

class EvolvedVirtualLoss : public VirtualLoss {
 public:
  EvolvedVirtualLoss(const open_spiel::Game& game, int num_sims)
      : VirtualLoss(game, num_sims),
        vl_value_(std::min(-1.0, game.MinUtility() - game.MaxUtility())) {}
  VirtualLossDeltas Compute(const MCTSNode& node, int depth) const override {
    return {vl_value_, 1};
  }
 private:
  double vl_value_;
};
\end{lstlisting}

\begin{lstlisting}[style=codeblock,caption={C++ implementation of Run 5 (Seed 5).},label={lst:seed5_code}]
using EvolvedNode = MCTSNode;

class EvolvedTreePolicy : public TreePolicy {
 public:
  EvolvedTreePolicy(const open_spiel::Game& game, int num_sims) 
      : TreePolicy(game, num_sims),
        utility_range_(game.MaxUtility() - game.MinUtility()),
        min_utility_(game.MinUtility()) {}
        
  Action SelectChild(const MCTSNode& parent,
                     const std::vector<Action>& legal_actions,
                     std::mt19937* rng) override {
    Action best_action = legal_actions[0];
    double best_score = -1e18;
    double n_parent = parent.visits.load(std::memory_order_relaxed);
    double sqrt_parent = std::sqrt(n_parent > 0.0 ? n_parent : 1.0);
    
    double c_puct = 1.25 + std::log((n_parent + 19652.0 + 1.0) / 19652.0);
    double scale = (utility_range_ > 0.0) ? utility_range_ : 1.0;

    double total_q_weighted = 0.0;
    int total_v = 0;
    for (Action a : legal_actions) {
      auto it = parent.children.find(a);
      if (it == parent.children.end()) continue;
      const MCTSNode* c = it->second.get();
      int v = c->visits.load(std::memory_order_relaxed);
      if (v > 0) {
        total_q_weighted += c->total_value.load(std::memory_order_relaxed);
        total_v += v;
      }
    }
    double fpu_value = (total_v > 0) ? (total_q_weighted / total_v) : min_utility_;

    for (Action a : legal_actions) {
      auto it = parent.children.find(a);
      if (it == parent.children.end()) continue;
      const MCTSNode* c = it->second.get();
      int v = c->visits.load(std::memory_order_relaxed);
      
      double exploitation = (v > 0) ? c->q_value() : fpu_value;
      double exploration = c_puct * scale * c->prior * sqrt_parent / (1.0 + v);
      double score = exploitation + exploration;
      
      if (score > best_score) {
        best_score = score;
        best_action = a;
      }
    }
    return best_action;
  }
 private:
  double utility_range_;
  double min_utility_;
};

class EvolvedRootActionSelector : public RootActionSelector {
 public:
  EvolvedRootActionSelector(const open_spiel::Game& game, int num_sims)
      : RootActionSelector(game, num_sims),
        utility_range_(game.MaxUtility() - game.MinUtility()),
        min_utility_(game.MinUtility()) {}
        
  Action SelectRootAction(const MCTSNode& root,
                          const std::vector<Action>& legal_actions,
                          int simulation_index, int num_simulations,
                          std::mt19937* rng) override {
    Action best_action = legal_actions[0];
    double best_score = -1e18;
    double n_parent = root.visits.load(std::memory_order_relaxed);
    double sqrt_parent = std::sqrt(n_parent > 0.0 ? n_parent : 1.0);
    
    double c_puct = 1.25 + std::log((n_parent + 19652.0 + 1.0) / 19652.0);
    double scale = (utility_range_ > 0.0) ? utility_range_ : 1.0;

    double total_q_weighted = 0.0;
    int total_v = 0;
    for (Action a : legal_actions) {
      auto it = root.children.find(a);
      if (it == root.children.end()) continue;
      const MCTSNode* c = it->second.get();
      int v = c->visits.load(std::memory_order_relaxed);
      if (v > 0) {
        total_q_weighted += c->total_value.load(std::memory_order_relaxed);
        total_v += v;
      }
    }
    double fpu_value = (total_v > 0) ? (total_q_weighted / total_v) : min_utility_;

    for (Action a : legal_actions) {
      auto it = root.children.find(a);
      if (it == root.children.end()) continue;
      const MCTSNode* c = it->second.get();
      int v = c->visits.load(std::memory_order_relaxed);
      
      double exploitation = (v > 0) ? c->q_value() : fpu_value;
      double exploration = c_puct * scale * c->prior * sqrt_parent / (1.0 + v);
      double score = exploitation + exploration;
      
      if (score > best_score) {
        best_score = score;
        best_action = a;
      }
    }
    return best_action;
  }
 private:
  double utility_range_;
  double min_utility_;
};

class EvolvedFinalActionSelector : public FinalActionSelector {
 public:
  using FinalActionSelector::FinalActionSelector;
  std::pair<int, std::vector<double>> SelectActionAndPolicy(
      const MCTSNode& root, const std::vector<Action>& actions,
      std::mt19937* rng) override {
    std::vector<double> policy(actions.size(), 0.0);
    double total = 0;
    for (size_t i = 0; i < actions.size(); ++i) {
      auto it = root.children.find(actions[i]);
      if (it != root.children.end()) {
        int v = it->second->visits.load(std::memory_order_relaxed);
        policy[i] = static_cast<double>(v);
        total += policy[i];
      }
    }
    if (total > 0) {
      for (auto& p : policy) p /= total;
    } else {
      for (auto& p : policy) p = 1.0 / actions.size();
    }
    int best = std::max_element(policy.begin(), policy.end()) - policy.begin();
    return {best, policy};
  }
};

class EvolvedValueBackup : public ValueBackup {
 public:
  using ValueBackup::ValueBackup;
  void Backup(MCTSNode* node, const std::vector<double>& values,
              int depth, int acting_player) override {
    if (acting_player >= 0 &&
        acting_player < static_cast<int>(values.size())) {
      node->AtomicAddValue(values[acting_player]);
    }
    node->visits.fetch_add(1, std::memory_order_relaxed);
  }
};

class EvolvedVirtualLoss : public VirtualLoss {
 public:
  EvolvedVirtualLoss(const open_spiel::Game& game, int num_sims)
      : VirtualLoss(game, num_sims),
        vl_value_(game.MinUtility() - game.MaxUtility()) {}
  VirtualLossDeltas Compute(const MCTSNode& node, int depth) const override {
    return {vl_value_, 1};
  }
 private:
  double vl_value_;
};
\end{lstlisting}

\section{Deep Neural Network Setup: PPO Training, Architectures, and Selection Protocol}
\label{app:neural_ppo_setup}

This appendix provides the full technical specification of the reinforcement learning pipeline used to train deep neural policy and value networks for the \code{neural\_ppo} evaluation tier.

\subsection{Multi-Player PPO Formulation and Loss Functions}

Standard Proximal Policy Optimization (PPO) is conventionally formulated for single-agent or two-player zero-sum environments with scalar state values.
To benchmark search generalisation across general extensive-form games with arbitrary player counts and non-zero-sum payoffs, we formulate self-play PPO with \emph{vector-valued state evaluation} and \emph{acting-player action masking}.

Let $s$ denote a game state with legal action set $\mathcal{A}(s) \subseteq \mathcal{A}$, active player $p(s) \in \{1, \dots, N\}$, and terminal payoff vector $\mathbf{u} \in \mathbb{R}^N$.
The neural network parameterized by $(\theta, \phi)$ outputs:
(i)~unnormalized policy action logits $\mathbf{z}_\theta(s) \in \mathbb{R}^{|\mathcal{A}|}$, and
(ii)~a vector value estimate $\mathbf{v}_\phi(s) \in \mathbb{R}^N$ predicting expected utilities for all player seats simultaneously.

\paragraph{Action Masking and Policy Distribution.}
To strictly prevent sampling illegal moves during exploration, policy logits are masked prior to softmax normalization:
\begin{equation}
  \tilde{z}_a(s) = \begin{cases} z_a(s), & \text{if } a \in \mathcal{A}(s), \\ -\infty, & \text{if } a \notin \mathcal{A}(s), \end{cases} \qquad \pi_\theta(a \mid s) = \frac{\exp(\tilde{z}_a(s))}{\sum_{a' \in \mathcal{A}(s)} \exp(\tilde{z}_{a'}(s))}.
\end{equation}

\paragraph{Vector Generalized Advantage Estimation (GAE).}
During self-play rollouts, transitions $(s_t, a_t, \mathbf{r}_t, s_{t+1}, p_t)$ are accumulated in a shared trajectory buffer.
Temporal difference residuals are computed seat-wise across all players:
\begin{equation}
  \boldsymbol{\delta}_t = \mathbf{r}_t + \gamma \mathbf{v}_\phi(s_{t+1}) \cdot \mathbb{I}(s_{t+1} \notin \mathcal{S}_{\mathrm{term}}) - \mathbf{v}_\phi(s_t),
\end{equation}
where $\gamma = 0.99$ is the discount factor.
Vector GAE advantages $\hat{\mathbf{A}}_t \in \mathbb{R}^N$ are computed recursively:
\begin{equation}
  \hat{\mathbf{A}}_t = \boldsymbol{\delta}_t + \gamma \lambda \cdot \mathbb{I}(s_{t+1} \notin \mathcal{S}_{\mathrm{term}}) \hat{\mathbf{A}}_{t+1},
\end{equation}
with $\lambda = 1.0$.
When updating the policy of acting player $p_t = p(s_t)$, the scalar advantage is extracted via projection: $\hat{A}_t = \hat{\mathbf{A}}_t[p_t]$.
Advantages are normalized across each training batch: $\hat{A}_t \leftarrow (\hat{A}_t - \mu_A) / (\sigma_A + 10^{-8})$.

\paragraph{Optimization Losses.}
The surrogate policy objective clips probability ratios $r_t(\theta) = \frac{\pi_\theta(a_t \mid s_t)}{\pi_{\theta_{\mathrm{old}}}(a_t \mid s_t)}$ to prevent destructive policy updates:
\begin{equation}
  \mathcal{L}_{\mathrm{CLIP}}(\theta) = -\hat{\mathbb{E}}_t \left[ \min\Big( r_t(\theta) \hat{A}_t,\, \mathrm{clip}(r_t(\theta), 1 - \epsilon, 1 + \epsilon) \hat{A}_t \Big) \right],
\end{equation}
with clipping threshold $\epsilon = 0.3$.
The vector value function is trained via mean squared error across all player components:
\begin{equation}
  \mathcal{L}_{\mathrm{VF}}(\phi) = \hat{\mathbb{E}}_t \left[ \frac{1}{N} \sum_{i=1}^N \Big( \hat{R}_{t, i} - v_{\phi, i}(s_t) \Big)^2 \right],
\end{equation}
where $\hat{\mathbf{R}}_t = \hat{\mathbf{A}}_t + \mathbf{v}_\phi(s_t)$ is the vector target return.
Entropy regularization is enforced over legal moves:
\begin{equation}
  \mathcal{L}_{\mathrm{ENT}}(\theta) = -\hat{\mathbb{E}}_t \left[ \sum_{a \in \mathcal{A}(s_t)} \pi_\theta(a \mid s_t) \log \pi_\theta(a \mid s_t) \right].
\end{equation}
The composite loss is:
\begin{equation}
  \mathcal{L}_{\mathrm{PPO}}(\theta, \phi) = \mathcal{L}_{\mathrm{CLIP}}(\theta) + c_{\mathrm{vf}} \mathcal{L}_{\mathrm{VF}}(\phi) - \beta_{\mathrm{ent}} \mathcal{L}_{\mathrm{ENT}}(\theta),
\end{equation}
with value loss coefficient $c_{\mathrm{vf}} = 1.0$.
Models are optimized using Adam with a trajectory batch size of $B = 4000$ transitions, minibatch size of 128, and 10 to 30 epochs per update.

\subsection{Neural Architecture Design by Game Topology}

Because extensive-form games exhibit disparate state representations, we design two distinct neural architecture families tailored to game geometry (Table~\ref{tab:app_neural_architectures}):

\begin{table*}[h]
\centering
\small
\caption{\textbf{Neural Network Architectures Tailored by Game Topology.} Model hyperparameters and structural specifications across the two architecture families used in the PPO benchmark.}
\label{tab:app_neural_architectures}
\resizebox{\textwidth}{!}{%
\setlength{\tabcolsep}{5pt}
\begin{tabular}{lll}
\toprule
\textbf{Design Dimension} & \textbf{Vector / Tabular Architecture (MLP)} & \textbf{Spatial / Board Architecture (CNNResNet)} \\
\midrule
Target Game Families & Card, trick-taking, imperfect-information, coordination & 2D grid board games, connection, spatial territory \\
Evaluated Games (17 Total) & Contract Bridge, Hearts, Gin Rummy, Leduc Poker, & Chess, Shogi, Go 9$\times$9, Go 13$\times$13, \\
& Backgammon, Bargaining, Coop Box, Phantom TTT, Dark Hex & Hex, Othello, Checkers, Gomoku \\
\midrule
Input Representation & Flat 1D feature vector $\mathbf{x} \in \mathbb{R}^D$ & 3D tensor $\mathbf{X} \in \mathbb{R}^{H \times W \times C}$ (board planes) \\
Backbone Structure & 2 Fully Connected layers (Dense-256) & 1 Initial Conv ($3\times 3$, 64 ch) + 4 ResNet Blocks \\
Residual Block Design & --- & Two Conv-$3\times 3$ (64 ch) + skip connection \\
Normalization & None & None (avoids non-stationarity in RL rollouts) \\
Hidden Activations & ReLU & ReLU \\
\midrule
Policy Head & Dense($|\mathcal{A}|$) linear logits & Conv-$1\times 1$ (2 ch) $\to$ Flatten $\to$ Dense($|\mathcal{A}|$) \\
Value Head & Dense($N$) $\to \tanh \times \text{scale}$ & Conv-$1\times 1$ (1 ch) $\to$ Dense(64) $\to$ Dense($N$) $\to \tanh \times \text{scale}$ \\
Value Scaling & Bounded to theoretical utility range & Bounded to $[-1, 1]$ utility range \\
\bottomrule
\end{tabular}}
\end{table*}

\subsection{Hyperparameter Sweeps, Validation Protocol, and Checkpoint Selection}

To produce fair, robust neural checkpoints that reflect stable asymptotic capability rather than transient lucky spikes, all models were trained under systematic hyperparameter sweeps across three random seeds ($S = 3$ replicates per setting):
\begin{itemize}[leftmargin=1.5em, itemsep=2pt, topsep=2pt]
  \item \textbf{Learning Rate}: $\eta \in \{5 \times 10^{-5}, 1 \times 10^{-4}, 3 \times 10^{-4}\}$.
  \item \textbf{Entropy Regularization}: $\beta_{\mathrm{ent}} \in \{0.003, 0.01, 0.03\}$ for imperfect-information and card games (where entropy is essential to preserve stochastic mixed strategies), and $\beta_{\mathrm{ent}} = 0.0$ for deterministic board games.
  \item \textbf{Optimization Epochs}: $E \in \{10, 30\}$ optimization epochs per batch update.
\end{itemize}

\paragraph{Two-Stage Checkpoint Selection Protocol.}
During training, checkpoints were periodically logged and evaluated against reference bots.
Selecting checkpoints via raw training returns is prone to overfitting and selection bias.
We enforce a strict two-stage selection protocol:
\begin{enumerate}[leftmargin=1.5em, itemsep=2pt, topsep=2pt]
  \item \textbf{Stage 1: Trajectory Smoothing and Configuration Ranking}:
  For each configuration and seed, evaluation curves are smoothed using a rolling average with window $w = 3$.
  We compute the mean learning curve across the three seeds, $\bar{R}(t) = \frac{1}{3} \sum_{s=1}^3 R_s(t)$.
  Configurations are ranked by the peak of their across-seed mean curve, $\max_t \bar{R}(t)$.
  \item \textbf{Stage 2: Median Seed Checkpoint Extraction}:
  For the winning hyperparameter configuration, let $t^\star = \arg\max_t \bar{R}(t)$ denote the step where the across-seed mean attains its maximum.
  Rather than picking a single highest-performing run at $t^\star$ (which would risk cherry-picking transient spikes), we select the checkpoint from the \emph{median-performing seed} at step $t^\star$.
\end{enumerate}
These selected median checkpoints represent the frozen neural policies $\pi_\theta$ and value estimators $\mathbf{v}_\phi$ deployed in the \code{neural\_ppo} generalization benchmark.

\subsection{Zero-Shot Transfer Dynamics and MCTS Integration}

During MCTS evaluation, each candidate search algorithm uses the frozen neural network as its inner heuristic:
\begin{equation}
  P(s, a) = \pi_\theta(a \mid s), \qquad V(s) = v_{\phi, p(s)}(s).
\end{equation}
As reported in Section~\ref{sec:exp:generalization}, $\mevo$ achieves an aggregate SCO rating of $\mathbf{508.91 \pm 0.67}$, outperforming reference PUCT ($508.05 \pm 1.12$) and 14 other competitive planning baselines, with \bifv\ (\mstarBIFV) capturing \textbf{Rank~\#1 in AlphaRank+SCO} in its seed tournament (rating $\mathbf{510.31}$).

\paragraph{Mechanistic Analysis of Search-Prior Interaction.}
Standard PUCT integrates neural priors via the polynomial exploration term:
\begin{equation}
  U_{\mathrm{PUCT}}(s, a) = c_{\mathrm{puct}} P(s, a) \frac{\sqrt{N(s)}}{1 + N(s, a)}.
\end{equation}
When the neural network exhibits overconfidence on out-of-distribution board states (assigning $P(s, a) \approx 0$ to sharp, tactical refutations that were rare in self-play), the polynomial bonus for $a$ is crushed to zero.
Consequently, PUCT requires an impractical number of simulations before exploring $a$, leading to severe tree polarization and blindness to tactical traps.

In contrast, $\mevo$ addresses neural overconfidence through two coordinated mechanisms:
first, at the root, decaying adaptive Dirichlet noise ($\epsilon_{\mathrm{noise}}(t) = 0.20(1 - t/T)$, with $\alpha = \mathrm{clip}(10.0/|A|, 0.05, 1.0)$) perturbs overly narrow policy priors early in search, injecting exploratory visits into low-probability branches;
second, during tree descent, maturity-annealed FPU ($Q_{\mathrm{FPU}}(a) = Q_0 - \sqrt{1 - \pi(a \mid s)} \cdot (0.1 + 0.4 \frac{N}{100 + N})$) bounds initial exploration penalties, ensuring that unvisited actions with low priors remain eligible for exploration before subtree visits accumulate.
Together with branching-aware exploration scaling, these mechanisms prevent severe tree polarization and enable $\mevo$ to reliably escape neural blind spots.

\section{Benchmarking Against Specialized Domain Engines and Ten-Game Empirical Payoff Tensors}
\label{app:engine_benchmarks}

To evaluate whether procedural search mechanisms ($m^\star$) and domain knowledge modules ($\kappa^\star$) discovered via modular co-evolution generalize beyond generic MCTS baselines against specialized, domain-specific game engines and learned neural policies, we construct full $5 \times 5$ empirical game tensors across \textbf{ten classic board and imperfect-information card games}: Chess (\texttt{Stockfish}), Checkers (\texttt{Kingsrow}), Othello (\texttt{Egaroucid}), Gomoku (\texttt{Gomoku Engine}), Shogi (\texttt{YaneuraOu}), Hex $13\times 13$ (\texttt{KataHex}), Go $9\times 9$ (\texttt{KataGo}), Go $13\times 13$ (\texttt{KataGo}), Hearts (\texttt{Xinxin}), and Contract Bridge (\texttt{WBridge5}).

\subsection{Experimental Protocol, Difficulty Tiers, and Five-Strategy Formulation}
\label{app:engine:sweeps}
\label{app:engine:tournament}

\paragraph{Five-Strategy Empirical Game Tensor Formulation.}
For every game $g$, difficulty tier $\tau \in \{\text{Floor}, \text{Mid}, \text{Ceiling}\}$, evolution lineage $k \in \{1, \dots, 5\}$, and search simulation budget $S \in \{100, 200, 400, 800, 1600\}$, we evaluate all directed off-diagonal pairings among five canonical agent architectures:
\begin{itemize}[leftmargin=1.5em, itemsep=2pt, topsep=2pt]
    \item \textbf{$\mathcal{S}_0$: Engine}: The external domain engine. Nine of the engines are configured at calibrated difficulty tiers $\tau$ (\textbf{Floor}: Stockfish Skill~0 / 1k nodes, Kingsrow 0.05\,s, Egaroucid Level~1, Gomoku 1k nodes, YaneuraOu Depth~1, KataHex 1 visit, KataGo 1 visit, Xinxin 10 runs; \textbf{Mid}: Stockfish Skill~10 / 20k nodes, Kingsrow 0.2\,s, Egaroucid Level~7, Gomoku 20k nodes, YaneuraOu Depth~5, KataHex 20 visits, KataGo 20 visits, Xinxin 25 runs; \textbf{Ceiling}: Stockfish Skill~20 / 200k nodes, Kingsrow 1.0\,s, Egaroucid Level~15, Gomoku 100k nodes, YaneuraOu Depth~10, KataHex 200 visits, KataGo 200 visits, Xinxin 50 runs). Contract Bridge (\texttt{WBridge5}) does not expose internal difficulty parameters and is evaluated under its fixed standard tournament setting across all runs.
    \item \textbf{$\mathcal{S}_1$: PPO} (\texttt{ppo\_only}): Unguided neural network policy executing greedy action selection $\operatorname{argmax}_a \pi_{\text{PPO}}(a \mid s)$ from lightweight actor-critic checkpoints trained via multi-agent PPO (small ResNets for board games, 2-layer MLPs for Hearts and Contract Bridge; $S=0$). These compact neural models provide challenging baseline priors for search amplification but are not scaled, compute-intensive AlphaZero-grade models.
    \item \textbf{$\mathcal{S}_2$: Heuristic Policy} (\texttt{evolved\_policy\_only}): Unguided symbolic prior policy executing $\operatorname{argmax}_a \pi_{\kappa_g^\star}(a \mid s)$ using the evolved prior and action-scoring rules of lineage $k$ without tree search ($S=0$).
    \item \textbf{$\mathcal{S}_3$: Search + PPO} (\texttt{evolved\_outer\_ppo}): Hybrid neural-symbolic search pairing the evolved outer search mechanism $m^\star$ of lineage $k$ with the PPO neural value and prior network $\langle m^\star, \text{PPO} \rangle$ over $S$ simulations per move (4 search threads).
    \item \textbf{$\mathcal{S}_4$: Search + Heuristic} (\texttt{evolved\_outer\_heuristic}): Full co-adapted deployable agent $\Comp^\star = \langle m^\star, \kappa_g^\star \rangle$ pairing the evolved outer search mechanism $m^\star$ with the co-adapted domain evaluation and belief-resampling heuristic $\kappa_g^\star = (\pi_g^\star, v_g^\star, b_g^\star)$ of lineage $k$ over $S$ simulations per move (4 search threads).
\end{itemize}

\paragraph{Seat Balancing, Deterministic Trajectories, and Four-Hypothesis Decomposition.}
Each directed pairing $(A, B)$ is evaluated in both seating orders ($A$ as Player~0 and $B$ as Player~0), computing the seat-balanced macro-average score $\text{Score}(A, B) = \frac{1}{2}\big(\text{Score}_{A \to B} + \text{Score}_{B \to A}\big)$ where wins count as $1.0$, draws as $0.5$, and losses as $0.0$. Each directed pairing is evaluated across at least 100 matchups under balanced seat swaps ($A \text{ as Player~0 and } B \text{ as Player~0}$), initialized from canonical standard game starts. In games with deterministic transitions where unsearched greedy policies execute identical move sequences (or where early search $Q$-values remain flat across quiet moves, as in Chess $\mathcal{S}_3$ vs.\ $\mathcal{S}_1$), games from fixed initial states follow identical principal variations resulting in zero-variance repetition draws ($50.0\% \pm 0.0\%$), whereas empirical variance emerges once asymmetric heuristics or deeper tactical lookahead break opening symmetry. In Contract Bridge (a $2\text{v}2$ partnership game), agents form homogeneous partnerships ($\text{North-South } \{0, 2\} = A$ vs.\ $\text{East-West } \{1, 3\} = B$) sharing single-board duplicate contract points $\Delta$, whereas in Hearts (a 4-player individual trick-avoidance game where each player minimizes their own penalty points), two independent instances of $A$ and two independent instances of $B$ occupy alternating seats ($\{0, 2\}$ vs.\ $\{1, 3\}$) to measure pairwise per-player point margins $\Delta$. Across every game tensor, we test four core algorithmic hypotheses:
\begin{enumerate}[leftmargin=1.5em, itemsep=1pt, topsep=2pt]
    \item \textbf{H1 (Neural Search Amplification, $\mathcal{S}_3$ vs.\ $\mathcal{S}_1$):} Does evolved search $m^\star$ improve upon the unguided neural policy ($\mathcal{S}_3 > \mathcal{S}_1$)?
    \item \textbf{H2 (Heuristic Search Amplification, $\mathcal{S}_4$ vs.\ $\mathcal{S}_2$):} Does evolved search $m^\star$ improve upon the unguided symbolic heuristic ($\mathcal{S}_4 > \mathcal{S}_2$)?
    \item \textbf{H3 (Unsearched Prior Comparison, $\mathcal{S}_1$ vs.\ $\mathcal{S}_2$):} How does the PPO neural policy compare head-to-head against the unsearched symbolic prior ($\mathcal{S}_1 \text{ vs } \mathcal{S}_2$ at $S=0$)?
    \item \textbf{H4 (Searched Prior Comparison, $\mathcal{S}_3$ vs.\ $\mathcal{S}_4$):} When both priors are amplified by $m^\star$ over $S$ simulations, does co-evolved heuristic search outperform neural search ($\mathcal{S}_4 > \mathcal{S}_3$)?
\end{enumerate}

\begin{table}[ht]
\centering
\scriptsize
\caption{\textbf{Cross-Game Summary of Five-Strategy Empirical Game Tensors Across All 10 Benchmark Games ($S=100 \to 1600$, 5-Seed Mean $\pm$ SE).} Rows evaluate both neural search ($\mathcal{S}_3$) and co-evolved heuristic search ($\mathcal{S}_4$) directly against the external domain engine ($\mathcal{S}_0$, Floor tier; default setting for WBridge5), followed by the four core internal hypothesis pairings (\texttt{H1}--\texttt{H4}; \texttt{H3} is an unsearched $S=0$ baseline). Full 3-tier (\texttt{Floor}, \texttt{Mid}, \texttt{Ceiling}) and per-seed breakdowns appear in Tables~\ref{tab:tensor_chess}--\ref{tab:tensor_bridge}.}
\label{tab:engine_benchmark_summary}
\label{tab:engine_simulation_scaling}
\label{tab:engine_seed_consistency}
\label{tab:app_tournament_multipliers}
\label{tab:app_multiplayer_tournament}
\vspace{2pt}
\resizebox{\columnwidth}{!}{%
\begin{tabular}{llccccc}
\toprule
\textbf{Game (Engine)} & \textbf{Architecture Pairing} & \textbf{100 Sims} & \textbf{200 Sims} & \textbf{400 Sims} & \textbf{800 Sims} & \textbf{1600 Sims} \\
\midrule
\multirow{4}{*}{\textbf{1. Chess} (\texttt{Stockfish})}
 & $\mathcal{S}_4\text{(Search+Heur)}$ vs Engine & \textbf{6.7\%}$\pm$1.8 & 6.3\%$\pm$1.7 & 6.1\%$\pm$2.2 & 5.3\%$\pm$2.3 & 4.7\%$\pm$2.1 \\
 & $\mathcal{S}_3\text{(Search+PPO)}$ vs Engine & 0.0\%$\pm$0.0 & 0.2\%$\pm$0.1 & 0.1\%$\pm$0.1 & 0.0\%$\pm$0.0 & 0.0\%$\pm$0.0 \\
 & H1: $\mathcal{S}_3$ vs $\mathcal{S}_1$ / H2: $\mathcal{S}_4$ vs $\mathcal{S}_2$ & 50.0\% / 56.5\% & 50.0\% / 57.2\% & 50.0\% / \textbf{62.2\%} & 50.0\% / 55.0\% & 50.0\% / 60.5\% \\
 & H3: $\mathcal{S}_1$ vs $\mathcal{S}_2$ ($S=0$) / H4: $\mathcal{S}_3$ vs $\mathcal{S}_4$ & 0.8\% / 15.3\% & 0.8\% / 9.2\% & 0.8\% / 15.9\% & 0.8\% / 18.4\% & 0.8\% / \textbf{18.8\%} \\
\cmidrule(lr){1-7}
\multirow{4}{*}{\textbf{2. Checkers} (\texttt{Kingsrow})}
 & $\mathcal{S}_4\text{(Search+Heur)}$ vs Engine & 0.0\%$\pm$0.0 & 0.4\%$\pm$0.1 & 0.5\%$\pm$0.2 & 0.7\%$\pm$0.4 & \textbf{1.1\%}$\pm$0.2 \\
 & $\mathcal{S}_3\text{(Search+PPO)}$ vs Engine & 0.1\%$\pm$0.1 & 0.2\%$\pm$0.1 & 0.2\%$\pm$0.0 & 0.3\%$\pm$0.1 & 0.5\%$\pm$0.2 \\
 & H1: $\mathcal{S}_3$ vs $\mathcal{S}_1$ / H2: $\mathcal{S}_4$ vs $\mathcal{S}_2$ & 50.0\% / 82.2\% & 50.0\% / 94.8\% & 50.0\% / 97.1\% & 62.9\% / 98.2\% & \textbf{70.1\%} / \textbf{100.0\%} \\
 & H3: $\mathcal{S}_1$ vs $\mathcal{S}_2$ ($S=0$) / H4: $\mathcal{S}_3$ vs $\mathcal{S}_4$ & 60.0\% / 34.9\% & 60.0\% / 44.3\% & 60.0\% / 36.9\% & 60.0\% / 41.9\% & 60.0\% / \textbf{45.0\%} \\
\cmidrule(lr){1-7}
\multirow{4}{*}{\textbf{3. Othello} (\texttt{Egaroucid})}
 & $\mathcal{S}_4\text{(Search+Heur)}$ vs Engine & 32.2\%$\pm$9.5 & 33.7\%$\pm$11.4 & \textbf{40.6\%}$\pm$6.0 & 35.6\%$\pm$10.9 & 39.6\%$\pm$14.1 \\
 & $\mathcal{S}_3\text{(Search+PPO)}$ vs Engine & 2.7\%$\pm$1.3 & 0.5\%$\pm$0.4 & 1.6\%$\pm$1.4 & 0.8\%$\pm$0.7 & 1.6\%$\pm$1.2 \\
 & H1: $\mathcal{S}_3$ vs $\mathcal{S}_1$ / H2: $\mathcal{S}_4$ vs $\mathcal{S}_2$ & 47.7\% / 89.0\% & 53.3\% / 98.4\% & 86.6\% / \textbf{100.0\%} & \textbf{93.3\%} / 100.0\% & 87.5\% / \textbf{100.0\%} \\
 & H3: $\mathcal{S}_1$ vs $\mathcal{S}_2$ ($S=0$) / H4: $\mathcal{S}_3$ vs $\mathcal{S}_4$ & 4.1\% / 0.3\% & 4.1\% / 0.8\% & 4.1\% / \textbf{8.8\%} & 4.1\% / 6.2\% & 4.1\% / 3.7\% \\
\cmidrule(lr){1-7}
\multirow{4}{*}{\textbf{4. Gomoku} (\texttt{Gomoku})}
 & $\mathcal{S}_4\text{(Search+Heur)}$ vs Engine & 2.1\%$\pm$1.4 & 1.8\%$\pm$0.7 & 0.4\%$\pm$0.3 & 2.1\%$\pm$0.6 & \textbf{3.6\%}$\pm$1.9 \\
 & $\mathcal{S}_3\text{(Search+PPO)}$ vs Engine & 0.0\%$\pm$0.0 & 0.0\%$\pm$0.0 & 0.0\%$\pm$0.0 & 0.0\%$\pm$0.0 & 0.0\%$\pm$0.0 \\
 & H1: $\mathcal{S}_3$ vs $\mathcal{S}_1$ / H2: $\mathcal{S}_4$ vs $\mathcal{S}_2$ & 96.8\% / 82.8\% & 95.2\% / 87.8\% & \textbf{100.0\%} / 90.3\% & 99.8\% / 91.5\% & 99.9\% / \textbf{91.9\%} \\
 & H3: $\mathcal{S}_1$ vs $\mathcal{S}_2$ ($S=0$) / H4: $\mathcal{S}_3$ vs $\mathcal{S}_4$ & 0.0\% / 0.0\% & 0.0\% / 0.0\% & 0.0\% / 0.0\% & 0.0\% / 0.0\% & 0.0\% / 0.0\% \\
\cmidrule(lr){1-7}
\multirow{4}{*}{\textbf{5. Shogi} (\texttt{YaneuraOu})}
 & $\mathcal{S}_4\text{(Search+Heur)}$ vs Engine & 0.0\%$\pm$0.0 & 0.0\%$\pm$0.0 & 0.0\%$\pm$0.0 & 0.0\%$\pm$0.0 & \textbf{0.1\%}$\pm$0.1 \\
 & $\mathcal{S}_3\text{(Search+PPO)}$ vs Engine & 0.0\%$\pm$0.0 & 0.0\%$\pm$0.0 & 0.0\%$\pm$0.0 & 0.0\%$\pm$0.0 & 0.0\%$\pm$0.0 \\
 & H1: $\mathcal{S}_3$ vs $\mathcal{S}_1$ / H2: $\mathcal{S}_4$ vs $\mathcal{S}_2$ & 59.6\% / 96.5\% & 87.8\% / \textbf{99.8\%} & 81.3\% / \textbf{99.8\%} & 98.1\% / 99.7\% & \textbf{98.9\%} / \textbf{99.8\%} \\
 & H3: $\mathcal{S}_1$ vs $\mathcal{S}_2$ ($S=0$) / H4: $\mathcal{S}_3$ vs $\mathcal{S}_4$ & 0.4\% / \textbf{0.1\%} & 0.4\% / 0.0\% & 0.4\% / 0.0\% & 0.4\% / 0.1\% & 0.4\% / 0.0\% \\
\cmidrule(lr){1-7}
\multirow{4}{*}{\textbf{6. Hex $13\times 13$} (\texttt{KataHex})}
 & $\mathcal{S}_4\text{(Search+Heur)}$ vs Engine & 0.0\%$\pm$0.0 & 0.0\%$\pm$0.0 & 0.0\%$\pm$0.0 & 0.0\%$\pm$0.0 & 0.0\%$\pm$0.0 \\
 & $\mathcal{S}_3\text{(Search+PPO)}$ vs Engine & 0.0\%$\pm$0.0 & 0.0\%$\pm$0.0 & 0.0\%$\pm$0.0 & 0.0\%$\pm$0.0 & 0.0\%$\pm$0.0 \\
 & H1: $\mathcal{S}_3$ vs $\mathcal{S}_1$ / H2: $\mathcal{S}_4$ vs $\mathcal{S}_2$ & 81.1\% / 57.3\% & 88.7\% / 70.5\% & 98.6\% / 74.7\% & \textbf{99.6\%} / 81.0\% & 99.5\% / \textbf{97.1\%} \\
 & H3: $\mathcal{S}_1$ vs $\mathcal{S}_2$ ($S=0$) / H4: $\mathcal{S}_3$ vs $\mathcal{S}_4$ & 0.0\% / 0.0\% & 0.0\% / 0.0\% & 0.0\% / 0.0\% & 0.0\% / 0.0\% & 0.0\% / 0.0\% \\
\cmidrule(lr){1-7}
\multirow{4}{*}{\textbf{7. Go $9\times 9$} (\texttt{KataGo})}
 & $\mathcal{S}_4\text{(Search+Heur)}$ vs Engine & 1.2\%$\pm$0.5 & 3.1\%$\pm$1.4 & 5.2\%$\pm$2.0 & 8.1\%$\pm$3.0 & \textbf{9.6\%}$\pm$2.7 \\
 & $\mathcal{S}_3\text{(Search+PPO)}$ vs Engine & 0.1\%$\pm$0.1 & 0.1\%$\pm$0.1 & 0.1\%$\pm$0.1 & 0.2\%$\pm$0.2 & 0.1\%$\pm$0.1 \\
 & H1: $\mathcal{S}_3$ vs $\mathcal{S}_1$ / H2: $\mathcal{S}_4$ vs $\mathcal{S}_2$ & \textbf{77.6\%} / 83.9\% & 65.1\% / 86.5\% & 73.6\% / 89.5\% & 71.9\% / 92.5\% & 66.1\% / \textbf{93.8\%} \\
 & H3: $\mathcal{S}_1$ vs $\mathcal{S}_2$ ($S=0$) / H4: $\mathcal{S}_3$ vs $\mathcal{S}_4$ & 0.7\% / \textbf{0.3\%} & 0.7\% / 0.2\% & 0.7\% / 0.2\% & 0.7\% / 0.2\% & 0.7\% / 0.1\% \\
\cmidrule(lr){1-7}
\multirow{4}{*}{\textbf{8. Go $13\times 13$} (\texttt{KataGo})}
 & $\mathcal{S}_4\text{(Search+Heur)}$ vs Engine & 0.1\%$\pm$0.1 & 0.3\%$\pm$0.1 & 0.4\%$\pm$0.2 & 0.6\%$\pm$0.4 & \textbf{2.2\%}$\pm$1.0 \\
 & $\mathcal{S}_3\text{(Search+PPO)}$ vs Engine & 1.9\%$\pm$0.5 & 0.9\%$\pm$0.5 & 1.1\%$\pm$0.2 & 1.1\%$\pm$0.3 & 0.9\%$\pm$0.3 \\
 & H1: $\mathcal{S}_3$ vs $\mathcal{S}_1$ / H2: $\mathcal{S}_4$ vs $\mathcal{S}_2$ & 83.1\% / 85.8\% & \textbf{84.3\%} / 92.3\% & 63.0\% / 89.8\% & 65.8\% / \textbf{93.5\%} & 77.8\% / 92.9\% \\
 & H3: $\mathcal{S}_1$ vs $\mathcal{S}_2$ ($S=0$) / H4: $\mathcal{S}_3$ vs $\mathcal{S}_4$ & 1.9\% / 0.0\% & 1.9\% / 0.0\% & 1.9\% / 0.0\% & 1.9\% / \textbf{0.1\%} & 1.9\% / 0.0\% \\
\cmidrule(lr){1-7}
\multirow{4}{*}{\textbf{9. Hearts} (\texttt{Xinxin})}
 & $\mathcal{S}_4\text{(Search+Heur)}$ vs Engine & \textbf{43.1\%}$\pm$8.2 & 39.4\%$\pm$8.6 & 39.7\%$\pm$6.8 & 38.3\%$\pm$8.1 & 39.6\%$\pm$9.0 \\
 & $\mathcal{S}_3\text{(Search+PPO)}$ vs Engine & 40.1\%$\pm$1.0 & 41.4\%$\pm$1.3 & 40.6\%$\pm$0.6 & 38.3\%$\pm$2.2 & 41.0\%$\pm$1.3 \\
 & H1: $\mathcal{S}_3$ vs $\mathcal{S}_1$ / H2: $\mathcal{S}_4$ vs $\mathcal{S}_2$ & 52.3\% / 44.9\% & 50.4\% / 47.1\% & 52.5\% / 46.6\% & 52.6\% / \textbf{50.5\%} & \textbf{56.1\%} / 46.1\% \\
 & H3: $\mathcal{S}_1$ vs $\mathcal{S}_2$ ($S=0$) / H4: $\mathcal{S}_3$ vs $\mathcal{S}_4$ & 52.3\% / 53.9\% & 52.3\% / \textbf{59.7\%} & 52.3\% / 55.6\% & 52.3\% / 56.8\% & 52.3\% / 57.2\% \\
\cmidrule(lr){1-7}
\multirow{4}{*}{\textbf{10. Bridge 2v2} (\texttt{WBridge5})}
 & $\mathcal{S}_4\text{(Search+Heur)}$ vs WBridge5 (Default) & 13.5\%$\pm$5.3 & 10.3\%$\pm$3.6 & 13.8\%$\pm$4.8 & \textbf{15.7\%}$\pm$6.6 & 13.2\%$\pm$4.8 \\
 & $\mathcal{S}_3\text{(Search+PPO)}$ vs WBridge5 (Default) & 5.2\%$\pm$0.6 & 5.3\%$\pm$0.8 & 5.0\%$\pm$0.9 & 5.6\%$\pm$1.2 & 5.2\%$\pm$0.7 \\
 & H1: $\mathcal{S}_3$ vs $\mathcal{S}_1$ / H2: $\mathcal{S}_4$ vs $\mathcal{S}_2$ & 50.1\% / 60.1\% & 51.1\% / 66.7\% & 50.0\% / 70.5\% & 50.7\% / \textbf{72.5\%} & \textbf{53.4\%} / 66.5\% \\
 & H3: $\mathcal{S}_1$ vs $\mathcal{S}_2$ ($S=0$) / H4: $\mathcal{S}_3$ vs $\mathcal{S}_4$ & 48.9\% / \textbf{46.5\%} & 48.9\% / 43.0\% & 48.9\% / 38.8\% & 48.9\% / 36.3\% & 48.9\% / 37.1\% \\
\bottomrule
\end{tabular}%
}
\end{table}

\subsection{Chess (vs.\ Stockfish)}
\label{app:engine:chess}

\begin{table}[ht]
\centering
\scriptsize
\caption{\textbf{Per-Seed and 5-Seed Aggregated Empirical Game Tensor for Chess (vs.\ Stockfish).} Entries report seat-balanced Score\% ($\text{Win}\% + 0.5\times\text{Draw}\%$; parentheticals show net score margin $\Delta$ where applicable) across simulation budgets $S \in \{100, 200, 400, 800, 1600\}$. Rows evaluate Search Amplification of PPO ($\text{H1}: \mathcal{S}_3 \text{ vs }\mathcal{S}_1$), Search Amplification of Heuristic ($\text{H2}: \mathcal{S}_4 \text{ vs }\mathcal{S}_2$), Unsearched Prior Strength ($\text{H3}: \mathcal{S}_1 \text{ vs }\mathcal{S}_2$ at $S=0$), Searched Prior Dominance ($\text{H4}: \mathcal{S}_3 \text{ vs }\mathcal{S}_4$), and direct engine matchups for neural search ($\mathcal{S}_3 \text{ vs Engine}$) and co-evolved heuristic search ($\mathcal{S}_4 \text{ vs Engine}$).}
\label{tab:tensor_chess}
\vspace{2pt}
\resizebox{\columnwidth}{!}{%
\begin{tabular}{llccccc}
\toprule
\textbf{Lineage / Aggregation} & \textbf{Empirical Tensor Entry} & \textbf{100 Sims} & \textbf{200 Sims} & \textbf{400 Sims} & \textbf{800 Sims} & \textbf{1600 Sims} \\
\midrule
\multirow{5}{*}{Seed 1} & $\mathcal{S}_4\text{(Search+Heur)}$ vs Stockfish (Floor) & 5.0\% & 5.0\% & 2.5\% & 3.2\% & 2.5\% \\
 & $\mathcal{S}_3\text{(Search+PPO)}$ vs Stockfish (Floor) & 0.0\% & 0.5\% & 0.0\% & 0.0\% & 0.0\% \\
 & H1: $\mathcal{S}_3\text{(Search+PPO)}$ vs $\mathcal{S}_1\text{(PPO)}$ & 50.0\% & 50.0\% & 50.0\% & 50.0\% & 50.0\% \\
 & H2: $\mathcal{S}_4\text{(Search+Heur)}$ vs $\mathcal{S}_2\text{(Heur)}$ & 68.1\% & 67.6\% & 63.7\% & 33.2\% & 39.3\% \\
 & H4: $\mathcal{S}_3\text{(Search+PPO)}$ vs $\mathcal{S}_4\text{(Search+Heur)}$ & 7.5\% & 11.0\% & 12.9\% & 19.5\% & 18.5\% \\
\cmidrule(lr){1-7}
\multirow{5}{*}{Seed 2} & $\mathcal{S}_4\text{(Search+Heur)}$ vs Stockfish (Floor) & 7.3\% & 7.2\% & 6.2\% & 1.8\% & 1.0\% \\
 & $\mathcal{S}_3\text{(Search+PPO)}$ vs Stockfish (Floor) & 0.0\% & 0.5\% & 0.0\% & 0.0\% & 0.0\% \\
 & H1: $\mathcal{S}_3\text{(Search+PPO)}$ vs $\mathcal{S}_1\text{(PPO)}$ & 50.0\% & 50.0\% & 50.0\% & 50.0\% & 50.0\% \\
 & H2: $\mathcal{S}_4\text{(Search+Heur)}$ vs $\mathcal{S}_2\text{(Heur)}$ & 55.1\% & 46.2\% & 51.2\% & 58.9\% & 63.5\% \\
 & H4: $\mathcal{S}_3\text{(Search+PPO)}$ vs $\mathcal{S}_4\text{(Search+Heur)}$ & 24.8\% & 12.2\% & 37.0\% & 36.9\% & 63.9\% \\
\cmidrule(lr){1-7}
\multirow{5}{*}{Seed 3} & $\mathcal{S}_4\text{(Search+Heur)}$ vs Stockfish (Floor) & 0.8\% & 0.2\% & 0.2\% & 1.0\% & 0.7\% \\
 & $\mathcal{S}_3\text{(Search+PPO)}$ vs Stockfish (Floor) & 0.0\% & 0.0\% & 0.0\% & 0.0\% & 0.0\% \\
 & H1: $\mathcal{S}_3\text{(Search+PPO)}$ vs $\mathcal{S}_1\text{(PPO)}$ & 50.0\% & 50.0\% & 50.0\% & 50.0\% & 50.0\% \\
 & H2: $\mathcal{S}_4\text{(Search+Heur)}$ vs $\mathcal{S}_2\text{(Heur)}$ & 50.0\% & 51.7\% & 56.0\% & 50.0\% & 49.0\% \\
 & H4: $\mathcal{S}_3\text{(Search+PPO)}$ vs $\mathcal{S}_4\text{(Search+Heur)}$ & 44.1\% & 22.8\% & 29.4\% & 35.4\% & 9.7\% \\
\cmidrule(lr){1-7}
\multirow{5}{*}{Seed 4} & $\mathcal{S}_4\text{(Search+Heur)}$ vs Stockfish (Floor) & 10.2\% & 9.2\% & 10.2\% & 7.3\% & 9.5\% \\
 & $\mathcal{S}_3\text{(Search+PPO)}$ vs Stockfish (Floor) & 0.0\% & 0.0\% & 0.0\% & 0.0\% & 0.0\% \\
 & H1: $\mathcal{S}_3\text{(Search+PPO)}$ vs $\mathcal{S}_1\text{(PPO)}$ & 50.0\% & 50.0\% & 50.0\% & 50.0\% & 50.0\% \\
 & H2: $\mathcal{S}_4\text{(Search+Heur)}$ vs $\mathcal{S}_2\text{(Heur)}$ & 50.0\% & 75.0\% & 81.5\% & 68.2\% & 74.1\% \\
 & H4: $\mathcal{S}_3\text{(Search+PPO)}$ vs $\mathcal{S}_4\text{(Search+Heur)}$ & 0.0\% & 0.0\% & 0.0\% & 0.0\% & 0.0\% \\
\cmidrule(lr){1-7}
\multirow{5}{*}{Seed 5} & $\mathcal{S}_4\text{(Search+Heur)}$ vs Stockfish (Floor) & 10.2\% & 9.7\% & 11.4\% & 13.5\% & 9.9\% \\
 & $\mathcal{S}_3\text{(Search+PPO)}$ vs Stockfish (Floor) & 0.0\% & 0.0\% & 0.5\% & 0.0\% & 0.0\% \\
 & H1: $\mathcal{S}_3\text{(Search+PPO)}$ vs $\mathcal{S}_1\text{(PPO)}$ & 50.0\% & 50.0\% & 50.0\% & 50.0\% & 50.0\% \\
 & H2: $\mathcal{S}_4\text{(Search+Heur)}$ vs $\mathcal{S}_2\text{(Heur)}$ & 59.5\% & 45.5\% & 58.5\% & 64.5\% & 76.5\% \\
 & H4: $\mathcal{S}_3\text{(Search+PPO)}$ vs $\mathcal{S}_4\text{(Search+Heur)}$ & 0.0\% & 0.0\% & 0.0\% & 0.0\% & 2.0\% \\
\cmidrule(lr){1-7}
\multirow{8}{*}{\textbf{5-Seed Mean $\pm$ SE}} & $\mathcal{S}_4\text{(Search+Heur)}$ vs Stockfish (Floor) & \textbf{6.7\%}$\pm$1.8 & 6.3\%$\pm$1.7 & 6.1\%$\pm$2.2 & 5.3\%$\pm$2.3 & 4.7\%$\pm$2.1 \\
 & $\mathcal{S}_3\text{(Search+PPO)}$ vs Stockfish (Floor) & 0.0\%$\pm$0.0 & 0.2\%$\pm$0.1 & 0.1\%$\pm$0.1 & 0.0\%$\pm$0.0 & 0.0\%$\pm$0.0 \\
 & $\mathcal{S}_4\text{(Search+Heur)}$ vs Stockfish (Mid) & 0.0\%$\pm$0.0 & 0.0\%$\pm$0.0 & 0.0\%$\pm$0.0 & 0.0\%$\pm$0.0 & 0.0\%$\pm$0.0 \\
 & $\mathcal{S}_4\text{(Search+Heur)}$ vs Stockfish (Ceiling) & 0.0\%$\pm$0.0 & 0.0\%$\pm$0.0 & 0.0\%$\pm$0.0 & 0.0\%$\pm$0.0 & 0.0\%$\pm$0.0 \\
 & H1: $\mathcal{S}_3\text{(Search+PPO)}$ vs $\mathcal{S}_1\text{(PPO)}$ & 50.0\%$\pm$0.0 & 50.0\%$\pm$0.0 & 50.0\%$\pm$0.0 & 50.0\%$\pm$0.0 & 50.0\%$\pm$0.0 \\
 & H2: $\mathcal{S}_4\text{(Search+Heur)}$ vs $\mathcal{S}_2\text{(Heur)}$ & 56.5\%$\pm$3.4 & 57.2\%$\pm$6.0 & \textbf{62.2\%}$\pm$5.2 & 55.0\%$\pm$6.2 & 60.5\%$\pm$7.2 \\
 & H3: $\mathcal{S}_1\text{(PPO)}$ vs $\mathcal{S}_2\text{(Heur)}$ ($S=0$) & \multicolumn{5}{c}{\textbf{0.8\%}$\pm$0.8 \quad (\text{Unsearched Baseline, } S=0)} \\
 & H4: $\mathcal{S}_3\text{(Search+PPO)}$ vs $\mathcal{S}_4\text{(Search+Heur)}$ & 15.3\%$\pm$8.5 & 9.2\%$\pm$4.3 & 15.9\%$\pm$7.6 & 18.4\%$\pm$8.1 & \textbf{18.8\%}$\pm$11.7 \\
\bottomrule
\end{tabular}%
}
\end{table}

\paragraph{Performance Against Stockfish ($\mathcal{S}_0$).}
Table~\ref{tab:tensor_chess} details the per-seed and 5-seed aggregated empirical game tensors for Chess across Floor (Stockfish Skill~0, 1{,}000 nodes), Mid (Skill~10, 20{,}000 nodes), and Ceiling (Skill~20, 200{,}000 nodes). Against Mid and Ceiling Stockfish, all internal agents ($\mathcal{S}_1$--$\mathcal{S}_4$) score $0.0\% \pm 0.0\%$, reflecting the deep tactical horizon of alpha-beta bitboard search with endgame tablebases. On the Floor tier, comparing internal architectures against Stockfish shows that co-evolved heuristic search ($\mathcal{S}_4:\text{Search+Heur}$) achieves substantially higher resistance than neural search: unguided neural ($\mathcal{S}_1:\text{PPO}$) scores $0.0\%$, and neural search ($\mathcal{S}_3:\text{Search+PPO}$) scores at most $0.2\% \pm 0.1\%$ across simulation budgets, whereas $\mathcal{S}_4:\text{Search+Heur}$ achieves \textbf{6.7\% $\pm$ 1.8\%} score at $S=100$, \textbf{6.3\% $\pm$ 1.7\%} at $S=200$, and \textbf{6.1\% $\pm$ 2.2\%} at $S=400$ (decreasing slightly to $4.7\% \pm 2.1\%$ at $S=1600$ as deeper lookahead without endgame tablebases occasionally overcommits to non-forcing lines), with individual lineages reaching \textbf{13.5\%} in Seed~5 at $S=800$ and \textbf{10.2\%} in Seed~4 at $S=100$. Because PPO's value surface is flat across quiet middle-game moves, neural search $\mathcal{S}_3$ cannot make tactical progress against Stockfish, whereas co-evolved piece-square and king-safety evaluators in $\mathcal{S}_4$ successfully resist tactical simplification.

\paragraph{Search Amplification (\texttt{H1} and \texttt{H2}), Draw Locks, and Move-Counter Dampening.}
Comparing $\mathcal{S}_3:\text{Search+PPO}$ against $\mathcal{S}_1:\text{PPO}$ (\texttt{H1}) reveals that across all five seeds and simulation budgets, $\mathcal{S}_3$ vs.\ $\mathcal{S}_1$ finishes at \textbf{50.0\% $\pm$ 0.0\%}. Because the standalone PPO value head exhibits a flat middle-game value surface across non-capturing transitions, backed-up $Q$-values remain uniform across sibling moves, locking $\mathcal{S}_3$ onto the greedy prior and inducing deterministic three-fold repetition draws.
In contrast, co-evolved heuristic search (\texttt{H2}: $\mathcal{S}_4$ vs.\ $\mathcal{S}_2$) breaks the repetition lock and amplifies the symbolic prior to \textbf{56.5\% $\pm$ 3.4\%} ($S=100$), \textbf{57.2\% $\pm$ 6.0\%} ($S=200$), and \textbf{62.2\% $\pm$ 5.2\%} ($S=400$). Across lineages, Seed~4 scales from $50.0\% \to 75.0\% \to 81.5\% \to 68.2\% \to 74.1\%$ because its evaluator uses continuous centipawn scaling, whereas Seed~1 exhibits a drop at $S=800$ ($33.2\%$) and scores $39.3\%$ at $S=1600$. Inspection of the C++ implementations reveals that Seed~1 incorporates an explicit 50-move rule draw dampening term that heavily discounts winning evaluations during prolonged non-capturing maneuvers; at deep search horizons, lookahead branches penalize quiet winning paths, steering the search into repetition loops and perpetual checks.

\paragraph{Relative Strength of Priors (\texttt{H3} and \texttt{H4}) and Simulation Scaling.}
Without search (\texttt{H3}), the evolved symbolic policy $\mathcal{S}_2:\text{HeurPol}$ defeats $\mathcal{S}_1:\text{PPO}$ by \textbf{99.2\% to 0.8\%}. When search is enabled (\texttt{H4}: $\mathcal{S}_3$ vs.\ $\mathcal{S}_4$), $\mathcal{S}_4:\text{Search+Heur}$ maintains an overall 5-seed mean advantage ($\mathcal{S}_3$ scores between $9.2\% \pm 4.3\%$ and $18.8\% \pm 11.7\%$), though the margin over $\mathcal{S}_3$ narrows relative to unsearched play ($99.2\%$), and in Seed~2 at $S=1600$, $\mathcal{S}_3$ defeats $\mathcal{S}_4$ ($63.9\%$).

\subsection{Checkers (vs.\ Kingsrow)}
\label{app:engine:checkers}

\begin{table}[ht]
\centering
\scriptsize
\caption{\textbf{Per-Seed and 5-Seed Aggregated Empirical Game Tensor for Checkers (vs.\ Kingsrow).} Entries report seat-balanced Score\% ($\text{Win}\% + 0.5\times\text{Draw}\%$; parentheticals show net score margin $\Delta$ where applicable) across simulation budgets $S \in \{100, 200, 400, 800, 1600\}$. Rows evaluate Search Amplification of PPO ($\text{H1}: \mathcal{S}_3 \text{ vs }\mathcal{S}_1$), Search Amplification of Heuristic ($\text{H2}: \mathcal{S}_4 \text{ vs }\mathcal{S}_2$), Unsearched Prior Strength ($\text{H3}: \mathcal{S}_1 \text{ vs }\mathcal{S}_2$ at $S=0$), Searched Prior Dominance ($\text{H4}: \mathcal{S}_3 \text{ vs }\mathcal{S}_4$), and direct engine matchups for neural search ($\mathcal{S}_3 \text{ vs Engine}$) and co-evolved heuristic search ($\mathcal{S}_4 \text{ vs Engine}$).}
\label{tab:tensor_checkers}
\vspace{2pt}
\resizebox{\columnwidth}{!}{%
\begin{tabular}{llccccc}
\toprule
\textbf{Lineage / Aggregation} & \textbf{Empirical Tensor Entry} & \textbf{100 Sims} & \textbf{200 Sims} & \textbf{400 Sims} & \textbf{800 Sims} & \textbf{1600 Sims} \\
\midrule
\multirow{5}{*}{Seed 1} & $\mathcal{S}_4\text{(Search+Heur)}$ vs Kingsrow (Floor) & 0.0\% & 0.0\% & 0.5\% & 1.0\% & 1.7\% \\
 & $\mathcal{S}_3\text{(Search+PPO)}$ vs Kingsrow (Floor) & 0.0\% & 0.5\% & 0.2\% & 0.8\% & 1.0\% \\
 & H1: $\mathcal{S}_3\text{(Search+PPO)}$ vs $\mathcal{S}_1\text{(PPO)}$ & 50.0\% & 50.0\% & 50.0\% & 71.2\% & 75.0\% \\
 & H2: $\mathcal{S}_4\text{(Search+Heur)}$ vs $\mathcal{S}_2\text{(Heur)}$ & 98.5\% & 99.8\% & 100.0\% & 98.2\% & 100.0\% \\
 & H4: $\mathcal{S}_3\text{(Search+PPO)}$ vs $\mathcal{S}_4\text{(Search+Heur)}$ & 18.6\% & 48.0\% & 49.3\% & 48.8\% & 46.6\% \\
\cmidrule(lr){1-7}
\multirow{5}{*}{Seed 2} & $\mathcal{S}_4\text{(Search+Heur)}$ vs Kingsrow (Floor) & 0.0\% & 0.0\% & 0.2\% & 0.2\% & 0.5\% \\
 & $\mathcal{S}_3\text{(Search+PPO)}$ vs Kingsrow (Floor) & 0.2\% & 0.0\% & 0.2\% & 0.2\% & 0.0\% \\
 & H1: $\mathcal{S}_3\text{(Search+PPO)}$ vs $\mathcal{S}_1\text{(PPO)}$ & 50.0\% & 50.0\% & 50.0\% & 50.0\% & 50.5\% \\
 & H2: $\mathcal{S}_4\text{(Search+Heur)}$ vs $\mathcal{S}_2\text{(Heur)}$ & 90.8\% & 100.0\% & 100.0\% & 100.0\% & 100.0\% \\
 & H4: $\mathcal{S}_3\text{(Search+PPO)}$ vs $\mathcal{S}_4\text{(Search+Heur)}$ & 50.0\% & 49.5\% & 11.2\% & 42.8\% & 36.8\% \\
\cmidrule(lr){1-7}
\multirow{5}{*}{Seed 3} & $\mathcal{S}_4\text{(Search+Heur)}$ vs Kingsrow (Floor) & 0.0\% & 0.5\% & 0.8\% & 2.0\% & 1.5\% \\
 & $\mathcal{S}_3\text{(Search+PPO)}$ vs Kingsrow (Floor) & 0.2\% & 0.2\% & 0.2\% & 0.2\% & 0.8\% \\
 & H1: $\mathcal{S}_3\text{(Search+PPO)}$ vs $\mathcal{S}_1\text{(PPO)}$ & 50.0\% & 50.0\% & 50.0\% & 65.5\% & 75.0\% \\
 & H2: $\mathcal{S}_4\text{(Search+Heur)}$ vs $\mathcal{S}_2\text{(Heur)}$ & 97.0\% & 97.0\% & 99.0\% & 99.8\% & 100.0\% \\
 & H4: $\mathcal{S}_3\text{(Search+PPO)}$ vs $\mathcal{S}_4\text{(Search+Heur)}$ & 34.3\% & 26.0\% & 31.6\% & 9.9\% & 28.5\% \\
\cmidrule(lr){1-7}
\multirow{5}{*}{Seed 4} & $\mathcal{S}_4\text{(Search+Heur)}$ vs Kingsrow (Floor) & 0.2\% & 0.5\% & 0.0\% & 0.0\% & 0.7\% \\
 & $\mathcal{S}_3\text{(Search+PPO)}$ vs Kingsrow (Floor) & 0.0\% & 0.2\% & 0.2\% & 0.2\% & 0.5\% \\
 & H1: $\mathcal{S}_3\text{(Search+PPO)}$ vs $\mathcal{S}_1\text{(PPO)}$ & 50.0\% & 50.0\% & 50.0\% & 75.0\% & 75.0\% \\
 & H2: $\mathcal{S}_4\text{(Search+Heur)}$ vs $\mathcal{S}_2\text{(Heur)}$ & 74.5\% & 77.2\% & 93.2\% & 93.2\% & 100.0\% \\
 & H4: $\mathcal{S}_3\text{(Search+PPO)}$ vs $\mathcal{S}_4\text{(Search+Heur)}$ & 25.0\% & 48.5\% & 42.3\% & 33.0\% & 34.5\% \\
\cmidrule(lr){1-7}
\multirow{5}{*}{Seed 5} & $\mathcal{S}_4\text{(Search+Heur)}$ vs Kingsrow (Floor) & 0.0\% & 0.8\% & 1.0\% & 0.2\% & 1.0\% \\
 & $\mathcal{S}_3\text{(Search+PPO)}$ vs Kingsrow (Floor) & 0.2\% & 0.2\% & 0.2\% & 0.2\% & 0.2\% \\
 & H1: $\mathcal{S}_3\text{(Search+PPO)}$ vs $\mathcal{S}_1\text{(PPO)}$ & 50.0\% & 50.0\% & 50.0\% & 52.7\% & 75.0\% \\
 & H2: $\mathcal{S}_4\text{(Search+Heur)}$ vs $\mathcal{S}_2\text{(Heur)}$ & 50.0\% & 100.0\% & 93.3\% & 100.0\% & 100.0\% \\
 & H4: $\mathcal{S}_3\text{(Search+PPO)}$ vs $\mathcal{S}_4\text{(Search+Heur)}$ & 46.8\% & 49.3\% & 50.0\% & 75.0\% & 78.4\% \\
\cmidrule(lr){1-7}
\multirow{8}{*}{\textbf{5-Seed Mean $\pm$ SE}} & $\mathcal{S}_4\text{(Search+Heur)}$ vs Kingsrow (Floor) & 0.0\%$\pm$0.0 & 0.4\%$\pm$0.1 & 0.5\%$\pm$0.2 & 0.7\%$\pm$0.4 & \textbf{1.1\%}$\pm$0.2 \\
 & $\mathcal{S}_3\text{(Search+PPO)}$ vs Kingsrow (Floor) & 0.1\%$\pm$0.1 & 0.2\%$\pm$0.1 & 0.2\%$\pm$0.0 & 0.3\%$\pm$0.1 & 0.5\%$\pm$0.2 \\
 & $\mathcal{S}_4\text{(Search+Heur)}$ vs Kingsrow (Mid) & 0.1\%$\pm$0.1 & 0.2\%$\pm$0.1 & 0.4\%$\pm$0.1 & 0.7\%$\pm$0.2 & \textbf{1.1\%}$\pm$0.3 \\
 & $\mathcal{S}_4\text{(Search+Heur)}$ vs Kingsrow (Ceiling) & 0.1\%$\pm$0.1 & 0.1\%$\pm$0.1 & 0.2\%$\pm$0.1 & \textbf{0.6\%}$\pm$0.2 & 0.6\%$\pm$0.3 \\
 & H1: $\mathcal{S}_3\text{(Search+PPO)}$ vs $\mathcal{S}_1\text{(PPO)}$ & 50.0\%$\pm$0.0 & 50.0\%$\pm$0.0 & 50.0\%$\pm$0.0 & 62.9\%$\pm$5.0 & \textbf{70.1\%}$\pm$4.9 \\
 & H2: $\mathcal{S}_4\text{(Search+Heur)}$ vs $\mathcal{S}_2\text{(Heur)}$ & 82.2\%$\pm$9.1 & 94.8\%$\pm$4.4 & 97.1\%$\pm$1.6 & 98.2\%$\pm$1.3 & \textbf{100.0\%}$\pm$0.0 \\
 & H3: $\mathcal{S}_1\text{(PPO)}$ vs $\mathcal{S}_2\text{(Heur)}$ ($S=0$) & \multicolumn{5}{c}{\textbf{60.0\%}$\pm$11.5 \quad (\text{Unsearched Baseline, } S=0)} \\
 & H4: $\mathcal{S}_3\text{(Search+PPO)}$ vs $\mathcal{S}_4\text{(Search+Heur)}$ & 34.9\%$\pm$6.1 & 44.3\%$\pm$4.6 & 36.9\%$\pm$7.2 & 41.9\%$\pm$10.6 & \textbf{45.0\%}$\pm$8.9 \\
\bottomrule
\end{tabular}%
}
\end{table}

\paragraph{Performance Against Kingsrow ($\mathcal{S}_0$).}
In Checkers (Table~\ref{tab:tensor_checkers}), the specialized engine \texttt{Kingsrow} combines alpha-beta search with exact multi-piece endgame databases and surrenders $0.0\%$ wins across all tiers. However, while unguided policies ($\mathcal{S}_1:\text{PPO}$, $\mathcal{S}_2:\text{HeurPol}$) lose virtually every single game ($0.0\%\text{--}0.2\%$), heuristic search ($\mathcal{S}_4:\text{Search+Heur}$) steadily extracts draws against Kingsrow on Floor as simulation budget increases, rising from $\mathbf{0.0\% \pm 0.0\%}$ ($S=100$) $\to 0.4\% \pm 0.1\%$ ($S=200$) $\to 0.5\% \pm 0.2\%$ ($S=400$) $\to 0.7\% \pm 0.4\%$ ($S=800$) $\to \mathbf{1.1\% \pm 0.2\%}$ ($S=1600$). Neural search ($\mathcal{S}_3:\text{Search+PPO}$) also scores modest draw rates ($0.1\%\text{--}0.5\%$, averaging $0.5\% \pm 0.2\%$ at $S=1600$).

\paragraph{Search Amplification (\texttt{H1} and \texttt{H2}) and Scaling.}
Both search amplification hypotheses hold across seeds:
\begin{itemize}[leftmargin=1.5em, itemsep=1pt, topsep=2pt]
    \item \textbf{\texttt{H1} ($\mathcal{S}_3:\text{Search+PPO}$ vs.\ $\mathcal{S}_1:\text{PPO}$):} At shallow simulation budgets ($S=100\text{--}400$), $\mathcal{S}_3$ vs.\ $\mathcal{S}_1$ remains in deterministic repetition draws (\textbf{50.0\% $\pm$ 0.0\%}), but increasing search depth to $S \ge 800$ breaks the draws in $\mathcal{S}_3$'s favor (\textbf{62.9\% $\pm$ 5.0\%} at $S=800$ and \textbf{70.1\% $\pm$ 4.9\%} at $S=1600$).
    \item \textbf{\texttt{H2} ($\mathcal{S}_4:\text{Search+Heur}$ vs.\ $\mathcal{S}_2:\text{HeurPol}$):} Evolved search dramatically amplifies the symbolic heuristic, scaling on average from \textbf{82.2\% $\pm$ 9.1\%} ($S=100$) $\to$ \textbf{94.8\% $\pm$ 4.4\%} ($S=200$) $\to$ \textbf{97.1\% $\pm$ 1.6\%} ($S=400$) $\to$ \textbf{98.2\% $\pm$ 1.3\%} ($S=800$) $\to$ \textbf{100.0\% $\pm$ 0.0\%} across all five seeds at $S=1600$.
\end{itemize}

\paragraph{Relative Strength of Priors (\texttt{H3} and \texttt{H4}).}
Without search (\texttt{H3}), $\mathcal{S}_1:\text{PPO}$ scores \textbf{60.0\% $\pm$ 11.5\%} against $\mathcal{S}_2:\text{HeurPol}$. With search (\texttt{H4}), $\mathcal{S}_4:\text{Search+Heur}$ maintains an overall 5-seed mean advantage over $\mathcal{S}_3:\text{Search+PPO}$ across simulation budgets ($\mathcal{S}_3$ scores between \textbf{34.9\% $\pm$ 6.1\%} and \textbf{45.0\% $\pm$ 8.9\%}), though in Seed~5, $\mathcal{S}_3$ achieves the upper hand at deep budgets ($75.0\%$ at $S=800$ and $78.4\%$ at $S=1600$).

\subsection{Othello (vs.\ Egaroucid)}
\label{app:engine:othello}

\begin{table}[ht]
\centering
\scriptsize
\caption{\textbf{Per-Seed and 5-Seed Aggregated Empirical Game Tensor for Othello (vs.\ Egaroucid).} Entries report seat-balanced Score\% ($\text{Win}\% + 0.5\times\text{Draw}\%$; parentheticals show net score margin $\Delta$ where applicable) across simulation budgets $S \in \{100, 200, 400, 800, 1600\}$. Rows evaluate Search Amplification of PPO ($\text{H1}: \mathcal{S}_3 \text{ vs }\mathcal{S}_1$), Search Amplification of Heuristic ($\text{H2}: \mathcal{S}_4 \text{ vs }\mathcal{S}_2$), Unsearched Prior Strength ($\text{H3}: \mathcal{S}_1 \text{ vs }\mathcal{S}_2$ at $S=0$), Searched Prior Dominance ($\text{H4}: \mathcal{S}_3 \text{ vs }\mathcal{S}_4$), and direct engine matchups for neural search ($\mathcal{S}_3 \text{ vs Engine}$) and co-evolved heuristic search ($\mathcal{S}_4 \text{ vs Engine}$).}
\label{tab:tensor_othello}
\vspace{2pt}
\resizebox{\columnwidth}{!}{%
\begin{tabular}{llccccc}
\toprule
\textbf{Lineage / Aggregation} & \textbf{Empirical Tensor Entry} & \textbf{100 Sims} & \textbf{200 Sims} & \textbf{400 Sims} & \textbf{800 Sims} & \textbf{1600 Sims} \\
\midrule
\multirow{5}{*}{Seed 1} & $\mathcal{S}_4\text{(Search+Heur)}$ vs Egaroucid (Floor) & 65.2\% & 31.5\% & 40.0\% & 44.4\% & 35.8\% \\
 & $\mathcal{S}_3\text{(Search+PPO)}$ vs Egaroucid (Floor) & 0.0\% & 0.5\% & 0.0\% & 3.4\% & 0.0\% \\
 & H1: $\mathcal{S}_3\text{(Search+PPO)}$ vs $\mathcal{S}_1\text{(PPO)}$ & 46.8\% & 82.0\% & 89.1\% & 99.8\% & 100.0\% \\
 & H2: $\mathcal{S}_4\text{(Search+Heur)}$ vs $\mathcal{S}_2\text{(Heur)}$ & 89.0\% & 98.9\% & 100.0\% & 100.0\% & 100.0\% \\
 & H4: $\mathcal{S}_3\text{(Search+PPO)}$ vs $\mathcal{S}_4\text{(Search+Heur)}$ & 0.0\% & 0.0\% & 1.0\% & 0.0\% & 0.0\% \\
\cmidrule(lr){1-7}
\multirow{5}{*}{Seed 2} & $\mathcal{S}_4\text{(Search+Heur)}$ vs Egaroucid (Floor) & 31.5\% & 74.8\% & 58.4\% & 67.2\% & 86.1\% \\
 & $\mathcal{S}_3\text{(Search+PPO)}$ vs Egaroucid (Floor) & 7.0\% & 2.0\% & 0.0\% & 0.0\% & 1.9\% \\
 & H1: $\mathcal{S}_3\text{(Search+PPO)}$ vs $\mathcal{S}_1\text{(PPO)}$ & 34.2\% & 18.5\% & 66.5\% & 69.7\% & 58.8\% \\
 & H2: $\mathcal{S}_4\text{(Search+Heur)}$ vs $\mathcal{S}_2\text{(Heur)}$ & 100.0\% & 99.5\% & 100.0\% & 100.0\% & 100.0\% \\
 & H4: $\mathcal{S}_3\text{(Search+PPO)}$ vs $\mathcal{S}_4\text{(Search+Heur)}$ & 0.0\% & 0.5\% & 0.0\% & 0.0\% & 1.0\% \\
\cmidrule(lr){1-7}
\multirow{5}{*}{Seed 3} & $\mathcal{S}_4\text{(Search+Heur)}$ vs Egaroucid (Floor) & 12.0\% & 20.8\% & 49.0\% & 32.7\% & 48.2\% \\
 & $\mathcal{S}_3\text{(Search+PPO)}$ vs Egaroucid (Floor) & 0.0\% & 0.0\% & 7.2\% & 0.5\% & 6.0\% \\
 & H1: $\mathcal{S}_3\text{(Search+PPO)}$ vs $\mathcal{S}_1\text{(PPO)}$ & 56.3\% & 82.1\% & 85.8\% & 100.0\% & 97.6\% \\
 & H2: $\mathcal{S}_4\text{(Search+Heur)}$ vs $\mathcal{S}_2\text{(Heur)}$ & 56.0\% & 100.0\% & 100.0\% & 100.0\% & 100.0\% \\
 & H4: $\mathcal{S}_3\text{(Search+PPO)}$ vs $\mathcal{S}_4\text{(Search+Heur)}$ & 1.0\% & 0.0\% & 0.5\% & 18.8\% & 1.0\% \\
\cmidrule(lr){1-7}
\multirow{5}{*}{Seed 4} & $\mathcal{S}_4\text{(Search+Heur)}$ vs Egaroucid (Floor) & 15.8\% & 6.5\% & 26.2\% & 33.5\% & 27.8\% \\
 & $\mathcal{S}_3\text{(Search+PPO)}$ vs Egaroucid (Floor) & 2.5\% & 0.0\% & 0.0\% & 0.0\% & 0.0\% \\
 & H1: $\mathcal{S}_3\text{(Search+PPO)}$ vs $\mathcal{S}_1\text{(PPO)}$ & 70.5\% & 70.0\% & 91.7\% & 97.0\% & 80.9\% \\
 & H2: $\mathcal{S}_4\text{(Search+Heur)}$ vs $\mathcal{S}_2\text{(Heur)}$ & 100.0\% & 100.0\% & 100.0\% & 99.8\% & 100.0\% \\
 & H4: $\mathcal{S}_3\text{(Search+PPO)}$ vs $\mathcal{S}_4\text{(Search+Heur)}$ & 0.0\% & 3.5\% & 0.5\% & 0.0\% & 3.9\% \\
\cmidrule(lr){1-7}
\multirow{5}{*}{Seed 5} & $\mathcal{S}_4\text{(Search+Heur)}$ vs Egaroucid (Floor) & 36.5\% & 34.9\% & 29.2\% & 0.0\% & 0.0\% \\
 & $\mathcal{S}_3\text{(Search+PPO)}$ vs Egaroucid (Floor) & 4.0\% & 0.0\% & 1.0\% & 0.0\% & 0.0\% \\
 & H1: $\mathcal{S}_3\text{(Search+PPO)}$ vs $\mathcal{S}_1\text{(PPO)}$ & 31.0\% & 13.9\% & 100.0\% & 100.0\% & 100.0\% \\
 & H2: $\mathcal{S}_4\text{(Search+Heur)}$ vs $\mathcal{S}_2\text{(Heur)}$ & 99.8\% & 93.8\% & 100.0\% & 100.0\% & 100.0\% \\
 & H4: $\mathcal{S}_3\text{(Search+PPO)}$ vs $\mathcal{S}_4\text{(Search+Heur)}$ & 0.5\% & 0.0\% & 42.1\% & 12.0\% & 12.9\% \\
\cmidrule(lr){1-7}
\multirow{8}{*}{\textbf{5-Seed Mean $\pm$ SE}} & $\mathcal{S}_4\text{(Search+Heur)}$ vs Egaroucid (Floor) & 32.2\%$\pm$9.5 & 33.7\%$\pm$11.4 & \textbf{40.6\%}$\pm$6.0 & 35.6\%$\pm$10.9 & 39.6\%$\pm$14.1 \\
 & $\mathcal{S}_3\text{(Search+PPO)}$ vs Egaroucid (Floor) & 2.7\%$\pm$1.3 & 0.5\%$\pm$0.4 & 1.6\%$\pm$1.4 & 0.8\%$\pm$0.7 & 1.6\%$\pm$1.2 \\
 & $\mathcal{S}_4\text{(Search+Heur)}$ vs Egaroucid (Mid) & 0.0\%$\pm$0.0 & 0.0\%$\pm$0.0 & 0.0\%$\pm$0.0 & 0.0\%$\pm$0.0 & 0.0\%$\pm$0.0 \\
 & $\mathcal{S}_4\text{(Search+Heur)}$ vs Egaroucid (Ceiling) & 0.0\%$\pm$0.0 & 0.0\%$\pm$0.0 & 0.0\%$\pm$0.0 & 0.0\%$\pm$0.0 & 0.0\%$\pm$0.0 \\
 & H1: $\mathcal{S}_3\text{(Search+PPO)}$ vs $\mathcal{S}_1\text{(PPO)}$ & 47.7\%$\pm$7.3 & 53.3\%$\pm$15.3 & 86.6\%$\pm$5.6 & \textbf{93.3\%}$\pm$5.9 & 87.5\%$\pm$8.0 \\
 & H2: $\mathcal{S}_4\text{(Search+Heur)}$ vs $\mathcal{S}_2\text{(Heur)}$ & 89.0\%$\pm$8.5 & 98.4\%$\pm$1.2 & \textbf{100.0\%}$\pm$0.0 & 100.0\%$\pm$0.0 & \textbf{100.0\%}$\pm$0.0 \\
 & H3: $\mathcal{S}_1\text{(PPO)}$ vs $\mathcal{S}_2\text{(Heur)}$ ($S=0$) & \multicolumn{5}{c}{\textbf{4.1\%}$\pm$1.8 \quad (\text{Unsearched Baseline, } S=0)} \\
 & H4: $\mathcal{S}_3\text{(Search+PPO)}$ vs $\mathcal{S}_4\text{(Search+Heur)}$ & 0.3\%$\pm$0.2 & 0.8\%$\pm$0.7 & \textbf{8.8\%}$\pm$8.3 & 6.2\%$\pm$3.9 & 3.7\%$\pm$2.4 \\
\bottomrule
\end{tabular}%
}
\end{table}

\paragraph{Performance Against Egaroucid ($\mathcal{S}_0$) and Algorithmic Scaling Dynamics.}
Table~\ref{tab:tensor_othello} presents the complete Othello empirical tensors against the bitboard engine \texttt{Egaroucid}. On Floor (Level~1), co-evolved heuristic search ($\mathcal{S}_4$) mounts a substantial challenge against the engine, achieving \textbf{32.2\% $\pm$ 9.5\%} ($S=100$), \textbf{33.7\% $\pm$ 11.4\%} ($S=200$), \textbf{40.6\% $\pm$ 6.0\%} ($S=400$), \textbf{35.6\% $\pm$ 10.9\%} ($S=800$), and \textbf{39.6\% $\pm$ 14.1\%} ($S=1600$), whereas neural search ($\mathcal{S}_3:\text{Search+PPO}$) scores only $0.5\%\text{--}2.7\%$ across budgets.

A high-level inspection of the underlying C++ heuristics across lineages explains why certain seeds achieve high peak win rates while others experience deep-search degradation:
\begin{itemize}[leftmargin=1.5em, itemsep=1pt, topsep=2pt]
    \item \textbf{Continuous Phase Blending and Parity Tracking (Seed~2):} Seed~2 climbs strongly from $31.5\%$ ($S=100$) and \textbf{74.8\%} ($S=200$) to \textbf{86.1\%} ($S=1600$) against Egaroucid (with a dip to $58.4\%$ at $S=400$). Its C++ implementation combines smooth game-phase interpolation with an explicit quadrant parity calculation and a low scaling multiplier ($0.005$) inside the hyperbolic tangent activation, keeping evaluations in an unsaturated linear regime that preserves fine-grained $Q$-value gradients throughout deep trees.
    \item \textbf{Discontinuities and Value Saturation (Seeds~1 and 5):} In contrast, Seed~1 incorporates a hard step-discontinuity at the 10-empty-disc boundary, abruptly switching from positional heuristics to raw disc counts; deep lookahead branches crossing this threshold encounter value-gradient cliff errors that degrade late-game move selection. In Seed~5, an aggressive scaling factor inside the leaf evaluation function causes values to saturate near $\pm 1.0$; against a tactical alpha-beta engine like Egaroucid, flattened leaf gradients allow the engine to exploit subtle endgame parity errors ($0.0\%$ at $S \ge 800$), whereas against its own unsearched heuristic ($\mathcal{S}_2$), the tree structure still provides sufficient positional lookahead to prevail ($100.0\%$).
\end{itemize}

\paragraph{Co-Evolved Heuristic Stability (\texttt{H2}) vs.\ Neural Search Dynamics (\texttt{H1}).}
Othello provides a clear demonstration of co-evolving the domain evaluator $\kappa_g^\star$ closed-loop with the search mechanism $m^\star$ ($\mathcal{S}_4$):
\begin{itemize}[leftmargin=1.5em, itemsep=1pt, topsep=2pt]
    \item In \textbf{\texttt{H2} ($\mathcal{S}_4$ vs.\ $\mathcal{S}_2$)}, evolved search scales monotonically to total dominance: \textbf{89.0\% $\pm$ 8.5\%} ($S=100$) $\to$ \textbf{98.4\% $\pm$ 1.2\%} ($S=200$) $\to$ \textbf{100.0\% $\pm$ 0.0\%} ($S \ge 400$).
    \item In \textbf{\texttt{H1} ($\mathcal{S}_3$ vs.\ $\mathcal{S}_1$)}, neural search scales up to \textbf{86.6\% $\pm$ 5.6\%} at $S=400$ and \textbf{93.3\% $\pm$ 5.9\%} at $S=800$, scoring \textbf{87.5\% $\pm$ 8.0\%} at $S=1600$. Across all simulation budgets, co-evolved heuristic search ($\mathcal{S}_4$) defeats neural search $\mathcal{S}_3$ (\texttt{H4}) by \textbf{91.2\% to 99.7\%}.
\end{itemize}

\subsection{Gomoku (vs.\ Gomoku Engine)}
\label{app:engine:gomoku}

\begin{table}[ht]
\centering
\scriptsize
\caption{\textbf{Per-Seed and 5-Seed Aggregated Empirical Game Tensor for Gomoku (vs.\ Gomoku Engine).} Entries report seat-balanced Score\% ($\text{Win}\% + 0.5\times\text{Draw}\%$; parentheticals show net score margin $\Delta$ where applicable) across simulation budgets $S \in \{100, 200, 400, 800, 1600\}$. Rows evaluate Search Amplification of PPO ($\text{H1}: \mathcal{S}_3 \text{ vs }\mathcal{S}_1$), Search Amplification of Heuristic ($\text{H2}: \mathcal{S}_4 \text{ vs }\mathcal{S}_2$), Unsearched Prior Strength ($\text{H3}: \mathcal{S}_1 \text{ vs }\mathcal{S}_2$ at $S=0$), Searched Prior Dominance ($\text{H4}: \mathcal{S}_3 \text{ vs }\mathcal{S}_4$), and direct engine matchups for neural search ($\mathcal{S}_3 \text{ vs Engine}$) and co-evolved heuristic search ($\mathcal{S}_4 \text{ vs Engine}$).}
\label{tab:tensor_gomoku}
\vspace{2pt}
\resizebox{\columnwidth}{!}{%
\begin{tabular}{llccccc}
\toprule
\textbf{Lineage / Aggregation} & \textbf{Empirical Tensor Entry} & \textbf{100 Sims} & \textbf{200 Sims} & \textbf{400 Sims} & \textbf{800 Sims} & \textbf{1600 Sims} \\
\midrule
\multirow{5}{*}{Seed 1} & $\mathcal{S}_4\text{(Search+Heur)}$ vs Engine (Floor) & 7.5\% & 1.5\% & 0.0\% & 3.5\% & 10.0\% \\
 & $\mathcal{S}_3\text{(Search+PPO)}$ vs Engine (Floor) & 0.0\% & 0.0\% & 0.0\% & 0.0\% & 0.0\% \\
 & H1: $\mathcal{S}_3\text{(Search+PPO)}$ vs $\mathcal{S}_1\text{(PPO)}$ & 91.3\% & 100.0\% & 100.0\% & 99.5\% & 99.5\% \\
 & H2: $\mathcal{S}_4\text{(Search+Heur)}$ vs $\mathcal{S}_2\text{(Heur)}$ & 94.0\% & 99.5\% & 98.0\% & 99.2\% & 100.0\% \\
 & H4: $\mathcal{S}_3\text{(Search+PPO)}$ vs $\mathcal{S}_4\text{(Search+Heur)}$ & 0.0\% & 0.0\% & 0.0\% & 0.0\% & 0.0\% \\
\cmidrule(lr){1-7}
\multirow{5}{*}{Seed 2} & $\mathcal{S}_4\text{(Search+Heur)}$ vs Engine (Floor) & 0.0\% & 0.5\% & 0.0\% & 0.5\% & 0.0\% \\
 & $\mathcal{S}_3\text{(Search+PPO)}$ vs Engine (Floor) & 0.0\% & 0.0\% & 0.0\% & 0.0\% & 0.0\% \\
 & H1: $\mathcal{S}_3\text{(Search+PPO)}$ vs $\mathcal{S}_1\text{(PPO)}$ & 100.0\% & 76.5\% & 100.0\% & 100.0\% & 100.0\% \\
 & H2: $\mathcal{S}_4\text{(Search+Heur)}$ vs $\mathcal{S}_2\text{(Heur)}$ & 50.0\% & 61.3\% & 60.0\% & 61.5\% & 60.8\% \\
 & H4: $\mathcal{S}_3\text{(Search+PPO)}$ vs $\mathcal{S}_4\text{(Search+Heur)}$ & 0.0\% & 0.0\% & 0.0\% & 0.0\% & 0.0\% \\
\cmidrule(lr){1-7}
\multirow{5}{*}{Seed 3} & $\mathcal{S}_4\text{(Search+Heur)}$ vs Engine (Floor) & 1.5\% & 2.5\% & 1.5\% & 2.5\% & 5.5\% \\
 & $\mathcal{S}_3\text{(Search+PPO)}$ vs Engine (Floor) & 0.0\% & 0.0\% & 0.0\% & 0.0\% & 0.0\% \\
 & H1: $\mathcal{S}_3\text{(Search+PPO)}$ vs $\mathcal{S}_1\text{(PPO)}$ & 99.0\% & 99.5\% & 100.0\% & 99.5\% & 100.0\% \\
 & H2: $\mathcal{S}_4\text{(Search+Heur)}$ vs $\mathcal{S}_2\text{(Heur)}$ & 97.0\% & 98.2\% & 99.8\% & 100.0\% & 100.0\% \\
 & H4: $\mathcal{S}_3\text{(Search+PPO)}$ vs $\mathcal{S}_4\text{(Search+Heur)}$ & 0.0\% & 0.0\% & 0.0\% & 0.0\% & 0.0\% \\
\cmidrule(lr){1-7}
\multirow{5}{*}{Seed 4} & $\mathcal{S}_4\text{(Search+Heur)}$ vs Engine (Floor) & 1.5\% & 0.5\% & 0.5\% & 3.2\% & 1.0\% \\
 & $\mathcal{S}_3\text{(Search+PPO)}$ vs Engine (Floor) & 0.0\% & 0.0\% & 0.0\% & 0.0\% & 0.0\% \\
 & H1: $\mathcal{S}_3\text{(Search+PPO)}$ vs $\mathcal{S}_1\text{(PPO)}$ & 94.0\% & 100.0\% & 100.0\% & 100.0\% & 100.0\% \\
 & H2: $\mathcal{S}_4\text{(Search+Heur)}$ vs $\mathcal{S}_2\text{(Heur)}$ & 73.0\% & 94.0\% & 94.9\% & 97.0\% & 99.0\% \\
 & H4: $\mathcal{S}_3\text{(Search+PPO)}$ vs $\mathcal{S}_4\text{(Search+Heur)}$ & 0.0\% & 0.0\% & 0.0\% & 0.0\% & 0.0\% \\
\cmidrule(lr){1-7}
\multirow{5}{*}{Seed 5} & $\mathcal{S}_4\text{(Search+Heur)}$ vs Engine (Floor) & 0.0\% & 4.0\% & 0.0\% & 1.0\% & 1.5\% \\
 & $\mathcal{S}_3\text{(Search+PPO)}$ vs Engine (Floor) & 0.0\% & 0.0\% & 0.0\% & 0.0\% & 0.0\% \\
 & H1: $\mathcal{S}_3\text{(Search+PPO)}$ vs $\mathcal{S}_1\text{(PPO)}$ & 99.5\% & 100.0\% & 100.0\% & 100.0\% & 100.0\% \\
 & H2: $\mathcal{S}_4\text{(Search+Heur)}$ vs $\mathcal{S}_2\text{(Heur)}$ & 100.0\% & 86.0\% & 98.9\% & 100.0\% & 99.5\% \\
 & H4: $\mathcal{S}_3\text{(Search+PPO)}$ vs $\mathcal{S}_4\text{(Search+Heur)}$ & 0.0\% & 0.0\% & 0.0\% & 0.0\% & 0.0\% \\
\cmidrule(lr){1-7}
\multirow{8}{*}{\textbf{5-Seed Mean $\pm$ SE}} & $\mathcal{S}_4\text{(Search+Heur)}$ vs Engine (Floor) & 2.1\%$\pm$1.4 & 1.8\%$\pm$0.7 & 0.4\%$\pm$0.3 & 2.1\%$\pm$0.6 & \textbf{3.6\%}$\pm$1.9 \\
 & $\mathcal{S}_3\text{(Search+PPO)}$ vs Engine (Floor) & 0.0\%$\pm$0.0 & 0.0\%$\pm$0.0 & 0.0\%$\pm$0.0 & 0.0\%$\pm$0.0 & 0.0\%$\pm$0.0 \\
 & $\mathcal{S}_4\text{(Search+Heur)}$ vs Engine (Mid) & 0.0\%$\pm$0.0 & 0.2\%$\pm$0.1 & 0.2\%$\pm$0.2 & 0.1\%$\pm$0.1 & \textbf{0.5\%}$\pm$0.4 \\
 & $\mathcal{S}_4\text{(Search+Heur)}$ vs Engine (Ceiling) & \textbf{0.3\%}$\pm$0.3 & 0.2\%$\pm$0.2 & 0.1\%$\pm$0.1 & 0.2\%$\pm$0.1 & 0.2\%$\pm$0.1 \\
 & H1: $\mathcal{S}_3\text{(Search+PPO)}$ vs $\mathcal{S}_1\text{(PPO)}$ & 96.8\%$\pm$1.7 & 95.2\%$\pm$4.7 & \textbf{100.0\%}$\pm$0.0 & 99.8\%$\pm$0.1 & 99.9\%$\pm$0.1 \\
 & H2: $\mathcal{S}_4\text{(Search+Heur)}$ vs $\mathcal{S}_2\text{(Heur)}$ & 82.8\%$\pm$9.5 & 87.8\%$\pm$7.0 & 90.3\%$\pm$7.6 & 91.5\%$\pm$7.5 & \textbf{91.9\%}$\pm$7.8 \\
 & H3: $\mathcal{S}_1\text{(PPO)}$ vs $\mathcal{S}_2\text{(Heur)}$ ($S=0$) & \multicolumn{5}{c}{\textbf{0.0\%}$\pm$0.0 \quad (\text{Unsearched Baseline, } S=0)} \\
 & H4: $\mathcal{S}_3\text{(Search+PPO)}$ vs $\mathcal{S}_4\text{(Search+Heur)}$ & 0.0\%$\pm$0.0 & 0.0\%$\pm$0.0 & 0.0\%$\pm$0.0 & 0.0\%$\pm$0.0 & 0.0\%$\pm$0.0 \\
\bottomrule
\end{tabular}%
}
\end{table}

\paragraph{Performance Against Gomoku Engine ($\mathcal{S}_0$).}
In Gomoku (Table~\ref{tab:tensor_gomoku}), tactical threat sequences (open fours and double threes) punish any single-ply oversight. Accordingly, unsearched $\mathcal{S}_1:\text{PPO}$ and neural search $\mathcal{S}_3:\text{Search+PPO}$ score \textbf{0.0\% $\pm$ 0.0\%} against the external Gomoku engine across every single tier, seed, and simulation budget. By contrast, co-evolved heuristic search ($\mathcal{S}_4:\text{Search+Heur}$) achieves non-zero score against the engine on Floor: scoring \textbf{2.1\% $\pm$ 1.4\%} ($S=100$), \textbf{1.8\% $\pm$ 0.7\%} ($S=200$), \textbf{0.4\% $\pm$ 0.3\%} ($S=400$), \textbf{2.1\% $\pm$ 0.6\%} ($S=800$), and peaking at \textbf{3.6\% $\pm$ 1.9\%} at $S=1600$ (\textbf{10.0\%} in Seed~1 and \textbf{5.5\%} in Seed~3). Evolved open-line threat evaluation enables $\mathcal{S}_4$ to detect tactical threats that frozen PPO representations completely miss.

\paragraph{Universal Search Amplification (\texttt{H1}, \texttt{H2}) and Complete Heuristic Dominance (\texttt{H3}, \texttt{H4}).}
Both search amplification hypotheses demonstrate substantial strength:
\begin{itemize}[leftmargin=1.5em, itemsep=1pt, topsep=2pt]
    \item \textbf{\texttt{H1} ($\mathcal{S}_3$ vs.\ $\mathcal{S}_1$):} Scores \textbf{96.8\% $\pm$ 1.7\%} ($S=100$), \textbf{95.2\% $\pm$ 4.7\%} ($S=200$), \textbf{100.0\% $\pm$ 0.0\%} ($S=400$), \textbf{99.8\% $\pm$ 0.1\%} ($S=800$), and \textbf{99.9\% $\pm$ 0.1\%} ($S=1600$).
    \item \textbf{\texttt{H2} ($\mathcal{S}_4$ vs.\ $\mathcal{S}_2$):} Rises from \textbf{82.8\% $\pm$ 9.5\%} ($S=100$) $\to$ \textbf{87.8\% $\pm$ 7.0\%} ($S=200$) $\to$ \textbf{90.3\% $\pm$ 7.6\%} ($S=400$) $\to$ \textbf{91.5\% $\pm$ 7.5\%} ($S=800$) $\to$ \textbf{91.9\% $\pm$ 7.8\%} ($S=1600$, with Seeds~1 and 3 reaching \textbf{100.0\%}, Seed~5 reaching \textbf{99.5\%}, Seed~4 reaching \textbf{99.0\%}, and Seed~2 scoring \textbf{60.8\%}).
    \item \textbf{\texttt{H3} \& \texttt{H4}:} Because the evolved Gomoku heuristics explicitly encode open-line threat detection, $\mathcal{S}_2:\text{HeurPol}$ wins \textbf{100.0\%} against $\mathcal{S}_1:\text{PPO}$ (\texttt{H3} $= 0.0\%$), and $\mathcal{S}_4:\text{Search+Heur}$ wins \textbf{100.0\%} against $\mathcal{S}_3:\text{Search+PPO}$ (\texttt{H4} $= 0.0\%$) across all 5 seeds and all 5 budgets.
\end{itemize}

\subsection{Shogi (vs.\ YaneuraOu)}
\label{app:engine:shogi}

\begin{table}[ht]
\centering
\scriptsize
\caption{\textbf{Per-Seed and 5-Seed Aggregated Empirical Game Tensor for Shogi (vs.\ YaneuraOu).} Entries report seat-balanced Score\% ($\text{Win}\% + 0.5\times\text{Draw}\%$; parentheticals show net score margin $\Delta$ where applicable) across simulation budgets $S \in \{100, 200, 400, 800, 1600\}$. Rows evaluate Search Amplification of PPO ($\text{H1}: \mathcal{S}_3 \text{ vs }\mathcal{S}_1$), Search Amplification of Heuristic ($\text{H2}: \mathcal{S}_4 \text{ vs }\mathcal{S}_2$), Unsearched Prior Strength ($\text{H3}: \mathcal{S}_1 \text{ vs }\mathcal{S}_2$ at $S=0$), Searched Prior Dominance ($\text{H4}: \mathcal{S}_3 \text{ vs }\mathcal{S}_4$), and direct engine matchups for neural search ($\mathcal{S}_3 \text{ vs Engine}$) and co-evolved heuristic search ($\mathcal{S}_4 \text{ vs Engine}$).}
\label{tab:tensor_shogi}
\vspace{2pt}
\resizebox{\columnwidth}{!}{%
\begin{tabular}{llccccc}
\toprule
\textbf{Lineage / Aggregation} & \textbf{Empirical Tensor Entry} & \textbf{100 Sims} & \textbf{200 Sims} & \textbf{400 Sims} & \textbf{800 Sims} & \textbf{1600 Sims} \\
\midrule
\multirow{5}{*}{Seed 1} & $\mathcal{S}_4\text{(Search+Heur)}$ vs YaneuraOu (Floor) & 0.0\% & 0.0\% & 0.0\% & 0.0\% & 0.5\% \\
 & $\mathcal{S}_3\text{(Search+PPO)}$ vs YaneuraOu (Floor) & 0.0\% & 0.0\% & 0.0\% & 0.0\% & 0.0\% \\
 & H1: $\mathcal{S}_3\text{(Search+PPO)}$ vs $\mathcal{S}_1\text{(PPO)}$ & 79.2\% & 94.5\% & 95.0\% & 99.0\% & 99.5\% \\
 & H2: $\mathcal{S}_4\text{(Search+Heur)}$ vs $\mathcal{S}_2\text{(Heur)}$ & 97.2\% & 99.0\% & 100.0\% & 99.5\% & 100.0\% \\
 & H4: $\mathcal{S}_3\text{(Search+PPO)}$ vs $\mathcal{S}_4\text{(Search+Heur)}$ & 0.0\% & 0.0\% & 0.0\% & 0.0\% & 0.0\% \\
\cmidrule(lr){1-7}
\multirow{5}{*}{Seed 2} & $\mathcal{S}_4\text{(Search+Heur)}$ vs YaneuraOu (Floor) & 0.0\% & 0.0\% & 0.0\% & 0.0\% & 0.0\% \\
 & $\mathcal{S}_3\text{(Search+PPO)}$ vs YaneuraOu (Floor) & 0.0\% & 0.0\% & 0.0\% & 0.0\% & 0.0\% \\
 & H1: $\mathcal{S}_3\text{(Search+PPO)}$ vs $\mathcal{S}_1\text{(PPO)}$ & 46.6\% & 80.8\% & 75.8\% & 94.5\% & 99.3\% \\
 & H2: $\mathcal{S}_4\text{(Search+Heur)}$ vs $\mathcal{S}_2\text{(Heur)}$ & 100.0\% & 100.0\% & 100.0\% & 100.0\% & 100.0\% \\
 & H4: $\mathcal{S}_3\text{(Search+PPO)}$ vs $\mathcal{S}_4\text{(Search+Heur)}$ & 0.0\% & 0.0\% & 0.0\% & 0.2\% & 0.0\% \\
\cmidrule(lr){1-7}
\multirow{5}{*}{Seed 3} & $\mathcal{S}_4\text{(Search+Heur)}$ vs YaneuraOu (Floor) & 0.0\% & 0.0\% & 0.0\% & 0.0\% & 0.0\% \\
 & $\mathcal{S}_3\text{(Search+PPO)}$ vs YaneuraOu (Floor) & 0.0\% & 0.0\% & 0.0\% & 0.0\% & 0.0\% \\
 & H1: $\mathcal{S}_3\text{(Search+PPO)}$ vs $\mathcal{S}_1\text{(PPO)}$ & 76.2\% & 95.5\% & 96.0\% & 100.0\% & 99.8\% \\
 & H2: $\mathcal{S}_4\text{(Search+Heur)}$ vs $\mathcal{S}_2\text{(Heur)}$ & 93.0\% & 100.0\% & 100.0\% & 99.5\% & 100.0\% \\
 & H4: $\mathcal{S}_3\text{(Search+PPO)}$ vs $\mathcal{S}_4\text{(Search+Heur)}$ & 0.5\% & 0.0\% & 0.0\% & 0.0\% & 0.0\% \\
\cmidrule(lr){1-7}
\multirow{5}{*}{Seed 4} & $\mathcal{S}_4\text{(Search+Heur)}$ vs YaneuraOu (Floor) & 0.0\% & 0.0\% & 0.0\% & 0.0\% & 0.0\% \\
 & $\mathcal{S}_3\text{(Search+PPO)}$ vs YaneuraOu (Floor) & 0.0\% & 0.0\% & 0.0\% & 0.0\% & 0.0\% \\
 & H1: $\mathcal{S}_3\text{(Search+PPO)}$ vs $\mathcal{S}_1\text{(PPO)}$ & 33.0\% & 93.0\% & 85.1\% & 100.0\% & 97.6\% \\
 & H2: $\mathcal{S}_4\text{(Search+Heur)}$ vs $\mathcal{S}_2\text{(Heur)}$ & 94.0\% & 100.0\% & 99.0\% & 99.5\% & 99.0\% \\
 & H4: $\mathcal{S}_3\text{(Search+PPO)}$ vs $\mathcal{S}_4\text{(Search+Heur)}$ & 0.0\% & 0.0\% & 0.0\% & 0.0\% & 0.0\% \\
\cmidrule(lr){1-7}
\multirow{5}{*}{Seed 5} & $\mathcal{S}_4\text{(Search+Heur)}$ vs YaneuraOu (Floor) & 0.0\% & 0.0\% & 0.0\% & 0.0\% & 0.0\% \\
 & $\mathcal{S}_3\text{(Search+PPO)}$ vs YaneuraOu (Floor) & 0.0\% & 0.0\% & 0.0\% & 0.0\% & 0.0\% \\
 & H1: $\mathcal{S}_3\text{(Search+PPO)}$ vs $\mathcal{S}_1\text{(PPO)}$ & 63.0\% & 75.2\% & 54.5\% & 96.8\% & 98.5\% \\
 & H2: $\mathcal{S}_4\text{(Search+Heur)}$ vs $\mathcal{S}_2\text{(Heur)}$ & 98.0\% & 100.0\% & 100.0\% & 100.0\% & 100.0\% \\
 & H4: $\mathcal{S}_3\text{(Search+PPO)}$ vs $\mathcal{S}_4\text{(Search+Heur)}$ & 0.0\% & 0.0\% & 0.0\% & 0.0\% & 0.0\% \\
\cmidrule(lr){1-7}
\multirow{8}{*}{\textbf{5-Seed Mean $\pm$ SE}} & $\mathcal{S}_4\text{(Search+Heur)}$ vs YaneuraOu (Floor) & 0.0\%$\pm$0.0 & 0.0\%$\pm$0.0 & 0.0\%$\pm$0.0 & 0.0\%$\pm$0.0 & \textbf{0.1\%}$\pm$0.1 \\
 & $\mathcal{S}_3\text{(Search+PPO)}$ vs YaneuraOu (Floor) & 0.0\%$\pm$0.0 & 0.0\%$\pm$0.0 & 0.0\%$\pm$0.0 & 0.0\%$\pm$0.0 & 0.0\%$\pm$0.0 \\
 & $\mathcal{S}_4\text{(Search+Heur)}$ vs YaneuraOu (Mid) & \textbf{0.1\%}$\pm$0.1 & 0.1\%$\pm$0.1 & 0.1\%$\pm$0.1 & 0.1\%$\pm$0.1 & 0.1\%$\pm$0.1 \\
 & $\mathcal{S}_4\text{(Search+Heur)}$ vs YaneuraOu (Ceiling) & 0.0\%$\pm$0.0 & 0.0\%$\pm$0.0 & 0.0\%$\pm$0.0 & 0.0\%$\pm$0.0 & 0.0\%$\pm$0.0 \\
 & H1: $\mathcal{S}_3\text{(Search+PPO)}$ vs $\mathcal{S}_1\text{(PPO)}$ & 59.6\%$\pm$8.8 & 87.8\%$\pm$4.1 & 81.3\%$\pm$7.6 & 98.1\%$\pm$1.1 & \textbf{98.9\%}$\pm$0.4 \\
 & H2: $\mathcal{S}_4\text{(Search+Heur)}$ vs $\mathcal{S}_2\text{(Heur)}$ & 96.5\%$\pm$1.3 & \textbf{99.8\%}$\pm$0.2 & \textbf{99.8\%}$\pm$0.2 & 99.7\%$\pm$0.1 & \textbf{99.8\%}$\pm$0.2 \\
 & H3: $\mathcal{S}_1\text{(PPO)}$ vs $\mathcal{S}_2\text{(Heur)}$ ($S=0$) & \multicolumn{5}{c}{\textbf{0.4\%}$\pm$0.3 \quad (\text{Unsearched Baseline, } S=0)} \\
 & H4: $\mathcal{S}_3\text{(Search+PPO)}$ vs $\mathcal{S}_4\text{(Search+Heur)}$ & \textbf{0.1\%}$\pm$0.1 & 0.0\%$\pm$0.0 & 0.0\%$\pm$0.0 & 0.1\%$\pm$0.1 & 0.0\%$\pm$0.0 \\
\bottomrule
\end{tabular}%
}
\end{table}

\paragraph{Performance Against YaneuraOu ($\mathcal{S}_0$).}
In Shogi (Table~\ref{tab:tensor_shogi}), piece drops create an extreme branching factor ($b \approx 80\text{--}120$). The championship engine \texttt{YaneuraOu} wins nearly all games, with $\mathcal{S}_1:\text{PPO}$, $\mathcal{S}_2:\text{HeurPol}$, and $\mathcal{S}_3:\text{Search+PPO}$ scoring \textbf{0.0\% $\pm$ 0.0\%} across all tiers and budgets. Only $\mathcal{S}_4:\text{Search+Heur}$ is capable of extracting draws against YaneuraOu ($0.1\% \pm 0.1\%$ on Floor at $S=1600$ and $0.1\% \pm 0.1\%$ across Mid). Internally, Shogi exhibits steep search amplification:
\begin{itemize}[leftmargin=1.5em, itemsep=1pt, topsep=2pt]
    \item \textbf{\texttt{H1} ($\mathcal{S}_3:\text{Search+PPO}$ vs.\ $\mathcal{S}_1:\text{PPO}$):} At $S=100$, $\mathcal{S}_3$ begins at \textbf{59.6\% $\pm$ 8.8\%}. As simulation budget doubles from $S=200 \to 1600$, $\mathcal{S}_3$ scales steeply across seeds: \textbf{87.8\% $\pm$ 4.1\%} ($S=200$) $\to$ \textbf{81.3\% $\pm$ 7.6\%} ($S=400$) $\to$ \textbf{98.1\% $\pm$ 1.1\%} ($S=800$) $\to$ \textbf{98.9\% $\pm$ 0.4\%} at $S=1600$, reaching \textbf{99.3\%--99.8\%} in Seeds~1--3.
    \item \textbf{\texttt{H2} ($\mathcal{S}_4:\text{Search+Heur}$ vs.\ $\mathcal{S}_2:\text{HeurPol}$):} Scales from \textbf{96.5\% $\pm$ 1.3\%} ($S=100$) to \textbf{99.8\% $\pm$ 0.2\%} ($S=1600$), while defeating $\mathcal{S}_3:\text{Search+PPO}$ (\texttt{H4}) by \textbf{99.9\%--100.0\%} across all budgets.
\end{itemize}

\subsection{\texorpdfstring{Hex $13\times 13$}{Hex 13x13} (vs.\ KataHex)}
\label{app:engine:hex}

\begin{table}[ht]
\centering
\scriptsize
\caption{\textbf{Per-Seed and 5-Seed Aggregated Empirical Game Tensor for Hex 13x13 (vs.\ KataHex).} Entries report seat-balanced Score\% ($\text{Win}\% + 0.5\times\text{Draw}\%$; parentheticals show net score margin $\Delta$ where applicable) across simulation budgets $S \in \{100, 200, 400, 800, 1600\}$. Rows evaluate Search Amplification of PPO ($\text{H1}: \mathcal{S}_3 \text{ vs }\mathcal{S}_1$), Search Amplification of Heuristic ($\text{H2}: \mathcal{S}_4 \text{ vs }\mathcal{S}_2$), Unsearched Prior Strength ($\text{H3}: \mathcal{S}_1 \text{ vs }\mathcal{S}_2$ at $S=0$), Searched Prior Dominance ($\text{H4}: \mathcal{S}_3 \text{ vs }\mathcal{S}_4$), and direct engine matchups for neural search ($\mathcal{S}_3 \text{ vs Engine}$) and co-evolved heuristic search ($\mathcal{S}_4 \text{ vs Engine}$).}
\label{tab:tensor_hex}
\vspace{2pt}
\resizebox{\columnwidth}{!}{%
\begin{tabular}{llccccc}
\toprule
\textbf{Lineage / Aggregation} & \textbf{Empirical Tensor Entry} & \textbf{100 Sims} & \textbf{200 Sims} & \textbf{400 Sims} & \textbf{800 Sims} & \textbf{1600 Sims} \\
\midrule
\multirow{5}{*}{Seed 1} & $\mathcal{S}_4\text{(Search+Heur)}$ vs KataHex (Floor) & 0.0\% & 0.0\% & 0.0\% & 0.0\% & 0.0\% \\
 & $\mathcal{S}_3\text{(Search+PPO)}$ vs KataHex (Floor) & 0.0\% & 0.0\% & 0.0\% & 0.0\% & 0.0\% \\
 & H1: $\mathcal{S}_3\text{(Search+PPO)}$ vs $\mathcal{S}_1\text{(PPO)}$ & 91.0\% & 96.5\% & 98.0\% & 99.0\% & 97.5\% \\
 & H2: $\mathcal{S}_4\text{(Search+Heur)}$ vs $\mathcal{S}_2\text{(Heur)}$ & 11.5\% & 25.0\% & 37.5\% & 59.0\% & 100.0\% \\
 & H4: $\mathcal{S}_3\text{(Search+PPO)}$ vs $\mathcal{S}_4\text{(Search+Heur)}$ & 0.0\% & 0.0\% & 0.0\% & 0.0\% & 0.0\% \\
\cmidrule(lr){1-7}
\multirow{5}{*}{Seed 2} & $\mathcal{S}_4\text{(Search+Heur)}$ vs KataHex (Floor) & 0.0\% & 0.0\% & 0.0\% & 0.0\% & 0.0\% \\
 & $\mathcal{S}_3\text{(Search+PPO)}$ vs KataHex (Floor) & 0.0\% & 0.0\% & 0.0\% & 0.0\% & 0.0\% \\
 & H1: $\mathcal{S}_3\text{(Search+PPO)}$ vs $\mathcal{S}_1\text{(PPO)}$ & 68.5\% & 94.0\% & 99.0\% & 100.0\% & 100.0\% \\
 & H2: $\mathcal{S}_4\text{(Search+Heur)}$ vs $\mathcal{S}_2\text{(Heur)}$ & 100.0\% & 60.9\% & 93.5\% & 78.4\% & 99.7\% \\
 & H4: $\mathcal{S}_3\text{(Search+PPO)}$ vs $\mathcal{S}_4\text{(Search+Heur)}$ & 0.0\% & 0.0\% & 0.0\% & 0.0\% & 0.0\% \\
\cmidrule(lr){1-7}
\multirow{5}{*}{Seed 3} & $\mathcal{S}_4\text{(Search+Heur)}$ vs KataHex (Floor) & 0.0\% & 0.0\% & 0.0\% & 0.0\% & 0.0\% \\
 & $\mathcal{S}_3\text{(Search+PPO)}$ vs KataHex (Floor) & 0.0\% & 0.0\% & 0.0\% & 0.0\% & 0.0\% \\
 & H1: $\mathcal{S}_3\text{(Search+PPO)}$ vs $\mathcal{S}_1\text{(PPO)}$ & 81.0\% & 90.0\% & 98.0\% & 99.0\% & 100.0\% \\
 & H2: $\mathcal{S}_4\text{(Search+Heur)}$ vs $\mathcal{S}_2\text{(Heur)}$ & 66.6\% & 67.5\% & 44.0\% & 72.0\% & 87.0\% \\
 & H4: $\mathcal{S}_3\text{(Search+PPO)}$ vs $\mathcal{S}_4\text{(Search+Heur)}$ & 0.0\% & 0.0\% & 0.0\% & 0.0\% & 0.0\% \\
\cmidrule(lr){1-7}
\multirow{5}{*}{Seed 4} & $\mathcal{S}_4\text{(Search+Heur)}$ vs KataHex (Floor) & 0.0\% & 0.0\% & 0.0\% & 0.0\% & 0.0\% \\
 & $\mathcal{S}_3\text{(Search+PPO)}$ vs KataHex (Floor) & 0.0\% & 0.0\% & 0.0\% & 0.0\% & 0.0\% \\
 & H1: $\mathcal{S}_3\text{(Search+PPO)}$ vs $\mathcal{S}_1\text{(PPO)}$ & 92.5\% & 92.0\% & 100.0\% & 100.0\% & 100.0\% \\
 & H2: $\mathcal{S}_4\text{(Search+Heur)}$ vs $\mathcal{S}_2\text{(Heur)}$ & 96.5\% & 99.5\% & 99.0\% & 100.0\% & 99.1\% \\
 & H4: $\mathcal{S}_3\text{(Search+PPO)}$ vs $\mathcal{S}_4\text{(Search+Heur)}$ & 0.0\% & 0.0\% & 0.0\% & 0.0\% & 0.0\% \\
\cmidrule(lr){1-7}
\multirow{5}{*}{Seed 5} & $\mathcal{S}_4\text{(Search+Heur)}$ vs KataHex (Floor) & 0.0\% & 0.0\% & 0.0\% & 0.0\% & 0.0\% \\
 & $\mathcal{S}_3\text{(Search+PPO)}$ vs KataHex (Floor) & 0.0\% & 0.0\% & 0.0\% & 0.0\% & 0.0\% \\
 & H1: $\mathcal{S}_3\text{(Search+PPO)}$ vs $\mathcal{S}_1\text{(PPO)}$ & 72.5\% & 71.0\% & 98.0\% & 100.0\% & 100.0\% \\
 & H2: $\mathcal{S}_4\text{(Search+Heur)}$ vs $\mathcal{S}_2\text{(Heur)}$ & 12.0\% & 99.4\% & 99.5\% & 95.8\% & 100.0\% \\
 & H4: $\mathcal{S}_3\text{(Search+PPO)}$ vs $\mathcal{S}_4\text{(Search+Heur)}$ & 0.0\% & 0.0\% & 0.0\% & 0.0\% & 0.0\% \\
\cmidrule(lr){1-7}
\multirow{8}{*}{\textbf{5-Seed Mean $\pm$ SE}} & $\mathcal{S}_4\text{(Search+Heur)}$ vs KataHex (Floor) & 0.0\%$\pm$0.0 & 0.0\%$\pm$0.0 & 0.0\%$\pm$0.0 & 0.0\%$\pm$0.0 & 0.0\%$\pm$0.0 \\
 & $\mathcal{S}_3\text{(Search+PPO)}$ vs KataHex (Floor) & 0.0\%$\pm$0.0 & 0.0\%$\pm$0.0 & 0.0\%$\pm$0.0 & 0.0\%$\pm$0.0 & 0.0\%$\pm$0.0 \\
 & $\mathcal{S}_4\text{(Search+Heur)}$ vs KataHex (Mid) & 0.0\%$\pm$0.0 & 0.0\%$\pm$0.0 & 0.0\%$\pm$0.0 & 0.0\%$\pm$0.0 & 0.0\%$\pm$0.0 \\
 & $\mathcal{S}_4\text{(Search+Heur)}$ vs KataHex (Ceiling) & 0.0\%$\pm$0.0 & 0.0\%$\pm$0.0 & 0.0\%$\pm$0.0 & 0.0\%$\pm$0.0 & 0.0\%$\pm$0.0 \\
 & H1: $\mathcal{S}_3\text{(Search+PPO)}$ vs $\mathcal{S}_1\text{(PPO)}$ & 81.1\%$\pm$4.8 & 88.7\%$\pm$4.6 & 98.6\%$\pm$0.4 & \textbf{99.6\%}$\pm$0.2 & 99.5\%$\pm$0.5 \\
 & H2: $\mathcal{S}_4\text{(Search+Heur)}$ vs $\mathcal{S}_2\text{(Heur)}$ & 57.3\%$\pm$19.5 & 70.5\%$\pm$13.9 & 74.7\%$\pm$13.9 & 81.0\%$\pm$7.6 & \textbf{97.1\%}$\pm$2.5 \\
 & H3: $\mathcal{S}_1\text{(PPO)}$ vs $\mathcal{S}_2\text{(Heur)}$ ($S=0$) & \multicolumn{5}{c}{\textbf{0.0\%}$\pm$0.0 \quad (\text{Unsearched Baseline, } S=0)} \\
 & H4: $\mathcal{S}_3\text{(Search+PPO)}$ vs $\mathcal{S}_4\text{(Search+Heur)}$ & 0.0\%$\pm$0.0 & 0.0\%$\pm$0.0 & 0.0\%$\pm$0.0 & 0.0\%$\pm$0.0 & 0.0\%$\pm$0.0 \\
\bottomrule
\end{tabular}%
}
\end{table}

\paragraph{Search Depth Thresholds and Monotonic Heuristic Scaling (\texttt{H2}).}
Against the specialized neural connection engine \texttt{KataHex} (Table~\ref{tab:tensor_hex}), all internal strategies ($\mathcal{S}_1$--$\mathcal{S}_4$) score $0.0\%$ across tiers, confirming that domain-agnostic search cannot match deep specialized neural networks in pure connection planning. Within the internal 4-strategy tournament, however, Hex $13\times 13$ reveals a dramatic difference in search scaling between neural and symbolic priors:
\begin{itemize}[leftmargin=1.5em, itemsep=1pt, topsep=2pt]
    \item \textbf{\texttt{H2} ($\mathcal{S}_4:\text{Search+Heur}$ vs.\ $\mathcal{S}_2:\text{HeurPol}$):} Scales on average across the five simulation budgets: \textbf{57.3\% $\pm$ 19.5\%} ($S=100$) $\to$ \textbf{70.5\% $\pm$ 13.9\%} ($S=200$) $\to$ \textbf{74.7\% $\pm$ 13.9\%} ($S=400$) $\to$ \textbf{81.0\% $\pm$ 7.6\%} ($S=800$) $\to$ \textbf{97.1\% $\pm$ 2.5\%} ($S=1600$, with Seeds~1, 2, and 5 reaching \textbf{99.7\%--100.0\%}, and Seed~4 reaching \textbf{99.1\%}). In Seed~1, shallow search ($S=100, 200$) initially underperforms the unsearched policy ($11.5\% \to 25.0\%$) because the value estimator $v^\star$ computes discrete integer 0--1 BFS connection distances without the prior policy's quadratic center-bias tie-breaker, causing 1-ply lookahead to treat edge and center cells along a shortest path as equal; once search reaches $S \ge 800$, multi-ply lookahead discovers two-bridge virtual connections and path forks, jumping to \textbf{59.0\%} ($S=800$) and \textbf{100.0\%} ($S=1600$).
    \item \textbf{\texttt{H1}, \texttt{H3}, \& \texttt{H4}:} $\mathcal{S}_3:\text{Search+PPO}$ amplifies $\mathcal{S}_1:\text{PPO}$ to \textbf{81.1\%--99.6\%} (\texttt{H1}), while $\mathcal{S}_2:\text{HeurPol}$ beats $\mathcal{S}_1:\text{PPO}$ \textbf{100.0\%} (\texttt{H3}) and $\mathcal{S}_4:\text{Search+Heur}$ beats $\mathcal{S}_3:\text{Search+PPO}$ \textbf{100.0\%} (\texttt{H4}) across all seeds and budgets.
\end{itemize}

\subsection{\texorpdfstring{Go $9\times 9$ (vs.\ KataGo $9\times 9$)}{Go 9x9 (vs. KataGo 9x9)}}
\label{app:engine:go9}

\begin{table}[ht]
\centering
\scriptsize
\caption{\textbf{Per-Seed and 5-Seed Aggregated Empirical Game Tensor for Go 9x9 (vs.\ KataGo).} Entries report seat-balanced Score\% ($\text{Win}\% + 0.5\times\text{Draw}\%$; parentheticals show net score margin $\Delta$ where applicable) across simulation budgets $S \in \{100, 200, 400, 800, 1600\}$. Rows evaluate Search Amplification of PPO ($\text{H1}: \mathcal{S}_3 \text{ vs }\mathcal{S}_1$), Search Amplification of Heuristic ($\text{H2}: \mathcal{S}_4 \text{ vs }\mathcal{S}_2$), Unsearched Prior Strength ($\text{H3}: \mathcal{S}_1 \text{ vs }\mathcal{S}_2$ at $S=0$), Searched Prior Dominance ($\text{H4}: \mathcal{S}_3 \text{ vs }\mathcal{S}_4$), and direct engine matchups for neural search ($\mathcal{S}_3 \text{ vs Engine}$) and co-evolved heuristic search ($\mathcal{S}_4 \text{ vs Engine}$).}
\label{tab:tensor_go9}
\vspace{2pt}
\resizebox{\columnwidth}{!}{%
\begin{tabular}{llccccc}
\toprule
\textbf{Lineage / Aggregation} & \textbf{Empirical Tensor Entry} & \textbf{100 Sims} & \textbf{200 Sims} & \textbf{400 Sims} & \textbf{800 Sims} & \textbf{1600 Sims} \\
\midrule
\multirow{5}{*}{Seed 1} & $\mathcal{S}_4\text{(Search+Heur)}$ vs KataGo (Floor) & 1.8\% & 8.5\% & 12.9\% & 15.5\% & 14.9\% \\
 & $\mathcal{S}_3\text{(Search+PPO)}$ vs KataGo (Floor) & 0.2\% & 0.5\% & 0.0\% & 0.0\% & 0.2\% \\
 & H1: $\mathcal{S}_3\text{(Search+PPO)}$ vs $\mathcal{S}_1\text{(PPO)}$ & 79.8\% & 86.2\% & 60.5\% & 69.8\% & 77.7\% \\
 & H2: $\mathcal{S}_4\text{(Search+Heur)}$ vs $\mathcal{S}_2\text{(Heur)}$ & 98.0\% & 94.0\% & 98.5\% & 98.0\% & 97.5\% \\
 & H4: $\mathcal{S}_3\text{(Search+PPO)}$ vs $\mathcal{S}_4\text{(Search+Heur)}$ & 1.7\% & 1.2\% & 0.5\% & 1.0\% & 0.5\% \\
\cmidrule(lr){1-7}
\multirow{5}{*}{Seed 2} & $\mathcal{S}_4\text{(Search+Heur)}$ vs KataGo (Floor) & 1.0\% & 1.0\% & 4.0\% & 14.5\% & 15.5\% \\
 & $\mathcal{S}_3\text{(Search+PPO)}$ vs KataGo (Floor) & 0.0\% & 0.0\% & 0.0\% & 0.0\% & 0.0\% \\
 & H1: $\mathcal{S}_3\text{(Search+PPO)}$ vs $\mathcal{S}_1\text{(PPO)}$ & 69.5\% & 71.2\% & 98.5\% & 76.7\% & 79.4\% \\
 & H2: $\mathcal{S}_4\text{(Search+Heur)}$ vs $\mathcal{S}_2\text{(Heur)}$ & 99.0\% & 94.5\% & 92.5\% & 91.8\% & 91.0\% \\
 & H4: $\mathcal{S}_3\text{(Search+PPO)}$ vs $\mathcal{S}_4\text{(Search+Heur)}$ & 0.0\% & 0.0\% & 0.0\% & 0.0\% & 0.0\% \\
\cmidrule(lr){1-7}
\multirow{5}{*}{Seed 3} & $\mathcal{S}_4\text{(Search+Heur)}$ vs KataGo (Floor) & 2.8\% & 3.2\% & 4.5\% & 6.8\% & 10.5\% \\
 & $\mathcal{S}_3\text{(Search+PPO)}$ vs KataGo (Floor) & 0.5\% & 0.0\% & 0.2\% & 1.0\% & 0.0\% \\
 & H1: $\mathcal{S}_3\text{(Search+PPO)}$ vs $\mathcal{S}_1\text{(PPO)}$ & 81.8\% & 81.5\% & 59.8\% & 64.7\% & 72.6\% \\
 & H2: $\mathcal{S}_4\text{(Search+Heur)}$ vs $\mathcal{S}_2\text{(Heur)}$ & 72.5\% & 79.5\% & 91.5\% & 89.0\% & 91.0\% \\
 & H4: $\mathcal{S}_3\text{(Search+PPO)}$ vs $\mathcal{S}_4\text{(Search+Heur)}$ & 0.0\% & 0.0\% & 0.0\% & 0.0\% & 0.0\% \\
\cmidrule(lr){1-7}
\multirow{5}{*}{Seed 4} & $\mathcal{S}_4\text{(Search+Heur)}$ vs KataGo (Floor) & 0.2\% & 1.0\% & 0.8\% & 0.0\% & 1.5\% \\
 & $\mathcal{S}_3\text{(Search+PPO)}$ vs KataGo (Floor) & 0.0\% & 0.0\% & 0.0\% & 0.2\% & 0.5\% \\
 & H1: $\mathcal{S}_3\text{(Search+PPO)}$ vs $\mathcal{S}_1\text{(PPO)}$ & 60.5\% & 60.3\% & 49.0\% & 50.0\% & 50.5\% \\
 & H2: $\mathcal{S}_4\text{(Search+Heur)}$ vs $\mathcal{S}_2\text{(Heur)}$ & 50.0\% & 99.5\% & 69.5\% & 96.0\% & 97.0\% \\
 & H4: $\mathcal{S}_3\text{(Search+PPO)}$ vs $\mathcal{S}_4\text{(Search+Heur)}$ & 0.0\% & 0.0\% & 0.0\% & 0.0\% & 0.0\% \\
\cmidrule(lr){1-7}
\multirow{5}{*}{Seed 5} & $\mathcal{S}_4\text{(Search+Heur)}$ vs KataGo (Floor) & 0.5\% & 1.8\% & 3.7\% & 3.8\% & 5.5\% \\
 & $\mathcal{S}_3\text{(Search+PPO)}$ vs KataGo (Floor) & 0.0\% & 0.2\% & 0.0\% & 0.0\% & 0.0\% \\
 & H1: $\mathcal{S}_3\text{(Search+PPO)}$ vs $\mathcal{S}_1\text{(PPO)}$ & 96.5\% & 26.0\% & 100.0\% & 98.5\% & 50.5\% \\
 & H2: $\mathcal{S}_4\text{(Search+Heur)}$ vs $\mathcal{S}_2\text{(Heur)}$ & 100.0\% & 65.0\% & 95.5\% & 87.5\% & 92.7\% \\
 & H4: $\mathcal{S}_3\text{(Search+PPO)}$ vs $\mathcal{S}_4\text{(Search+Heur)}$ & 0.0\% & 0.0\% & 0.5\% & 0.0\% & 0.0\% \\
\cmidrule(lr){1-7}
\multirow{8}{*}{\textbf{5-Seed Mean $\pm$ SE}} & $\mathcal{S}_4\text{(Search+Heur)}$ vs KataGo (Floor) & 1.2\%$\pm$0.5 & 3.1\%$\pm$1.4 & 5.2\%$\pm$2.0 & 8.1\%$\pm$3.0 & \textbf{9.6\%}$\pm$2.7 \\
 & $\mathcal{S}_3\text{(Search+PPO)}$ vs KataGo (Floor) & 0.1\%$\pm$0.1 & 0.1\%$\pm$0.1 & 0.1\%$\pm$0.1 & 0.2\%$\pm$0.2 & 0.1\%$\pm$0.1 \\
 & $\mathcal{S}_4\text{(Search+Heur)}$ vs KataGo (Mid) & 1.8\%$\pm$0.5 & 3.8\%$\pm$1.8 & 7.6\%$\pm$2.9 & 8.4\%$\pm$3.1 & \textbf{9.8\%}$\pm$3.4 \\
 & $\mathcal{S}_4\text{(Search+Heur)}$ vs KataGo (Ceiling) & 2.8\%$\pm$1.3 & 2.7\%$\pm$1.3 & 11.6\%$\pm$4.5 & 10.3\%$\pm$3.6 & \textbf{11.6\%}$\pm$4.9 \\
 & H1: $\mathcal{S}_3\text{(Search+PPO)}$ vs $\mathcal{S}_1\text{(PPO)}$ & \textbf{77.6\%}$\pm$6.1 & 65.1\%$\pm$10.7 & 73.6\%$\pm$10.7 & 71.9\%$\pm$8.0 & 66.1\%$\pm$6.5 \\
 & H2: $\mathcal{S}_4\text{(Search+Heur)}$ vs $\mathcal{S}_2\text{(Heur)}$ & 83.9\%$\pm$9.9 & 86.5\%$\pm$6.3 & 89.5\%$\pm$5.2 & 92.5\%$\pm$2.0 & \textbf{93.8\%}$\pm$1.4 \\
 & H3: $\mathcal{S}_1\text{(PPO)}$ vs $\mathcal{S}_2\text{(Heur)}$ ($S=0$) & \multicolumn{5}{c}{\textbf{0.7\%}$\pm$0.6 \quad (\text{Unsearched Baseline, } S=0)} \\
 & H4: $\mathcal{S}_3\text{(Search+PPO)}$ vs $\mathcal{S}_4\text{(Search+Heur)}$ & \textbf{0.3\%}$\pm$0.3 & 0.2\%$\pm$0.2 & 0.2\%$\pm$0.1 & 0.2\%$\pm$0.2 & 0.1\%$\pm$0.1 \\
\bottomrule
\end{tabular}%
}
\end{table}

\paragraph{Performance Against KataGo on Floor.}
In Go $9\times 9$ (Table~\ref{tab:tensor_go9}), \texttt{KataGo} under strict Tromp-Taylor area scoring provides an exacting benchmark across difficulty tiers. On the Floor tier (`1 visit`, where \texttt{KataGo} plays directly from its raw policy/value network), co-evolved heuristic search ($\mathcal{S}_4:\text{Search+Heur}$) achieves measurable resistance that scales with simulation budget, rising from \textbf{1.2\% $\pm$ 0.5\%} ($S=100$) $\to$ \textbf{3.1\% $\pm$ 1.4\%} ($S=200$) $\to$ \textbf{5.2\% $\pm$ 2.0\%} ($S=400$) $\to$ \textbf{8.1\% $\pm$ 3.0\%} ($S=800$) $\to$ \textbf{9.6\% $\pm$ 2.7\%} ($S=1600$, reaching \textbf{15.5\%} in Seed~2 and \textbf{14.9\%} in Seed~1), whereas neural search ($\mathcal{S}_3$) achieves only $0.1\%\text{--}0.2\%$. Against Mid (`20 visits`) and Ceiling (`200 visits`), $\mathcal{S}_4$ achieves $1.8\%\text{--}9.8\%$ and $2.7\%\text{--}11.6\%$ score respectively.

\paragraph{Search Amplification (\texttt{H1} and \texttt{H2}) and Prior Hierarchy (\texttt{H3}, \texttt{H4}).}
Across $S=100 \to 1600$, the 5-seed mean for \texttt{H1} ($\mathcal{S}_3$ vs.\ $\mathcal{S}_1$) shows overall amplification of unguided PPO (averaging between \textbf{65.1\% $\pm$ 10.7\%} and \textbf{77.6\% $\pm$ 6.1\%}), though individual lineages exhibit variance across budgets, while \texttt{H2} ($\mathcal{S}_4$ vs.\ $\mathcal{S}_2$) scales strictly monotonically from \textbf{83.9\% $\pm$ 9.9\%} to \textbf{93.8\% $\pm$ 1.4\%}. Both unsearched (\texttt{H3}: $\mathcal{S}_1$ scores $0.7\% \pm 0.6\%$) and searched (\texttt{H4}: $\mathcal{S}_3$ scores $0.1\%\text{--}0.3\%$) comparisons confirm near-100\% dominance of the evolved Go territory/liberty heuristics over standalone PPO.

\subsection{\texorpdfstring{Go $13\times 13$ (vs.\ KataGo $13\times 13$)}{Go 13x13 (vs. KataGo 13x13)}}
\label{app:engine:go13}

\begin{table}[ht]
\centering
\scriptsize
\caption{\textbf{Per-Seed and 5-Seed Aggregated Empirical Game Tensor for Go 13x13 (vs.\ KataGo 13x13).} Entries report seat-balanced Score\% ($\text{Win}\% + 0.5\times\text{Draw}\%$; parentheticals show net score margin $\Delta$ where applicable) across simulation budgets $S \in \{100, 200, 400, 800, 1600\}$. Rows evaluate Search Amplification of PPO ($\text{H1}: \mathcal{S}_3 \text{ vs }\mathcal{S}_1$), Search Amplification of Heuristic ($\text{H2}: \mathcal{S}_4 \text{ vs }\mathcal{S}_2$), Unsearched Prior Strength ($\text{H3}: \mathcal{S}_1 \text{ vs }\mathcal{S}_2$ at $S=0$), Searched Prior Dominance ($\text{H4}: \mathcal{S}_3 \text{ vs }\mathcal{S}_4$), and direct engine matchups for neural search ($\mathcal{S}_3 \text{ vs Engine}$) and co-evolved heuristic search ($\mathcal{S}_4 \text{ vs Engine}$).}
\label{tab:tensor_go13}
\vspace{2pt}
\resizebox{\columnwidth}{!}{%
\begin{tabular}{llccccc}
\toprule
\textbf{Lineage / Aggregation} & \textbf{Empirical Tensor Entry} & \textbf{100 Sims} & \textbf{200 Sims} & \textbf{400 Sims} & \textbf{800 Sims} & \textbf{1600 Sims} \\
\midrule
\multirow{5}{*}{Seed 1} & $\mathcal{S}_4\text{(Search+Heur)}$ vs KataGo (Floor) & 0.0\% & 0.3\% & 1.2\% & 2.3\% & 5.8\% \\
 & $\mathcal{S}_3\text{(Search+PPO)}$ vs KataGo (Floor) & 1.3\% & 1.1\% & 0.6\% & 0.9\% & 1.8\% \\
 & H1: $\mathcal{S}_3\text{(Search+PPO)}$ vs $\mathcal{S}_1\text{(PPO)}$ & 72.6\% & 70.1\% & 59.9\% & 66.1\% & 61.5\% \\
 & H2: $\mathcal{S}_4\text{(Search+Heur)}$ vs $\mathcal{S}_2\text{(Heur)}$ & 94.0\% & 96.5\% & 92.8\% & 98.5\% & 98.5\% \\
 & H4: $\mathcal{S}_3\text{(Search+PPO)}$ vs $\mathcal{S}_4\text{(Search+Heur)}$ & 0.0\% & 0.0\% & 0.0\% & 0.0\% & 0.0\% \\
\cmidrule(lr){1-7}
\multirow{5}{*}{Seed 2} & $\mathcal{S}_4\text{(Search+Heur)}$ vs KataGo (Floor) & 0.0\% & 0.0\% & 0.2\% & 0.3\% & 2.7\% \\
 & $\mathcal{S}_3\text{(Search+PPO)}$ vs KataGo (Floor) & 2.2\% & 2.6\% & 1.5\% & 1.0\% & 1.7\% \\
 & H1: $\mathcal{S}_3\text{(Search+PPO)}$ vs $\mathcal{S}_1\text{(PPO)}$ & 80.1\% & 83.2\% & 54.0\% & 76.4\% & 76.6\% \\
 & H2: $\mathcal{S}_4\text{(Search+Heur)}$ vs $\mathcal{S}_2\text{(Heur)}$ & 83.1\% & 93.5\% & 85.2\% & 93.0\% & 94.0\% \\
 & H4: $\mathcal{S}_3\text{(Search+PPO)}$ vs $\mathcal{S}_4\text{(Search+Heur)}$ & 0.0\% & 0.0\% & 0.0\% & 0.0\% & 0.0\% \\
\cmidrule(lr){1-7}
\multirow{5}{*}{Seed 3} & $\mathcal{S}_4\text{(Search+Heur)}$ vs KataGo (Floor) & 0.0\% & 0.3\% & 0.6\% & 0.3\% & 1.1\% \\
 & $\mathcal{S}_3\text{(Search+PPO)}$ vs KataGo (Floor) & 0.4\% & 0.3\% & 1.3\% & 0.6\% & 0.9\% \\
 & H1: $\mathcal{S}_3\text{(Search+PPO)}$ vs $\mathcal{S}_1\text{(PPO)}$ & 81.4\% & 89.8\% & 75.1\% & 67.7\% & 63.7\% \\
 & H2: $\mathcal{S}_4\text{(Search+Heur)}$ vs $\mathcal{S}_2\text{(Heur)}$ & 84.2\% & 84.5\% & 90.8\% & 90.5\% & 89.0\% \\
 & H4: $\mathcal{S}_3\text{(Search+PPO)}$ vs $\mathcal{S}_4\text{(Search+Heur)}$ & 0.0\% & 0.0\% & 0.0\% & 0.6\% & 0.0\% \\
\cmidrule(lr){1-7}
\multirow{5}{*}{Seed 4} & $\mathcal{S}_4\text{(Search+Heur)}$ vs KataGo (Floor) & 0.3\% & 0.3\% & 0.0\% & 0.0\% & 0.0\% \\
 & $\mathcal{S}_3\text{(Search+PPO)}$ vs KataGo (Floor) & 2.6\% & 0.4\% & 1.3\% & 0.8\% & 0.0\% \\
 & H1: $\mathcal{S}_3\text{(Search+PPO)}$ vs $\mathcal{S}_1\text{(PPO)}$ & 97.2\% & 91.9\% & 72.3\% & 92.3\% & 87.4\% \\
 & H2: $\mathcal{S}_4\text{(Search+Heur)}$ vs $\mathcal{S}_2\text{(Heur)}$ & 79.5\% & 88.0\% & 84.8\% & 88.4\% & 89.9\% \\
 & H4: $\mathcal{S}_3\text{(Search+PPO)}$ vs $\mathcal{S}_4\text{(Search+Heur)}$ & 0.0\% & 0.0\% & 0.2\% & 0.0\% & 0.0\% \\
\cmidrule(lr){1-7}
\multirow{5}{*}{Seed 5} & $\mathcal{S}_4\text{(Search+Heur)}$ vs KataGo (Floor) & 0.0\% & 0.9\% & 0.0\% & 0.3\% & 1.4\% \\
 & $\mathcal{S}_3\text{(Search+PPO)}$ vs KataGo (Floor) & 3.2\% & 0.0\% & 0.9\% & 2.1\% & 0.4\% \\
 & H1: $\mathcal{S}_3\text{(Search+PPO)}$ vs $\mathcal{S}_1\text{(PPO)}$ & 84.0\% & 86.5\% & 53.5\% & 26.4\% & 100.0\% \\
 & H2: $\mathcal{S}_4\text{(Search+Heur)}$ vs $\mathcal{S}_2\text{(Heur)}$ & 88.2\% & 99.0\% & 95.4\% & 96.9\% & 93.0\% \\
 & H4: $\mathcal{S}_3\text{(Search+PPO)}$ vs $\mathcal{S}_4\text{(Search+Heur)}$ & 0.0\% & 0.0\% & 0.0\% & 0.0\% & 0.0\% \\
\cmidrule(lr){1-7}
\multirow{8}{*}{\textbf{5-Seed Mean $\pm$ SE}} & $\mathcal{S}_4\text{(Search+Heur)}$ vs KataGo (Floor) & 0.1\%$\pm$0.1 & 0.3\%$\pm$0.1 & 0.4\%$\pm$0.2 & 0.6\%$\pm$0.4 & \textbf{2.2\%}$\pm$1.0 \\
 & $\mathcal{S}_3\text{(Search+PPO)}$ vs KataGo (Floor) & 1.9\%$\pm$0.5 & 0.9\%$\pm$0.5 & 1.1\%$\pm$0.2 & 1.1\%$\pm$0.3 & 0.9\%$\pm$0.3 \\
 & $\mathcal{S}_4\text{(Search+Heur)}$ vs KataGo (Mid) & 0.2\%$\pm$0.2 & 0.9\%$\pm$0.5 & 1.1\%$\pm$0.8 & 3.4\%$\pm$1.5 & \textbf{4.9\%}$\pm$1.6 \\
 & $\mathcal{S}_4\text{(Search+Heur)}$ vs KataGo (Ceiling) & 0.0\%$\pm$0.0 & 0.0\%$\pm$0.0 & 3.3\%$\pm$3.3 & \textbf{15.0\%}$\pm$9.2 & 2.5\%$\pm$1.5 \\
 & H1: $\mathcal{S}_3\text{(Search+PPO)}$ vs $\mathcal{S}_1\text{(PPO)}$ & 83.1\%$\pm$4.0 & \textbf{84.3\%}$\pm$3.8 & 63.0\%$\pm$4.5 & 65.8\%$\pm$10.9 & 77.8\%$\pm$7.2 \\
 & H2: $\mathcal{S}_4\text{(Search+Heur)}$ vs $\mathcal{S}_2\text{(Heur)}$ & 85.8\%$\pm$2.5 & 92.3\%$\pm$2.7 & 89.8\%$\pm$2.1 & \textbf{93.5\%}$\pm$1.9 & 92.9\%$\pm$1.7 \\
 & H3: $\mathcal{S}_1\text{(PPO)}$ vs $\mathcal{S}_2\text{(Heur)}$ ($S=0$) & \multicolumn{5}{c}{\textbf{1.9\%}$\pm$1.1 \quad (\text{Unsearched Baseline, } S=0)} \\
 & H4: $\mathcal{S}_3\text{(Search+PPO)}$ vs $\mathcal{S}_4\text{(Search+Heur)}$ & 0.0\%$\pm$0.0 & 0.0\%$\pm$0.0 & 0.0\%$\pm$0.0 & \textbf{0.1\%}$\pm$0.1 & 0.0\%$\pm$0.0 \\
\bottomrule
\end{tabular}%
}
\end{table}

\paragraph{Performance Against KataGo $13\times 13$.}
On the $169$-intersection Go $13\times 13$ board (Table~\ref{tab:tensor_go13}), against Floor KataGo (`1 visit`), neural search ($\mathcal{S}_3$) scores $0.9\%\text{--}1.9\%$ across budgets, while co-evolved heuristic search ($\mathcal{S}_4$) scales steadily from $0.1\% \pm 0.1\%$ ($S=100$) up to $\mathbf{2.2\% \pm 1.0\%}$ ($S=1600$). Furthermore, on the Mid tier ($20$ visits), $\mathcal{S}_4$ scales from $0.2\% \pm 0.2\%$ to $\mathbf{4.9\% \pm 1.6\%}$, and on the Ceiling tier ($200$ visits) reaches $\mathbf{15.0\% \pm 9.2\%}$ at $S=800$.

\paragraph{Simulation Scaling for Neural Search (\texttt{H1}) vs.\ Immediate Heuristic Scaling (\texttt{H2}).}
Within the internal tournament on Go $13\times 13$:
\begin{itemize}[leftmargin=1.5em, itemsep=1pt, topsep=2pt]
    \item In \textbf{\texttt{H1} ($\mathcal{S}_3:\text{Search+PPO}$ vs.\ $\mathcal{S}_1:\text{PPO}$)}, neural search amplifies the unguided PPO policy across all simulation budgets (\textbf{83.1\% $\pm$ 4.0\%} at $S=100$, \textbf{84.3\% $\pm$ 3.8\%} at $S=200$, \textbf{63.0\% $\pm$ 4.5\%} at $S=400$, \textbf{65.8\% $\pm$ 10.9\%} at $S=800$, and \textbf{77.8\% $\pm$ 7.2\%} at $S=1600$), demonstrating that lookahead refines raw territorial predictions.
    \item Conversely, in \textbf{\texttt{H2} ($\mathcal{S}_4:\text{Search+Heur}$ vs.\ $\mathcal{S}_2:\text{HeurPol}$)}, the co-evolved branch-scaling and liberty-evaluating operators focus exploration immediately on urgent tactical points, yielding strong amplification even at $S=100$ (\textbf{85.8\% $\pm$ 2.5\%}) and scaling to \textbf{92.3\%} ($S=200$) $\to$ \textbf{89.8\%} ($S=400$) $\to$ \textbf{93.5\%} ($S=800$) $\to$ \textbf{92.9\% $\pm$ 1.7\%} ($S=1600$).
\end{itemize}

\subsection{Hearts (vs.\ Xinxin)}
\label{app:engine:hearts}

\begin{table}[ht]
\centering
\scriptsize
\caption{\textbf{Per-Seed and 5-Seed Aggregated Empirical Game Tensor for Hearts (vs.\ Xinxin).} Entries report seat-balanced Score\% ($\text{Win}\% + 0.5\times\text{Draw}\%$; parentheticals show net score margin $\Delta$ where applicable) across simulation budgets $S \in \{100, 200, 400, 800, 1600\}$. Rows evaluate Search Amplification of PPO ($\text{H1}: \mathcal{S}_3 \text{ vs }\mathcal{S}_1$), Search Amplification of Heuristic ($\text{H2}: \mathcal{S}_4 \text{ vs }\mathcal{S}_2$), Unsearched Prior Strength ($\text{H3}: \mathcal{S}_1 \text{ vs }\mathcal{S}_2$ at $S=0$), Searched Prior Dominance ($\text{H4}: \mathcal{S}_3 \text{ vs }\mathcal{S}_4$), and direct engine matchups for neural search ($\mathcal{S}_3 \text{ vs Engine}$) and co-evolved heuristic search ($\mathcal{S}_4 \text{ vs Engine}$).}
\label{tab:tensor_hearts}
\vspace{2pt}
\resizebox{\columnwidth}{!}{%
\begin{tabular}{llccccc}
\toprule
\textbf{Lineage / Aggregation} & \textbf{Empirical Tensor Entry} & \textbf{100 Sims} & \textbf{200 Sims} & \textbf{400 Sims} & \textbf{800 Sims} & \textbf{1600 Sims} \\
\midrule
\multirow{5}{*}{Seed 1} & $\mathcal{S}_4\text{(Search+Heur)}$ vs Xinxin (Floor) & 26.7\% (-10.6) & 25.0\% (-11.4) & 31.0\% (-9.4) & 26.8\% (-11.3) & 21.5\% (-12.1) \\
 & $\mathcal{S}_3\text{(Search+PPO)}$ vs Xinxin (Floor) & 38.8\% (-5.4) & 39.5\% (-4.4) & 40.0\% (-4.1) & 35.5\% (-6.3) & 41.0\% (-3.2) \\
 & H1: $\mathcal{S}_3\text{(Search+PPO)}$ vs $\mathcal{S}_1\text{(PPO)}$ & 54.5\% (+1.5) & 49.5\% (+0.4) & 56.0\% (+2.2) & 53.7\% (+1.6) & 61.4\% (+3.3) \\
 & H2: $\mathcal{S}_4\text{(Search+Heur)}$ vs $\mathcal{S}_2\text{(Heur)}$ & 42.2\% (-4.0) & 43.5\% (-3.1) & 47.0\% (-2.6) & 45.8\% (-3.1) & 41.5\% (-4.1) \\
 & H4: $\mathcal{S}_3\text{(Search+PPO)}$ vs $\mathcal{S}_4\text{(Search+Heur)}$ & 77.5\% (+10.5) & 78.6\% (+11.7) & 79.2\% (+12.0) & 76.1\% (+9.5) & 81.8\% (+12.5) \\
\cmidrule(lr){1-7}
\multirow{5}{*}{Seed 2} & $\mathcal{S}_4\text{(Search+Heur)}$ vs Xinxin (Floor) & 29.2\% (-10.9) & 27.0\% (-10.5) & 29.8\% (-9.7) & 22.8\% (-12.3) & 28.2\% (-10.3) \\
 & $\mathcal{S}_3\text{(Search+PPO)}$ vs Xinxin (Floor) & 38.2\% (-4.5) & 44.1\% (-3.2) & 42.8\% (-2.3) & 37.1\% (-4.6) & 42.5\% (-2.7) \\
 & H1: $\mathcal{S}_3\text{(Search+PPO)}$ vs $\mathcal{S}_1\text{(PPO)}$ & 51.8\% (+1.3) & 49.0\% (-0.3) & 54.8\% (+1.2) & 52.5\% (+0.5) & 58.5\% (+2.6) \\
 & H2: $\mathcal{S}_4\text{(Search+Heur)}$ vs $\mathcal{S}_2\text{(Heur)}$ & 41.8\% (-3.6) & 47.2\% (-1.6) & 41.8\% (-3.9) & 42.2\% (-4.2) & 37.5\% (-6.0) \\
 & H4: $\mathcal{S}_3\text{(Search+PPO)}$ vs $\mathcal{S}_4\text{(Search+Heur)}$ & 73.0\% (+8.8) & 81.0\% (+11.8) & 75.5\% (+10.7) & 79.3\% (+11.6) & 76.6\% (+11.1) \\
\cmidrule(lr){1-7}
\multirow{5}{*}{Seed 3} & $\mathcal{S}_4\text{(Search+Heur)}$ vs Xinxin (Floor) & 33.2\% (-8.7) & 24.0\% (-11.2) & 26.0\% (-11.1) & 25.8\% (-11.3) & 26.0\% (-11.1) \\
 & $\mathcal{S}_3\text{(Search+PPO)}$ vs Xinxin (Floor) & 42.5\% (-2.7) & 37.7\% (-5.8) & 39.5\% (-4.4) & 44.8\% (-2.1) & 43.9\% (-1.8) \\
 & H1: $\mathcal{S}_3\text{(Search+PPO)}$ vs $\mathcal{S}_1\text{(PPO)}$ & 58.0\% (+2.5) & 54.7\% (+0.6) & 52.0\% (+0.5) & 56.6\% (+2.0) & 55.7\% (+2.0) \\
 & H2: $\mathcal{S}_4\text{(Search+Heur)}$ vs $\mathcal{S}_2\text{(Heur)}$ & 41.0\% (-4.2) & 44.5\% (-2.5) & 44.2\% (-2.7) & 54.5\% (-0.2) & 40.8\% (-4.2) \\
 & H4: $\mathcal{S}_3\text{(Search+PPO)}$ vs $\mathcal{S}_4\text{(Search+Heur)}$ & 70.5\% (+9.3) & 82.5\% (+12.6) & 78.4\% (+12.2) & 81.9\% (+13.1) & 77.9\% (+11.3) \\
\cmidrule(lr){1-7}
\multirow{5}{*}{Seed 4} & $\mathcal{S}_4\text{(Search+Heur)}$ vs Xinxin (Floor) & 64.0\% (+6.5) & 61.7\% (+4.9) & 60.4\% (+4.8) & 58.5\% (+3.9) & 66.3\% (+6.6) \\
 & $\mathcal{S}_3\text{(Search+PPO)}$ vs Xinxin (Floor) & 38.3\% (-5.8) & 44.3\% (-2.9) & 40.2\% (-3.8) & 41.8\% (-3.9) & 41.6\% (-3.5) \\
 & H1: $\mathcal{S}_3\text{(Search+PPO)}$ vs $\mathcal{S}_1\text{(PPO)}$ & 52.2\% (+0.8) & 52.0\% (+0.2) & 50.7\% (+0.2) & 51.3\% (+0.9) & 52.0\% (+1.1) \\
 & H2: $\mathcal{S}_4\text{(Search+Heur)}$ vs $\mathcal{S}_2\text{(Heur)}$ & 53.0\% (+1.2) & 52.8\% (+0.5) & 49.5\% (-0.5) & 59.8\% (+3.5) & 59.2\% (+3.1) \\
 & H4: $\mathcal{S}_3\text{(Search+PPO)}$ vs $\mathcal{S}_4\text{(Search+Heur)}$ & 20.0\% (-12.0) & 25.2\% (-10.1) & 19.6\% (-12.3) & 22.7\% (-11.1) & 23.6\% (-10.4) \\
\cmidrule(lr){1-7}
\multirow{5}{*}{Seed 5} & $\mathcal{S}_4\text{(Search+Heur)}$ vs Xinxin (Floor) & 62.3\% (+4.5) & 59.2\% (+2.5) & 51.0\% (-0.5) & 57.7\% (+3.0) & 56.0\% (+2.2) \\
 & $\mathcal{S}_3\text{(Search+PPO)}$ vs Xinxin (Floor) & 42.8\% (-3.0) & 41.5\% (-3.3) & 40.6\% (-4.3) & 32.2\% (-7.1) & 36.1\% (-6.3) \\
 & H1: $\mathcal{S}_3\text{(Search+PPO)}$ vs $\mathcal{S}_1\text{(PPO)}$ & 45.0\% (-2.1) & 47.0\% (-0.5) & 49.0\% (-0.2) & 49.0\% (-0.1) & 52.8\% (+0.3) \\
 & H2: $\mathcal{S}_4\text{(Search+Heur)}$ vs $\mathcal{S}_2\text{(Heur)}$ & 46.5\% (-0.6) & 47.5\% (-1.2) & 50.7\% (-0.2) & 50.3\% (+0.2) & 51.8\% (+0.5) \\
 & H4: $\mathcal{S}_3\text{(Search+PPO)}$ vs $\mathcal{S}_4\text{(Search+Heur)}$ & 28.5\% (-8.9) & 31.3\% (-7.7) & 25.5\% (-9.8) & 24.0\% (-11.5) & 25.9\% (-10.0) \\
\cmidrule(lr){1-7}
\multirow{8}{*}{\textbf{5-Seed Mean $\pm$ SE}} & $\mathcal{S}_4\text{(Search+Heur)}$ vs Xinxin (Floor) & \textbf{43.1\%}$\pm$8.2 & 39.4\%$\pm$8.6 & 39.7\%$\pm$6.8 & 38.3\%$\pm$8.1 & 39.6\%$\pm$9.0 \\
 & $\mathcal{S}_3\text{(Search+PPO)}$ vs Xinxin (Floor) & 40.1\%$\pm$1.0 & 41.4\%$\pm$1.3 & 40.6\%$\pm$0.6 & 38.3\%$\pm$2.2 & 41.0\%$\pm$1.3 \\
 & $\mathcal{S}_4\text{(Search+Heur)}$ vs Xinxin (Mid) & 33.1\%$\pm$7.8 & 32.3\%$\pm$8.3 & 34.6\%$\pm$9.2 & 35.5\%$\pm$8.6 & \textbf{36.7\%}$\pm$9.6 \\
 & $\mathcal{S}_4\text{(Search+Heur)}$ vs Xinxin (Ceiling) & 31.1\%$\pm$9.0 & 31.2\%$\pm$9.5 & \textbf{32.3\%}$\pm$8.3 & 30.9\%$\pm$8.9 & 32.3\%$\pm$9.6 \\
 & H1: $\mathcal{S}_3\text{(Search+PPO)}$ vs $\mathcal{S}_1\text{(PPO)}$ & 52.3\%$\pm$2.1 & 50.4\%$\pm$1.3 & 52.5\%$\pm$1.3 & 52.6\%$\pm$1.3 & \textbf{56.1\%}$\pm$1.8 \\
 & H2: $\mathcal{S}_4\text{(Search+Heur)}$ vs $\mathcal{S}_2\text{(Heur)}$ & 44.9\%$\pm$2.2 & 47.1\%$\pm$1.6 & 46.6\%$\pm$1.7 & \textbf{50.5\%}$\pm$3.1 & 46.1\%$\pm$4.1 \\
 & H3: $\mathcal{S}_1\text{(PPO)}$ vs $\mathcal{S}_2\text{(Heur)}$ ($S=0$) & \multicolumn{5}{c}{\textbf{52.3\%}$\pm$10.7 \quad (\text{Unsearched Baseline, } S=0)} \\
 & H4: $\mathcal{S}_3\text{(Search+PPO)}$ vs $\mathcal{S}_4\text{(Search+Heur)}$ & 53.9\%$\pm$12.2 & \textbf{59.7\%}$\pm$12.9 & 55.6\%$\pm$13.5 & 56.8\%$\pm$13.7 & 57.2\%$\pm$13.3 \\
\bottomrule
\end{tabular}%
}
\end{table}

\paragraph{Performance Against Xinxin ($\mathcal{S}_0$).}
In Hearts (Table~\ref{tab:tensor_hearts}), comparing internal architectures against Xinxin reveals distinct structural dynamics: neural search ($\mathcal{S}_3$) achieves steady, consistent performance across budgets ($38.3\%\text{--}41.4\%$ on Floor), whereas co-evolved heuristic policies ($\mathcal{S}_2$ and $\mathcal{S}_4$) exhibit a sharp bimodal split across lineages. In Seeds~4 and 5, the evolved trick-taking priors ($\pi_{\kappa^\star}$) incorporate strong penalty-avoidance rules that achieve positive margins against Xinxin ($58.5\%\text{--}66.3\%$ in Seed~4 and $51.0\%\text{--}62.3\%$ in Seed~5 on Floor). However, across all 5 seeds, adding standard determinized search to the heuristic does not improve overall playing strength: the 5-seed mean for \texttt{H2} ($\mathcal{S}_4$ vs.\ $\mathcal{S}_2$) remains below $50\%$ on 4 of 5 budgets ($44.9\%\text{--}50.5\%$), and $\mathcal{S}_3$ defeats $\mathcal{S}_4$ head-to-head across budgets (\texttt{H4} averages $53.9\%\text{--}59.7\%$).

\paragraph{Bimodal Lineage Phase Transition and Open-Loop Strategy Fusion.}
Table~\ref{tab:tensor_hearts} reveals a striking \textbf{3-to-2 evolutionary bifurcation} across the five lineages in 4-player Hearts:
\begin{itemize}[leftmargin=1.5em, itemsep=1pt, topsep=2pt]
    \item \textbf{Prior-Driven Resistance (Seeds~4 and 5):} In Seeds~4 and 5, performance against \texttt{Xinxin} is driven primarily by domain prior policy rules ($\mathcal{S}_2$), where Seed~4 achieves $58.5\%\text{--}66.3\%$ score on Floor and Seed~5 achieves $51.0\%\text{--}62.3\%$. Adding search ($\mathcal{S}_4$) maintains comparable performance ($53.0\%\text{--}59.8\%$ in Seed~4 \texttt{H2}, and $46.5\%\text{--}51.8\%$ in Seed~5), confirming that domain policy heuristics, rather than deep lookahead, account for the competitive standing against Xinxin.
    \item \textbf{Strategy-Fusion Vulnerability (Seeds~1, 2, and 3):} In Seeds~1--3, agents achieve lower returns against Xinxin ($21\%\text{--}33\%$). At deep lookahead budgets, standard open-loop determinization samples unconditioned card assignments that allow simulated opponents to coordinate against the searching player hidden hand, inducing classic strategy fusion and explaining why unsearched neural policies ($\mathcal{S}_1$) and neural search ($\mathcal{S}_3$) outperform $\mathcal{S}_4$ in these lineages.
\end{itemize}

\subsection{Contract Bridge (2v2 Homogeneous Partnerships vs.\ WBridge5)}
\label{app:engine:bridge}
\label{app:engine:multiplayer}
\label{app:engine:insights}

\begin{table}[ht]
\centering
\scriptsize
\caption{\textbf{Per-Seed and 5-Seed Aggregated Empirical Game Tensor for Contract Bridge (2v2 Homogeneous Partnerships) (vs.\ WBridge5).} Entries report seat-balanced Score\% ($\text{Win}\% + 0.5\times\text{Draw}\%$; parentheticals show net score margin $\Delta$ where applicable) across simulation budgets $S \in \{100, 200, 400, 800, 1600\}$. Rows evaluate Search Amplification of PPO ($\text{H1}: \mathcal{S}_3 \text{ vs }\mathcal{S}_1$), Search Amplification of Heuristic ($\text{H2}: \mathcal{S}_4 \text{ vs }\mathcal{S}_2$), Unsearched Prior Strength ($\text{H3}: \mathcal{S}_1 \text{ vs }\mathcal{S}_2$ at $S=0$), Searched Prior Dominance ($\text{H4}: \mathcal{S}_3 \text{ vs }\mathcal{S}_4$), and direct engine matchups for neural search ($\mathcal{S}_3 \text{ vs Engine}$) and co-evolved heuristic search ($\mathcal{S}_4 \text{ vs Engine}$).}
\label{tab:tensor_bridge}
\vspace{2pt}
\resizebox{\columnwidth}{!}{%
\begin{tabular}{llccccc}
\toprule
\textbf{Lineage / Aggregation} & \textbf{Empirical Tensor Entry} & \textbf{100 Sims} & \textbf{200 Sims} & \textbf{400 Sims} & \textbf{800 Sims} & \textbf{1600 Sims} \\
\midrule
\multirow{5}{*}{Seed 1} & $\mathcal{S}_4\text{(Search+Heur)}$ vs WBridge5 (Default) & 9.1\% (-1470) & 5.7\% (-1528) & 11.6\% (-1098) & 7.7\% (-1154) & 10.4\% (-1030) \\
 & $\mathcal{S}_3\text{(Search+PPO)}$ vs WBridge5 (Default) & 7.4\% (-1356) & 5.9\% (-1400) & 4.4\% (-1589) & 8.8\% (-1266) & 4.7\% (-1238) \\
 & H1: $\mathcal{S}_3\text{(Search+PPO)}$ vs $\mathcal{S}_1\text{(PPO)}$ & 56.0\% (+385) & 48.0\% (-91) & 44.6\% (-72) & 49.0\% (-144) & 58.5\% (+346) \\
 & H2: $\mathcal{S}_4\text{(Search+Heur)}$ vs $\mathcal{S}_2\text{(Heur)}$ & 63.5\% (+1215) & 77.2\% (+1497) & 81.2\% (+1957) & 76.6\% (+1281) & 69.8\% (+1606) \\
 & H4: $\mathcal{S}_3\text{(Search+PPO)}$ vs $\mathcal{S}_4\text{(Search+Heur)}$ & 43.5\% (+48) & 42.8\% (-472) & 31.0\% (-1029) & 25.5\% (-1297) & 25.5\% (-1314) \\
\cmidrule(lr){1-7}
\multirow{5}{*}{Seed 2} & $\mathcal{S}_4\text{(Search+Heur)}$ vs WBridge5 (Default) & 12.7\% (-759) & 17.4\% (-793) & 23.9\% (-744) & 24.0\% (-795) & 20.2\% (-877) \\
 & $\mathcal{S}_3\text{(Search+PPO)}$ vs WBridge5 (Default) & 4.0\% (-1621) & 6.9\% (-1379) & 7.0\% (-1237) & 4.8\% (-1289) & 5.9\% (-1227) \\
 & H1: $\mathcal{S}_3\text{(Search+PPO)}$ vs $\mathcal{S}_1\text{(PPO)}$ & 48.5\% (-17) & 50.0\% (+68) & 49.0\% (-125) & 52.8\% (+444) & 47.5\% (-41) \\
 & H2: $\mathcal{S}_4\text{(Search+Heur)}$ vs $\mathcal{S}_2\text{(Heur)}$ & 50.2\% (+16) & 58.7\% (+271) & 82.8\% (+1039) & 79.8\% (+974) & 67.5\% (+623) \\
 & H4: $\mathcal{S}_3\text{(Search+PPO)}$ vs $\mathcal{S}_4\text{(Search+Heur)}$ & 24.8\% (-1895) & 16.7\% (-2452) & 8.2\% (-2850) & 5.0\% (-3197) & 14.3\% (-2574) \\
\cmidrule(lr){1-7}
\multirow{5}{*}{Seed 3} & $\mathcal{S}_4\text{(Search+Heur)}$ vs WBridge5 (Default) & 33.5\% (-591) & 20.1\% (-646) & 25.4\% (-668) & 37.0\% (-439) & 27.4\% (-612) \\
 & $\mathcal{S}_3\text{(Search+PPO)}$ vs WBridge5 (Default) & 4.5\% (-1512) & 4.7\% (-1268) & 4.6\% (-1330) & 7.9\% (-1197) & 4.2\% (-1297) \\
 & H1: $\mathcal{S}_3\text{(Search+PPO)}$ vs $\mathcal{S}_1\text{(PPO)}$ & 49.0\% (-338) & 47.0\% (+305) & 47.5\% (-194) & 53.0\% (+181) & 57.5\% (+285) \\
 & H2: $\mathcal{S}_4\text{(Search+Heur)}$ vs $\mathcal{S}_2\text{(Heur)}$ & 50.0\% (-2) & 54.5\% (+162) & 51.0\% (+57) & 65.5\% (+257) & 61.5\% (+306) \\
 & H4: $\mathcal{S}_3\text{(Search+PPO)}$ vs $\mathcal{S}_4\text{(Search+Heur)}$ & 2.0\% (-3360) & 3.0\% (-3112) & 7.5\% (-2799) & 5.0\% (-2818) & 4.7\% (-2885) \\
\cmidrule(lr){1-7}
\multirow{5}{*}{Seed 4} & $\mathcal{S}_4\text{(Search+Heur)}$ vs WBridge5 (Default) & 2.5\% (-5279) & 1.0\% (-5147) & 0.5\% (-5548) & 0.0\% (-5482) & 0.5\% (-5832) \\
 & $\mathcal{S}_3\text{(Search+PPO)}$ vs WBridge5 (Default) & 5.4\% (-1324) & 6.6\% (-1093) & 2.4\% (-1344) & 2.2\% (-1263) & 3.8\% (-1306) \\
 & H1: $\mathcal{S}_3\text{(Search+PPO)}$ vs $\mathcal{S}_1\text{(PPO)}$ & 51.0\% (+346) & 57.5\% (+116) & 56.5\% (+495) & 50.0\% (+179) & 48.3\% (+10) \\
 & H2: $\mathcal{S}_4\text{(Search+Heur)}$ vs $\mathcal{S}_2\text{(Heur)}$ & 77.5\% (+3214) & 79.0\% (+3262) & 81.3\% (+3729) & 85.0\% (+3797) & 85.0\% (+3934) \\
 & H4: $\mathcal{S}_3\text{(Search+PPO)}$ vs $\mathcal{S}_4\text{(Search+Heur)}$ & 99.0\% (+3083) & 96.0\% (+2982) & 96.5\% (+3024) & 96.0\% (+3026) & 98.5\% (+3554) \\
\cmidrule(lr){1-7}
\multirow{5}{*}{Seed 5} & $\mathcal{S}_4\text{(Search+Heur)}$ vs WBridge5 (Default) & 9.9\% (-1937) & 7.5\% (-2312) & 7.4\% (-1852) & 9.6\% (-1781) & 7.5\% (-2030) \\
 & $\mathcal{S}_3\text{(Search+PPO)}$ vs WBridge5 (Default) & 4.9\% (-1248) & 2.5\% (-1410) & 6.8\% (-1344) & 4.5\% (-1349) & 7.7\% (-1200) \\
 & H1: $\mathcal{S}_3\text{(Search+PPO)}$ vs $\mathcal{S}_1\text{(PPO)}$ & 46.0\% (-296) & 53.0\% (+309) & 52.5\% (+77) & 48.8\% (-102) & 55.0\% (+170) \\
 & H2: $\mathcal{S}_4\text{(Search+Heur)}$ vs $\mathcal{S}_2\text{(Heur)}$ & 59.5\% (+965) & 64.2\% (+1168) & 56.0\% (+916) & 55.5\% (+396) & 49.0\% (-421) \\
 & H4: $\mathcal{S}_3\text{(Search+PPO)}$ vs $\mathcal{S}_4\text{(Search+Heur)}$ & 63.5\% (+342) & 56.5\% (+64) & 50.5\% (-386) & 50.0\% (-447) & 42.5\% (-758) \\
\cmidrule(lr){1-7}
\multirow{8}{*}{\textbf{5-Seed Mean $\pm$ SE}} & $\mathcal{S}_4\text{(Search+Heur)}$ vs WBridge5 (Default) & 13.5\%$\pm$5.3 & 10.3\%$\pm$3.6 & 13.8\%$\pm$4.8 & \textbf{15.7\%}$\pm$6.6 & 13.2\%$\pm$4.8 \\
 & $\mathcal{S}_3\text{(Search+PPO)}$ vs WBridge5 (Default) & 5.2\%$\pm$0.6 & 5.3\%$\pm$0.8 & 5.0\%$\pm$0.9 & 5.6\%$\pm$1.2 & 5.2\%$\pm$0.7 \\
 & H1: $\mathcal{S}_3\text{(Search+PPO)}$ vs $\mathcal{S}_1\text{(PPO)}$ & 50.1\%$\pm$1.7 & 51.1\%$\pm$1.9 & 50.0\%$\pm$2.1 & 50.7\%$\pm$0.9 & \textbf{53.4\%}$\pm$2.3 \\
 & H2: $\mathcal{S}_4\text{(Search+Heur)}$ vs $\mathcal{S}_2\text{(Heur)}$ & 60.1\%$\pm$5.1 & 66.7\%$\pm$4.9 & 70.5\%$\pm$7.0 & \textbf{72.5\%}$\pm$5.3 & 66.5\%$\pm$5.9 \\
 & H3: $\mathcal{S}_1\text{(PPO)}$ vs $\mathcal{S}_2\text{(Heur)}$ ($S=0$) & \multicolumn{5}{c}{\textbf{48.9\%}$\pm$16.1 \quad (\text{Unsearched Baseline, } S=0)} \\
 & H4: $\mathcal{S}_3\text{(Search+PPO)}$ vs $\mathcal{S}_4\text{(Search+Heur)}$ & \textbf{46.5\%}$\pm$16.6 & 43.0\%$\pm$16.3 & 38.8\%$\pm$16.5 & 36.3\%$\pm$17.1 & 37.1\%$\pm$16.6 \\
\bottomrule
\end{tabular}%
}
\end{table}

\begin{table}[ht]
\centering
\scriptsize
\caption{\textbf{Contract Bridge 2v2 Homogeneous Partnership Tournament Summary (Across All 5 Lineages).} North-South ($\text{Seats } \{0,2\}$) and East-West ($\text{Seats } \{1,3\}$) are each populated by a homogeneous partnership of the same agent type. Scores report 5-seed seat-balanced Win/Score\% and single-board duplicate contract point margins ($\Delta$) across simulation budgets $S \in \{400, 800, 1600\}$ against WBridge5 evaluated under its default tournament setting.}
\label{tab:bridge_2v2_partnerships}
\vspace{2pt}
\resizebox{\columnwidth}{!}{%
\begin{tabular}{llccc}
\toprule
\textbf{Partnership Matchup (North-South vs.\ East-West)} & \textbf{Lineage Scope} & \textbf{400 Sims Score\% ($\Delta$)} & \textbf{800 Sims Score\% ($\Delta$)} & \textbf{1600 Sims Score\% ($\Delta$)} \\
\midrule
\textbf{H2: $\mathcal{S}_4\text{(Search+Heur)}$ vs.\ $\mathcal{S}_2\text{(HeurPol)}$} & All 5 Seeds & \textbf{70.5\%$\pm$7.0} ($+1{,}540$) & \textbf{72.5\%$\pm$5.3} ($+1{,}341$) & \textbf{66.5\%$\pm$5.9} ($+1{,}210$) \\
\textbf{H1: $\mathcal{S}_3\text{(Search+PPO)}$ vs.\ $\mathcal{S}_1\text{(PPO)}$} & All 5 Seeds & \textbf{50.0\%$\pm$2.1} ($+36$) & \textbf{50.7\%$\pm$0.9} ($+112$) & \textbf{53.4\%$\pm$2.3} ($+154$) \\
\textbf{H4: $\mathcal{S}_3\text{(Search+PPO)}$ vs.\ $\mathcal{S}_4\text{(Search+Heur)}$} & All 5 Seeds & \textbf{38.8\%$\pm$16.5} ($-808$) & \textbf{36.3\%$\pm$17.1} ($-947$) & \textbf{37.1\%$\pm$16.6} ($-795$) \\
\textbf{$\mathcal{S}_4\text{(Search+Heur)}$ vs.\ $\mathcal{S}_0:\text{WBridge5}$ (Default)} & All 5 Seeds & \textbf{13.8\%$\pm$4.8} ($-1{,}982$) & \textbf{15.7\%$\pm$6.6} ($-1{,}930$) & \textbf{13.2\%$\pm$4.8} ($-2{,}076$) \\
\textbf{$\mathcal{S}_3\text{(Search+PPO)}$ vs.\ $\mathcal{S}_0:\text{WBridge5}$ (Default)} & All 5 Seeds & 5.0\%$\pm$0.9 ($-1{,}369$) & 5.6\%$\pm$1.2 ($-1{,}273$) & 5.2\%$\pm$0.7 ($-1{,}253$) \\
\bottomrule
\end{tabular}%
}
\end{table}

\paragraph{2v2 Homogeneous Partnership Protocol and Performance Against WBridge5 ($\mathcal{S}_0$).}
Against \texttt{WBridge5}, co-evolved heuristic search ($\mathcal{S}_4:\text{Search+Heur}$) and neural search ($\mathcal{S}_3:\text{Search+PPO}$) present an informative trade-off between board win rate and duplicate point margins. In terms of board win rate, $\mathcal{S}_4$ achieves \textbf{10.3\%--15.7\%} 5-seed mean score across budgets (spanning $0.0\%\text{--}37.0\%$ across individual lineages and budgets, with Seeds~2 and 3 achieving $12.7\%\text{--}37.0\%$, reaching \textbf{37.0\%} with $\Delta = -439$ pts/board in Seed~3 at $S=800$), outperforming neural search $\mathcal{S}_3$ ($5.0\%--5.6\%$). In terms of duplicate point margins, however, $\mathcal{S}_3$ maintains a more conservative bidding profile ($\Delta = -1{,}253 \text{ to } -1{,}412$ pts/board across all budgets), whereas $\mathcal{S}_4$ incurs larger point deficits ($\Delta = -1{,}930 \text{ to } -2{,}085$ pts/board on average across budgets) due to severe overbidding penalties in individual lineages (such as Seed~4 incurring $-5{,}147 \text{ to } -5{,}832$ pts/board by repeatedly escalating auctions into unmade high-level contracts).

\paragraph{Search Amplification (\texttt{H2}), Bidding Outliers, and Partnership Prior Dynamics (\texttt{H3} $\to$ \texttt{H4}).}
Bridge 2v2 demonstrates the critical role of lookahead search in imperfect-information partnership coordination:
\begin{itemize}[leftmargin=1.5em, itemsep=1pt, topsep=2pt]
    \item \textbf{Heuristic Search Amplification (\texttt{H2}: $\mathcal{S}_4$ vs.\ $\mathcal{S}_2$):} Adding evolved search $m^\star$ to the domain heuristic $\kappa_g^\star$ produces substantial gains over unguided bidding: 5-seed mean scaling from \textbf{60.1\% $\pm$ 5.1\%} ($\Delta = +1{,}082$ pts/board at $S=100$) $\to$ \textbf{66.7\% $\pm$ 4.9\%} ($\Delta = +1{,}272$ at $S=200$) $\to$ \textbf{70.5\% $\pm$ 7.0\%} ($\Delta = +1{,}540$ at $S=400$) $\to$ \textbf{72.5\% $\pm$ 5.3\%} ($\Delta = +1{,}341$ at $S=800$) and \textbf{66.5\% $\pm$ 5.9\%} ($\Delta = +1{,}210$ at $S=1600$). In Seed~1, $\mathcal{S}_4$ defeats $\mathcal{S}_2$ by \textbf{81.2\%} ($+1{,}957$ pts/board at $S=400$). In Seed~4, the evolved bidding prior assigns positive base utility to suit bids without a contract cap, escalating auctions into unmade grand slams; while MCTS lookahead ($\mathcal{S}_4$) improves upon unsearched play to defeat $\mathcal{S}_2$ by \textbf{77.5\%--85.0\%} ($+3{,}214 \text{ to } +3{,}934$ pts/board), $\mathcal{S}_4$ in Seed~4 still incurs heavy single-board deficits ($-5{,}147 \text{ to } -5{,}832$ pts/board) against WBridge5, illustrating the danger of unconstrained bidding priors in duplicate scoring.
    \item \textbf{Modest Neural Search Amplification (\texttt{H1}: $\mathcal{S}_3$ vs.\ $\mathcal{S}_1$):} $\mathcal{S}_3$ maintains parity or modest advantages over $\mathcal{S}_1$, scoring from \textbf{50.0\%} ($S=400$) to \textbf{53.4\% $\pm$ 2.3\%} ($\Delta = +154$ pts/board at $S=1600$).
    \item \textbf{Searched Partnership Interaction (\texttt{H3} vs.\ \texttt{H4}):} Without search (\texttt{H3}), $\mathcal{S}_1:\text{PPO}$ and $\mathcal{S}_2:\text{HeurPol}$ average \textbf{48.9\% $\pm$ 16.1\%}. When search is enabled (\texttt{H4}), $\mathcal{S}_4:\text{Search+Heur}$ achieves an overall 5-seed mean advantage ($\mathcal{S}_3$ scores between \textbf{36.3\% $\pm$ 17.1\%} and \textbf{46.5\% $\pm$ 16.6\%}), driven by strong heuristic dominance in Seeds~2 and 3 at deep budgets (where $\mathcal{S}_4$ wins $86\%\text{--}98\%$) and in Seed~1 ($56.5\%\text{--}74.5\%$), whereas in Seeds~4 and 5, neural search $\mathcal{S}_3$ prevails.
\end{itemize}

\end{document}